\RequirePackage{fix-cm}
\documentclass[twocolumn, natbib]{svjour3}          % twocolumn

\smartqed  % flush right qed marks, e.g. at end of proof
\usepackage{times}
\usepackage{epsfig}
\usepackage{graphicx}
\usepackage{amsmath}
\usepackage{amssymb}
\usepackage{booktabs}
\usepackage{algorithm}
\usepackage{algorithmic}
\usepackage{multirow}
\usepackage{subfigure}
\usepackage{url}
\usepackage{rotate}
\usepackage{bm}
\usepackage{setspace}
\usepackage{color}
\usepackage{natbib}
\setcitestyle{authoryear,open={(},close={)}} % cite style
\usepackage[pagebackref=true,breaklinks=true,colorlinks,citecolor=blue,bookmarks=false]{hyperref}
\usepackage{colortbl}

\begin{document}
	
	\title{Bilevel Fast Scene Adaptation for Low-Light Image Enhancement}
	
	
	
	\author{Long Ma$^1$\and
		Dian Jin$^2$\and
		Nan An$^3$\and
		Jinyuan Liu${^4}$\and
		Xin Fan$^{1}$\and
		Zhongxuan Luo$^{3}$ \and
		Risheng Liu$^{1}$
	}
	
	\authorrunning{International Journal of Computer Vision}
	
	\institute{
		Corresponding author: Risheng Liu \at
		\email{\textit{rsliu@dlut.edu.cn}}\\ 
		\and
		$^1$ DUT-RU International School of Information Science \& Engineering, Dalian University of Technology, Dalian, 116024, China. \\
		$^2$ Xiaomi AI Lab, Beijing, 100085, China.\\
		$^3$ School of Software Technology, Dalian University of Technology, Dalian, 116024, China.\\
		$^4$ School of Mechanical Engineering, Dalian University of Technology, Dalian, 116024, China.\\
	}
	
	\date{Received: date / Accepted: date}

	\maketitle
	\begin{abstract}
		Enhancing images in low-light scenes is a challenging but widely concerned task in the computer vision. The mainstream learning-based methods mainly acquire the enhanced model by learning the data distribution from the specific scenes, causing poor adaptability (even failure) when meeting real-world scenarios that have never been encountered before. 
		The main obstacle lies in the modeling conundrum from distribution discrepancy across different scenes. To remedy this, we first explore relationships between diverse low-light scenes based on statistical analysis, i.e., the network parameters of the encoder trained in different data distributions are close. We introduce the bilevel paradigm to model the above latent correspondence from the perspective of hyperparameter optimization. A bilevel learning framework is constructed to endow the scene-irrelevant generality of the encoder towards diverse scenes (i.e., freezing the encoder in the adaptation and testing phases). Further, we define a reinforced bilevel learning framework to provide a meta-initialization for scene-specific decoder to further ameliorate visual quality. Moreover, to improve the practicability, we establish a Retinex-induced architecture with adaptive denoising and apply our built learning framework to acquire its parameters by using two training losses including supervised and unsupervised forms. Extensive experimental evaluations on multiple datasets verify our adaptability and competitive performance against existing state-of-the-art works. The code and datasets will be available at \url{https://github.com/vis-opt-group/BL}.  
		
	\end{abstract}
	

	\begin{figure*}
		\centering
		\begin{tabular}{c@{\extracolsep{0.35em}}c@{\extracolsep{0.35em}}c@{\extracolsep{0.35em}}c}
			\includegraphics[width=0.24\linewidth]{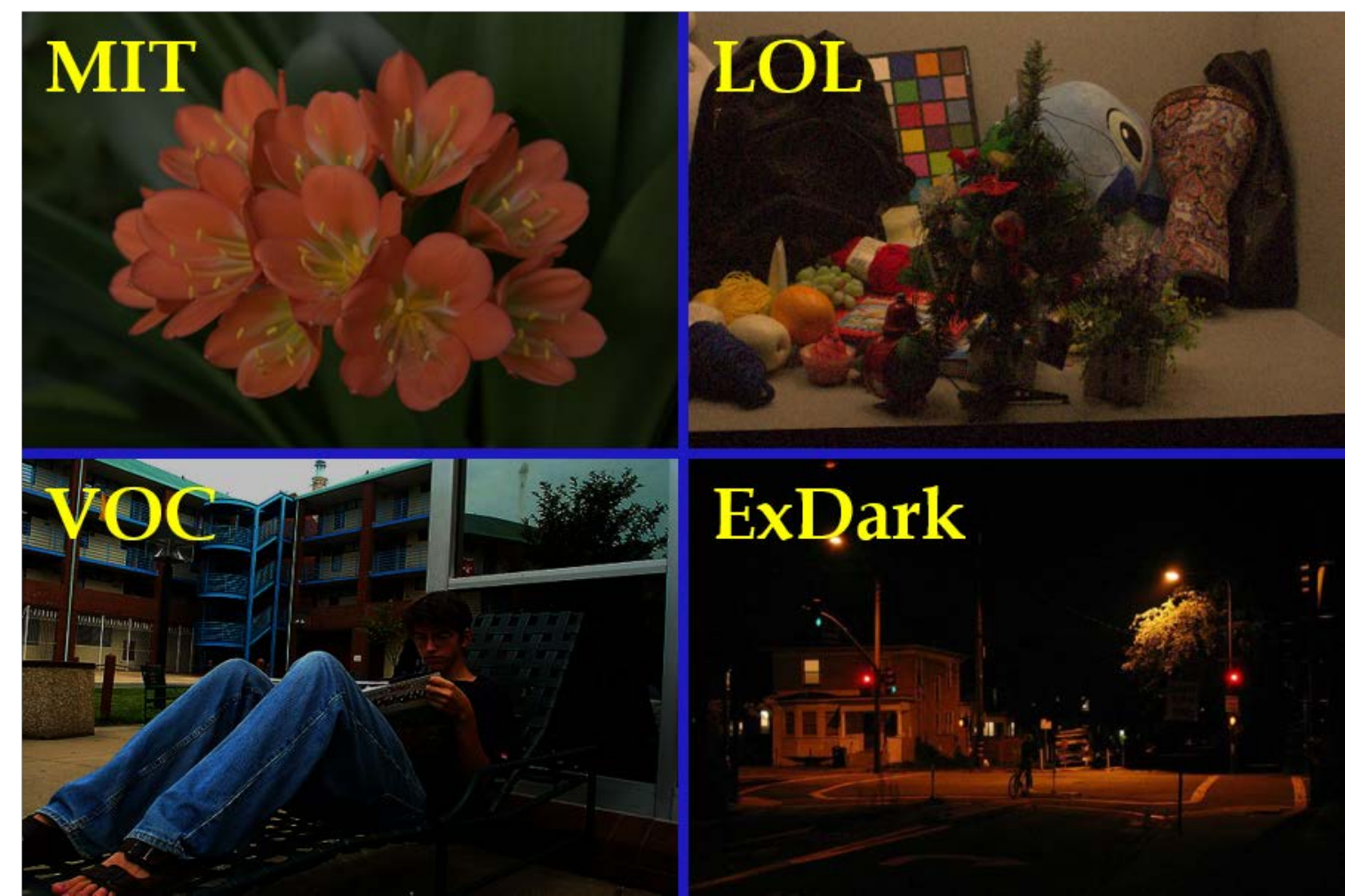}&
			\includegraphics[width=0.24\linewidth]{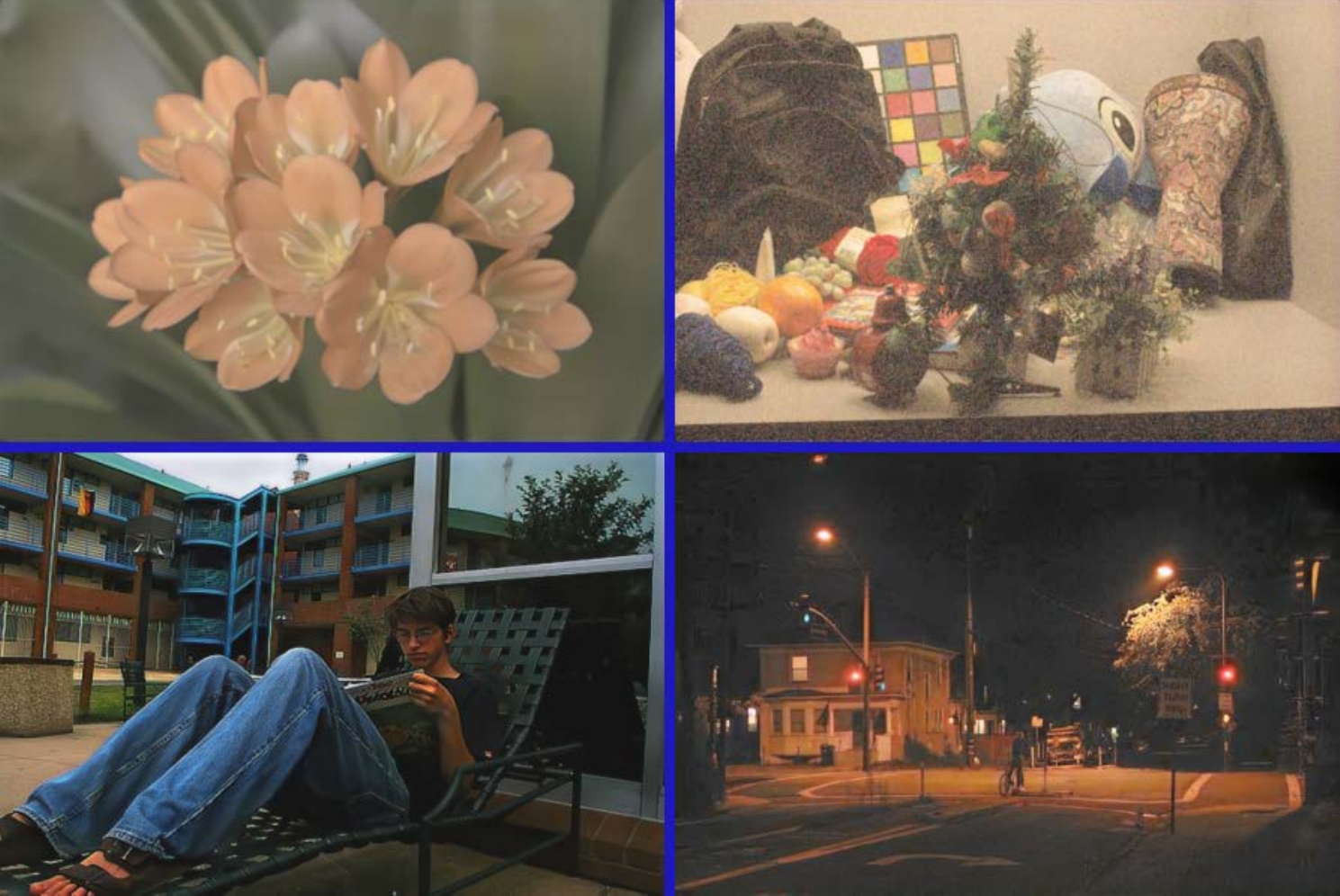}&
			\includegraphics[width=0.24\linewidth]{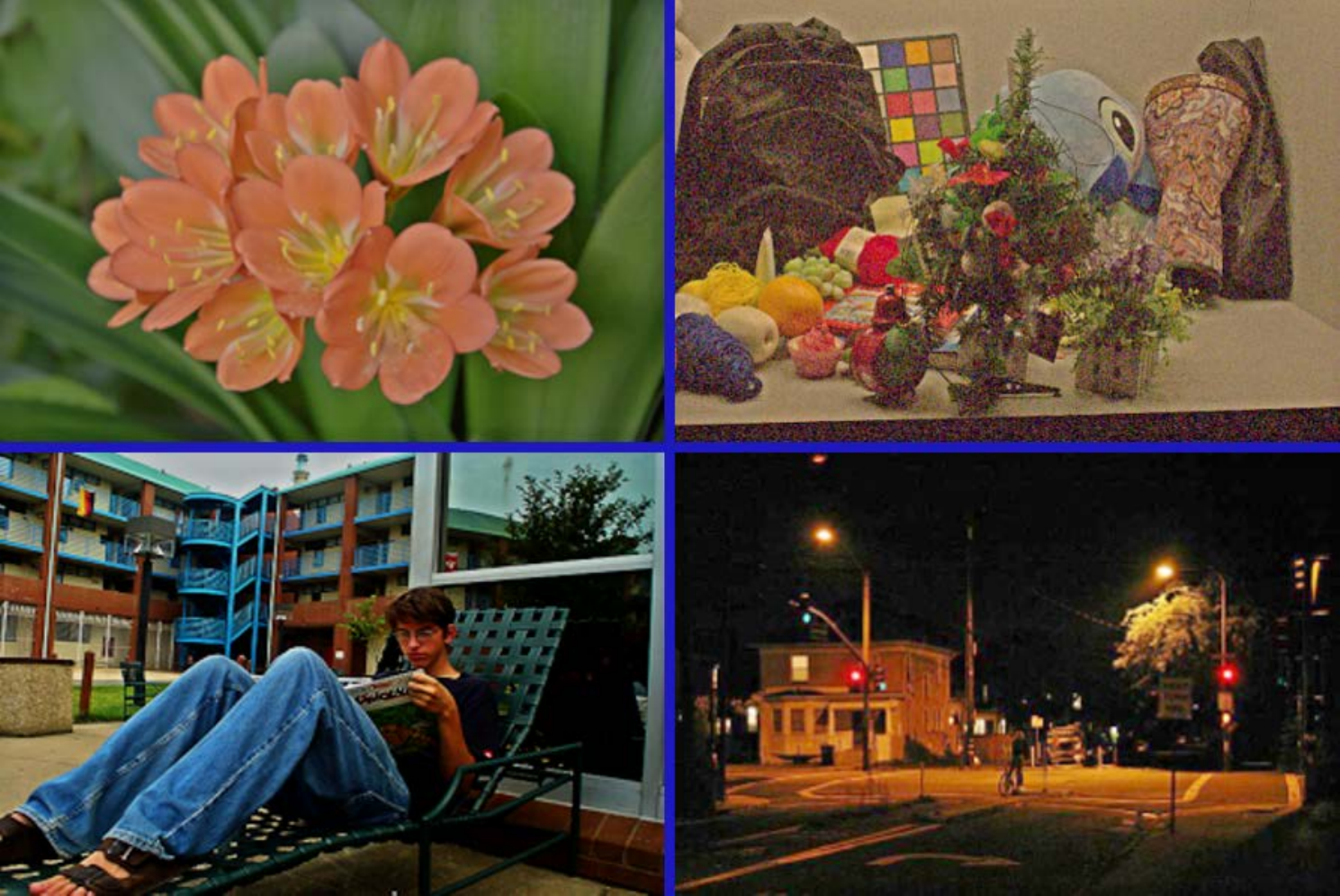}&	
			\includegraphics[width=0.24\textwidth]{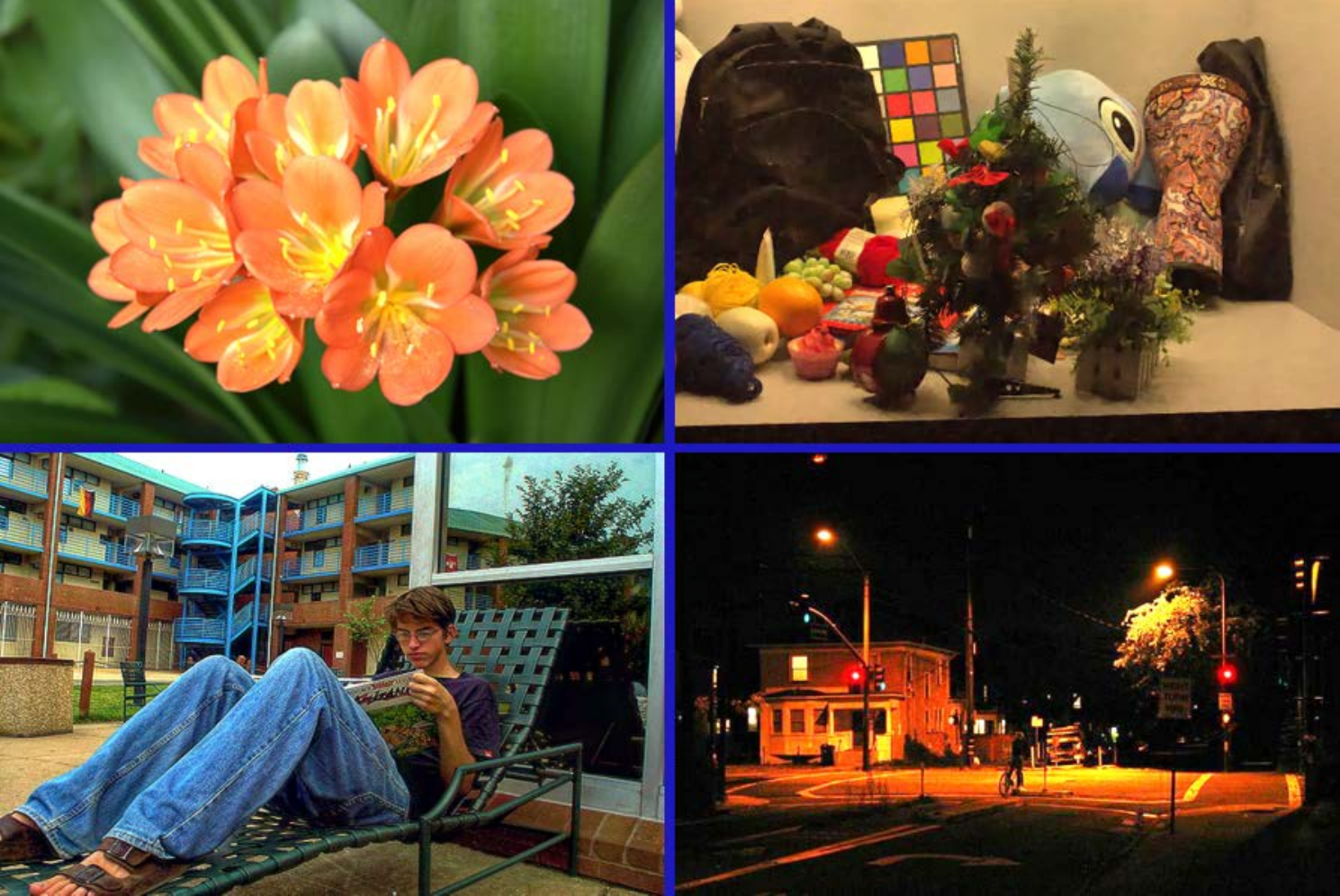}\\
			\footnotesize Input&\footnotesize FIDE&\footnotesize SCL&\footnotesize Ours\\
		\end{tabular}
		%	\vspace{-0.4cm}
		\caption{Visual comparison among two representative recently-proposed methods (including FIDE~\citep{xu2020learning} and SCL~\citep{liang2022semantically}) and our proposed algorithm on four different benchmarks (including MIT~\citep{fivek}, LOL~\citep{Chen2018Retinex}, VOC~\citep{lv2021attention}, ExDark~\citep{Exdark_dataset}). Obviously, our method realizes the best visual quality on all challenging scenarios.}
		\label{fig:firstfig}
	\end{figure*}
	
	\section{Introduction}
	Images captured in low-light conditions tend to appear problems like poor contrast and high noise, which seriously affect image quality. Unfortunately, there are many inevitable challenging shooting conditions. Low-light images reduce the visual quality as well as hinder the following industrial application content. Specifically, it affects some high-level tasks, e.g. detection~\citep{wang2022unsupervised,liang2021recurrent,cui2021multitask}, segmentation~\citep{wu2021one,sakaridis2019guided,gao2022cross} and tracking~\citep{ye2022unsupervised}. Therefore, the study of low-light image enhancement has strong practical significance. 
	
	Benefiting from the flourishing development of Convolutional Neural Networks (CNNs)~\citep{gu2018recent}, constructing CNNs for low-light image enhancement~\citep{Chen2018Retinex,zhang2019kindling,ma2022toward} has become the principal measure. Data is the foundation stone for learning a CNN model and can be divided into with and without reference images, corresponding to the supervised~\citep{xu2020learning,zheng2021adaptive} and unsupervised~\citep{li2021learning, jiang2019enlightengan} learning paradigms which are two patterns in CNN-based low-light image enhancement. 
	However, a common issue with these existing techniques is that they are just effective enough for images that possess a similar distribution to the training dataset since the regular end-to-end learning paradigm. It causes that these works usually need to spend too much energy in learning a completely new model for unseen low-light scenes. In other words, they have a limited solving capacity and cannot adapt to other low-light scenes quickly and effectively. 
	
	To endow the fast adaptation for the learned model, a series of works~\citep{lee2020self,park2020fast,chi2021test} based on meta-learning~\citep{finn2017model} have been developed for different low-level vision tasks (\textit{have not been investigated in the field of low-light image enhancement}). 
	Most of them explicitly define meta-initialization~\citep{liu2021investigating} learning as the pathway to realizing fast adaptation. Meta-initialization learning is to acquire a general initialization of network parameters from multiple tasks, then just perform the fine-tuning process when meeting the new task. Indeed, these works perform a fast adaptation ability towards new tasks (commonly defined as the new scene that never encountered before). However, the above approach of achieving fast adaptation is just to connect multiple tasks from the initial status, ignoring the unified representation in the feature space to abandon the opportunity of a shortcut for fast adaptation.  
	
	In this work,  we develop a new bilevel learning scheme for fast adaptation by bridging the gap between low-light scenes in the learning procedure. As shown in Fig.~\ref{fig:firstfig}, we can easily observe that our proposed method realizes the best visual quality on different challenging scenarios against other advanced methods. It indicates that we indeed achieve the goal of adapting scenarios.
	% Concretely, taking Retinex into consideration, we first design a Retinex-induced encoder-decoder and introduce an adaptive denoising mechanism for covering more cases. Then from the hyperparameter learning perspective, we establish a novel bilevel learning scheme to realize the fast adaptation for general low-light scenarios. A sequence of comparisons and analysis are performed to verify our superiority and effectiveness.
	Our main contributions can be concluded as the following five-folds. 
	\begin{itemize}
		\item To the best of our knowledge, we are the first to focus on fast adaptation for low-light image enhancement from the view of hyperparameter optimization, to ameliorate the adaptability towards unknown scenes. 
		\item We propose to use bilevel paradigm to model the latent scene-irrelevant correspondence which presents that the parameters are close between encoders trained in different data distributions based on statistical exploration.
		\item We design a new Bilevel Learning (BL) framework to learn a scene-irrelevant encoder for freezing its parameters when meeting unknown scenes. It reduces training costs to support fast adaptation nicely.  
		\item Considering the indefinite initialization of scene-specific decoder, we establish a Reinforced Bilevel Learning (RBL) framework to provide a meta-initialization to further improve adapting efficiency. 
		\item To handle different challenging scenes, we build a new Retinex-induced encoder-decoder architecture with an adaptive denoising mechanism and define two training patterns including supervised and unsupervised forms.   
		%	\item Extensive experiments on four datasets are conducted to indicate our superiority against other state-of-the-art methods. Elaborated analytical verifications are also performed to reveal our effectiveness.
	\end{itemize}
	
	This work is extended from our preliminary version~\citep{jin2021bridging} with the promotion from three key aspects. 
	\begin{itemize}
		\item We provide an elaborate analysis about the research motivation by presenting the distribution discrepancy and exchanging learned architectures. 
		\item We develop a reinforced bilevel learning to energizing the initialization of the scene-specific decoder for further improving performances. 
		\item Sufficient comparisons on more benchmarks with recent advanced methods and detailed analyses are performed to prove our effectiveness. 
	\end{itemize}

	\begin{figure*}[t]
		\centering
		\begin{tabular}{c}
			\includegraphics[width=0.98\linewidth]{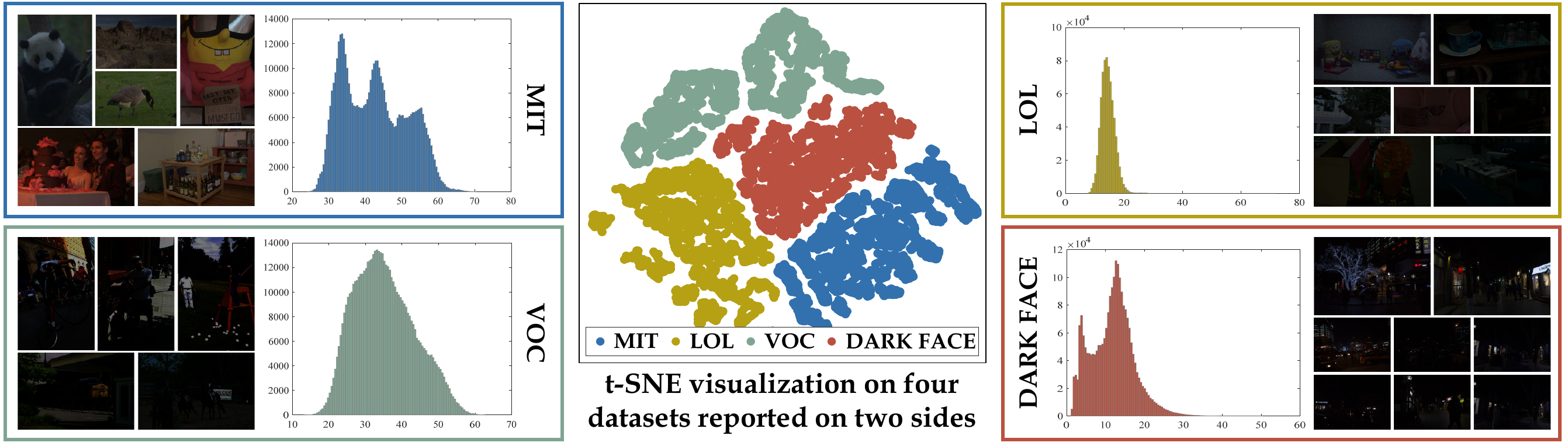}\\
		\end{tabular}
		\caption{Demonstrating distribution discrepancy of different datasets. On two sides, we randomly sampled 100 low-light images each from four datasets to plot the histogram. The middle subfigure plots the t-SNE distribution~\citep{van2008visualizing} on these four datasets. }
		\label{fig: datadiscrepancy}
	\end{figure*}
	
	\section{Related Works}
	In this part, we make a comprehensive review of related works including CNNs for low-light image enhancement, and fast adaptation for low-level vision. 
	
	\textbf{CNNs for low-light image enhancement.}$\;$
	Recently, accounting for the rapid development of deep learning, advanced algorithms in the field of low-light image enhancement continued to emerge. Many CNN-based methods have achieved surprising results in low-light image enhancement. According to different learning patterns with different scene adaptation abilities, existing works can be roughly divided into two categories: supervised and unsupervised learning. 
	
	Supervised learning is the most common paradigm which benefits from the development of the paired low-/normal-light datasets (e.g., LOL~\citep{Chen2018Retinex}, MIT~\citep{MIT_Adobe_5K}, LSRW~\citep{hai2021r2rnet}, and VOC~\citep{lv2021attention}).  
	RetinexNet~\citep{Chen2018Retinex} provided an end-to-end framework that combined Retinex theory with deep networks. KinD~\citep{zhang2019kindling} adjusted RetinexNet to estimate the illumination and added a series of training losses. DeepUPE~\citep{Wang_2019_CVPR} could adapt to the complex illumination of the ground truth by learning the mapping between low-light images and illumination. FIDE~\citep{xu2020learning} proposed to restore image objects in the low-frequency layer, and enhanced high-frequency details on the restored image. DSNet~\citep{zhao2021deep} developed a symmetrical deep network by combining an invertible feature transformer and two pairs of pre-trained encoder-decoder. The work in~\citep{zheng2021adaptive} proposed an adaptive unfolding total variation network to realize noise reduction and detail preservation. Wu~\emph{et al.}~\citep{wu2022uretinex} designed a URetinex-Net by unfolding the Retinex-based optimization and learning data-dependent priors. DCC-Net~\citep{zhang2022deep} followed the ``divide and conquer'' collaborative strategy to construct a deep color consistent network by separately restoring gray images and color distribution. 
	However, limited to the specific distribution from the paired data, the main drawback of this paradigm is the generalization capabilities of models.

	\begin{figure*}[t]
		\centering
		\footnotesize
		\begin{tabular}{c@{\extracolsep{1em}}c}
			\includegraphics[width=0.48\linewidth]{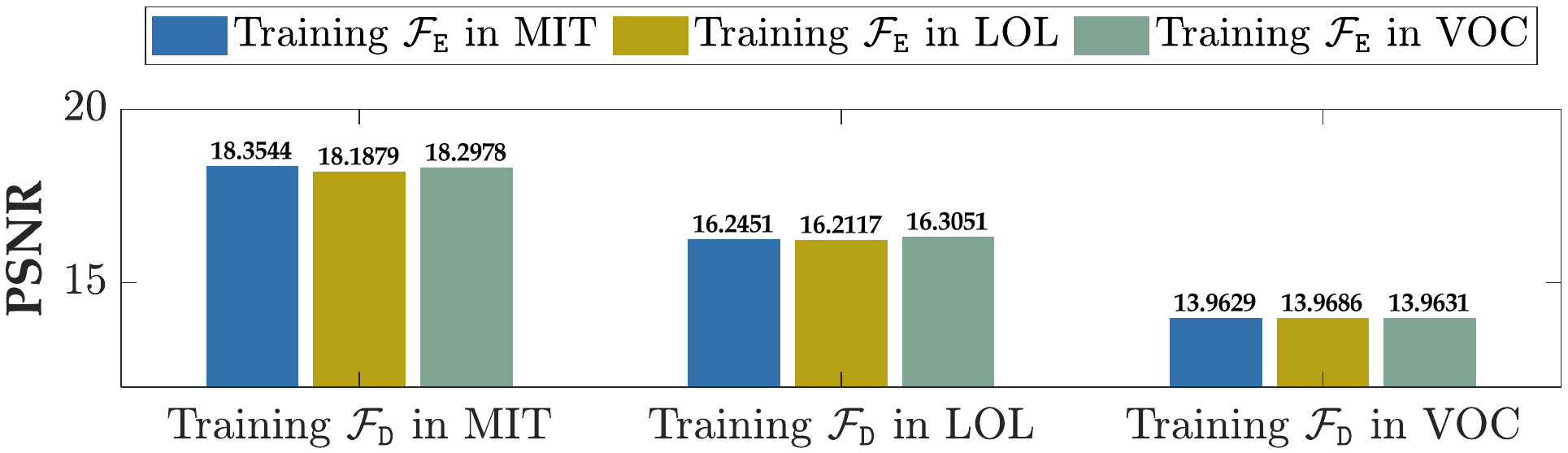}&
			\includegraphics[width=0.48\linewidth]{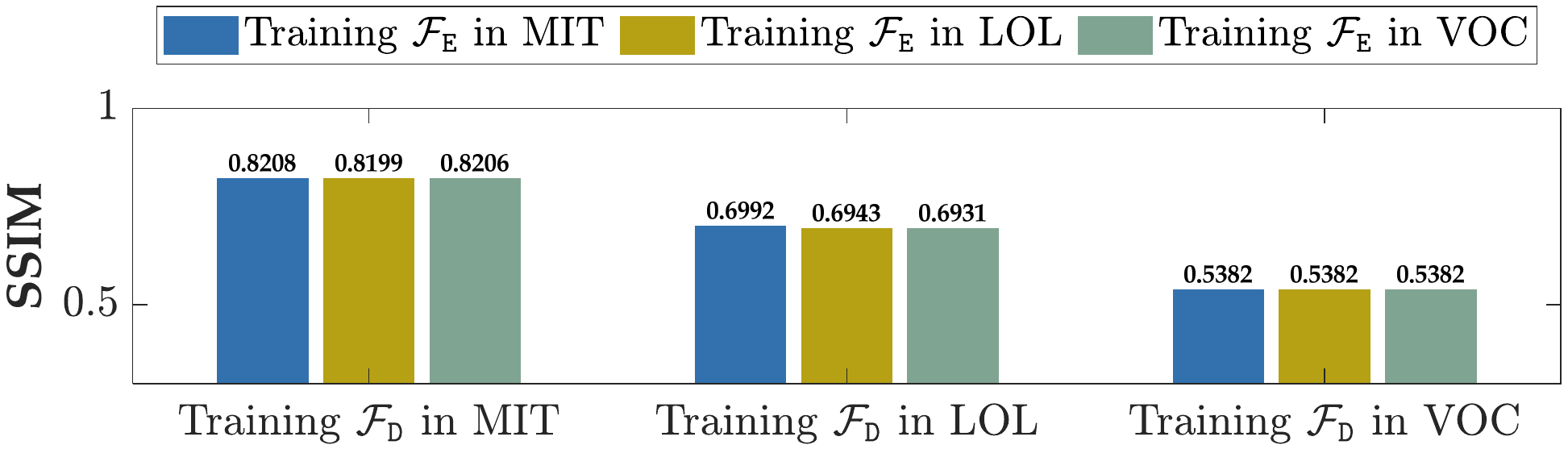}\\
			\multicolumn{2}{c}{(a) Numerical results in full-reference metrics }\\
			\includegraphics[width=0.48\linewidth]{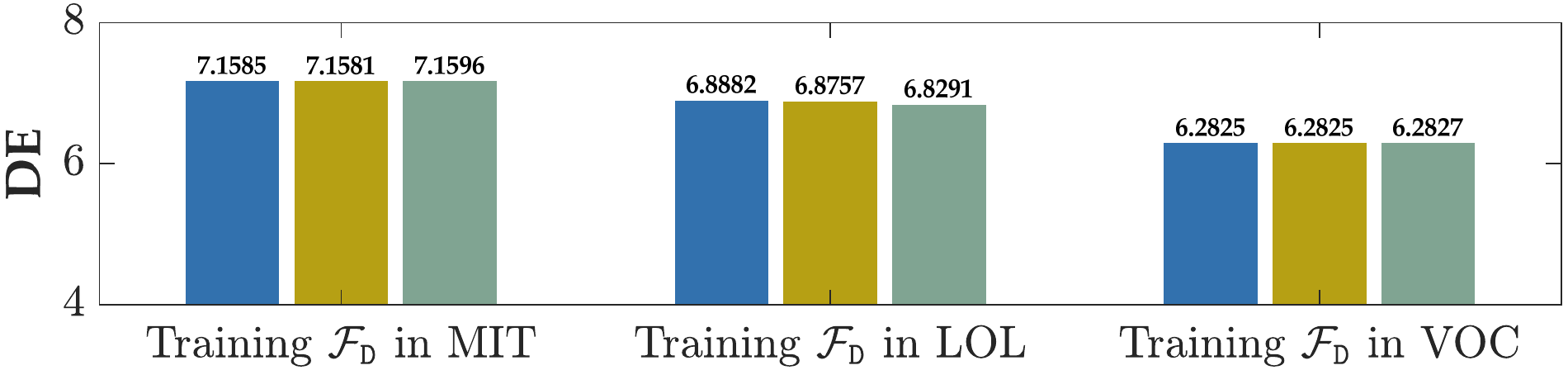}&
			\includegraphics[width=0.48\linewidth]{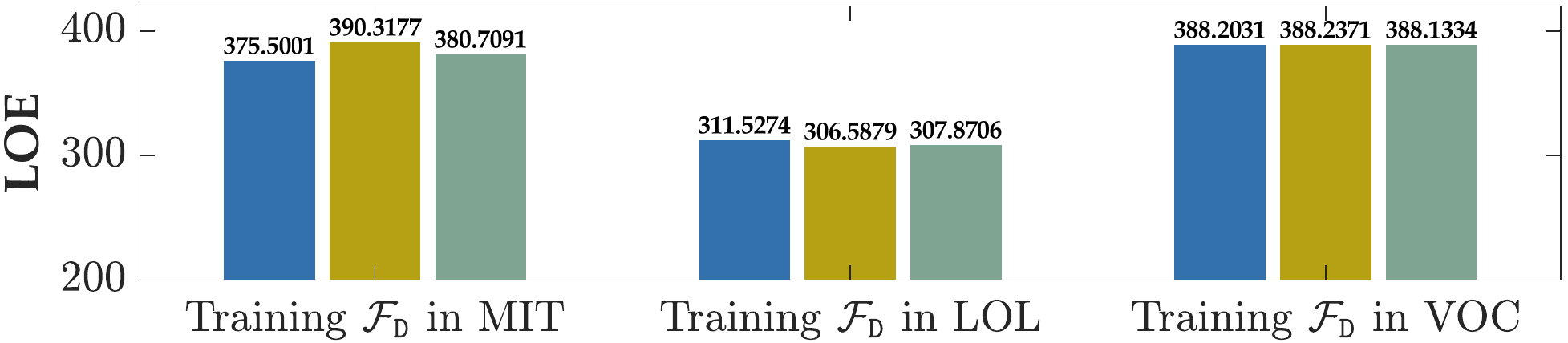}\\
			\multicolumn{2}{c}{(b) Numerical results in no-reference metrics }\\
		\end{tabular}
		\caption{Quantitative scores after exchanging the encoder and decoder trained in different datasets. Note that the testing dataset for each case is the same one used in training $\mathcal{F}_{\mathtt{D}}$. }
		\label{fig: Motivation}
	\end{figure*}
	
	Unsupervised learning focuses on learning the model without the support of data with reference. 
	EnGAN~\citep{jiang2019enlightengan} first raised that unpaired training scheme based on generative-adversarial mode could be introduced into low-light image enhancement, which enhanced the generalization performance of the algorithm to a certain extent. Nevertheless, it was unstable and prone to tail shadow and color bias. The work in~\citep{ma2021learning} constructed a context-sensitive decomposition network and also adopted the pattern of generative-adversarial. 
	ZeroDCE~\citep{Guo_2020_ZeroDCE}, an unsupervised method, designed a lightweight deep network to estimate pixels and high-order curves for dynamic range adjustment of a given image. However, introducing a series of training losses weakened the excellent generalization of the unsupervised learning without the paired/unpaired data. 
	The work in~\citep{liang2021semantically} developed a semantically contrastive learning mechanism by three constraints of contrastive learning, semantic brightness consistency, and feature preservation to guarantee the consistencies of color, texture, and exposure. Ma~\emph{et al.}~\citep{ma2022toward} built a self-calibrated illumination learning framework for unsupervised low-light image enhancement.

	%A pre-trained network can not be applied to all the images collected from different scenes and different devices. It is because they only fit the current dataset in the training process, and lack a certain degree of generalization. At present, the only way to solve this problem is to construct multiple training sets so as to fit different test scenarios. Obviously, because the test scenarios are numerous and complex, it is unrealistic to construct training sets according to each scenario. Therefore, the problem restricts the development of deep learning methods.
	In summary, most above-mentioned low-light image enhancement algorithms were still addressing one or some specific issues in one single scene from the training dataset. They are difficult to adapt to other challenging scenes which contain unseen distributionss. In other words, the similarities and differences between various scenes should be exploited fully to improve the model's generalization. 
	
	\textbf{Fast adaptation for low-level vision.}$\;$
	Currently, the adaptability of models in different scenes has received unprecedented attention. Methods with fast adaptation and less training costs are becoming more and more popular. The emergent meta-learning~\citep{wang2021variational,vuorio2019multimodal,liu2021investigating,gao2022curvature} has attracted broad attention in various fields on the strength of powerful ability of ``learning to learn''. 
	Especially, the popular Model-Agnostic Meta-Learning (MAML)~\citep{finn2017model} learns the meta-initialization from multiple tasks to receive a fast adaptation property towards the new task, which has been widely applied to low-level vision field sto realize the fast adaptation. 
	
	Lee~\emph{et al.}~\citep{lee2020self} combined the internal statistics of a natural image used for exploiting the self-similarity, and a MAML-based meta-learning paradigm to construct a fast adaptation denoising algorithm. The paper in~\citep{park2020fast} applied the meta-learning for super-resolution and utilized the patch-recurrence property of the natural image to boost the performance. The work in~\citep{chi2021test} introduced a self-reconstruction auxiliary task for primary deblurring and designed the meta-auxiliary learning to endow the fast adaptation ability for dynamic scene deblurring. 
	To solve the difficulty of learning a general dehazing model on multiple datasets, a multi-domain learning approach for dehazing was constructed in~\citep{liu2022towards} by designing the helper network for boosting the performance during the test time and introducing the meta-learning paradigm to intensify the generality of the helper network. 
	Choi~\emph{et al.}~\citep{choi2021test} proved that the meta-learning framework could be easily applied to any video frame interpolation network with only a few fine-tunings.
	
	Although many low-level vision fields had begun to consider the problem of model generalization capabilities as described above, because of the diversity of low-light scenes, the low-light image enhancement field never had a good application in this regard.

	\section{Exploring Relationships among Low-Light Scenes}
	Here, we explore the relationship between diverse low-light data distributions from two aspects. On one hand, we first present the difference among diverse datasets from a statistical perspective. On the other hand, we excavate the latent relationship among datasets from the trained model. 
	
	We know that the low-light scenes are miscellaneous including the appearance, level of luminance, and so on. Here we consider four different datasets including MIT~\citep{fivek}, LOL~\citep{Chen2018Retinex}, VOC~\citep{lv2021attention}, and DARK FACE~\citep{yang2020advancing}. They are with different objects, scenes, and levels of luminance. As shown in Fig.~\ref{fig: datadiscrepancy}, we plot the low-light examples and distribution for these datasets, and their relationship in the same dimension. There exists an apparent distribution discrepancy between these datasets, that is to say, it is extremely difficult to establish a unified model for adapting them. It actually reveals why most of existing works need to retrain their whole model in the unseen scenes. 
	
	A question worth pondering is whether there exists one possibility to bridge these various datasets. Here we would like to explore the relationship among models trained by different datasets. We define the enhancement architecture as an encoder-decoder\footnote{More details can be found in Sec.~\ref{sec: architecture}.} architecture~\citep{ronneberger2015u}, formulated as $\mathcal{F}=\mathcal{F}_{\mathtt{E}}\cup\mathcal{F}_{\mathtt{D}}$ ($\mathcal{F}$: encoder-decoder, $\mathcal{F}_{\mathtt{E}}$: encoder, and $\mathcal{F}_{\mathtt{D}}$: decoder). We adopt three various datasets (including MIT, LOL, and VOC) presented in Fig.~\ref{fig: datadiscrepancy} to analyze this process. 
	
	To be concrete, we first train three distribution-specific models on all these datasets. We adopt the MSE loss to train it and the numbers of training and testing samples follow the settings of the adaptation stage in Table~\ref{table: benchmarks}.
	In the testing phase for the specific dataset (e.g., MIT), we perform the results of three versions including the original model trained on MIT, and the other two models by substituting the original encoder as the trained on other datasets (i.e., LOL and VOC). By performing this procedure on these three datasets and calculating four metrics consisting of two full-reference metrics (i.e., PSNR and SSIM), and two no-reference metrics (i.e., DE~\citep{shannon1948mathematical} and LOE~\citep{wang2013naturalness}), we plot these numerical results in Fig.~\ref{fig: Motivation}. Evidently, although the encoder and decoder are not trained in the same dataset, these scores on different settings all approach the same state. That is to say, the encoder has a unified representation that is scene-irrelevant.
	% Further, we exhibit the learned features among different settings in the same testing sample. As shown in Fig.~\ref{label}, 
	
\begin{figure}
		\centering
		\begin{tabular}{c}
			\includegraphics[width=0.95\linewidth]{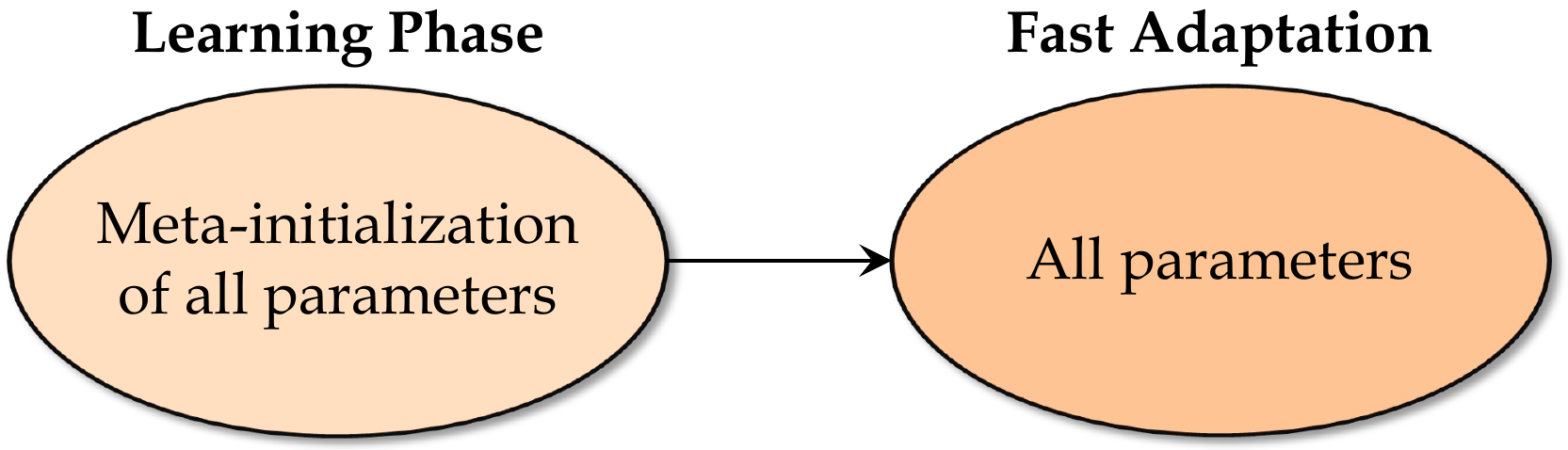}\\
			\footnotesize (a) Traditional fast adaptation scheme\\
			\includegraphics[width=0.95\linewidth]{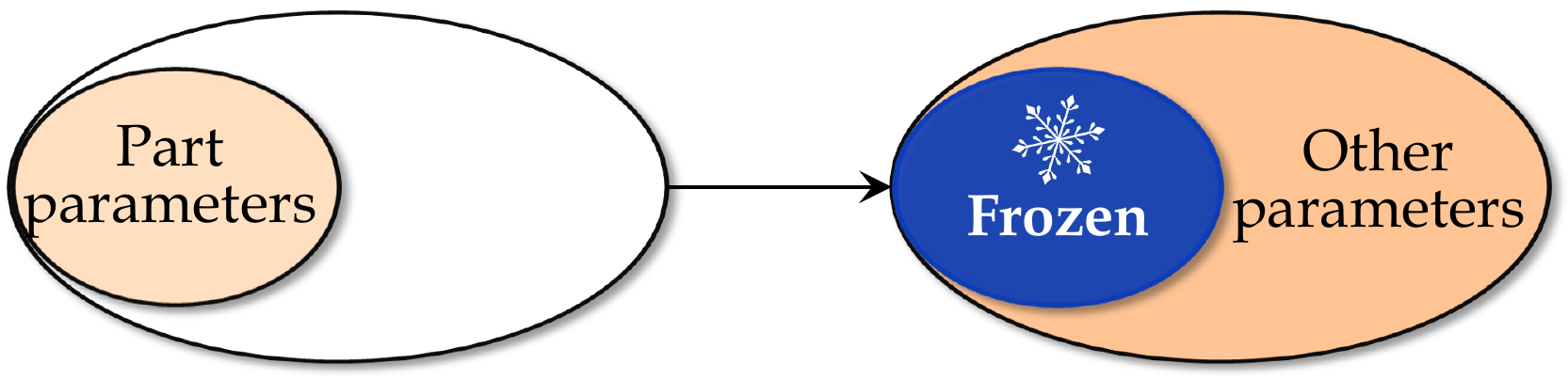}\\
			\footnotesize (b) Our built fast adaptation scheme \\
		\end{tabular}
		\caption{Comparing different fast adaptation schemes. (a) represents the traditional scheme \citep{lee2020self,park2020fast,choi2021test,liu2022towards} based on meta-learning for fast adaptation. (b) is our built fast adaptation scheme based on bilevel learning. In contrast, the latter can largely reduce the computational burden in the adaptation stage by just fine-tuning part parameters, rather than all parameters. }
		\label{fig: learningmechanisms}
	\end{figure}

	Actually, the phenomenon above manifests there exists a pattern to alleviate the transferred burden towards unseen distributions. It exactly coincides with the goal of fast adaptation in the field of low-level vision~\citep{choi2021test,liu2022towards}. However, a substantive difference lies in the regular fast adaptation mostly learns a meta-initialization based on the meta-learning mechanism~\citep{lee2020self,park2020fast,choi2021test,liu2022towards}, i.e., \textit{all parameters need to be finetuned when applied to unknown scenes, see Fig.~\ref{fig: learningmechanisms} (a)}. Different from it, our finding above is to provide a generic encoder whose parameters are frozen when meeting unknown scenes, i.e., \textit{part of parameters are frozen, and rest need to be finetuned,  see Fig.~\ref{fig: learningmechanisms} (b)}. Briefly, the desired fast adaptation ability is more challenging than traditional fast adaptation.

	In the following, based on the observations above, we introduce bilevel learning for fast adaptation to improve the model generality and reduce the burden of transferring.

	%Different from existing works that train the designed architecture by the regular back-propagation, we would like to introduce a new learning procedure to endow the ability of fast adaptation for our designed Retinex-induced encoder-decoder from the perspective hyperparameter optimization~\citep{feurer2019hyperparameter,falkner2018bohb}. 

	\section{Fast Adaptation via Bilevel Learning}\label{sec: FA}
	This section first provides a modeling perspective from hyperparameter optimization for fast adaptation among diverse datasets, then constructs a bilevel learning scheme and its reinforced version for handling the built model. 
	
	%\begin{algorithm}[t]
	%	\caption{Bilevel Learning (BL)}\label{alg:BL}
	%	\begin{algorithmic}[1]
		%		\REQUIRE 
		%		The training and validation sets $\mathcal{D}_{\mathtt{tr}}$ and $\mathcal{D}_{\mathtt{val}}$, random initialization $\mathbf{u}$ and $\mathbf{v}$, and other hyper-parameters.
		%		\ENSURE The desired optimal solution $\mathbf{u}^*$. 
		%		\WHILE{not converged}
		%		\STATE  $\hat{\mathbf{v}} \leftarrow$ $\mathbf{v} - \nabla_{\mathbf{v}}f(\mathbf{u},\mathbf{v}; \mathcal{D}_\mathtt{tr})$; // Update $\mathbf{v}$.
		%		\STATE $\hat{\mathbf{u}} \leftarrow \mathbf{u} - \nabla_{\mathbf{u}}F(\mathbf{u},\mathbf{v};\mathcal{D}_{\mathtt{val}})$; // Update $\mathbf{u}$.
		%		%		\STATE Calculate $\nabla_{\mathbf{u}}F(\mathbf{u},\mathbf{v};\mathcal{D}_{\mathtt{val}})$ with Eq.~\eqref{eq: forth} and Eq.~\eqref{eq: fifth}.
		%		%		\ENDWHILE
		%		\ENDWHILE
		%		%		\STATE // Update  and  for TM.
		%		\RETURN $\mathbf{u}^*$.
		%	\end{algorithmic}
	%\end{algorithm}
	
	\subsection{Modeling with Bilevel Paradigm}
	In essence, many vision problems can be modeled as the specific optimization model. Fortunately, our desired fast adaptation scheme shown in Fig.~\ref{fig: learningmechanisms} (b) can be exactly presented as the hyperparameter optimization. Hyperparameter optimization~\citep{feurer2019hyperparameter,falkner2018bohb,liu2021investigating} is an emergent concept, which is defined as \textit{finding a tuple of hyperparameters that yields an optimal model which minimizes a predefined loss function on given independent data}\footnote{\url{https://en.wikipedia.org/wiki/Hyperparameter_optimization}.}. 
	In our designed scheme, the parameters in the learning phase can be viewed as the hyperparameters, and the other parameters can be viewed as the parameters. Then the hyperparameter optimization can be written as the following bilevel model
	\begin{equation}\label{eq: HO}
		\begin{aligned}
			&\min_{\mathbf{u}\in\mathcal{U}}\;F(\mathbf{u},\mathbf{v};\mathcal{D}_{\mathtt{val}}),\\
			&{s.t.}\;\mathbf{v}\in\mathcal{S}(\mathbf{u}),\;\mathcal{S}(\mathbf{u}):=\arg\min_{\mathbf{v}}\;f(\mathbf{u},\mathbf{v}; \mathcal{D}_\mathtt{tr}),
		\end{aligned}
	\end{equation}
	where $\mathbf{u}$ and $\mathbf{v}$ represent the hyperprameters and parameters, respectively. $\mathcal{S}(\mathbf{u})$ is the set of solution of the lower-level problem. 
	$F(\cdot)$ and $f(\cdot)$ are the objective for the upper and lower level, respectively. 
	$\mathcal{D}=\mathcal{D}_\mathtt{tr}\cup\mathcal{D}_{\mathtt{val}}$ denotes that the given dataset $\mathcal{D}$ is split into the training set $\mathcal{D}_{\mathtt{tr}}$ and validation set $\mathcal{D}_{\mathtt{val}}$. 
	Eq.~\eqref{eq: HO} actually presents an explicit relationship between hyperparameters and parameters. 
	%
	%\begin{algorithm}[t]
	%	\caption{Reinforced Bilevel Learning (RBL)}\label{alg:RBL}
	%	\begin{algorithmic}[1]
		%		\REQUIRE ss
		%		The training and validation sets $\mathcal{D}_{\mathtt{tr}}$ and $\mathcal{D}_{\mathtt{val}}$, random initialization $\mathbf{u}$ and $\mathbf{v}$, and other hyper-parameters.
		%		\ENSURE The desired optimal solution $\mathbf{u}^*$ and $\tilde{\mathbf{v}}^*$. 
		%		\WHILE{not converged}
		%		\STATE  $\hat{\mathbf{v}}$ $\leftarrow$ $\tilde{\mathbf{v}} - \nabla_{\tilde{\mathbf{v}}}f(\mathbf{u},\tilde{\mathbf{v}}; \mathcal{D}_\mathtt{tr})$; // Update $\mathbf{v}$.
		%		\STATE $\hat{\mathbf{u}} \leftarrow \mathbf{u} - \nabla_{\mathbf{u}}F(\mathbf{u},\mathbf{v};\mathcal{D}_{\mathtt{val}})$; // Update $\mathbf{u}$.
		%		%		\STATE Calculate $\nabla_{\mathbf{u}}F(\mathbf{u},\mathbf{v};\mathcal{D}_{\mathtt{val}})$ with Eq.~\eqref{eq: forth} and Eq.~\eqref{eq: fifth}.
		%		\STATE $\overline{\mathbf{v}} \leftarrow \tilde{\mathbf{v}} - \nabla_{\tilde{\mathbf{v}}}F({\tilde{\mathbf{v}}},\mathbf{v};\mathcal{D}_{\mathtt{val}})$; // Update $\tilde{\mathbf{v}}$.
		%		%		\STATE Calculate $\mathbf{G}_\mathbf{\tilde{\mathbf{v}}}$ with Eq.~\eqref{eq:seven}.
		%		\ENDWHILE
		%		\RETURN $\mathbf{u}^*$, $\tilde{\mathbf{v}}^*$.
		%	\end{algorithmic}
	%\end{algorithm}
	
	From the perspective of hyperparameter optimization, we hope to obtain a fast adaptation ability by defining a hyper-network that consists of hyperparameters in our designed architecture. To be specific, we know that the encoder extracts features from diverse inputs, and the decoder reconstructs features to output the desired targets. Actually, these features between encoder and decoder are compactly relevant to low-light inputs, that is to say, the encoder characterizes the representation of diverse inputs from different scenarios. The results reported in Fig.~\ref{fig: Motivation} can also verify the observation. Additionally, inspired by the work in~\citep{franceschi2018bilevel}, here we define the encoder as the scene-irrelevant hyper-network that learns the hyperparameter in Eq.~\eqref{eq: HO}. Then the decoder can be viewed as a scene-specific network, which learns the parameters in Eq.~\eqref{eq: HO}. This procedure presents that 
	\begin{equation}
		\mathbf{u}=\bm{\Theta}_{\mathcal{F}_{\mathtt{E}}},
		\mathbf{v}=\bm{\Theta}_{\mathcal{F}_{\mathtt{D}}}, 
	\end{equation}
	where $\mathcal{F}=\mathcal{F}_{\mathtt{E}}\cup\mathcal{F}_{\mathtt{D}}$ represents the Retinex-induced encoder-decoder\footnote{More details can be found in Sec.~\ref{sec: architecture}.} $\mathcal{F}$ consists of encoder $\mathcal{F}_{\mathtt{E}}$ and decoder $\mathcal{F}_{\mathtt{D}}$.
	As for the definition of $F$ and $f$, we adopt our defined loss for Retinex-induced encoder-decoder described in the above.

	In this way, we can successfully model our designed architecture by using the bilevel model based on hyperparameter optimization, to acquire the ability of fast adaptation. How to solve it become our next concentration.

	\subsection{Bilevel Learning Framework}
	As for solving Eq.~\eqref{eq: HO}, on one hand, it has been proved that the bilevel optimization model is extremely complex and difficult to solve~\citep{franceschi2018bilevel,liu2020generic}. Even though we can utilize the existing solving scheme~\citep{liu2020generic,liu2021value,liu2020investigating} for the bilevel model to solve our model. But it needs to consume a huge computational burden. This is why existing bilevel techniques just can be applied to some simple tasks, e.g., few-shot learning~\citep{simon2020adaptive, ziko2020laplacian}. 
	
	To solve Eq.~\eqref{eq: HO} efficiently, following the solving manner for neural architecture search as presented in~\citep{liu2018darts}, we adopt a simple approximation scheme to solve it, formulated as 
	\begin{equation}
		\nabla_{\mathbf{u}}F(\mathbf{u},\mathbf{v};\mathcal{D}_{\mathtt{val}})\approx\nabla_{\mathbf{u}}F(\mathbf{u},\mathbf{v}-\xi \nabla_{\mathbf{v}} f(\mathbf{u},\mathbf{v}; \mathcal{D}_\mathtt{tr});\mathcal{D}_{\mathtt{val}}), 
	\end{equation}
	where $\xi$ is a learning rate for a step of lower-level optimization. This way is to approximate the hyperparameter $\mathbf{u}$ using only a single training step, without solving lower-level problem accurately.  
	
	Furthermore, an inevitable phenomenon is that the second order gradient will appears after applying the chain rule, formulated as 
	\begin{equation}\label{eq: forth}
		\nabla_{\mathbf{u}}F(\mathbf{u},\mathbf{v}';\mathcal{D}_{\mathtt{val}})-\xi \nabla_{\mathbf{u},\mathbf{v}}^2  f(\mathbf{u},\mathbf{v}; \mathcal{D}_\mathtt{tr})\nabla_{\mathbf{v}'}F(\mathbf{u},\mathbf{v}';\mathcal{D}_{\mathtt{val}}), 
	\end{equation} 
	where $\mathbf{v}'=\mathbf{v}-\xi \nabla_{\mathbf{v}} f(\mathbf{u},\mathbf{v}; \mathcal{D}_\mathtt{tr})$ denotes the weights for a one-step forward model. In the above equation, there appears an expensive matrix-vector product which increases the complexity for this formulation. Fortunately, it can be substantially reduced by using the finite difference approximation, expressed as
	\begin{equation}\label{eq: fifth}
		\begin{aligned}
			&\;\;\;\;\nabla_{\mathbf{u},\mathbf{v}}^2  f(\mathbf{u},\mathbf{v}; \mathcal{D}_\mathtt{tr})\nabla_{\mathbf{v}'}F(\mathbf{u},\mathbf{v}';\mathcal{D}_{\mathtt{val}}),\\
			&\approx \frac{\nabla_{\mathbf{u}}f(\mathbf{u},\mathbf{v}^{+}; \mathcal{D}_\mathtt{tr})-\nabla_{\mathbf{u}}f(\mathbf{u},\mathbf{v}^{-}; \mathcal{D}_\mathtt{tr})}{2\epsilon},
		\end{aligned}
	\end{equation}
	where $\epsilon$ is a small constant, $\mathbf{v}^{\pm}=\mathbf{v}\pm\epsilon\nabla_{\mathbf{v}'}F(\mathbf{u},\mathbf{v}';\mathcal{D}_{\mathtt{val}})$. In this way, solving Eq.~\eqref{eq: forth} requires only two forward passes for the weights and two backward passes for $\mathbf{u}$. 
	
	In general, our established bilevel learning strategy contains three stages, i.e., learning scene-irrelevant encoder, fast adaptation to the scene that has never been encountered before, and testing process.

	\subsection{Reinforced Bilevel Learning Framework}
	Indeed, our above-built bilevel learning paradigm has satisfied the demand for fast adaptation by learning a scene-irrelevant encoder, but an inevitable issue is that the decoder needs to learn from scratch in the fast adaptation stage, leading to the adaptation difficulty. That is to say, can we provide a better, unified initialization for the decoder to adapt to various scenes better?
	
	Following the work in~\citep{liu2021towards,liu2021investigating}, here we introduce the meta-initialization process $\tilde{\mathbf{v}}= \bm{\Theta}^{0}_{\mathcal{F}_{\mathtt{D}}}$, to impose the meta-learning ability to decoder in the bilevel learning phase. To be specific, we define the following bilevel optimization 
	\begin{equation}\label{eq:VS}
		\begin{aligned}
			&\min_{\tilde{\mathbf{v}}\in\mathcal{V}}\;H(\tilde{\mathbf{v}},\mathbf{v};\mathcal{D}_{\mathtt{val}}),\\
			&{s.t.}\;\mathbf{v}\in\mathcal{R}(\tilde{\mathbf{v}}), \;\mathcal{R}(\tilde{\mathbf{v}}):=\arg\min_{\tilde{\mathbf{v}}}\;h(\tilde{\mathbf{v}},\mathbf{v};\mathcal{D}_{\mathtt{tr}}),\\
			%	\arg\min_{\mathbf{v}}\;f(\mathbf{u},\mathbf{v}; \mathcal{D}_\mathtt{tr}),
		\end{aligned}
	\end{equation}
	where $H(\cdot)$ and $h(\cdot)$ keep the same meaning with $F(\cdot)$ and $f(\cdot)$, respectively. $\mathcal{R}(\tilde{\mathbf{v}})$ is the set of solution of the lower-level problem. 
	Actually, the above-built model has the same form as Eq.~\eqref{eq: HO}, that is to say, we can use the same solving strategy for this model. It can be formulated as 
	\begin{equation}\label{eq:seven}
		\begin{aligned}
			&\nabla_{\tilde{\mathbf{v}}}H(\tilde{\mathbf{v}},\mathbf{v};\mathcal{D}_{\mathtt{val}})\approx\\
			&\nabla_{\tilde{\mathbf{v}}}H(\tilde{\mathbf{v}},\mathbf{v}';\mathcal{D}_{\mathtt{val}})
			-\frac{\nabla_{\tilde{\mathbf{v}}}h(\tilde{\mathbf{v}},\mathbf{v}^{+}; \mathcal{D}_\mathtt{tr})-\nabla_{\tilde{\mathbf{v}}}h(\tilde{\mathbf{v}},\mathbf{v}^{-}; \mathcal{D}_\mathtt{tr})}{2\epsilon},\\
		\end{aligned}
	\end{equation}
	where $\mathbf{v}'=\mathbf{v}-\xi \nabla_{\mathbf{v}} h(\tilde{\mathbf{v}},\mathbf{v}; \mathcal{D}_\mathtt{tr})$, $\mathbf{v}^{\pm}=\mathbf{v}\pm\epsilon\nabla_{\mathbf{v}'}H(\tilde{\mathbf{v}},\mathbf{v}';\mathcal{D}_{\mathtt{val}})$, $\epsilon$ is a small constant.

	\begin{figure}[t]
		\centering
		\begin{tabular}{c}
			\includegraphics[width=0.97\linewidth]{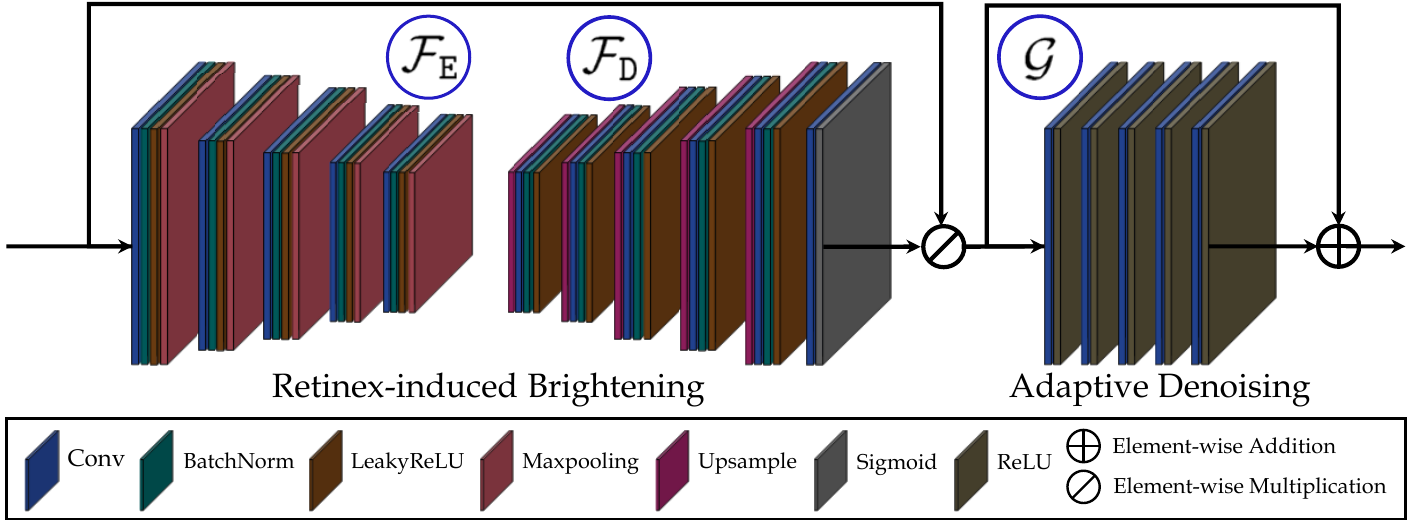}\\
		\end{tabular}
		%	\vspace{-0.4cm}
		\caption{The overall network architecture, which contains Retinex-induced brightening and adaptive denoising. The rectangle box below shows the operations used in our constructed architecture.}
		\label{fig: Architecture}
	\end{figure}
	
	\section{Network Architecture and Training Loss}\label{sec: architecture}
	In this section, we introduce a Retinex-induced brightening to satisfy the task demand according to the domain knowledge. Then we construct a new adaptive denoising mechanism to handle the noises. Finally, we introduce the training loss for these two parts.

	\subsection{Retinex-induced Brightening Architecture}
	Retinex theory~\citep{land1971lightness} is the well-known and commonly-approved model for low-light image enhancement. This theory says that a given low-light image can be decomposed as the illumination and reflectance. The formulation can be written as $\mathbf{y}=\mathbf{x}\odot\mathbf{z}$, where $\mathbf{y}$, $\mathbf{x}$, and $\mathbf{z}$ represent the low-light observation, illumination, and reflectance, respectively. $\odot$ is the element-wise multiplication. As described in existing works~\citep{guo2017lime,zhang2018high}, the illumination component is the key for addressing this task. Following the consensus, we establish a direct mapping between low-light and normal-light image as  
	\begin{equation}\label{eq: Retinexinduced}
		\left\{
		\begin{aligned}
			\mathbf{x}&= \mathcal{F}(\mathbf{y}; \bm{\Theta}_\mathcal{F}),\\
			\mathbf{z}&= \mathbf{y}\oslash\mathbf{x},\\
		\end{aligned}
		\right. 
	\end{equation}
	where $\mathcal{F}=\mathcal{F}_{\mathtt{E}}\cup\mathcal{F}_{\mathtt{D}}$ represents the Retinex-induced encoder-decoder~\citep{ronneberger2015u} consists of encoder $\mathcal{F}_{\mathtt{E}}$ and decoder $\mathcal{F}_{\mathtt{D}}$ with the parameters $\bm{\Theta}_\mathcal{F}$ to convert the low-light input to illumination. $\oslash$ denotes the element-wise division operation. In this way, we successfully perform domain knowledge in the network design to obtain the basic performance support.
	
	As for the architecture of encoder-decoder, the encoder consists of four down-sampling steps and a block (i.e. a convolutional layer, a batch normalization layer and a Leaky ReLU layer), and each step is composed of a block and a max pooling layer. Meanwhile, the decoder part consists of four upsampling steps, a block and a sigmoid activation function layer. Each upsampling step is made up of a block and an upsampling layer.

	\subsection{Adaptive Denoising Architecture}
	Indeed, we can directly perform Eq.~\eqref{eq: Retinexinduced} to realize the low-light image enhancement. However, this way implicitly contains an assumption, i.e., noises/artifacts do not exist in the reflectance. It will cause uncontrollable performance in some unknown and challenging low-light scenarios (maybe exist noises/artifacts). To handle these scenarios, we further construct a denoising mechanism which can be described as 
	\begin{equation}
		\hat{\mathbf{z}} = \mathbf{z}-\mathcal{G}(\mathbf{z}; \bm{\Theta}_\mathcal{G}),
	\end{equation}
	where $\hat{\mathbf{z}}$ represents the final output after denoising. $\mathcal{G}$ represents the noise estimation network with the parameters $\bm{\Theta}_\mathcal{G}$, whose architecture contains five convolutional layers, and each of the layer is followed by a ReLU activation function layer. An estimated noise map is generated by the module. Then the denoising effect can be achieved by subtracting the enhanced image from the noise map.

	\textit{Actually, as for training $\mathcal{G}$, we can directly adopt image pairs (low-light input contains noises) to train it. But if doing this, the model just addresses this case and loses the generalization. To this end, we propose to establish an adaptive denoising model by introducing a new adversarial loss for different types of data. The specific form of this loss function can be found in the upcoming section. }

	\begin{table}[t]
		\renewcommand\arraystretch{1.6}	
		\setlength{\tabcolsep}{1.5mm}
		\footnotesize
		\centering
		\caption{Benchmarks description.}
		\begin{tabular}{|c|cc|cc|c|}
			\hline
			\multirow{2}{*}{Benchmarks}&\multicolumn{2}{c|}{Seen-paired}&\multicolumn{2}{c|}{Unseen-paired}&Unseen-unpaired\\
			\cline{2-6}
			~&MIT&LOL&LSRW&VOC&DARKFACE\\
			\hline 
			Learning&$\surd$&$\surd$&$\times$&$\times$&$\times$\\
			\rowcolor[gray]{0.9}(Numbers)&(500)&(500)&(---)&(---)&(---)\\
			\hline
			Adaptation&$\surd$&$\surd$&$\surd$&$\surd$&$\surd$\\
			\rowcolor[gray]{0.9}(Numbers)&(500)&(500)&(500)&(500)&(500)\\
			\hline
			Testing&$\surd$&$\surd$&$\surd$&$\surd$&$\surd$\\
			\rowcolor[gray]{0.9}(Numbers)&(100)&(100)&(50)&(100)&(100)\\
			\hline
		\end{tabular}
		\label{table: benchmarks}
	\end{table}

	\subsection{Training Loss for Brightening Process}
	Here, we define two types of training loss for brightening from the data perspective. If the targeted datasets contain the reference images, we can utilize the supervised loss. Otherwise, we adopt the unsupervised loss to satisfy the case of no reference images. 
	
	\textbf{Supervised Loss.}
	We utilize the following MSE loss to train the brightening architecture so as to improve the enhancement capability.
	\begin{equation}\label{mse_loss}
		\mathcal{L}_{\mathtt{su}}=\|\mathbf{z}-\mathbf{z}_\mathtt{gt}\|^2,
	\end{equation}
	where $\mathbf{z}$ represents the reflection obtained by the encoder-decoder estimation of the input image, while $\mathbf{z}_\mathtt{gt}$ represents the ground truth image.

	\textbf{Unsupervised Loss.}
	The more common and challenging scenes are that the datasets only contain low-light observations, without the corresponding reference images. In this case, we adopt the unsupervised training loss following the work in~\citep{liu2021retinex}, represented as
	\begin{equation}\label{uns_loss}
		\mathcal{L}_{\mathtt{uns}}=\lambda\|\mathbf{x}-\mathbf{y}\|^2+\sum_{i=1}^{N}\sum_{j\in\mathcal{N}(i)}w_{i,j}|\mathbf{x}^{t}_{i}-\mathbf{x}^{t}_{j}|,
	\end{equation}
	where the first and second terms are fidelity and smoothing loss, respectively. The fidelity loss aims to guarantee the pixel-level consistency of the low-light input $\mathbf{y}$ and illumination $\mathbf{x}$. As for the smoothing loss, we adopt the spatially-variant $\ell_1$-norm~\citep{fan2018image} as the smoothness term in this paper, where $N$ and $i$ indicate the total number of pixels and the $i$-th pixel. $w_{i,j}$ represents the weight, whose formulated form is  $w_{i,j}=\exp\Big(-\frac{\sum_{c}((\mathbf{y}_{i,c}+\mathbf{s}_{i,c}^{t-1})-(\mathbf{y}_{j,c}+\mathbf{s}_{j,c}^{t-1}))^2}{2\sigma^2}\Big)$, where $c$ denotes image channel in the YUV color space. $\sigma=0.1$ is the standard deviations for the Gaussian kernels. The hyper-parameter $\lambda$ is empirically set to 0.2.
	
	%{\color{cyan}where the first term constrain the illumination by the low-light input. The second term represents a smoothness constraint with norm~\citep{fan2018image} for the illumination, where $N$ is the total number of pixels. $i$ is the $i$-th pixel. $\mathcal{N}(i)$ denotes the adjacent pixels of $i$ in its $5\times5$ window. $w_{i,j}$ represents the weight, whose formulated form is  $w_{i,j}=\exp\Big(-\frac{\sum_{c}((\mathbf{y}_{i,c}+\mathbf{s}_{i,c}^{t-1})-(\mathbf{y}_{j,c}+\mathbf{s}_{j,c}^{t-1}))^2}{2\sigma^2}\Big)$, where $c$ denotes image channel in the YUV color space. $\sigma=0.1$ is the standard deviations for the Gaussian kernels. }
	
	\begin{table*}[t]
		\renewcommand\arraystretch{1.5}	
		\setlength{\tabcolsep}{0.87mm}
		\footnotesize
		\centering
		\caption{Quantitative results (Metrics with reference: PSNR, SSIM, and LPIPS; Metrics without reference: DE, LOE, and NIQE) on MIT and LOL datasets. The best result is in bold red whereas the second best one is in bold blue.}
		~
		\begin{tabular}{|c|c||cccccccccc||cc|}
			\hline
			\multicolumn{2}{|c||}{Metrics}   & RetinexNet& DeepUPE & KinD & EnGAN & FIDE & DRBN & ZeroDCE &RUAS & UTVNet &SCL& BL & RBL\\
			\hline 
			\multirow{6}*{MIT} & PSNR$\uparrow$  & 12.9500 &18.3862 & 16.2148 & 15.3271 & 14.9522 & 15.2089 & 15.5432 &18.5549&16.3470&16.4040& \color{blue}{\textbf{20.1299}} &\color{red}{\textbf{20.6759}}\\
			\cline{2-14}
			& SSIM$\uparrow$  & 0.5996 & 0.7922& 0.7243 & 0.7247 & 0.6489 & 0.6684 & 0.7232 &0.7683&0.7351&0.7721& \color{blue}{\textbf{0.8413}}& \color{red}{\textbf{0.8352}}\\
			\cline{2-14}
			& LPIPS$\downarrow$  & 0.3654&0.1970& 0.2535 & 0.2376 & 0.3368 & 0.3153 & 0.2191&0.1721&0.2214&0.1881 &\color{blue}{\textbf{0.1799}}  & \color{red}{\textbf{0.1631}}\\  
			\cline{2-14}      
			&DE$\uparrow$ &6.1320&7.0559&6.6879&7.0327&6.7724&6.6012&6.2838&7.2471&6.6270&6.2947&\color{blue}{\textbf{7.2521}}& \color{red}{\textbf{7.3109}}\\       
			\cline{2-14}
			& LOE$\downarrow$ &1206.40&187.73&502.84&844.57 &552.99&678.45&475.23&341.22&272.64&563.90&\color{blue}{\textbf{183.42}}&  \color{red}{\textbf{171.09}}\\  	
			\cline{2-14}
			& NIQE$\downarrow$ &5.6853&4.1581&4.5200&5.0463 &5.5359&5.0958&4.2899&4.1312&4.4722&\color{red}{\textbf{4.0044}}&\color{blue}{\textbf{4.0318}}&  4.4554\\        
			\hline	\hline	
			\multirow{6}*{LOL} & PSNR$\uparrow$ & 14.2995& 18.1951 & 17.3540 & 15.3148 & 18.9835 & 19.3976& 18.4158 &16.0052&20.0077&15.4014&\color{blue}{\textbf{20.4271}}& \color{red}{\textbf{20.6905}}\\
			\cline{2-14}
			& SSIM$\uparrow$  & 0.4965 & 0.6617 & 0.7181 & 0.6163 & 0.7518& 0.7223 & 0.7245 &0.6803&\color{red}{\textbf{0.8565}}&0.6644 & 0.7331&\color{blue}{\textbf{0.7728}}\\
			\cline{2-14}
			& LPIPS$\downarrow$  & 0.5429& 0.3383 & 0.1518 & 0.3088 & 0.2141 & 0.2520 & 0.3126 &0.2384&0.2891&0.3004&\color{red}{\textbf{0.1305}} &  \color{blue}{\textbf{0.1443}}\\ 
			\cline{2-14}      
			&DE$\uparrow$ &6.9049&6.1240&\color{red}{\textbf{7.2118}}&6.9634 &6.5978&6.9074&6.6565&6.9763&6.7385&6.3470&6.6595&  \color{blue}{\textbf{7.0123}}\\       	
			\cline{2-14}
			& LOE$\downarrow$ &857.25&\color{blue}{\textbf{228.41}}&481.32&551.13 &1410.60&619.19&\color{red}{\textbf{204.56}}&326.82&346.78&239.48&314.77& 285.61\\      
			\cline{2-14}
			& NIQE$\downarrow$ &9.4275&7.5147&4.8798&5.0888 &4.9493&4.5934&5.0578&4.6483&5.4438&7.7964&\color{red}{\textbf{4.5289}}& \color{blue}{\textbf{4.5554}}\\
			\hline					
		\end{tabular}
		\label{table: MIT_LOL}
	\end{table*}

	\subsection{Adversarial Loss for Denoising Process}
	During the training process, in order to obtain the capability of adaptive denoising, we utilize a dataset with noisy images (e.g., LOL) and a dataset without noise (e.g., MIT) to train the network. We aim to obtain a denosing network with the ability to automatically identify noise. The denoising module distinguishes images that generated by the previous networks. Nevertheless, the previous encoder and decoder are trained noise-insensitive (not care if there is noise on the input image), which significantly increase the difficulty of identifying noise for the denoising module. Therefore, we need a reliable training loss to guide the denoising module.
	
	Following the manner presented in~\citep{sindagi2020prior}, we define a new adversarial loss to learn an adaptive denoising module, which is able to distinguish image with/without noises. This loss can be formulated as 
	\begin{equation}\label{adv_loss}
		\mathcal{L}_{adv}^{l}=\|\hat{\mathbf{z}}^{l}-\mathbf{z}_\mathtt{gt}^{l}\|^2, l = a,b,
	\end{equation}
	where $\left\{\hat{\mathbf{z}}=\hat{\mathbf{z}}^{a}\cup\hat{\mathbf{z}}^{b},\mathbf{z}_\mathtt{gt}=\mathbf{z}_\mathtt{gt}^{a}\cup\mathbf{z}_\mathtt{gt}^{b}\right\}$ represents that the training pairs $\left\{\mathbf{z},\mathbf{z}_\mathtt{gt}\right\}$ can be split to two groups, i.e., low-light image pairs with noises $\left\{\mathbf{z}^{a},\mathbf{z}_\mathtt{gt}^{a}\right\}$ and low-light image pairs without noises $\left\{\mathbf{z}^{b},\mathbf{z}_\mathtt{gt}^{b}\right\}$. 
	
	\begin{figure*}[!htb]
		\centering
		\begin{tabular}{c@{\extracolsep{0.3em}}c@{\extracolsep{0.3em}}c@{\extracolsep{0.3em}}c@{\extracolsep{0.3em}}c@{\extracolsep{0.3em}}c}
			\includegraphics[width=0.156\linewidth]{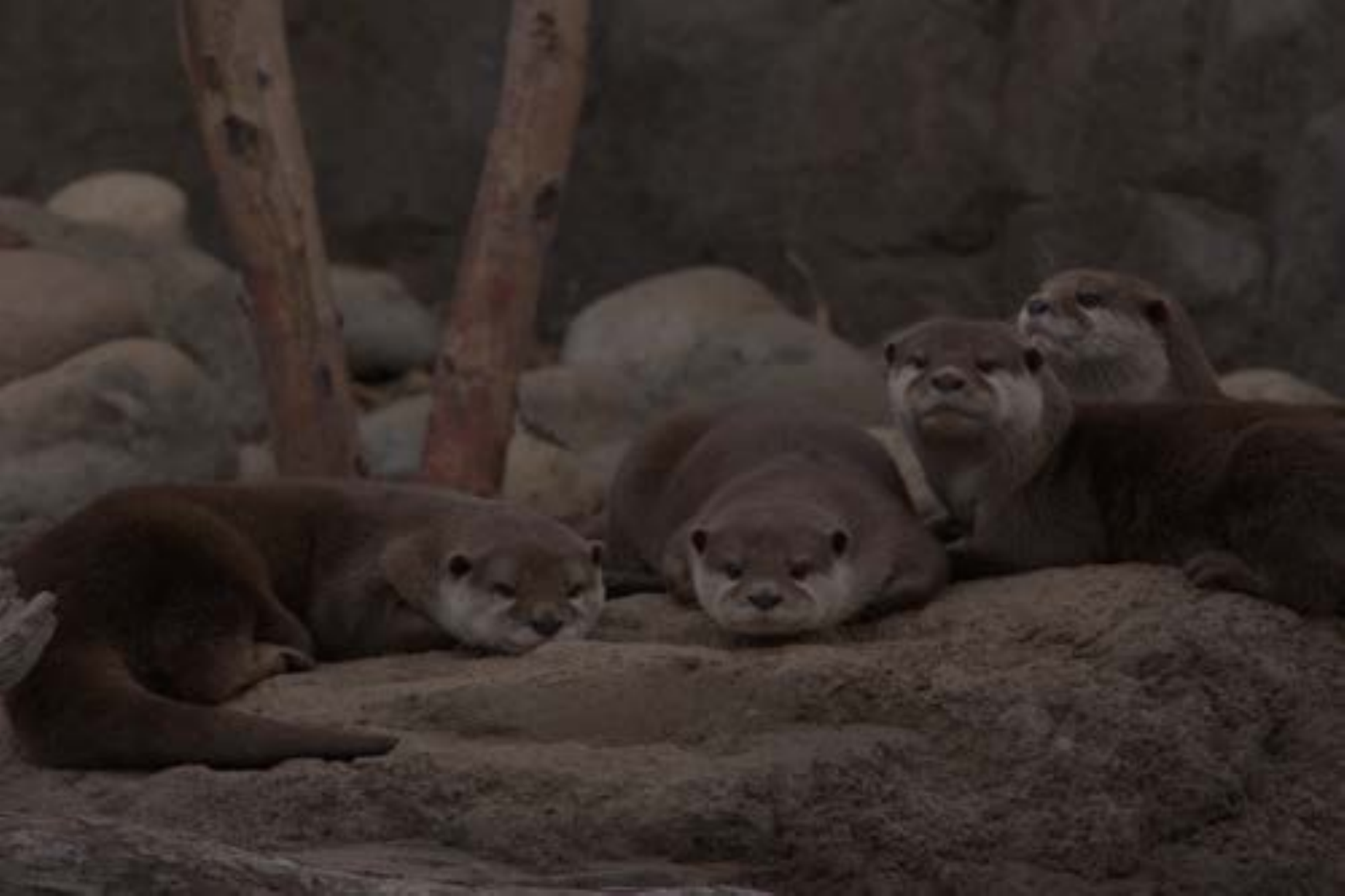}&
			\includegraphics[width=0.156\linewidth]{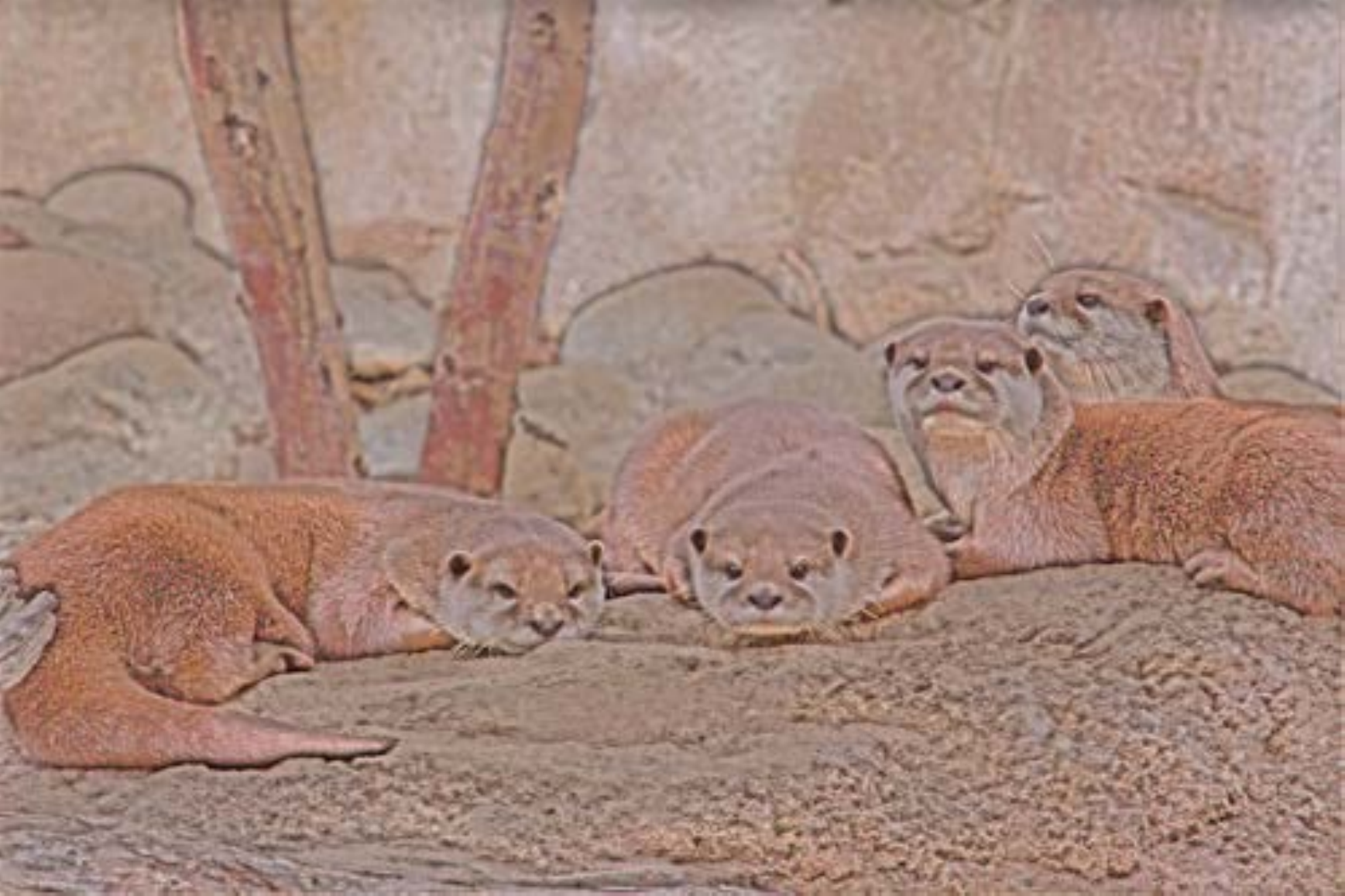}&
			\includegraphics[width=0.156\linewidth]{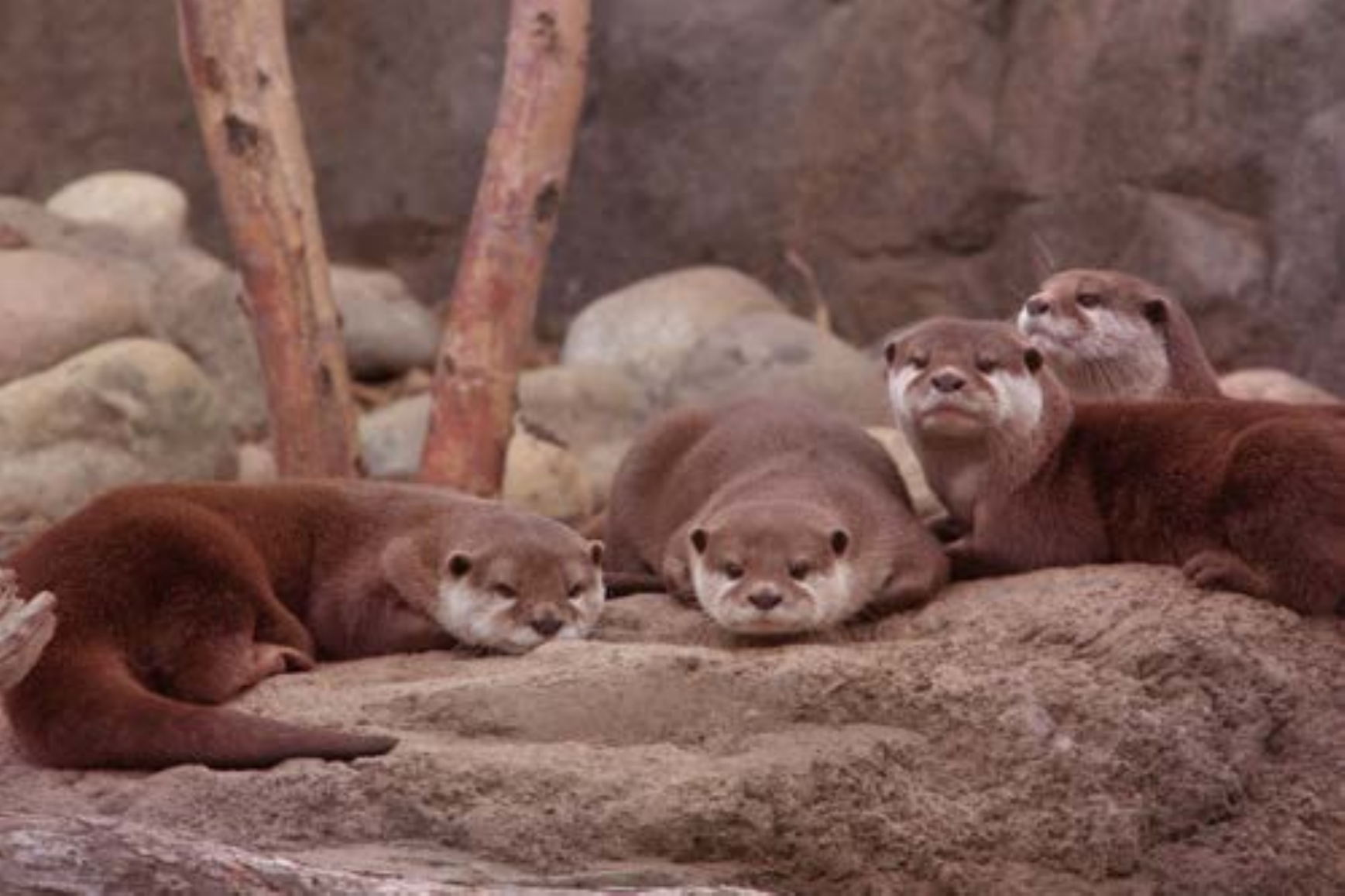}&
			\includegraphics[width=0.156\linewidth]{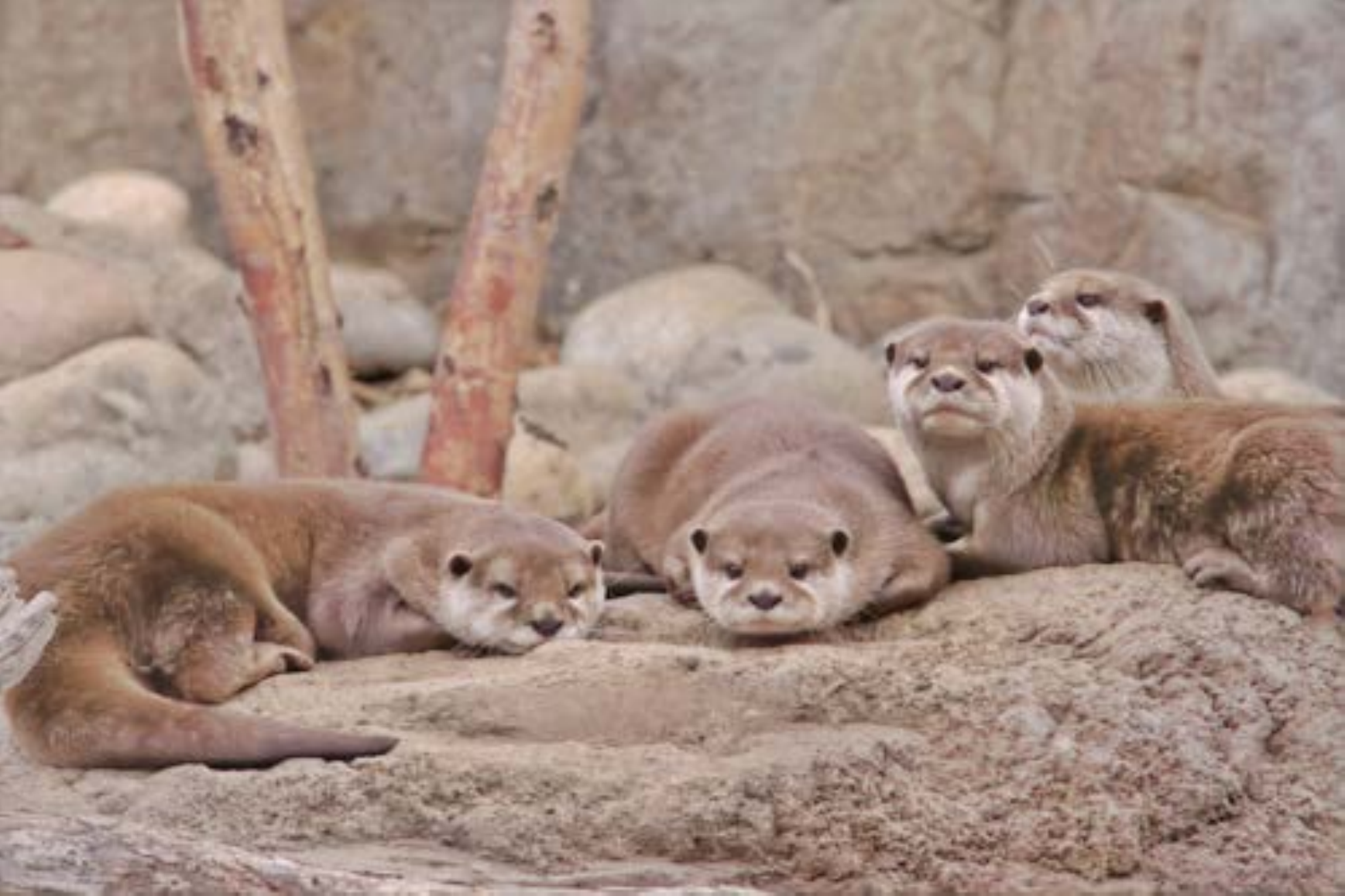}&
			\includegraphics[width=0.156\linewidth]{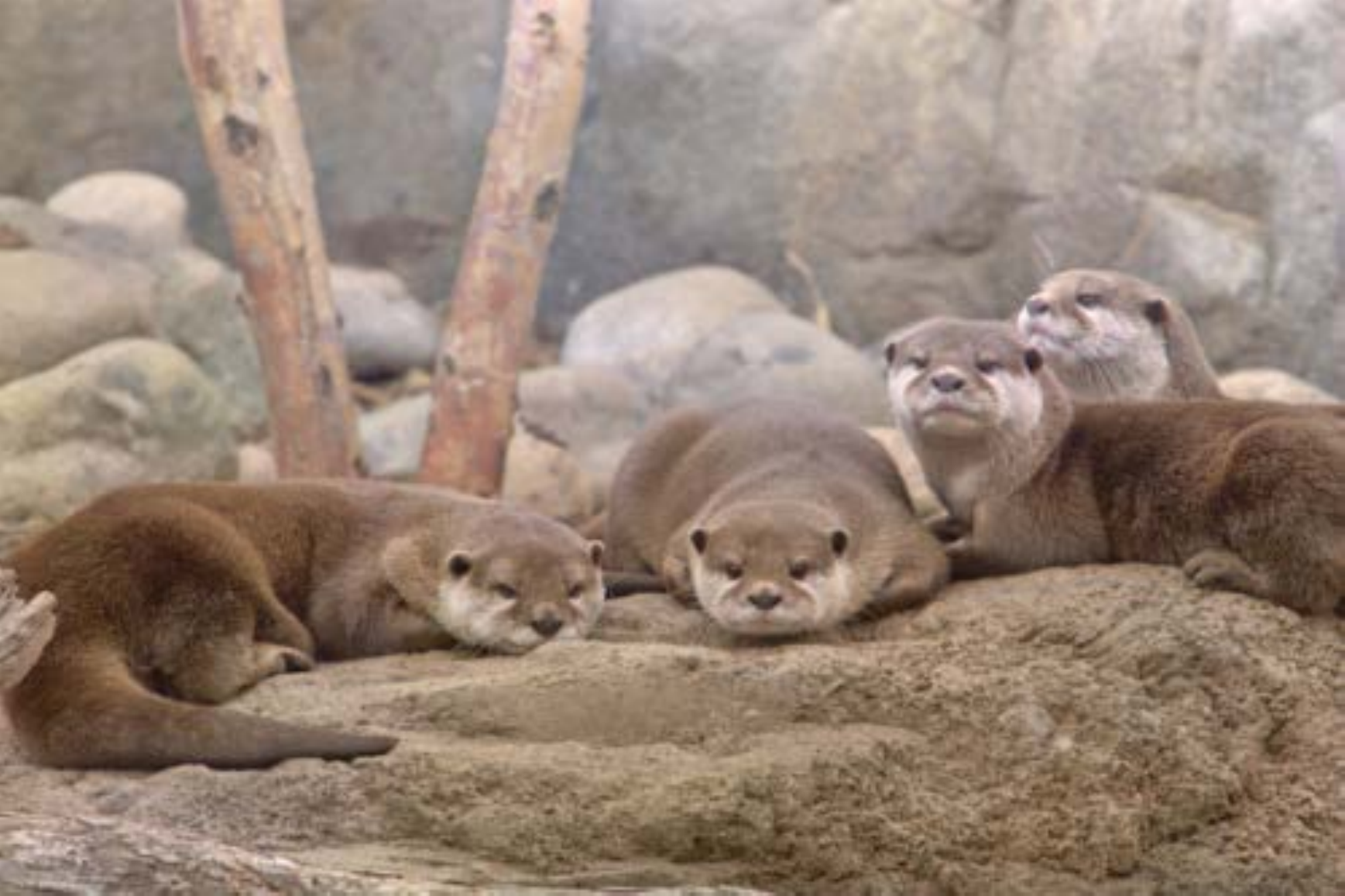}&
			\includegraphics[width=0.156\linewidth]{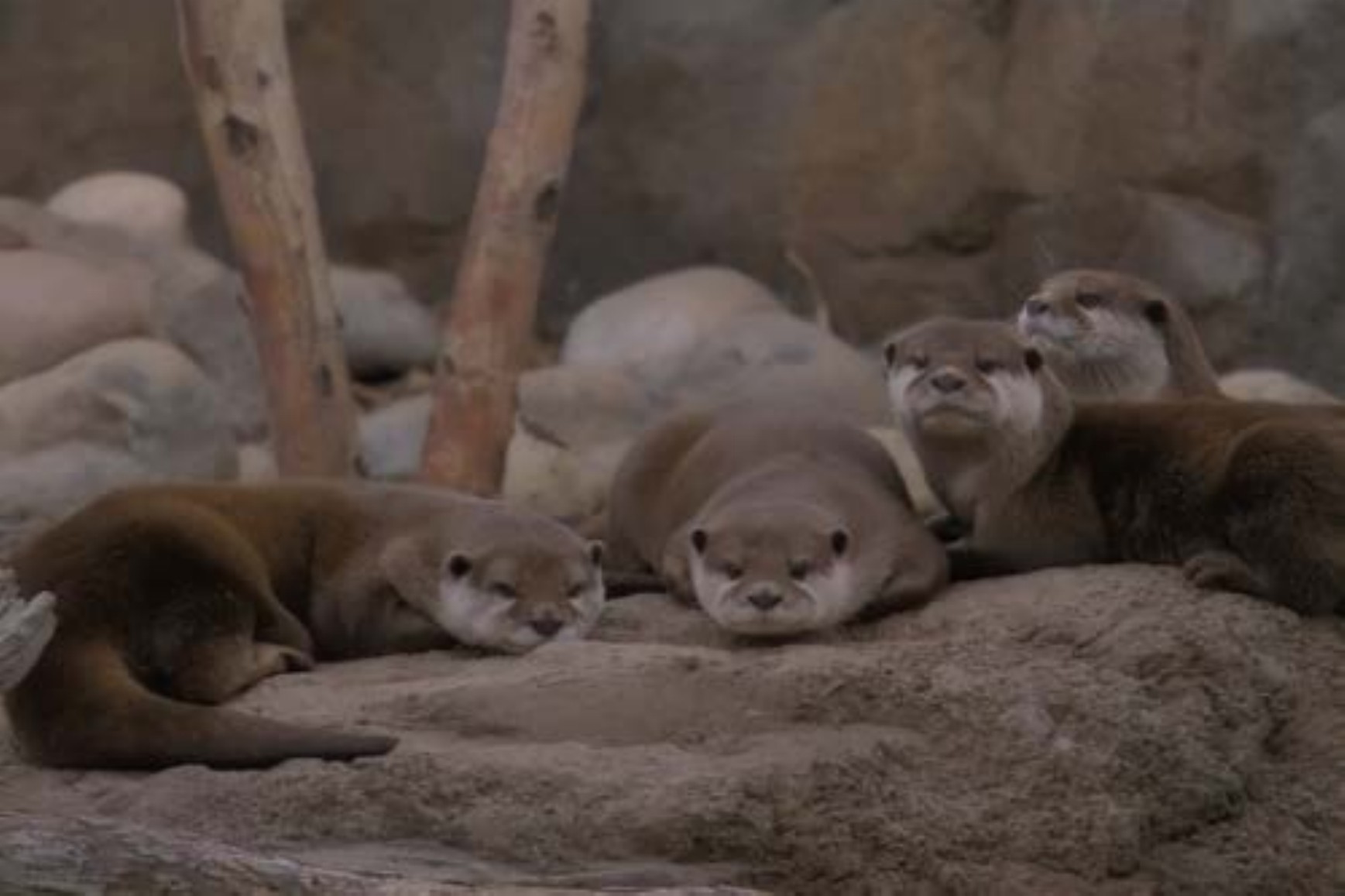}\\
			\footnotesize Input&\footnotesize RetinexNet&\footnotesize DeepUPE&\footnotesize KinD&\footnotesize EnGAN&\footnotesize FIDE\\			
			\includegraphics[width=0.156\linewidth]{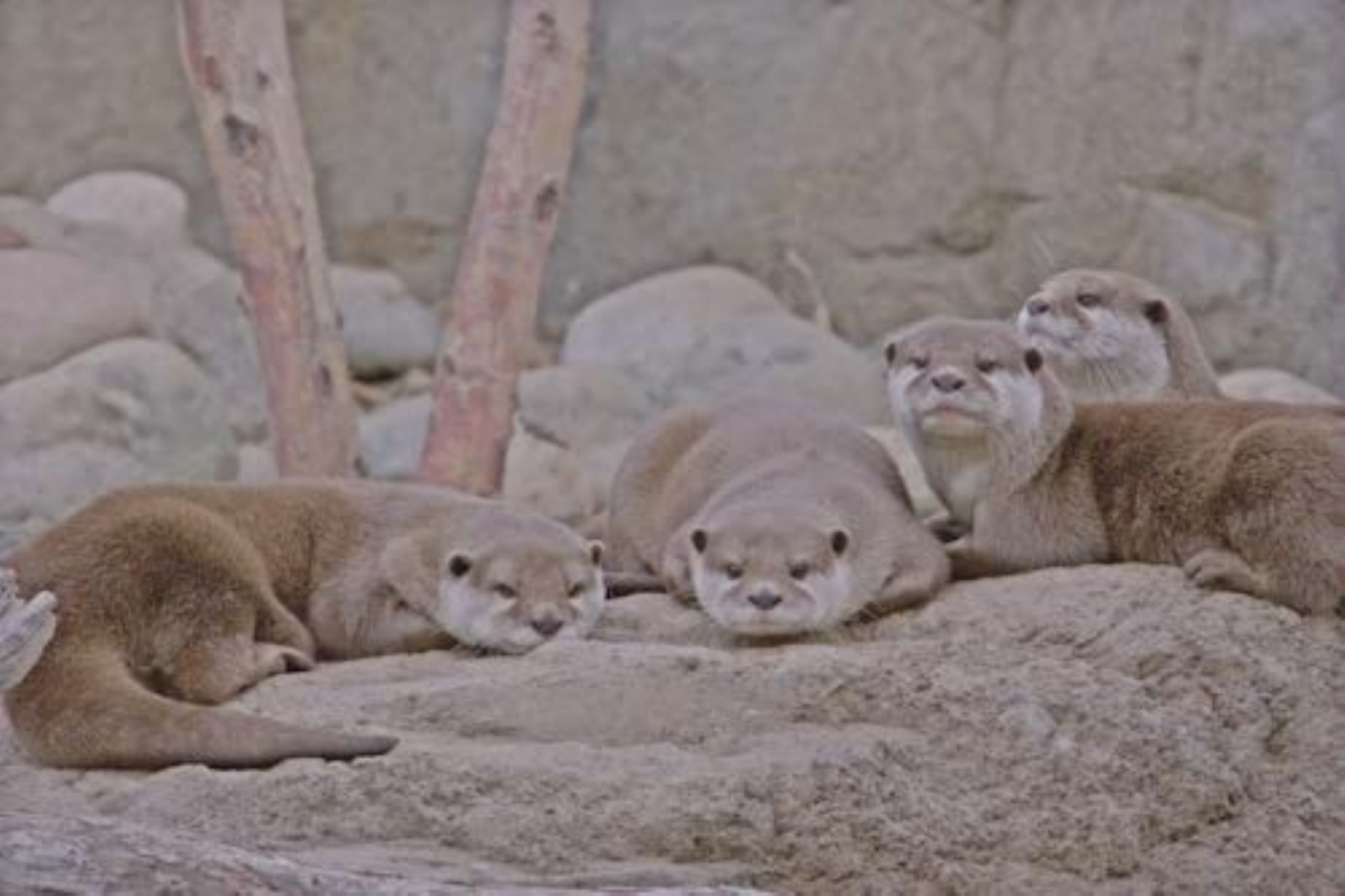}&
			\includegraphics[width=0.156\linewidth]{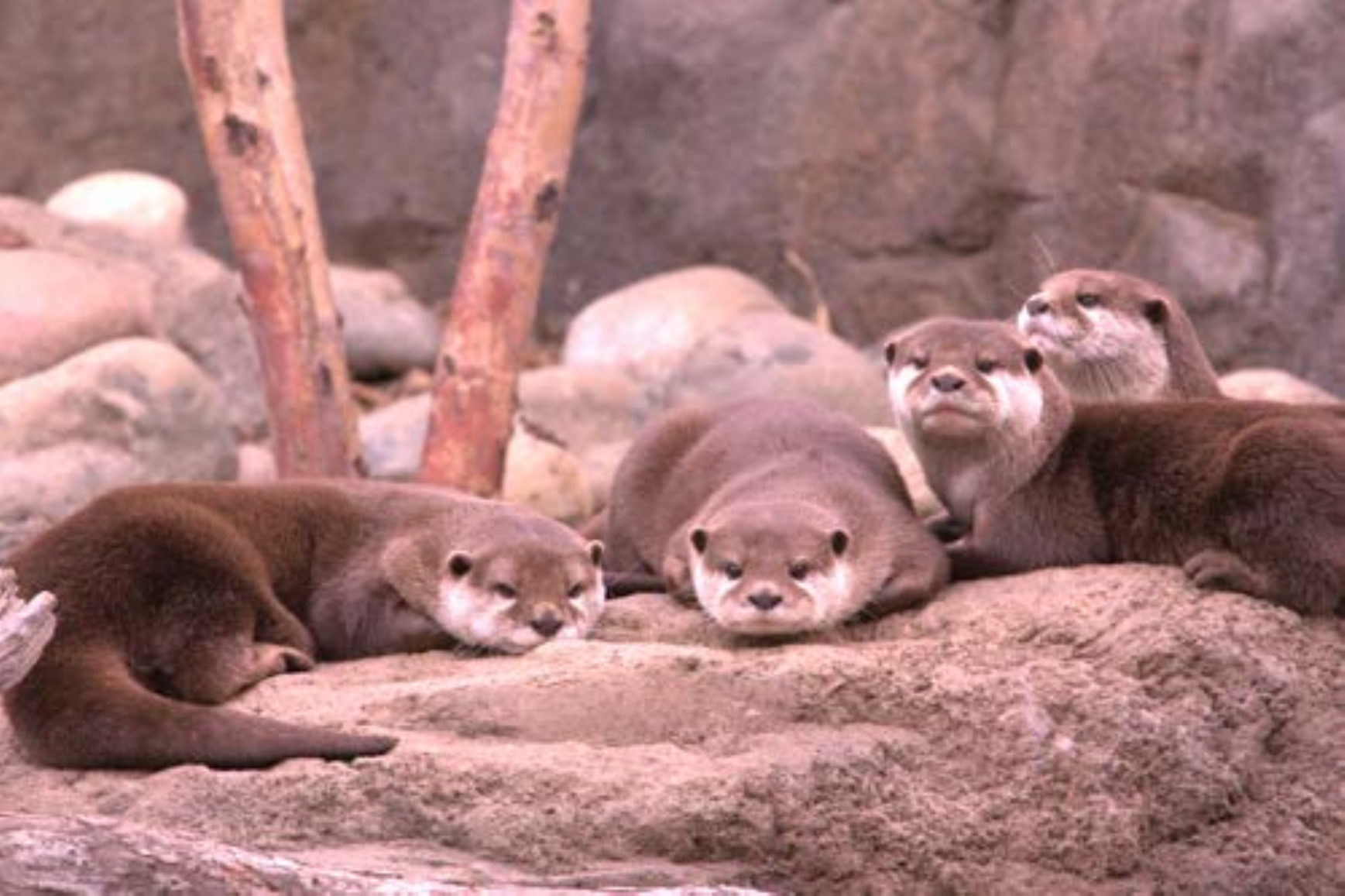}&
			\includegraphics[width=0.156\linewidth]{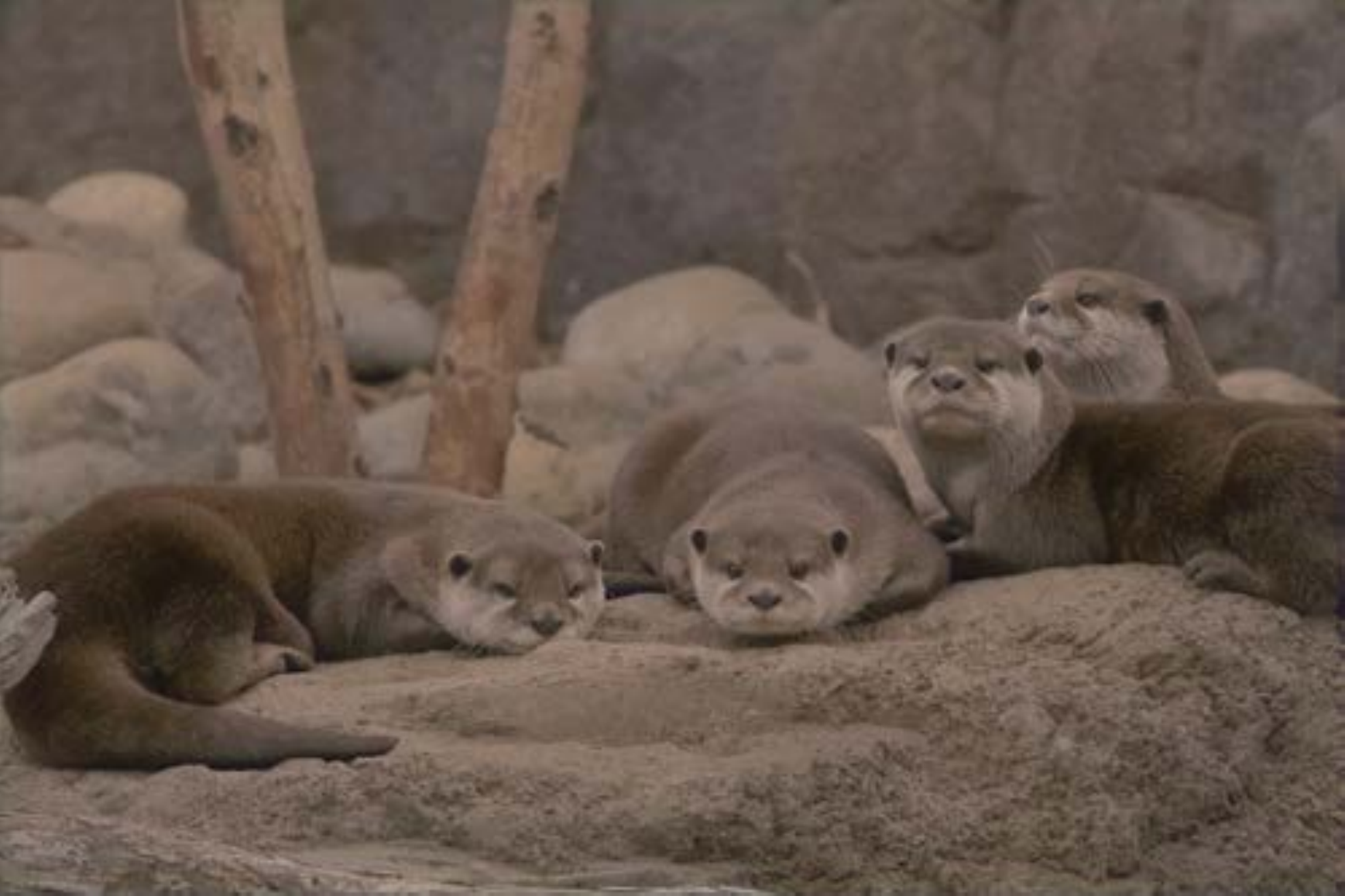}&
			\includegraphics[width=0.156\linewidth]{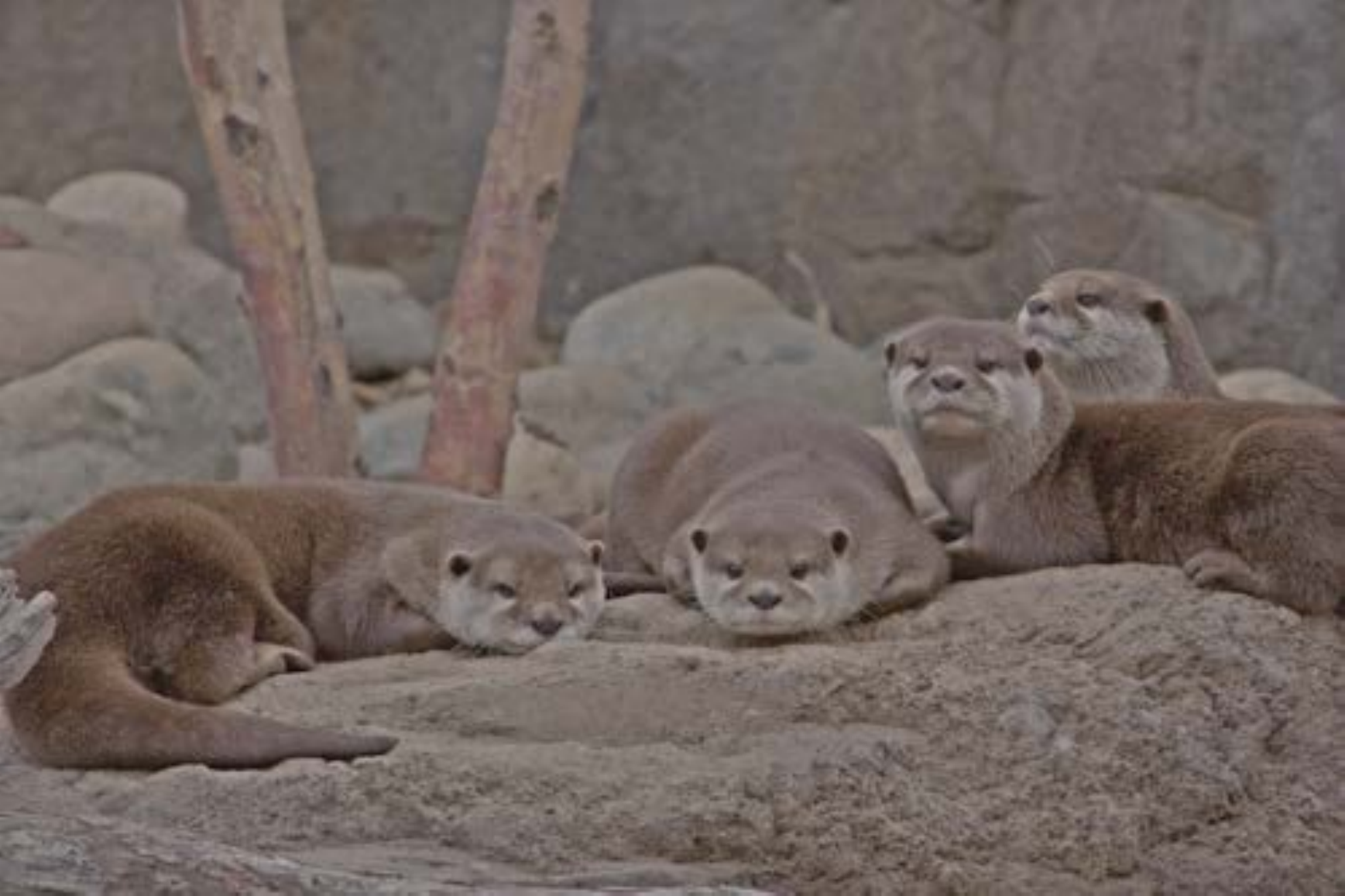}&
			\includegraphics[width=0.156\linewidth]{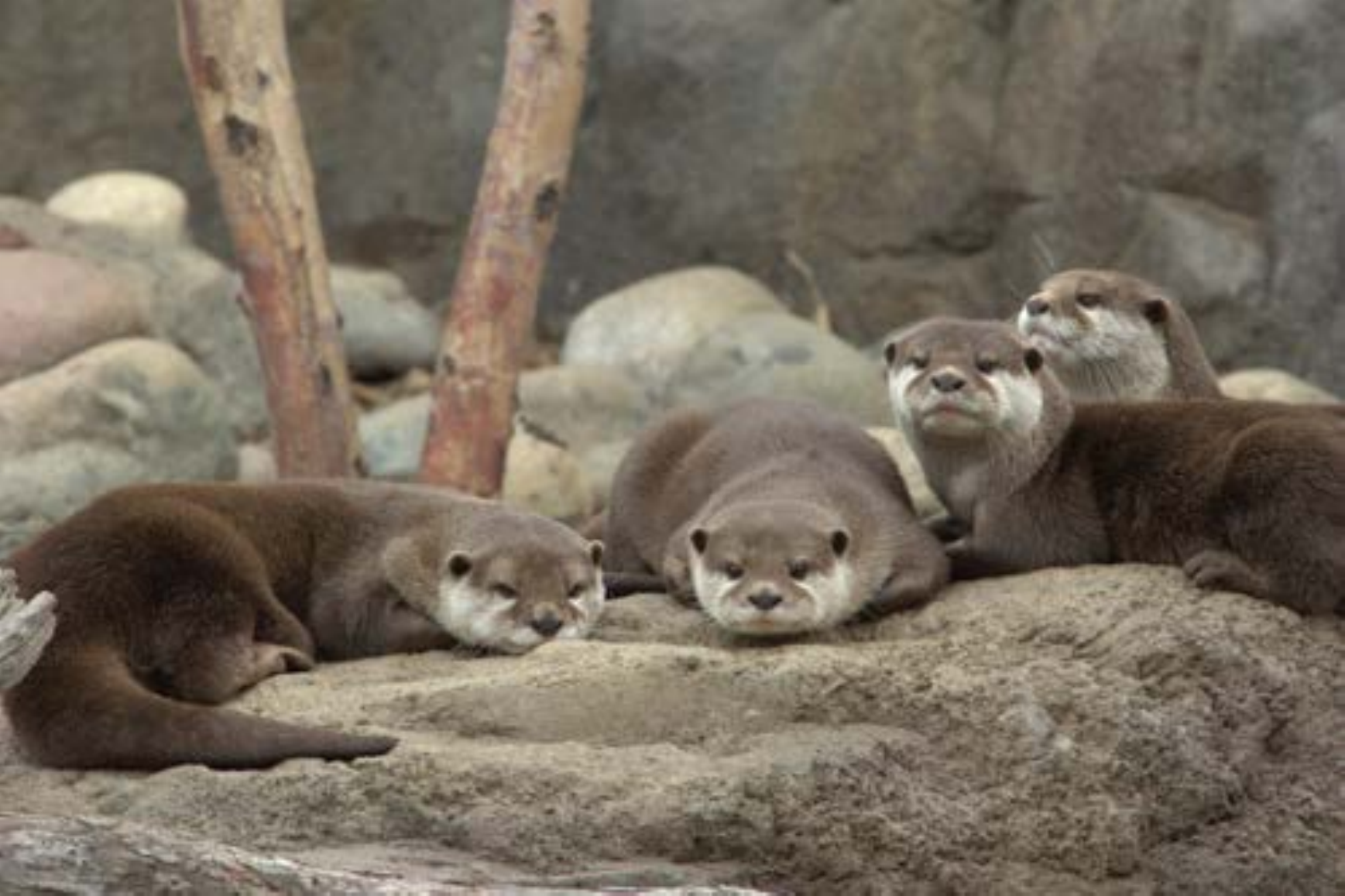}&
			\includegraphics[width=0.156\linewidth]{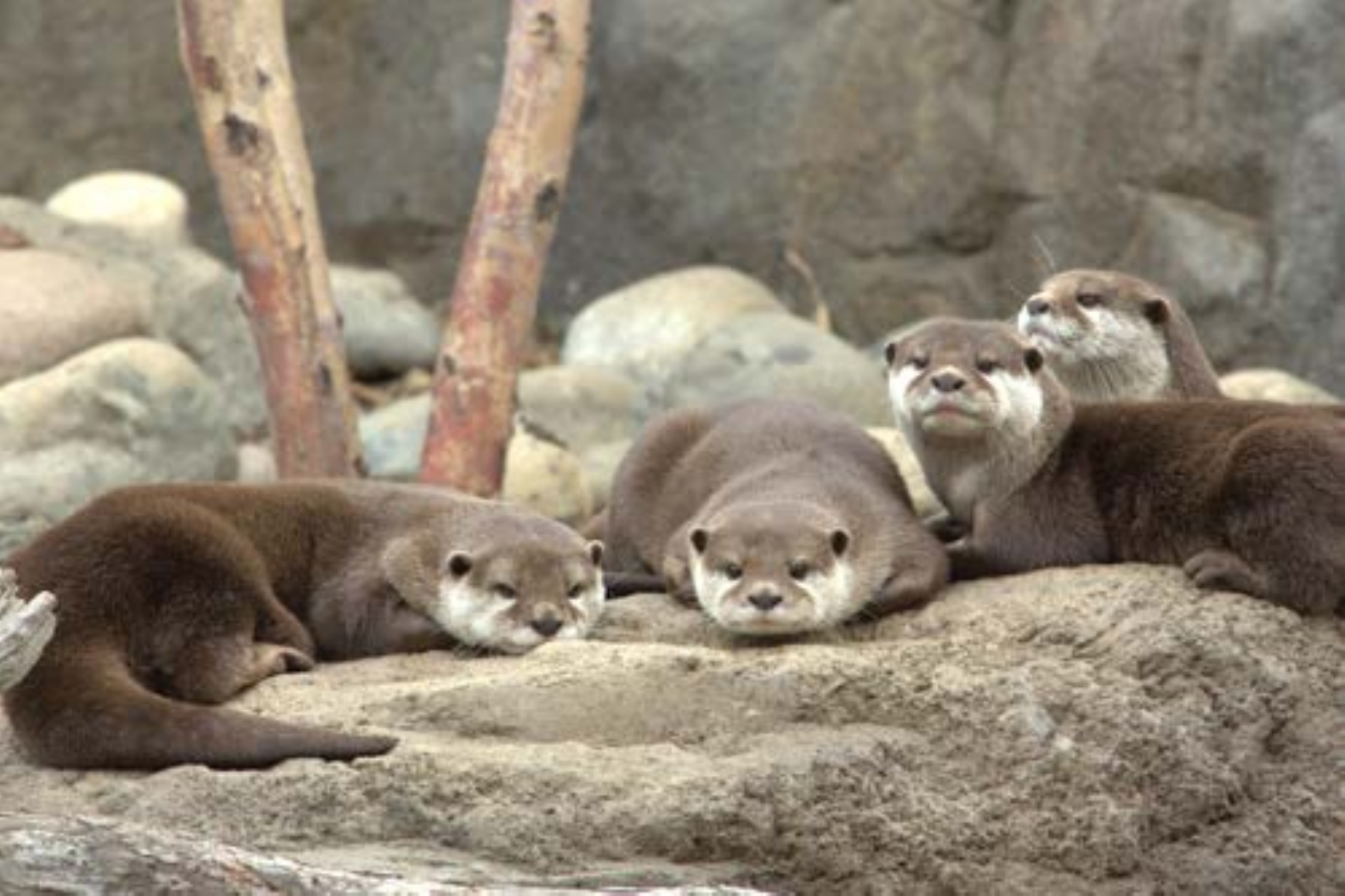}\\
			\footnotesize ZeroDCE&\footnotesize RUAS&\footnotesize UTVNet&\footnotesize SCL&\footnotesize \textbf{BL}&\footnotesize \textbf{RBL}\\
			\multicolumn{6}{c}{\footnotesize (a) Visual comparison among different methods on the MIT dataset~\citep{MIT_Adobe_5K}}\\
			\includegraphics[width=0.156\linewidth]{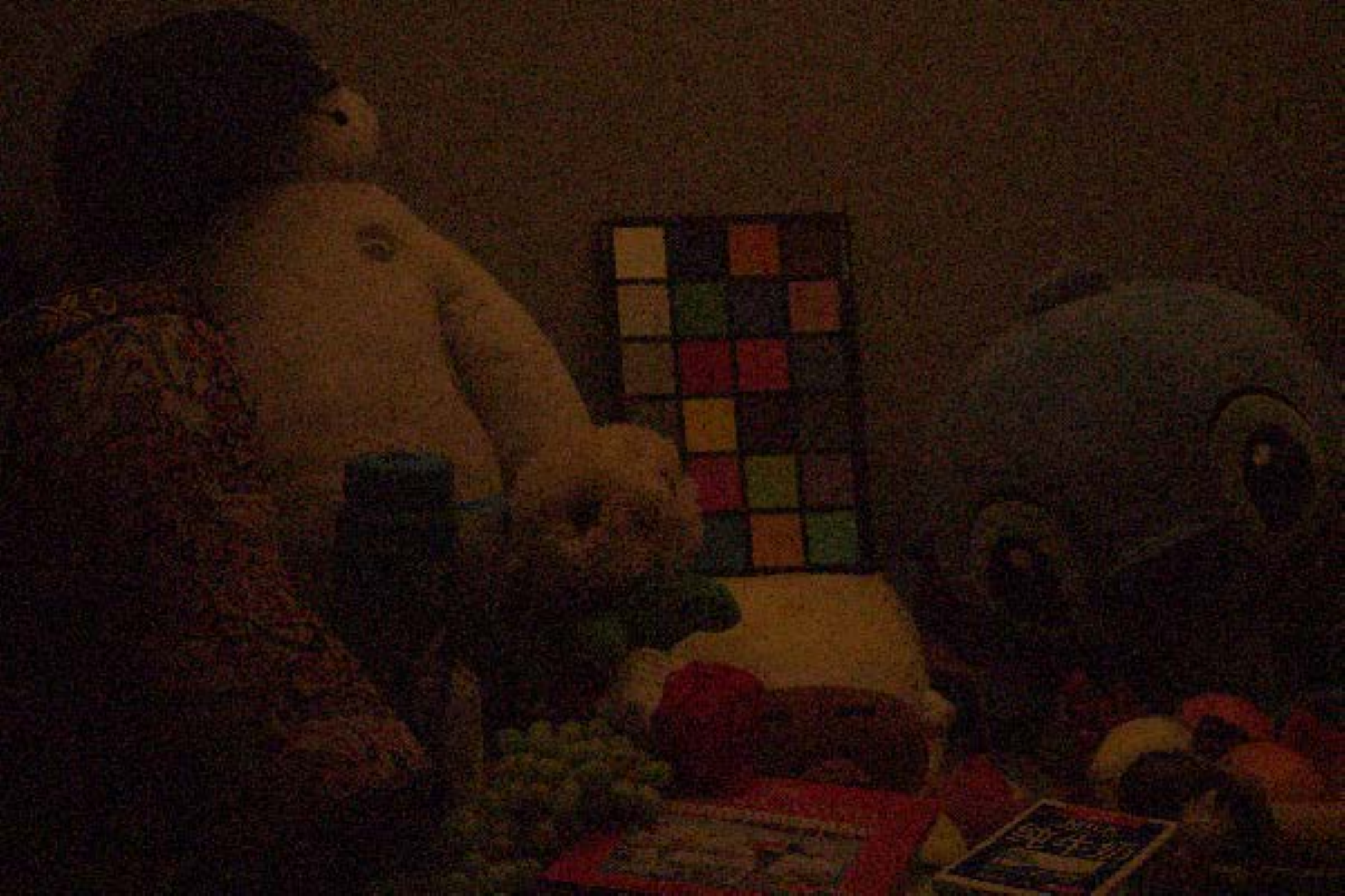}&
			\includegraphics[width=0.156\linewidth]{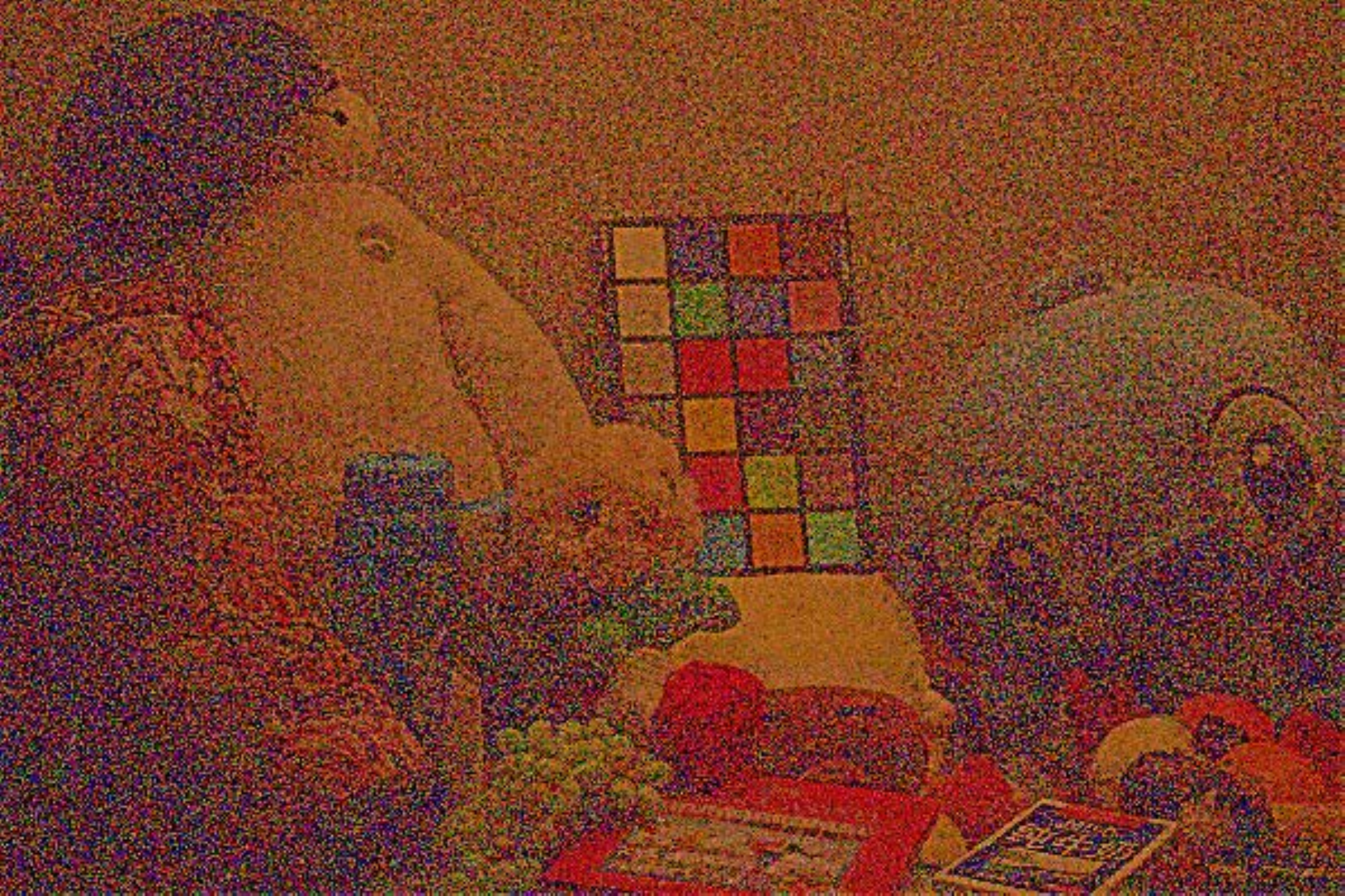}&
			\includegraphics[width=0.156\linewidth]{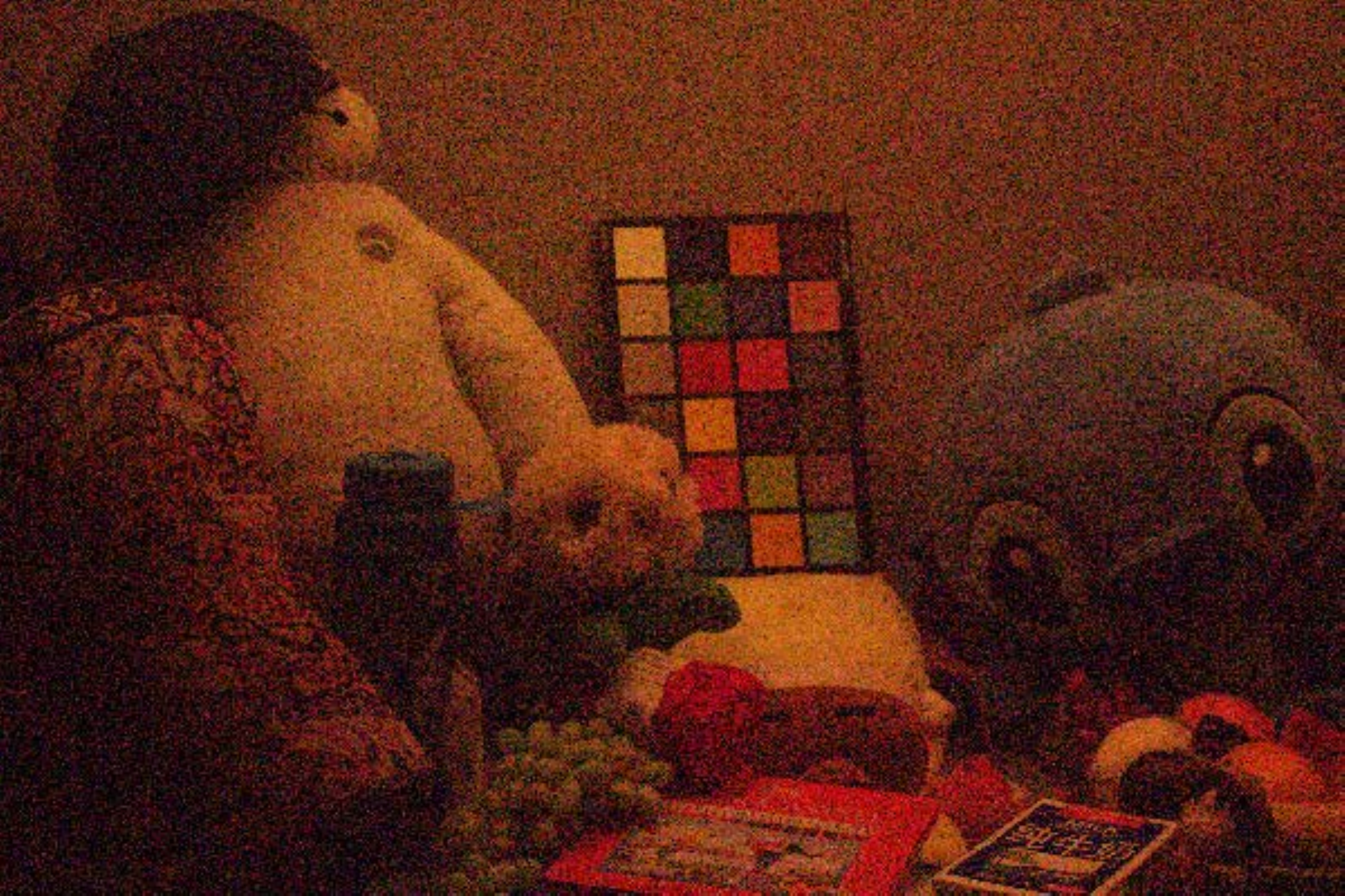}&
			\includegraphics[width=0.156\linewidth]{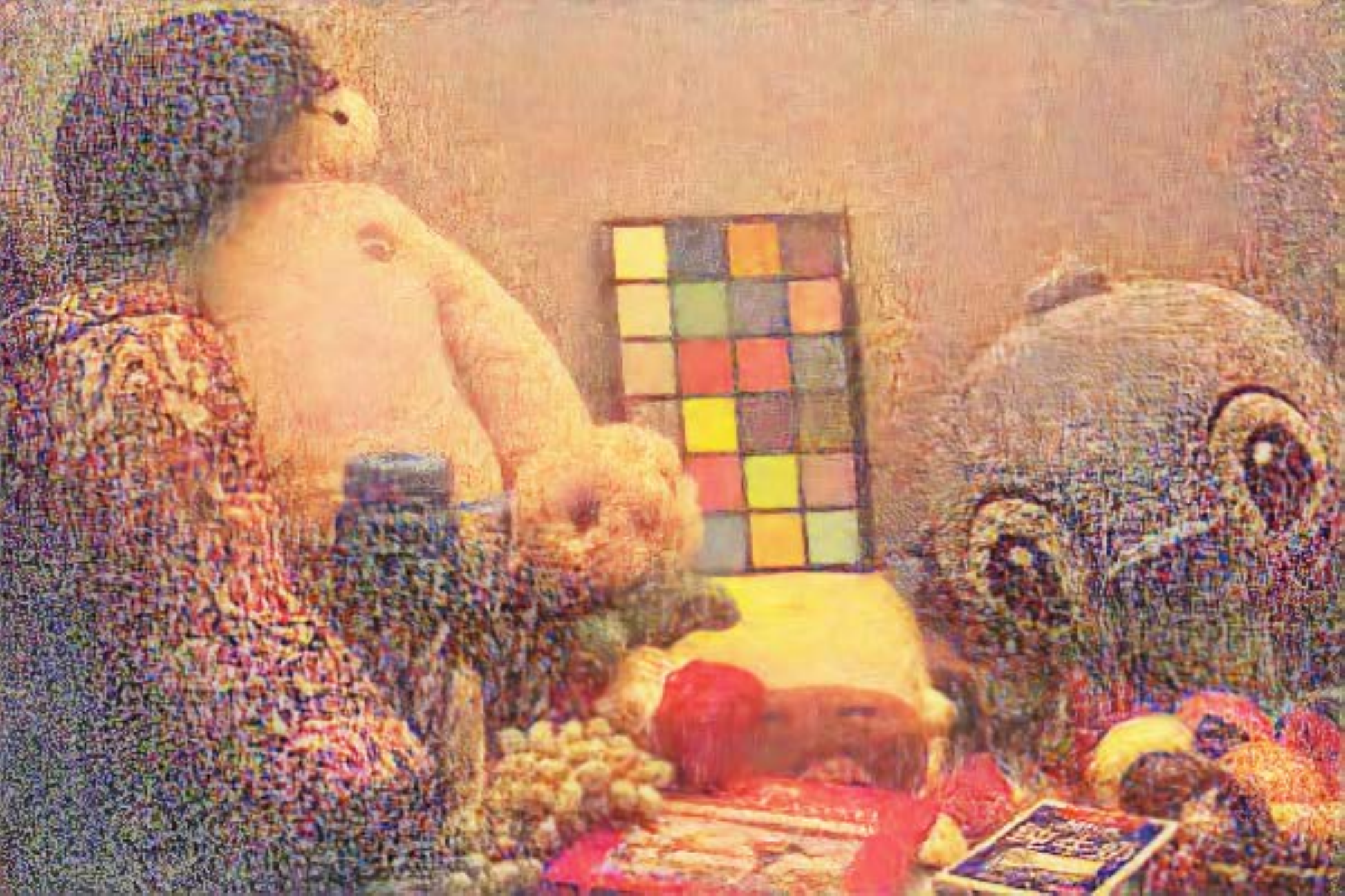}&
			\includegraphics[width=0.156\linewidth]{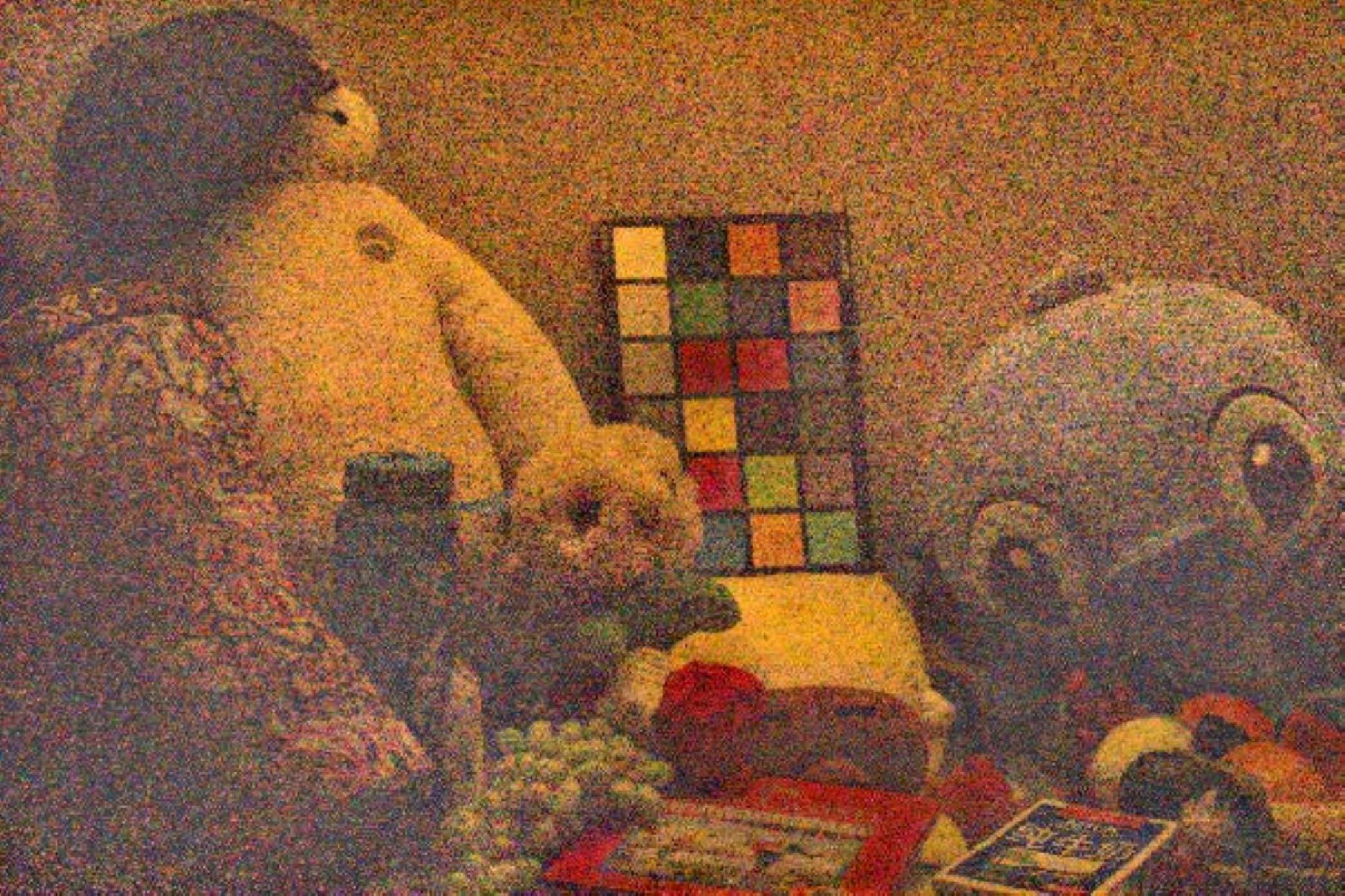}&
			\includegraphics[width=0.156\linewidth]{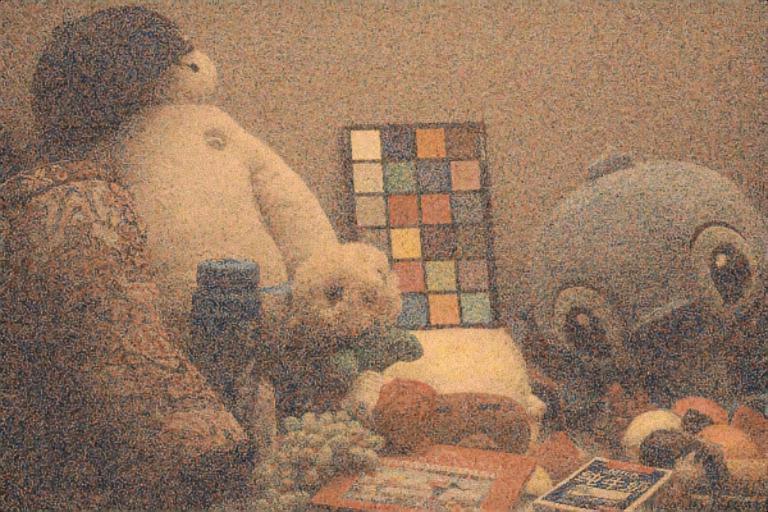}\\
			\footnotesize Input&\footnotesize RetinexNet&\footnotesize DeepUPE&\footnotesize KinD&\footnotesize EnGAN&\footnotesize FIDE\\	\includegraphics[width=0.156\linewidth]{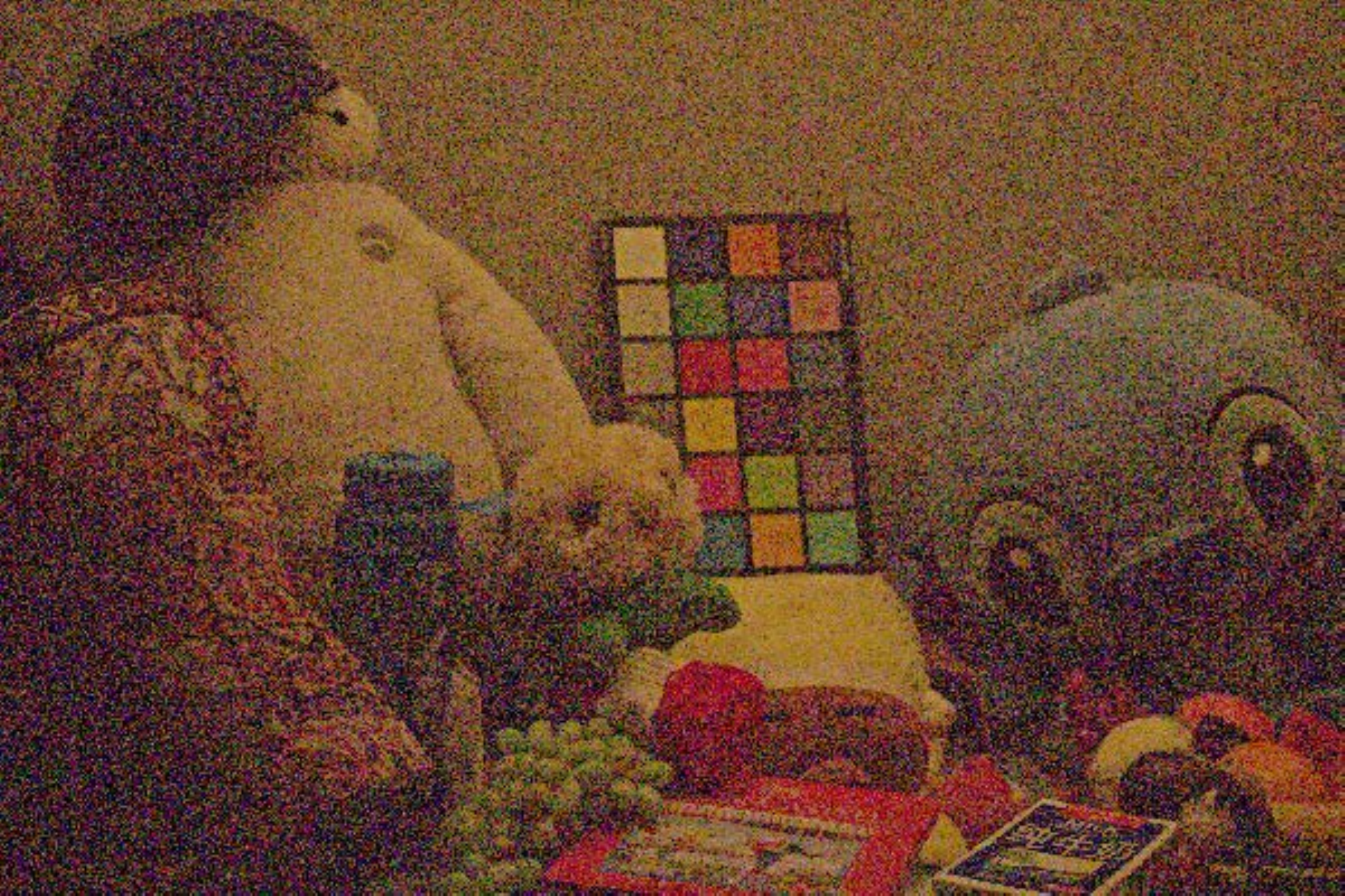}&
			\includegraphics[width=0.156\linewidth]{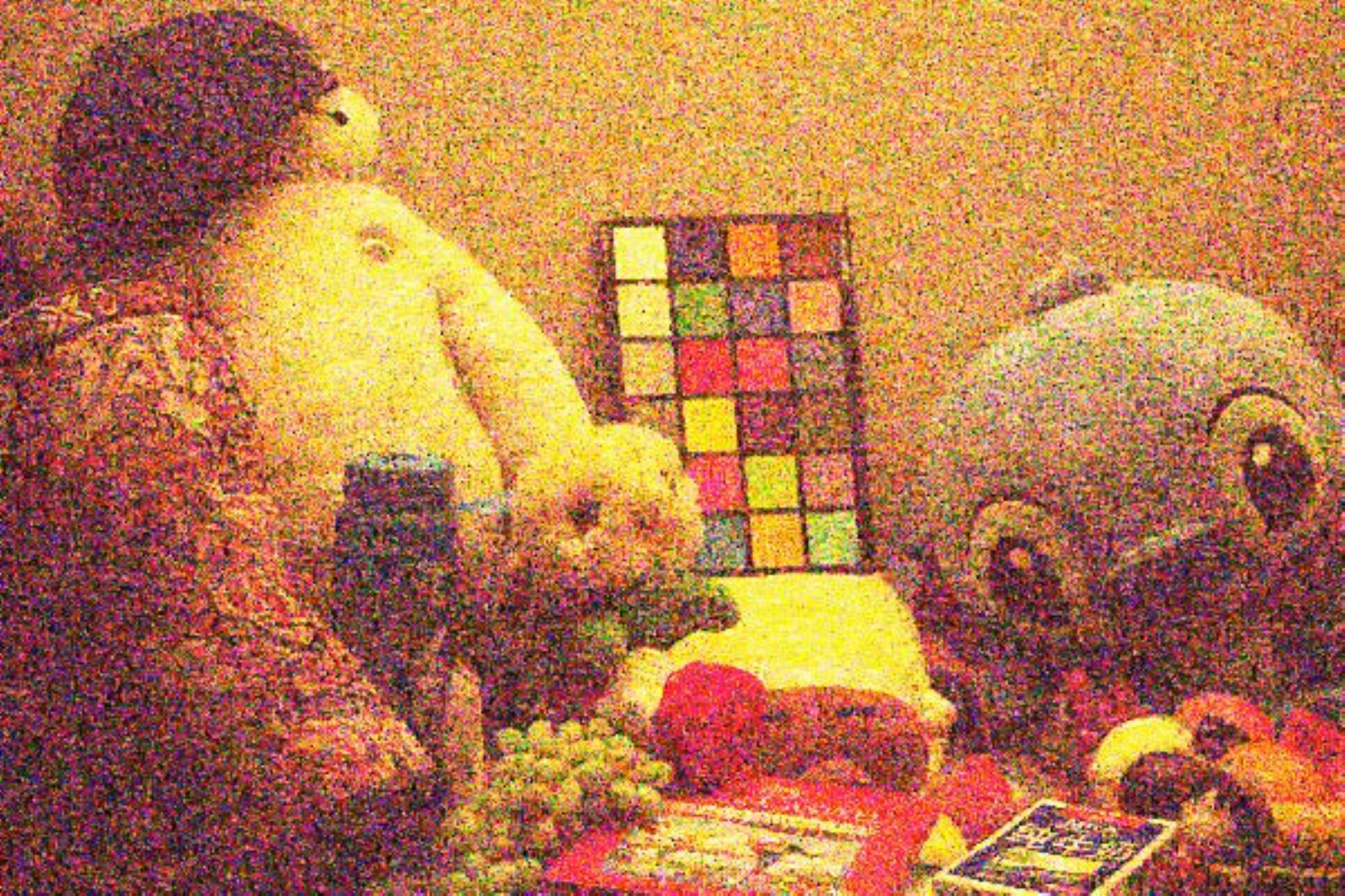}&
			\includegraphics[width=0.156\linewidth]{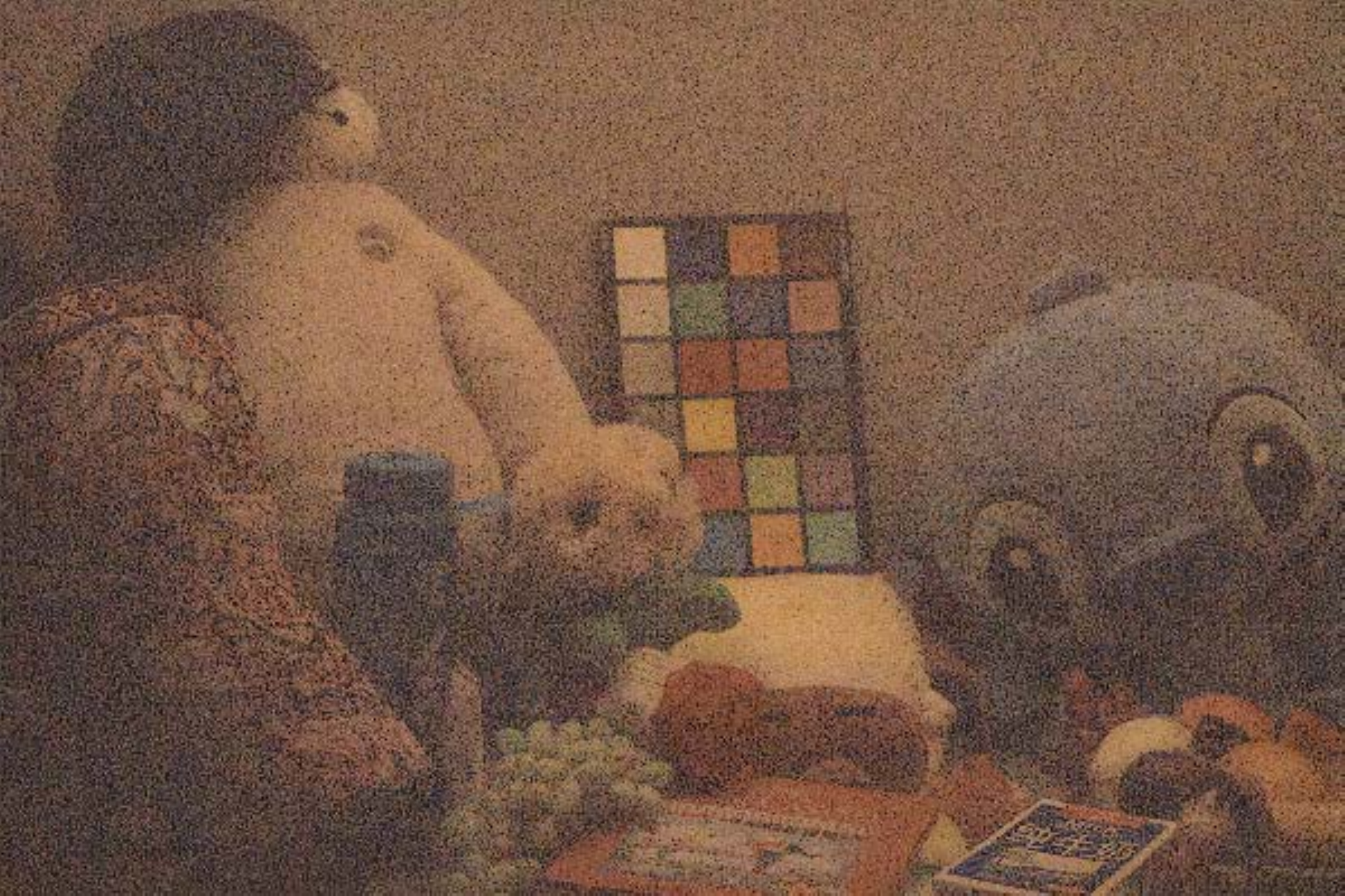}&
			\includegraphics[width=0.156\linewidth]{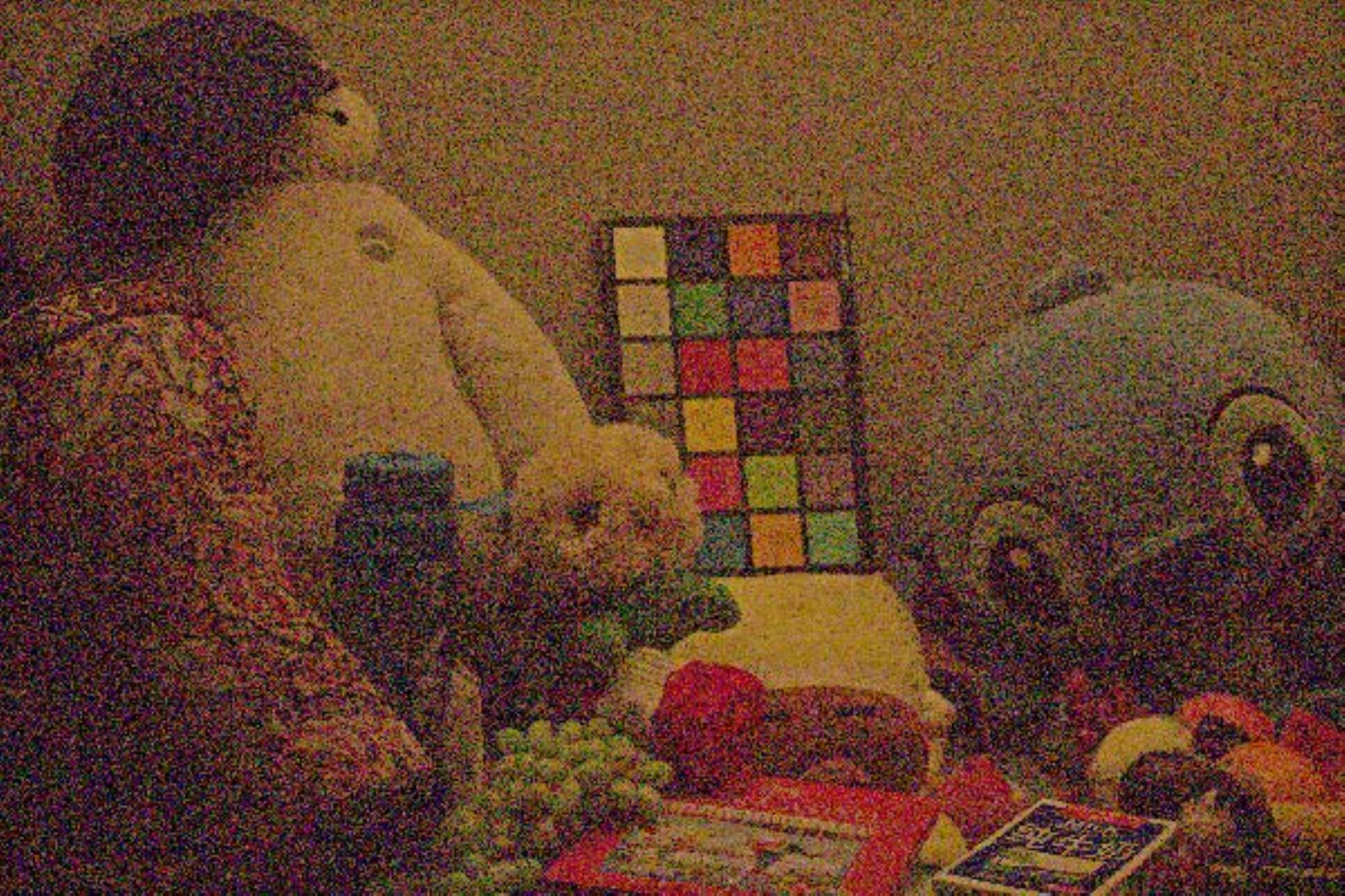}&
			\includegraphics[width=0.156\linewidth]{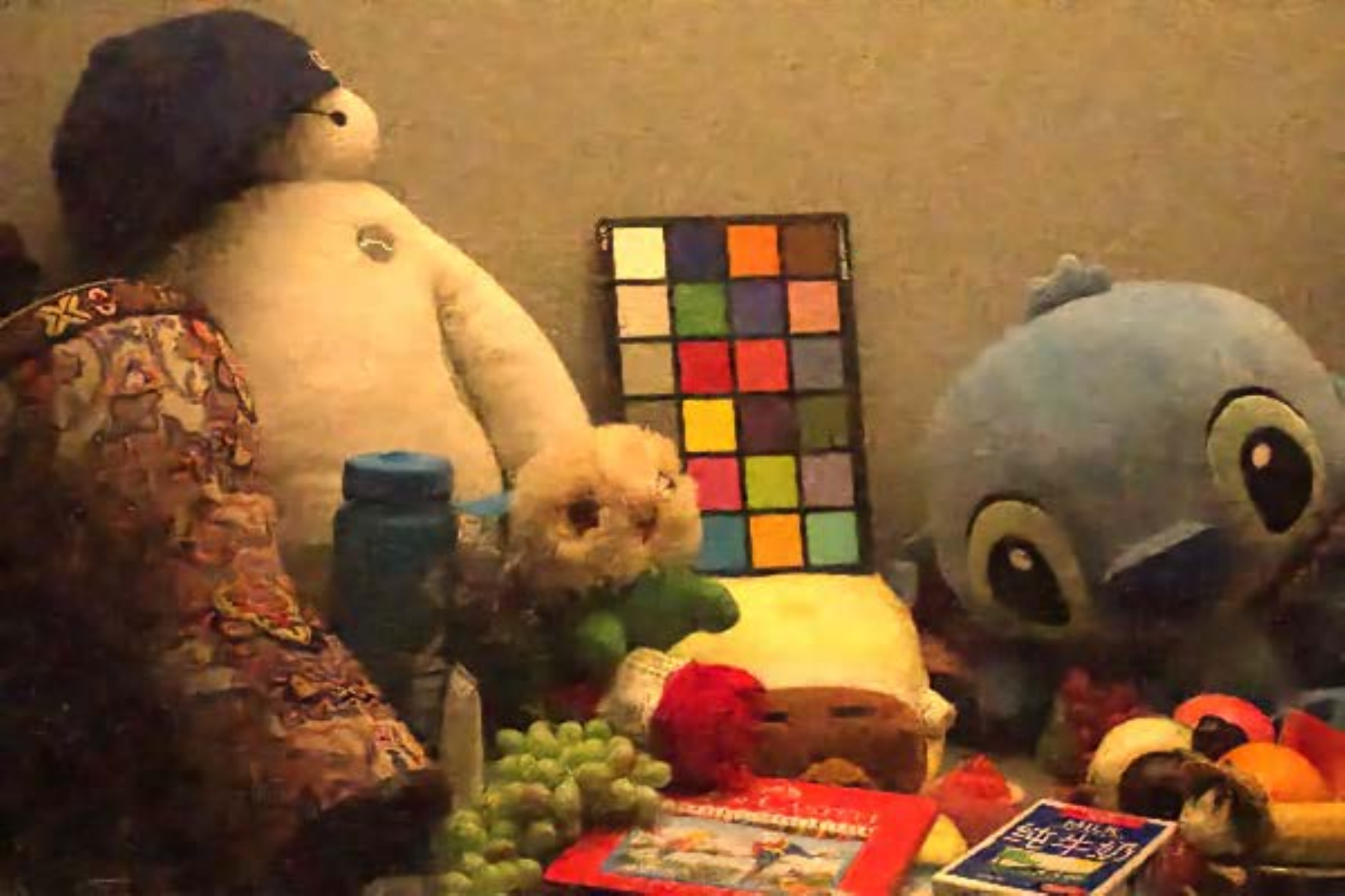}&
			\includegraphics[width=0.156\linewidth]{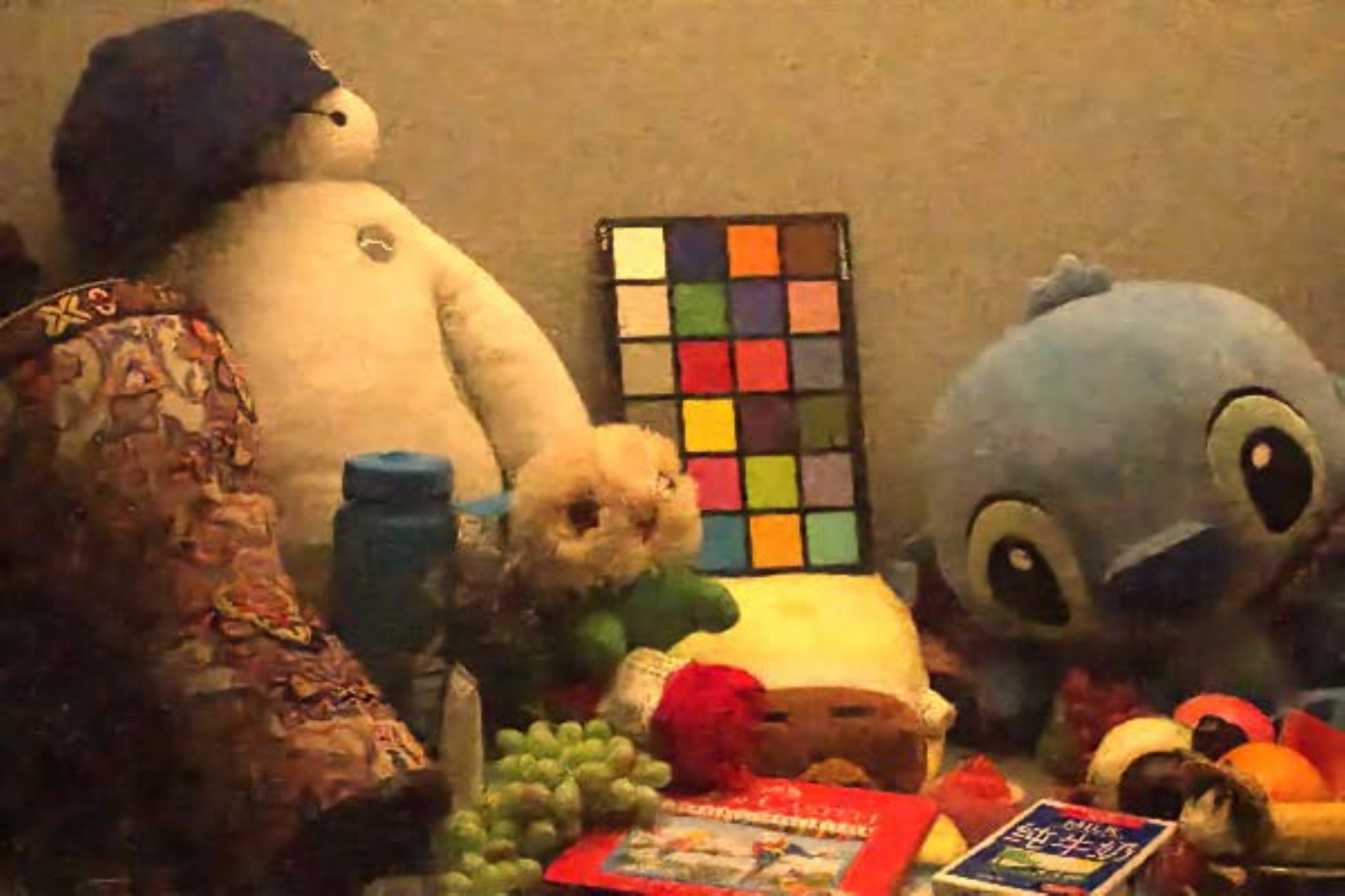}\\
			\footnotesize ZeroDCE&\footnotesize RUAS&\footnotesize UTVNet&\footnotesize SCL&\footnotesize \textbf{BL}&\footnotesize \textbf{RBL}\\
			\multicolumn{6}{c}{\footnotesize (b) Visual comparison among different methods on the LOL dataset~\citep{Chen2018Retinex}}\\
		\end{tabular}
		\caption{Visual results of state-of-the-art methods and our two versions (BL and RBL) on different datasets.}
		\label{fig: MIT_LOL}
		%	\vspace{-1em}
	\end{figure*}
	
	\section{Experimental Results}
	In this section, we first introduced the implementation details related to the experiments. Subsequently, aiming to comprehensively validate the superiority of our method, we provided subjective and objective comparison results and analysis between the proposed method and state-of-the-art methods in the field, conducted in multiple testing cases, including seen-paired, unseen-paired, unseen-unpaired datasets.
	
	%Furthermore, we make a series of comparison among our algorithm and other state-of-art algorithms on various datasets. Lastly, we provide analyses of our proposed method. 
	%-------------------------------------------------------------------------

	\subsection{Implementation Details}
	Here, we introduced implementation details from three aspects, including benchmarks description, compared methods and metrics, and parameters setting.

	\textbf{Benchmarks description.}$\;$
	We adopted five representative datasets (including MIT~\citep{fivek}, LOL~\citep{Chen2018Retinex}, LSRW~\citep{hai2021r2rnet}, VOC~\citep{lv2021attention}, DARKFACE~\citep{yang2020advancing}) to comprehensively evaluate the performance. As shown in Table~\ref{table: benchmarks}, we executed the learning phase (BL and RBL) by using MIT (without noises) and LOL (with noises) datasets to define $\bm{\Theta}_{\mathcal{F}_{\mathtt{E}}}$ and $\bm{\Theta}_{\mathcal{G}}$. Then all four datasets were performed in the adaptation (for defining $\bm{\Theta}_{\mathcal{F}_{\mathtt{D}}}$) and testing phases. The adopted numbers of different stages for each dataset were also reported in Table~\ref{table: benchmarks}. Notice that we used the same setting in the learning and adaptation stages. Finally, we also made evaluations on some challenging scenarios.
	
	\textbf{Compared methods and metrics.}
	We compared the proposed method with a series of recently proposed state-of-the-art approaches, including RetinexNet~\citep{Chen2018Retinex}, DeepUPE~\citep{wang2019underexposed}, KinD~\citep{zhang2019kindling}, EnGAN~\citep{jiang2019enlightengan}, UTVNet~\citep{zheng2021adaptive}, ZeroDCE~\citep{li2021learning}, FIDE~\citep{xu2020learning}, SCL~\citep{liang2022semantically}, RUAS~\citep{liu2021retinex}. We utilized several mainstream metrics with reference (including PSNR, SSIM, and LPIPS~\citep{LPIPS_2018}), and metrics without reference (including DE~\citep{shannon1948mathematical}, LOE~\citep{wang2013naturalness} and NIQE~\citep{mittal2012making}) to measure the quality of output images.
	
	\textbf{Parameters setting.} During the learning and adaptation phases, we used the ADAM optimizer~\citep{kingma2014adam} with parameters $\beta_1=0.5$, $\beta_2=0.999$, and $\epsilon=1\times10^{-3}$. The minibatch size was set to 8 and learning rate was initialized to $1\times10^{-3}$. Besides, all our experiments was completed with PyTorch framework on NVIDIA TITAN XP.
	%{\color{cyan}
		%	\textbf{Parameters setting.} Our learning, adaptation, and testing phases were completed in the PyTorch framework on the computer with NVIDIA TITAN XP. In the learning and adaptation pahse, we used the ADAM optimizer~\citep{kingma2014adam} with parameters $\beta_1=xx$, $\beta_2=xx$, and $\epsilon=10^{-xx}$. The minibatch size was set to 8. The learning rate was initialized to $10^{-xx}$. The training epoch number was set to xxx.
		%}

	\subsection{Evaluations on Seen-Paired Benchmarks}
	In this part, we aimed at verifying the effectiveness of our proposed method in a series of standard datasets which contained the reference images. The first part provided numerical results on datasets that were used for learning, it follows the regular pattern, i.e., acquiring the overall parameters by using the same data distribution with the testing phase. The second part presented related subjective results.
	
	\textbf{Quantitative comparison.}$\;$ First and foremost, quantitative results on the MIT and LOL datasets were reported in Table ~\ref{table: MIT_LOL}. It is evident that the proposed BL and RBL methods consistently achieved top rankings (almost all are the best or second best) in the majority of the performance metrics among other methods. Furthermore, the proposed extended version demonstrated significant performance improvement compared to BL, providing evidence of the effectiveness and superiority of the reinforced bilevel learning scheme.
	
	%It is because our algorithm possessed the denoising capability and tends to generate images with vivid color, which are more in line with natural scenes.
	
	\textbf{Qualitative comparison.}$\;$ 
	Visual comparison with state-of-the-art methods were presented in Fig.~\ref{fig: MIT_LOL}. It could be seen that the results of other methods exhibit noticeable underexposure or overexposure. Some methods (RetinexNet and RUAS) even demonstrated color distortion issues. In comparison, our method achieved superior performance on the MIT dataset, demonstrating more appropriate brightness and color in contrast to other methods. Furthermore, the results on the LOL dataset provided evidence that the proposed method significantly removed noise and exhibits clear advantages over the comparative methods.
	
	\begin{table*}[ht]
		\renewcommand\arraystretch{1.5}	
		\setlength{\tabcolsep}{0.85mm}
		\footnotesize
		\centering
		\caption{Quantitative results (Metrics with reference: PSNR, SSIM, and LPIPS; Metrics without reference: DE, LOE, and NIQE) on LSRW and VOC datasets. The best result is in bold red whereas the second best one is in bold blue.}
		~
		\begin{tabular}{|c|c||cccccccccc||cc|}
			\hline
			\multicolumn{2}{|c||}{Metrics}   & RetinexNet& DeepUPE & KinD & EnGAN & FIDE & DRBN & ZeroDCE &RUAS & UTVNet &SCL& BL & RBL\\
			\hline 
			\multirow{6}*{LSRW} & PSNR$\uparrow$  & 15.9062 &16.8890 & 16.4717 & 16.3106 & \color{red}{\textbf{17.6694}}& 16.1497 & 15.8337 &14.4372& \color{blue}{\textbf{16.4771}}&16.1327&14.8726&15.3892\\
			\cline{2-14}
			& SSIM$\uparrow$  & 0.3725 & 0.5125& 0.4929 & 0.4697 & 0.5485 & 0.5422 & 0.4664 &0.4276&\color{blue}{\textbf{0.6673}}&0.5710& 0.5933& \color{red}{\textbf{0.6761}}\\
			\cline{2-14}
			& LPIPS$\downarrow$  & 0.4326&0.3466&0.3371&0.3299&0.3351&0.3347&\color{blue}{\textbf{0.3174}}&0.3726&\color{red}{\textbf{0.2913}}&0.3623&0.3844&0.3633\\ 
			\cline{2-14}
			&DE$\uparrow$ &6.9392&6.7712&\color{blue}{\textbf{7.0368}}&6.6692 &6.8745&\color{red}{\textbf{7.2051}}&6.8729&5.6056&7.0330&6.4815&6.8634&  6.8293\\      
			\cline{2-14}
			& LOE$\downarrow$ &591.28&339.02&379.90&248.19&221.94&755.13&\color{blue}{\textbf{219.13}}&357.41&219.45&235.80&255.79&\color{red}{\textbf{215.92}}\\   
			\cline{2-14}
			& NIQE$\downarrow$ &4.1479&3.9816&\color{blue}{\textbf{3.6636}}&3.7754&4.3277&4.5500&3.7183&4.1687&4.0145&3.8422&4.0179&\color{red}{\textbf{3.5265}}\\        
			\hline	\hline
			\multirow{6}*{VOC} & PSNR$\uparrow$ &15.4087&15.7017&16.3160&10.7134&15.8569&15.7868&14.1782&\color{blue}{\textbf{17.0125}}&16.9024&14.8902&16.7280&\color{red}{\textbf{17.1492}}\\
			\cline{2-14}
			& SSIM$\uparrow$  & 0.5971&0.5619&0.6564&0.2985&0.6101&0.5209&0.4412&\color{red}{\textbf{0.6738}}&0.6539&0.5516&0.6325&\color{blue}{\textbf{0.6477}}\\
			\cline{2-14}
			& LPIPS$\downarrow$  &0.2489&0.2260&0.2260&0.2073&0.4002&0.2165&0.3034&0.3852&0.2239&\color{red}{\textbf{0.1797}}&\color{blue}{\textbf{0.1987}}&0.2103\\ 
			\cline{2-14} 
			& DE$\uparrow$ &\color{blue}{\textbf{7.2555}}&6.6517&\color{red}{\textbf{7.2906}}&5.7221&7.0681&6.8160&6.7553&6.9088&7.2089&6.3099&6.9586&6.8790\\   
			\cline{2-14}
			& LOE$\downarrow$ &845.26&485.59&389.41&311.41&408.34&591.20&423.78&340.24&423.47&435.74&\color{blue}{\textbf{246.98}}&\color{red}{\textbf{233.04}}\\           
			\cline{2-14}   
			& NIQE$\downarrow$ &6.0860&5.0727&\color{red}{\textbf{4.6222}}&5.1635&5.7456&4.7397&4.8665&4.7712&\color{blue}{\textbf{4.7071}}&4.8824&5.3080&4.8916\\
			\hline					
		\end{tabular}
		\label{table: LSRW_VOC}
	\end{table*}
	\begin{figure*}[!htb]
		\centering
		\begin{tabular}{c@{\extracolsep{0.3em}}c@{\extracolsep{0.3em}}c@{\extracolsep{0.3em}}c@{\extracolsep{0.3em}}c@{\extracolsep{0.3em}}c}
			\includegraphics[width=0.156\linewidth]{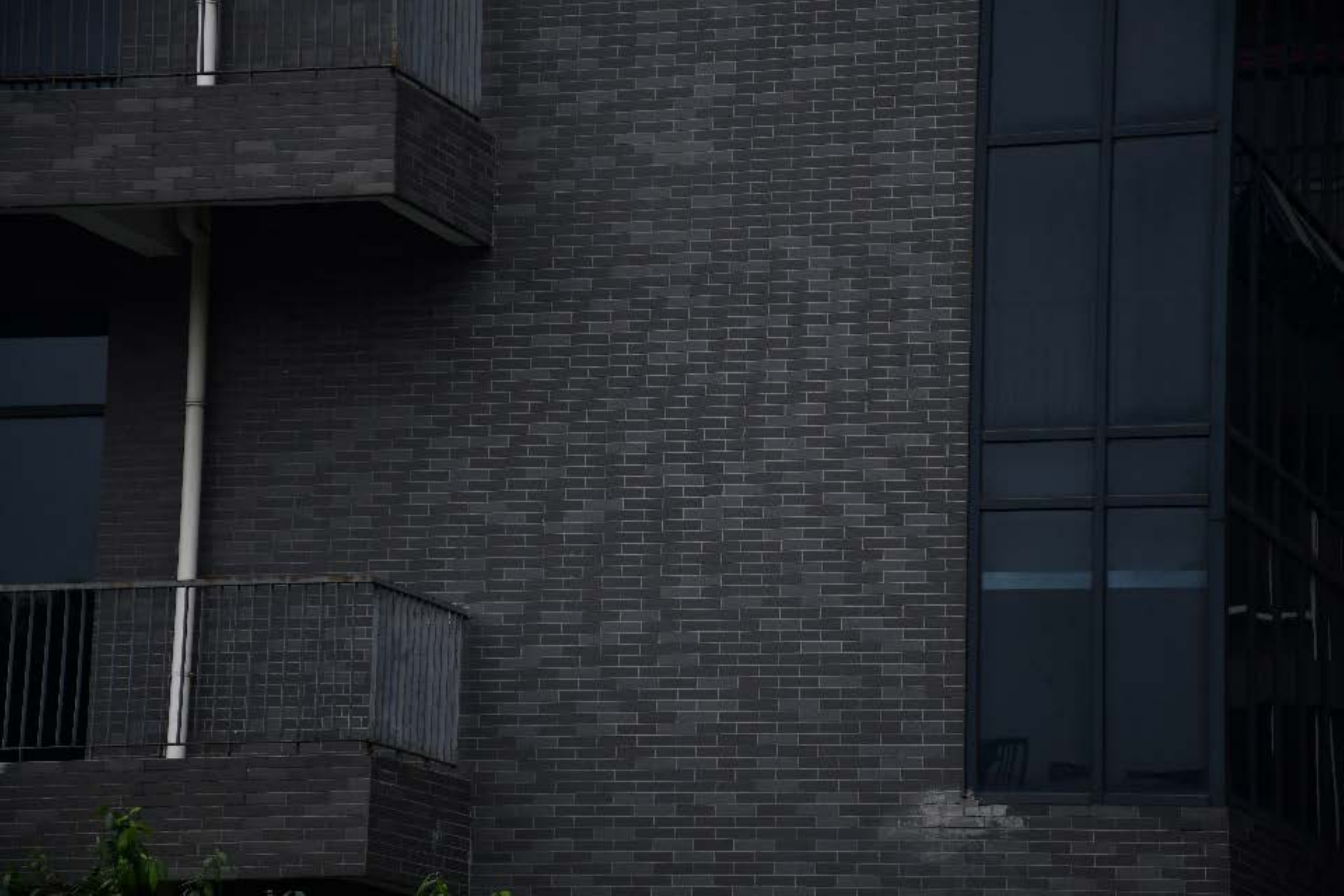}&
			\includegraphics[width=0.156\linewidth]{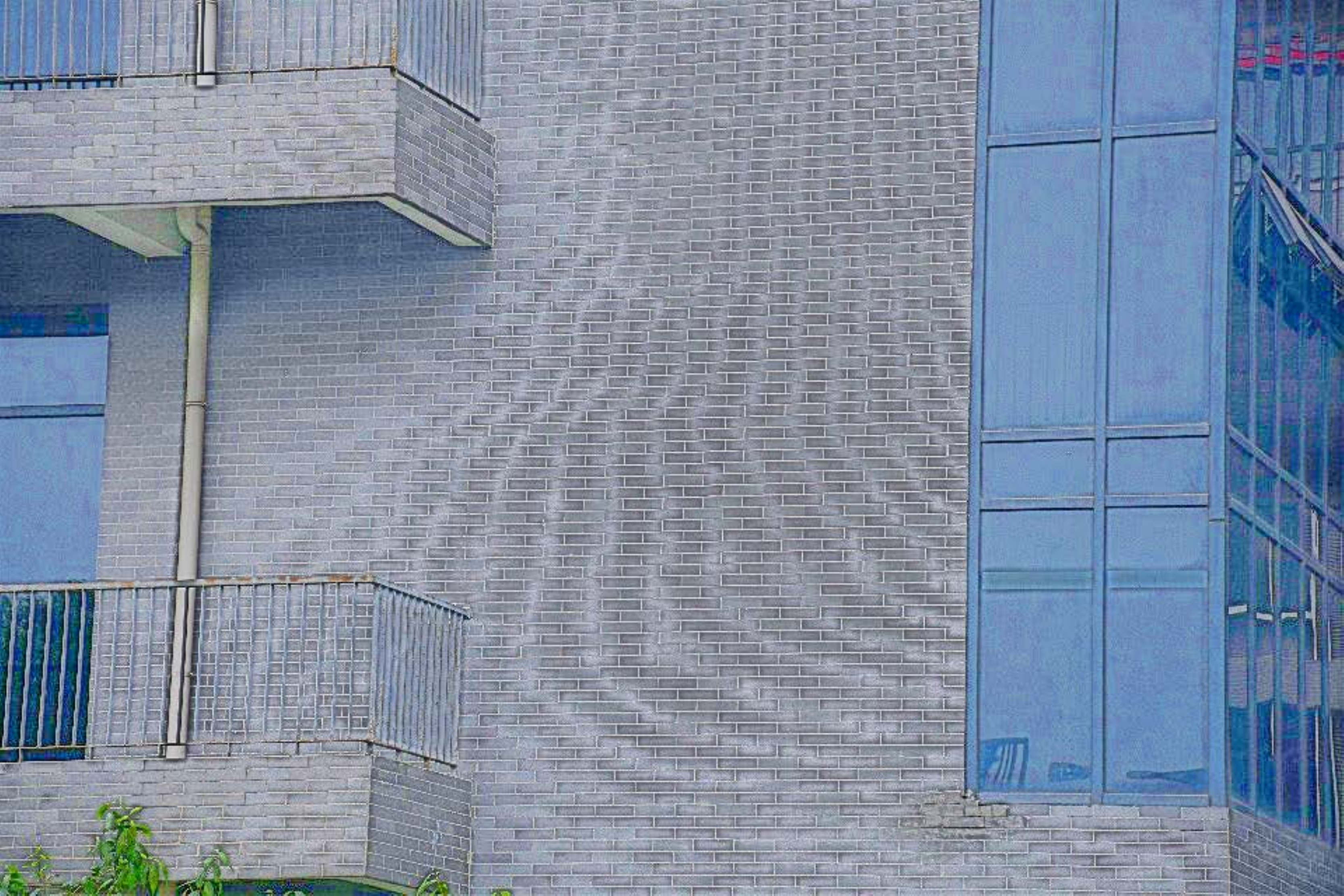}&
			\includegraphics[width=0.156\linewidth]{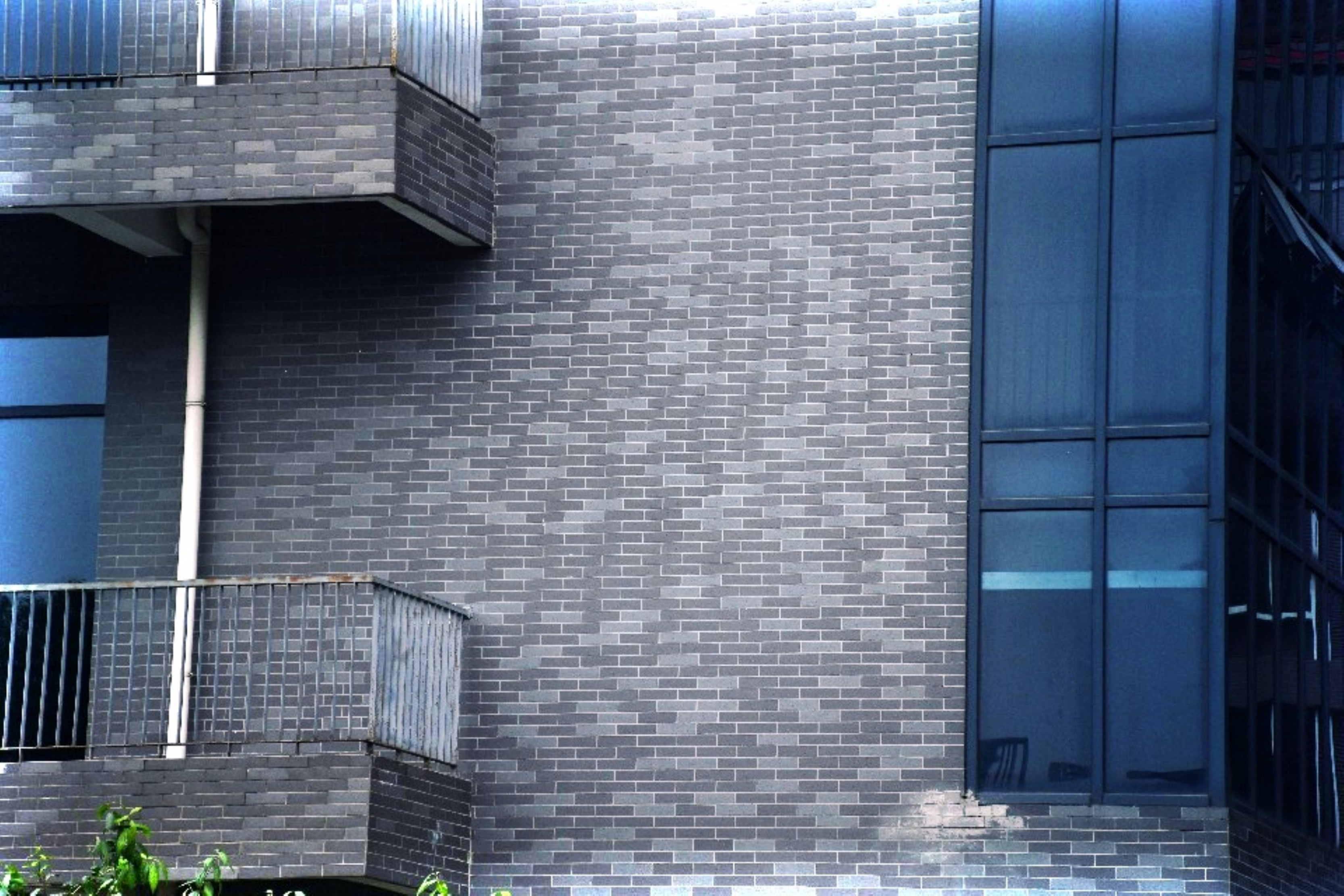}&
			\includegraphics[width=0.156\linewidth]{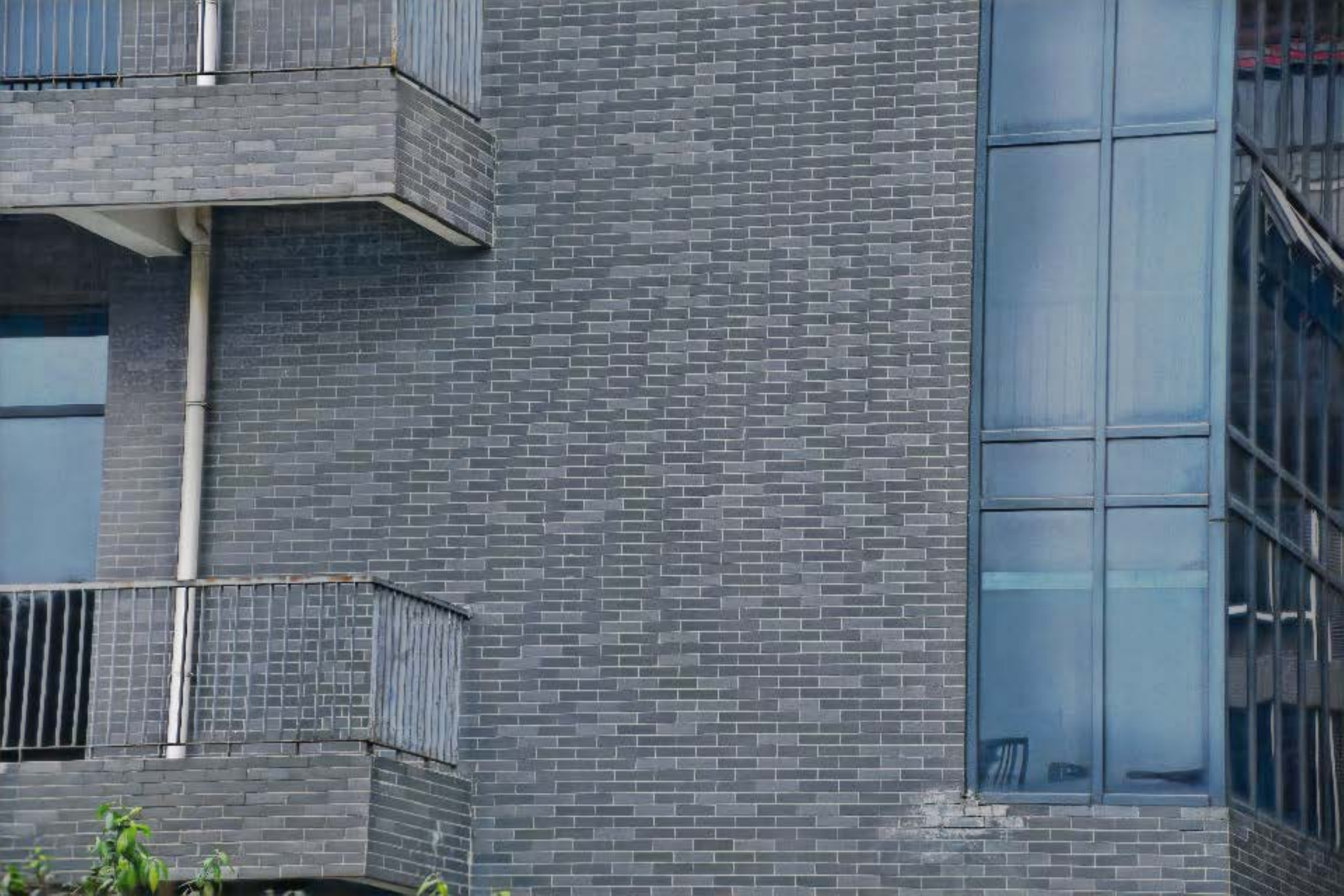}&
			\includegraphics[width=0.156\linewidth]{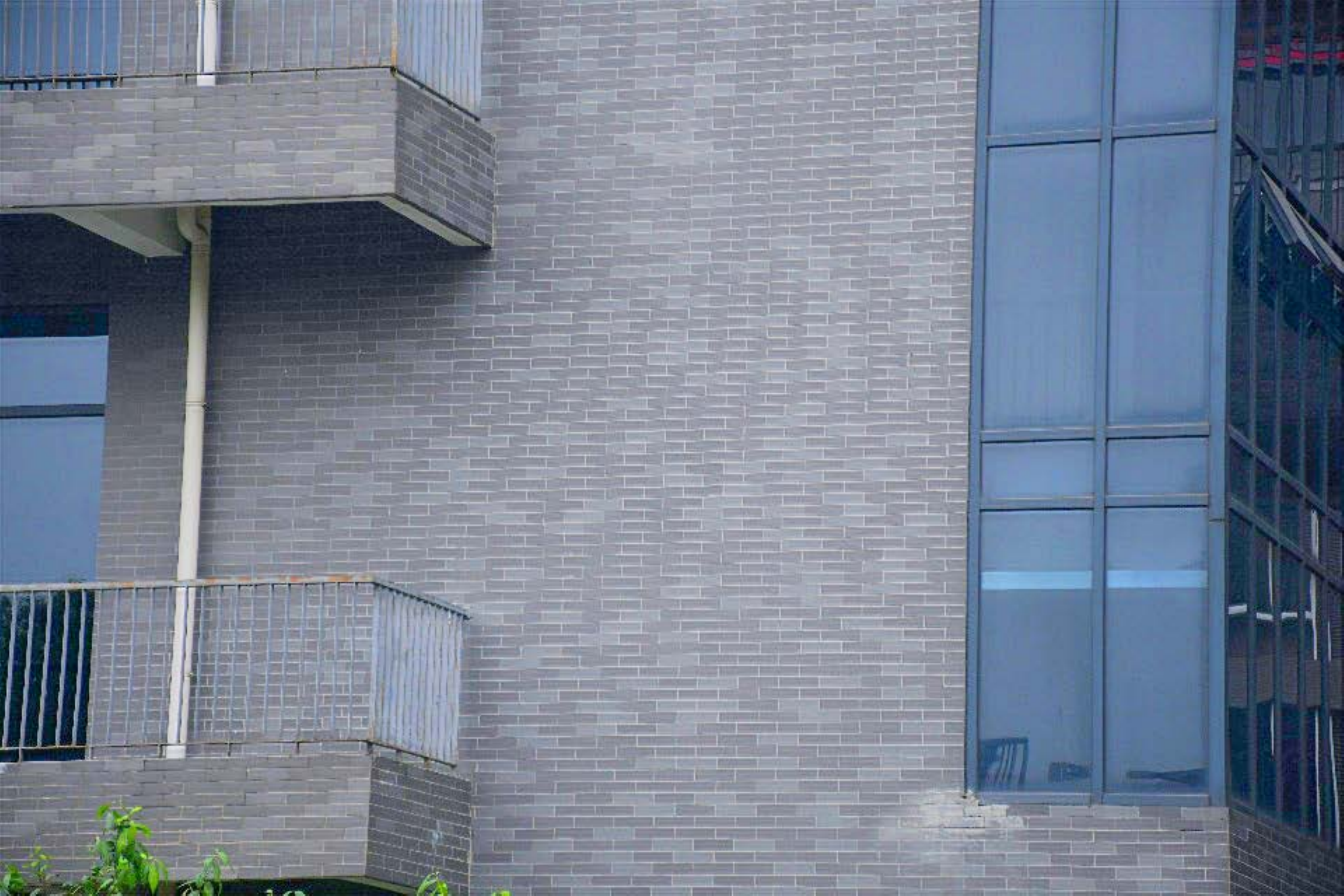}&
			\includegraphics[width=0.156\linewidth]{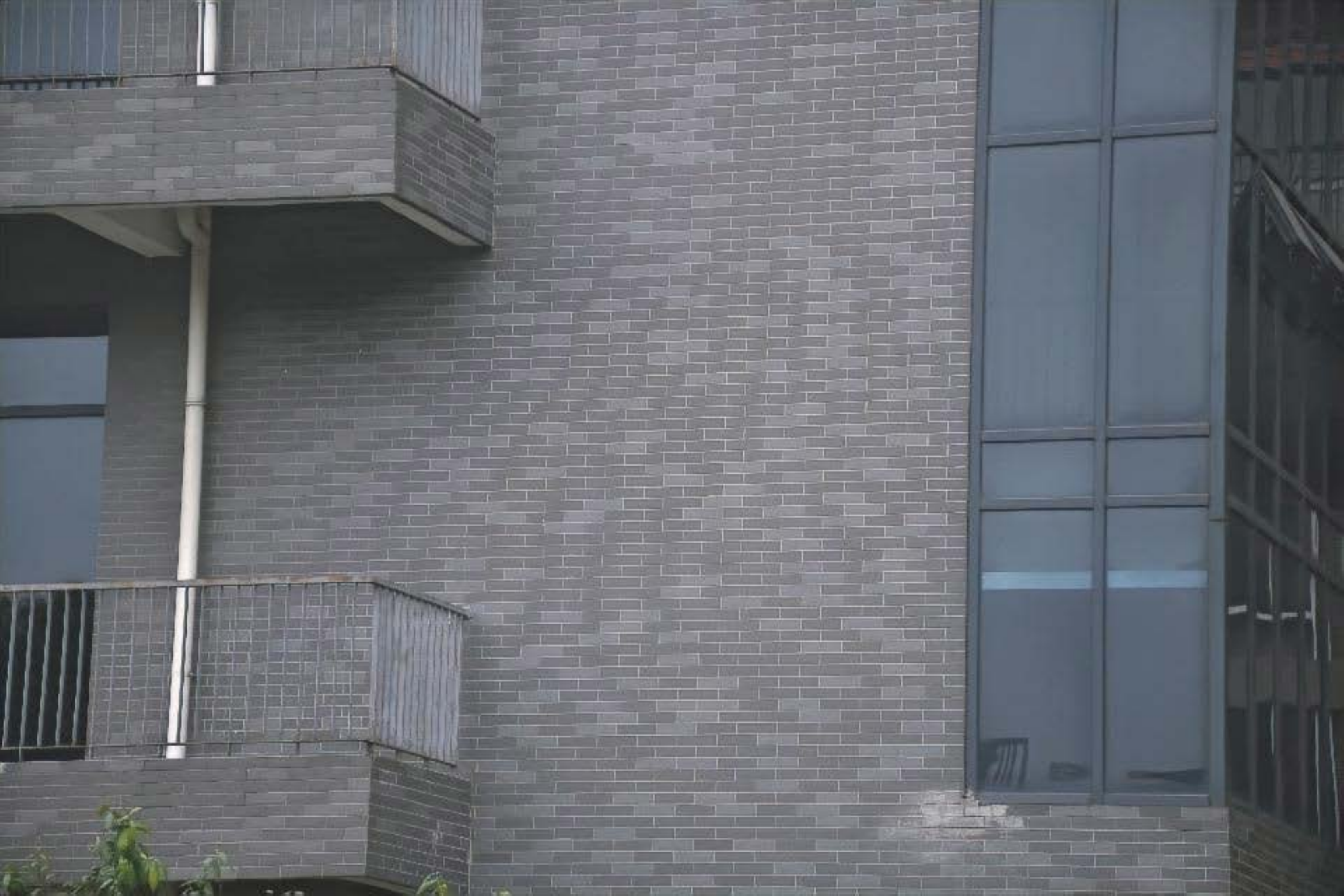}\\
			\footnotesize Input&\footnotesize RetinexNet&\footnotesize DeepUPE&\footnotesize KinD&\footnotesize EnGAN&\footnotesize FIDE\\
			\includegraphics[width=0.156\linewidth]{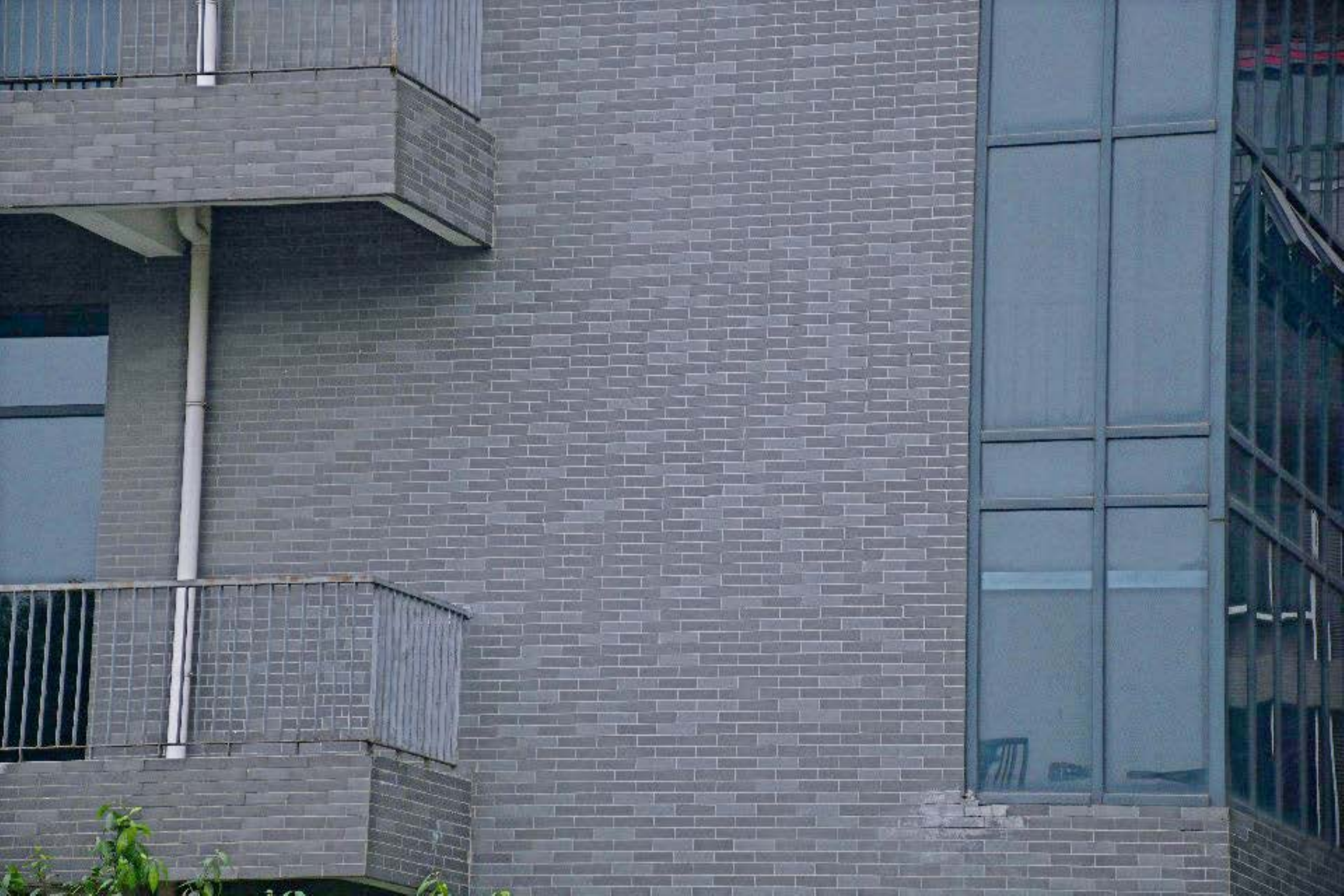}&
			\includegraphics[width=0.156\linewidth]{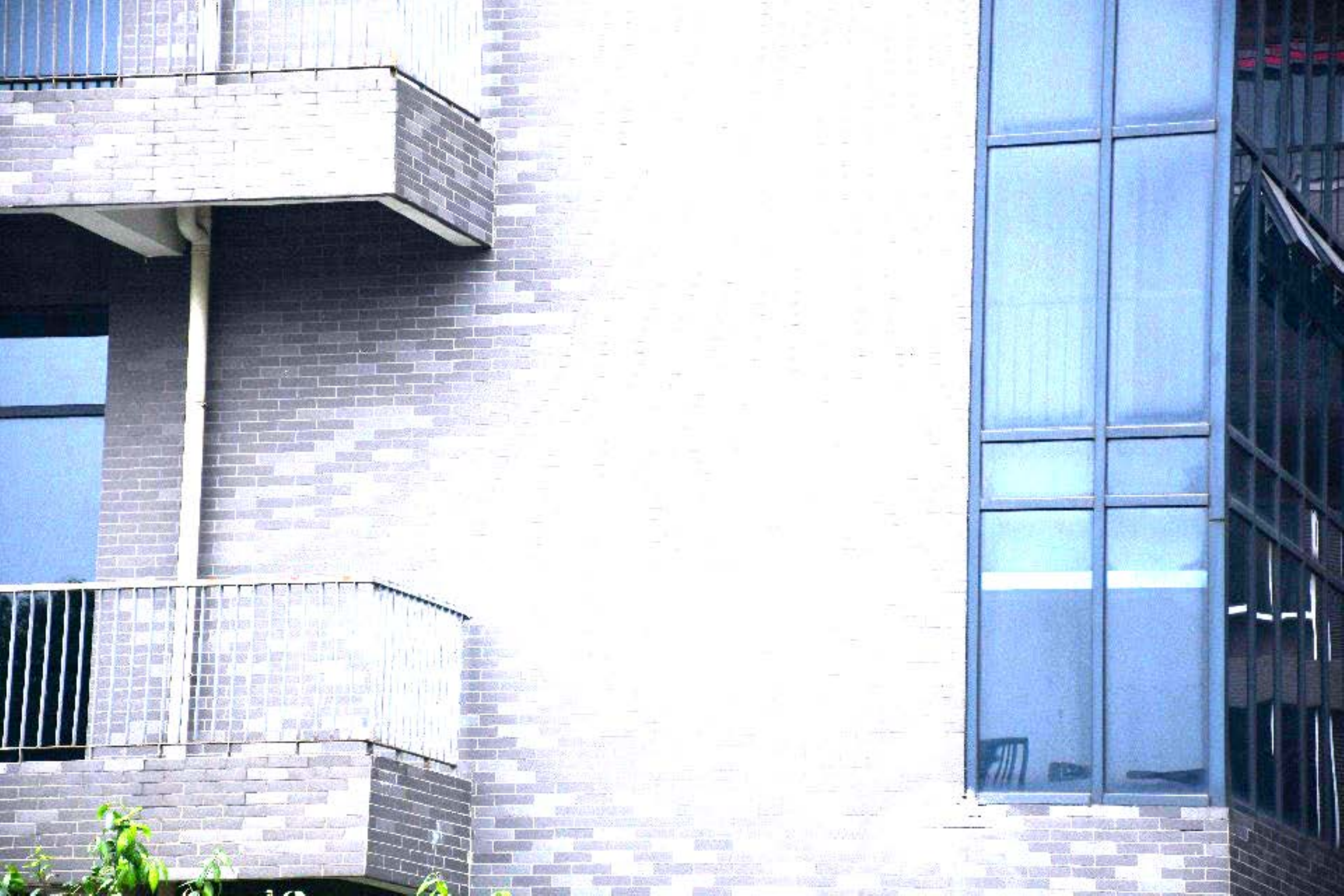}&
			\includegraphics[width=0.156\linewidth]{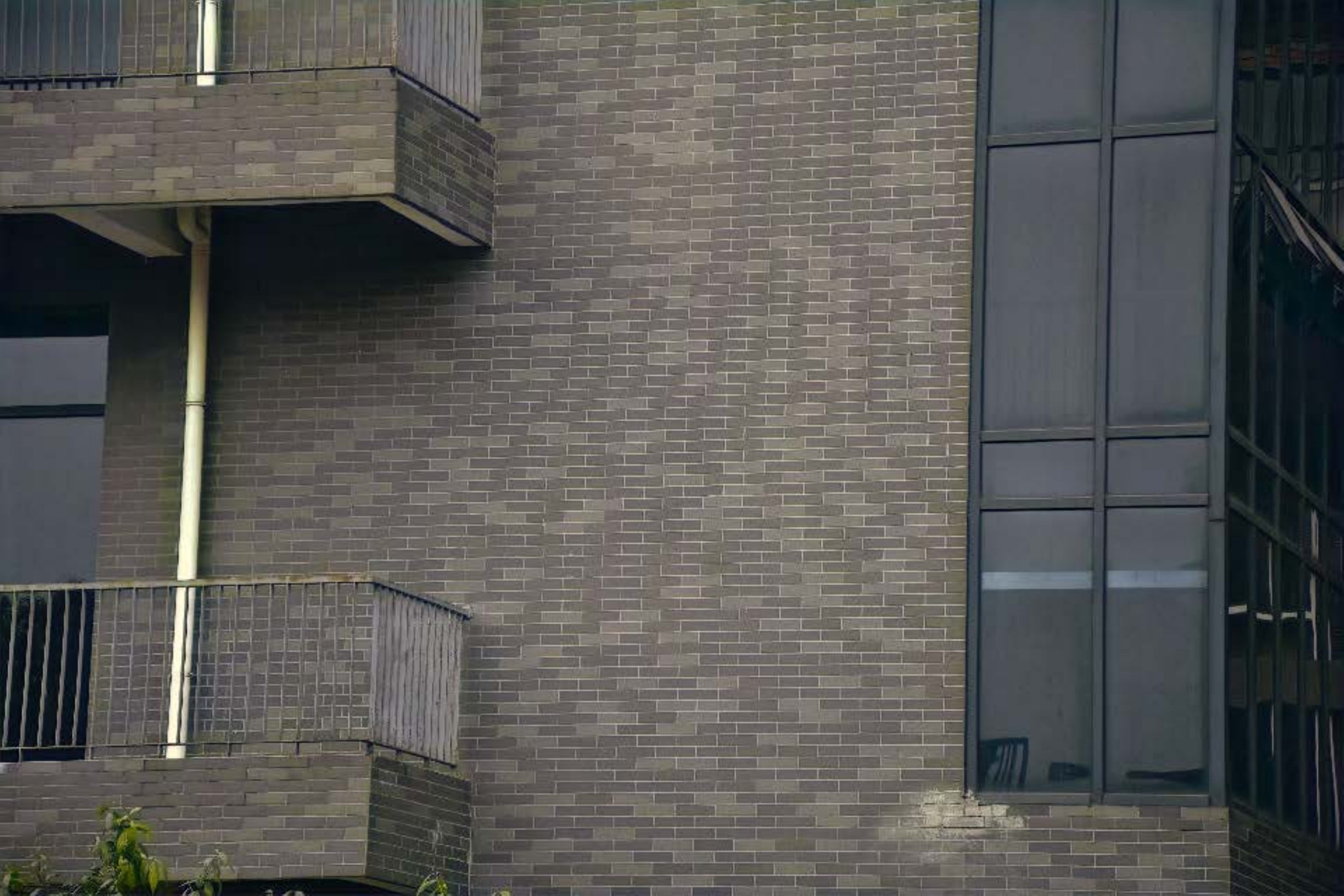}&
			\includegraphics[width=0.156\linewidth]{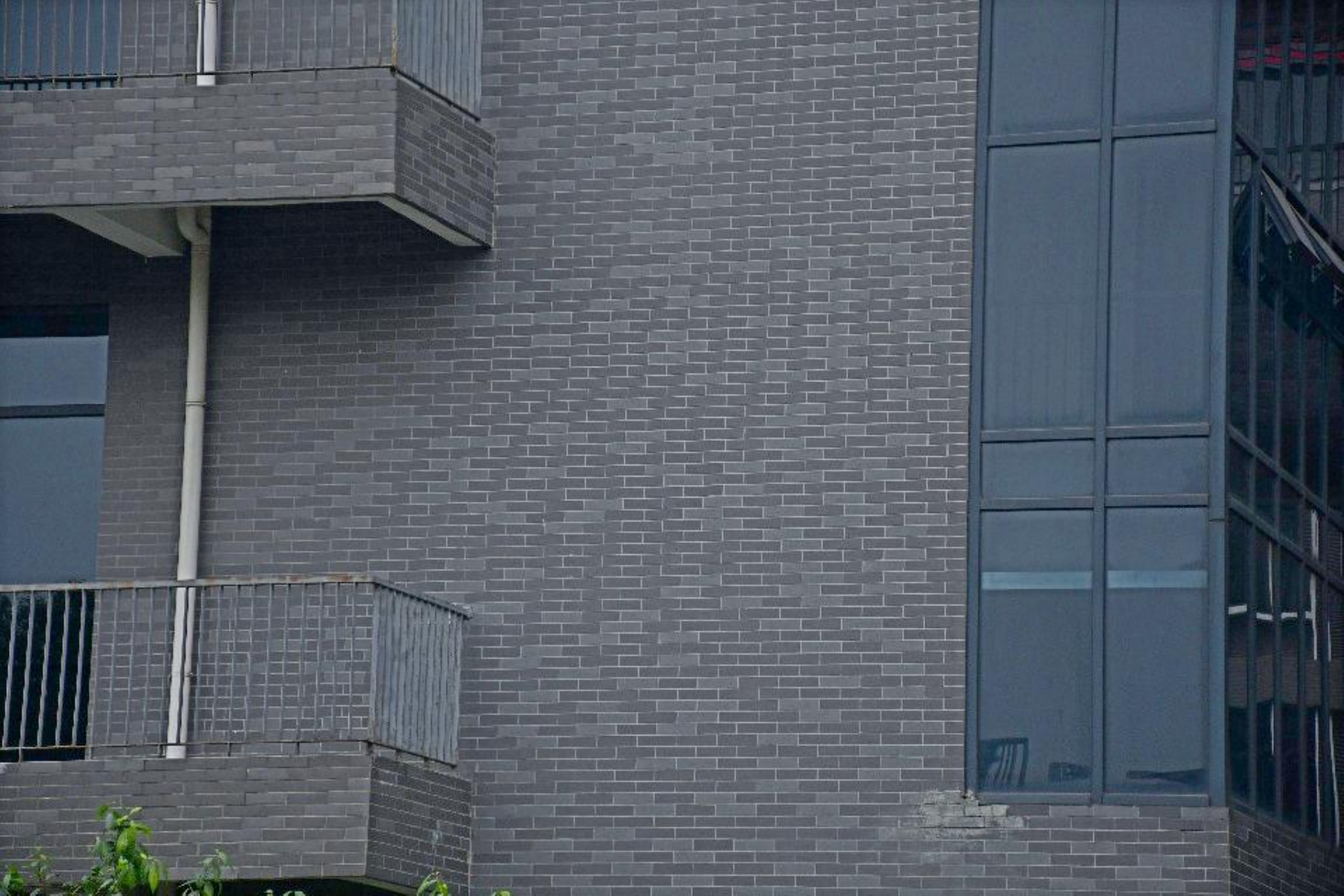}&
			\includegraphics[width=0.156\linewidth]{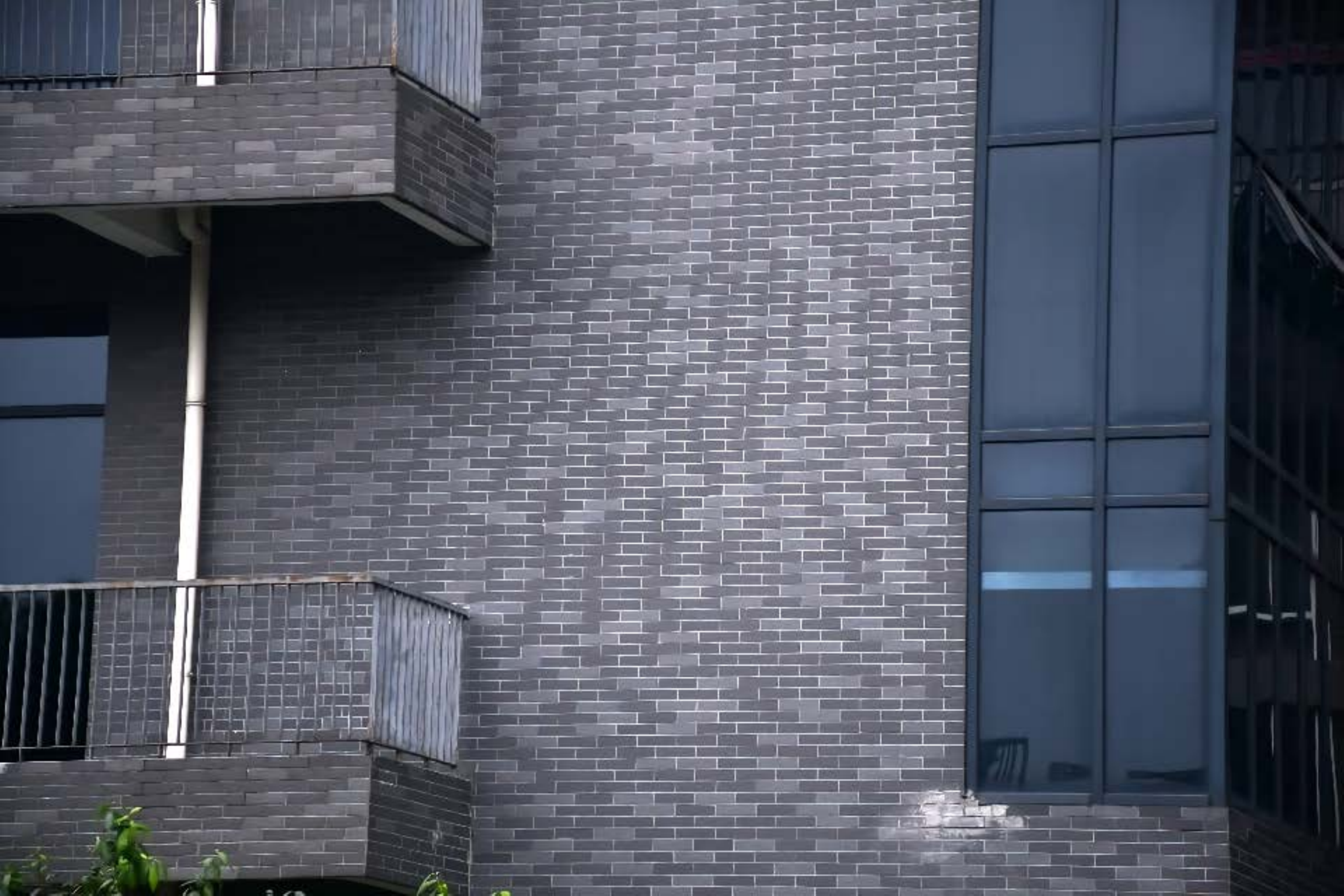}&
			\includegraphics[width=0.156\linewidth]{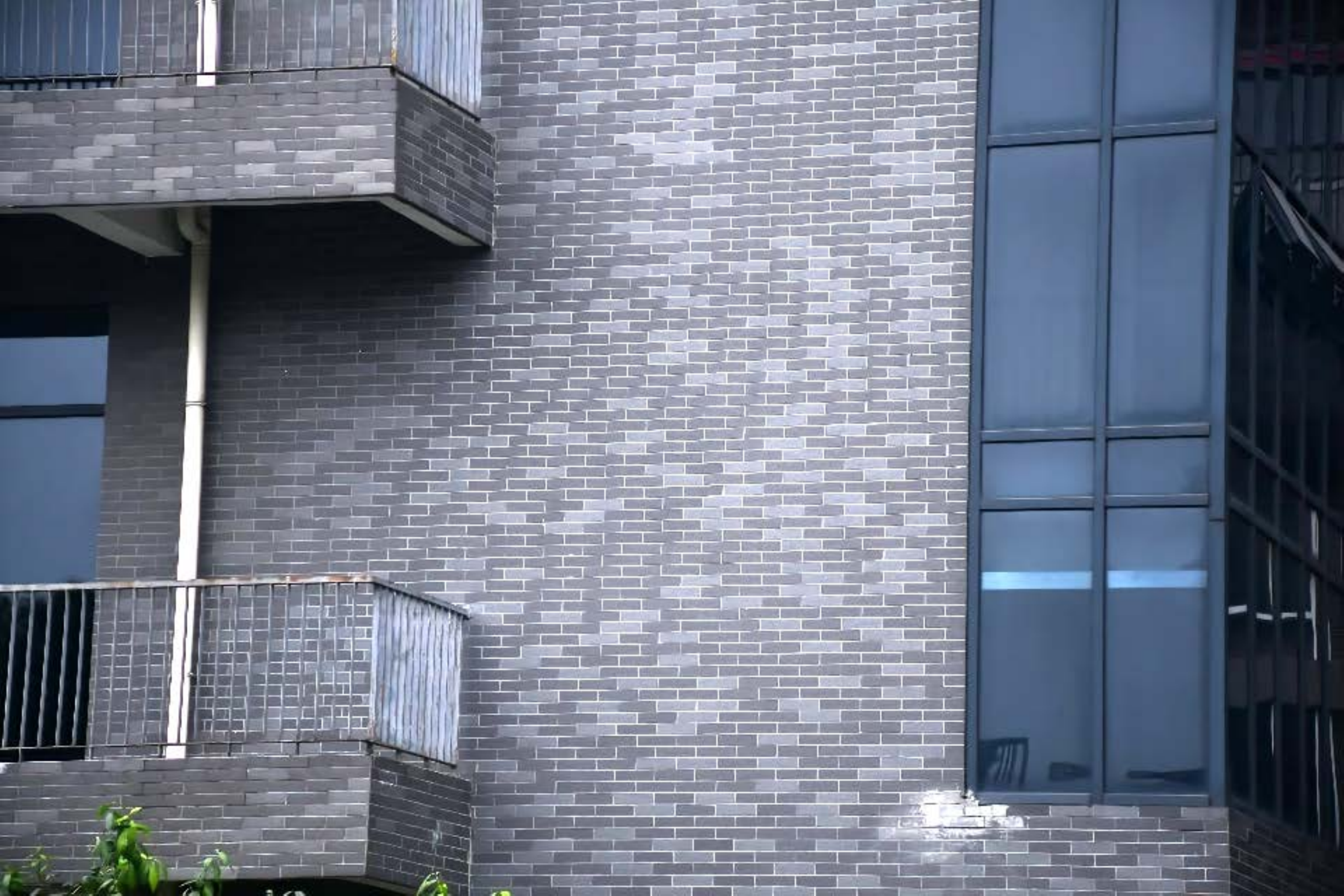}\\
			\footnotesize ZeroDCE&\footnotesize RUAS&\footnotesize UTVNet&\footnotesize SCL&\footnotesize \textbf{BL}&\footnotesize \textbf{RBL}\\
			\multicolumn{6}{c}{\footnotesize (a) Visual comparison among different methods on the LSRW dataset~\citep{hai2021r2rnet}}\\
			\includegraphics[width=0.156\linewidth]{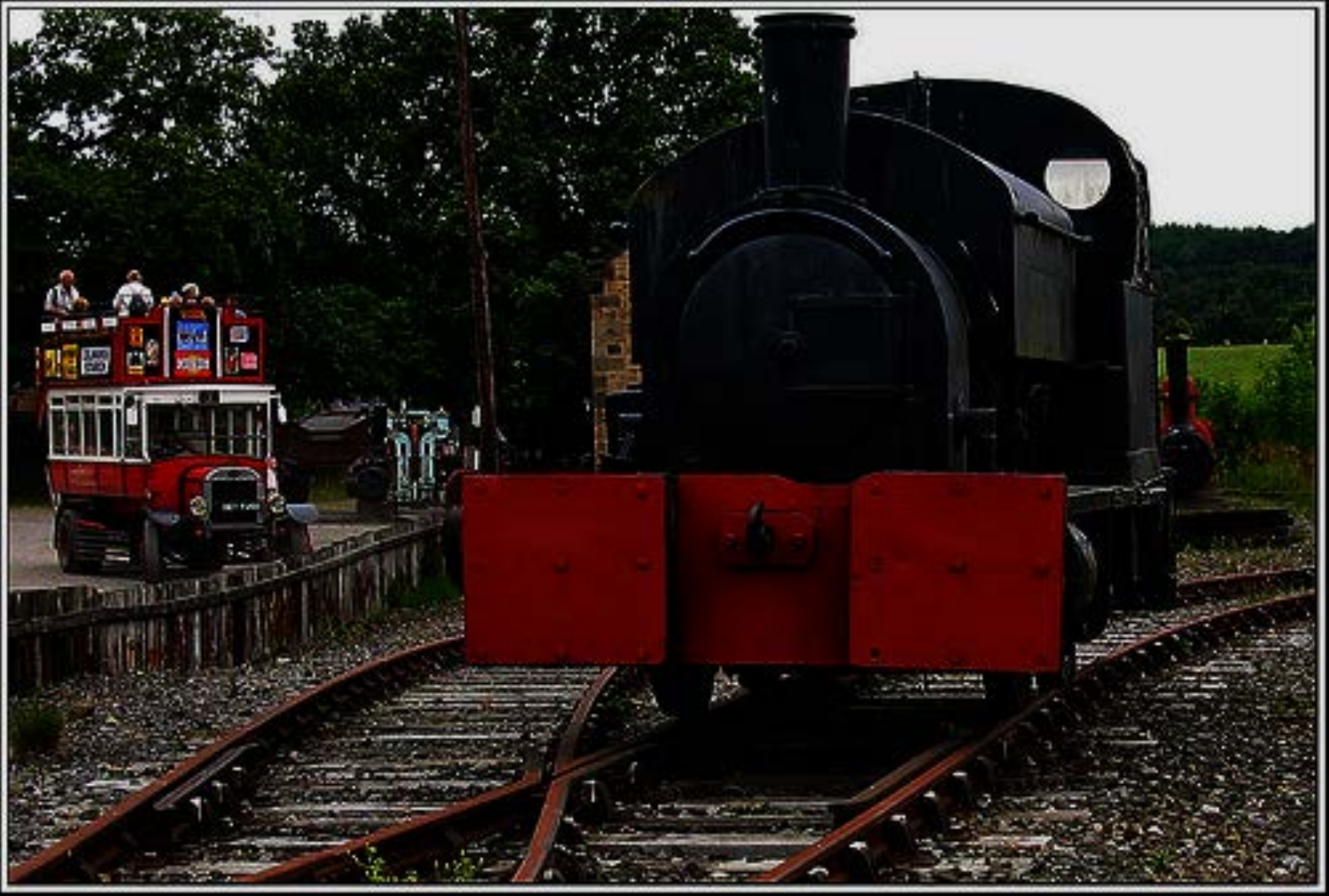}&
			\includegraphics[width=0.156\linewidth]{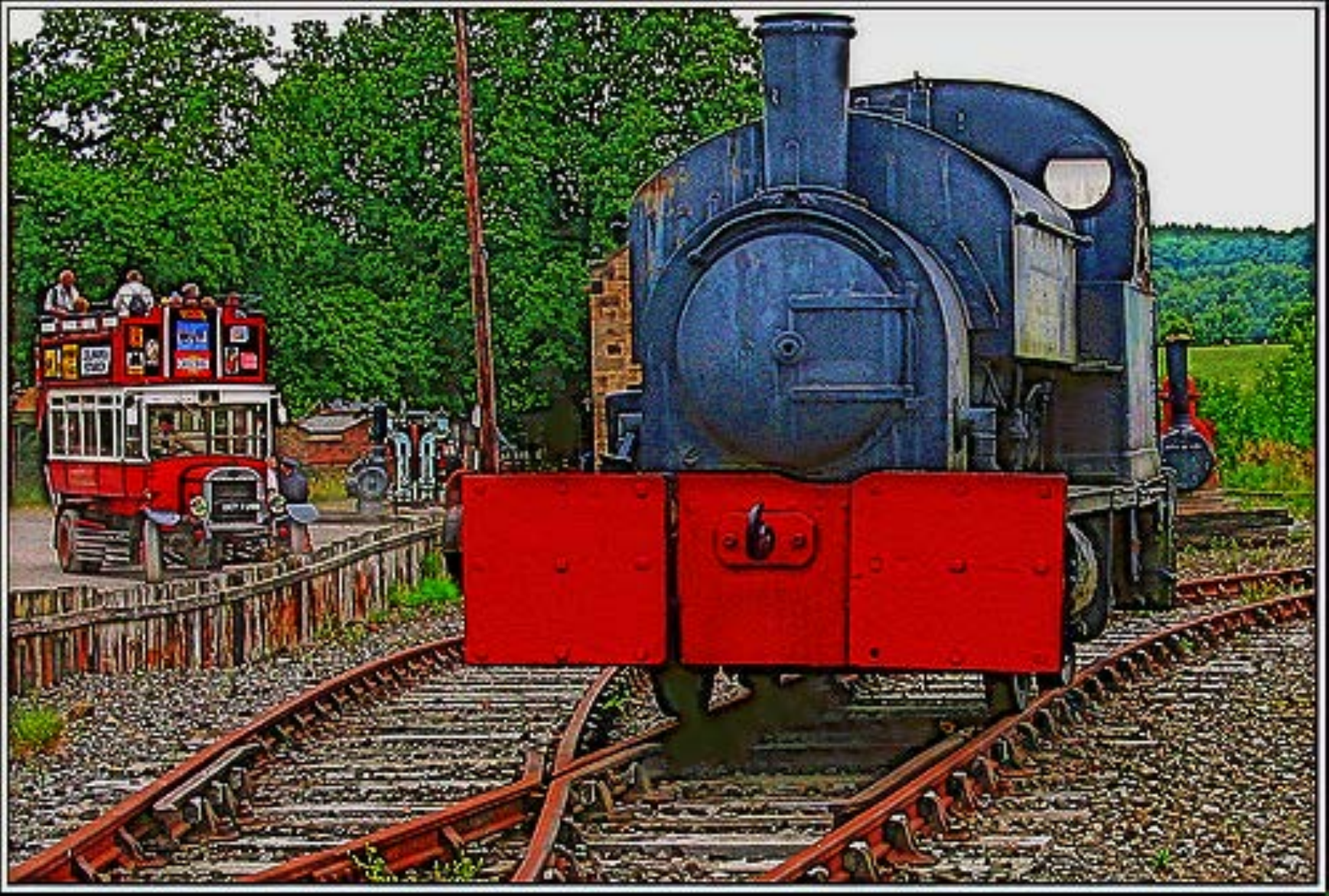}&
			\includegraphics[width=0.156\linewidth]{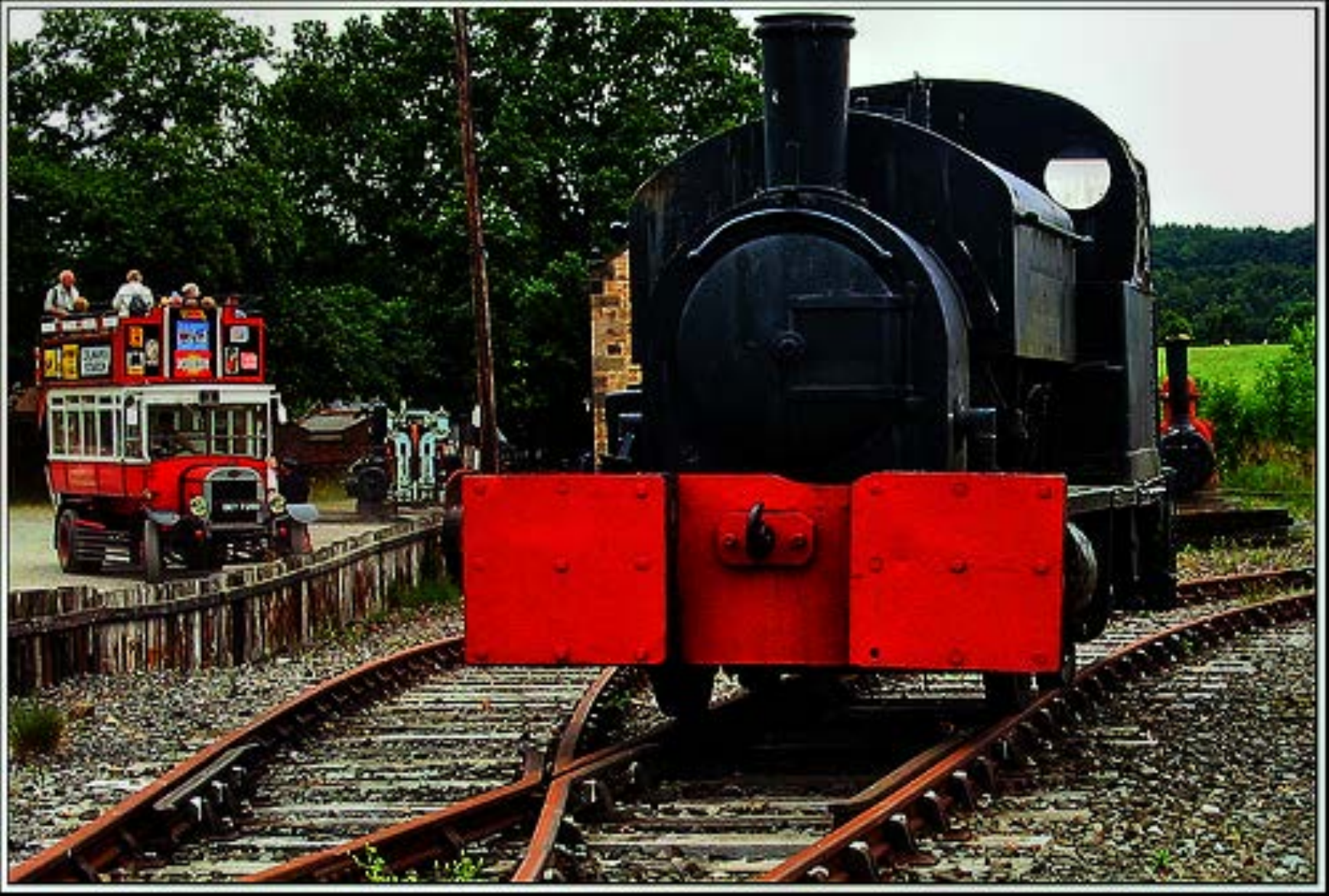}&
			\includegraphics[width=0.156\linewidth]{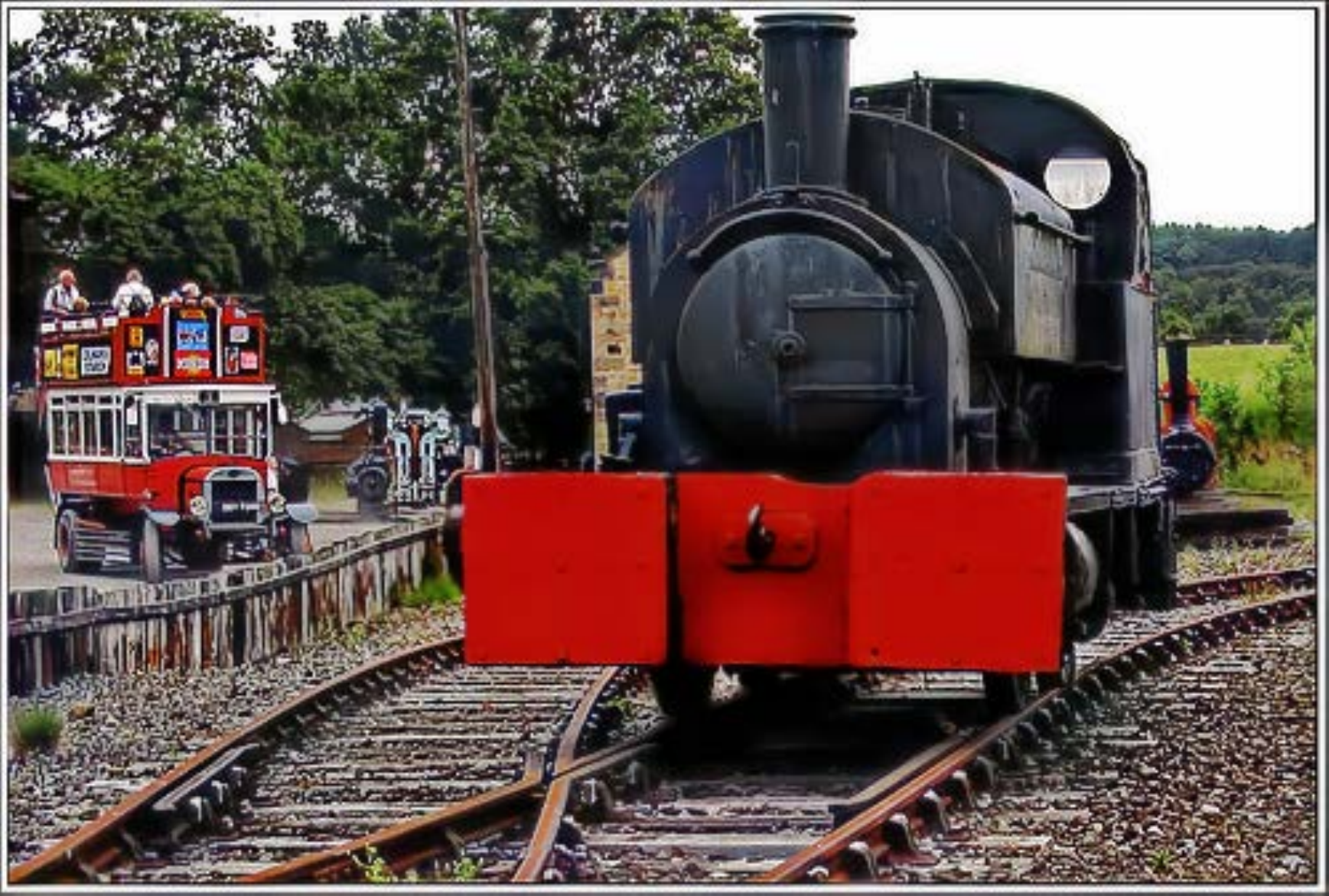}&
			\includegraphics[width=0.156\linewidth]{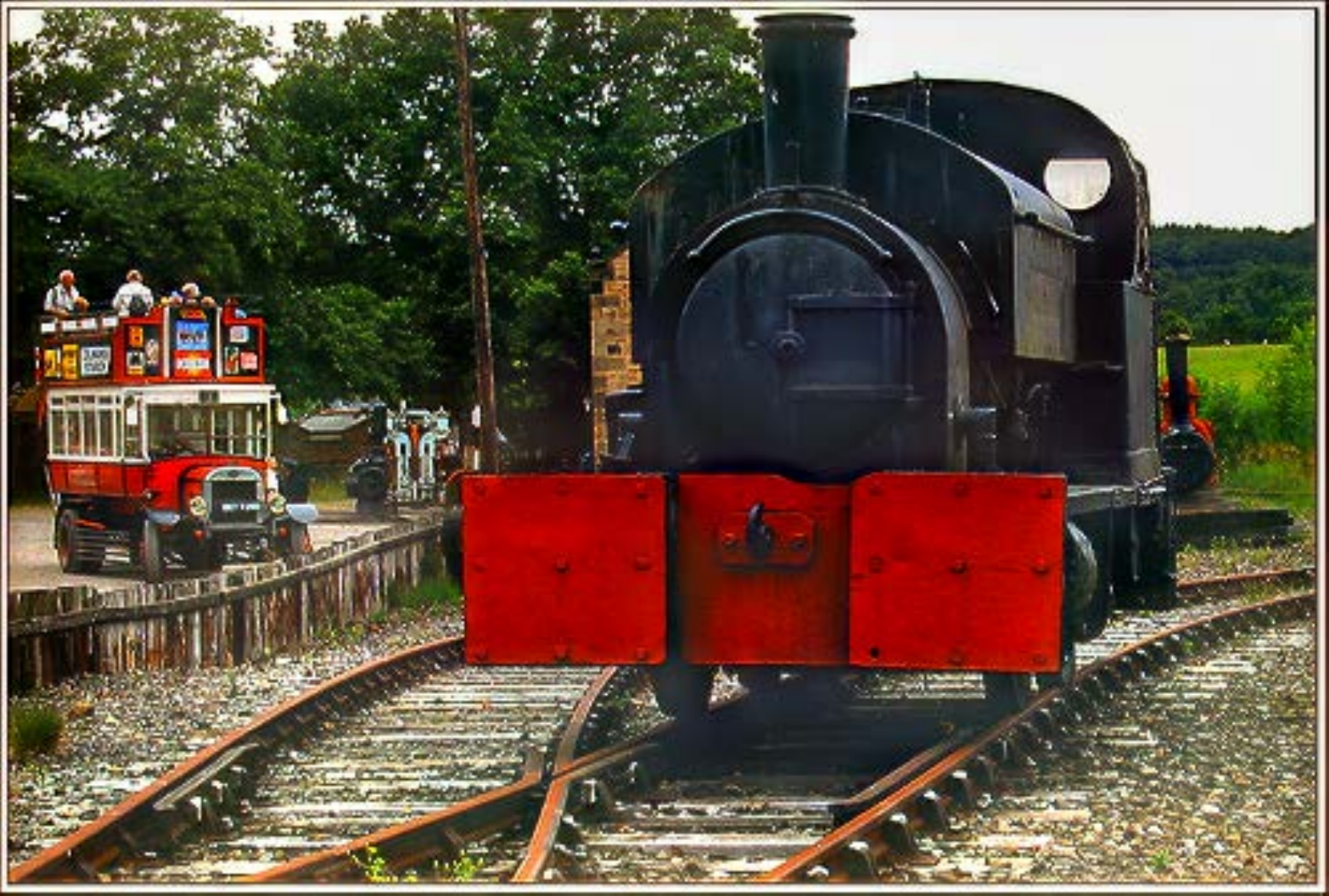}&
			\includegraphics[width=0.156\linewidth]{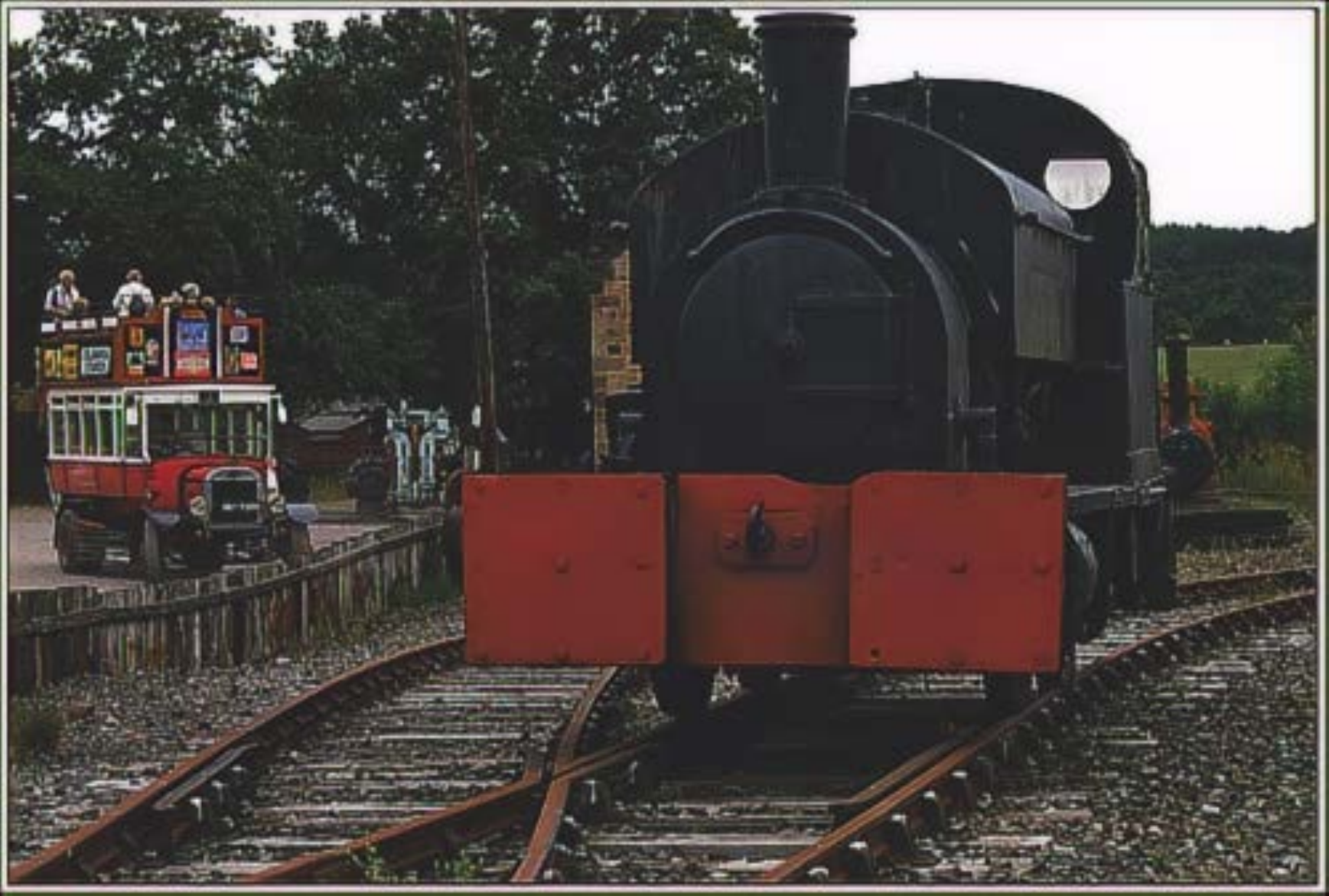}\\
			\footnotesize Input&\footnotesize RetinexNet&\footnotesize DeepUPE&\footnotesize KinD&\footnotesize EnGAN&\footnotesize FIDE\\
			\includegraphics[width=0.156\linewidth]{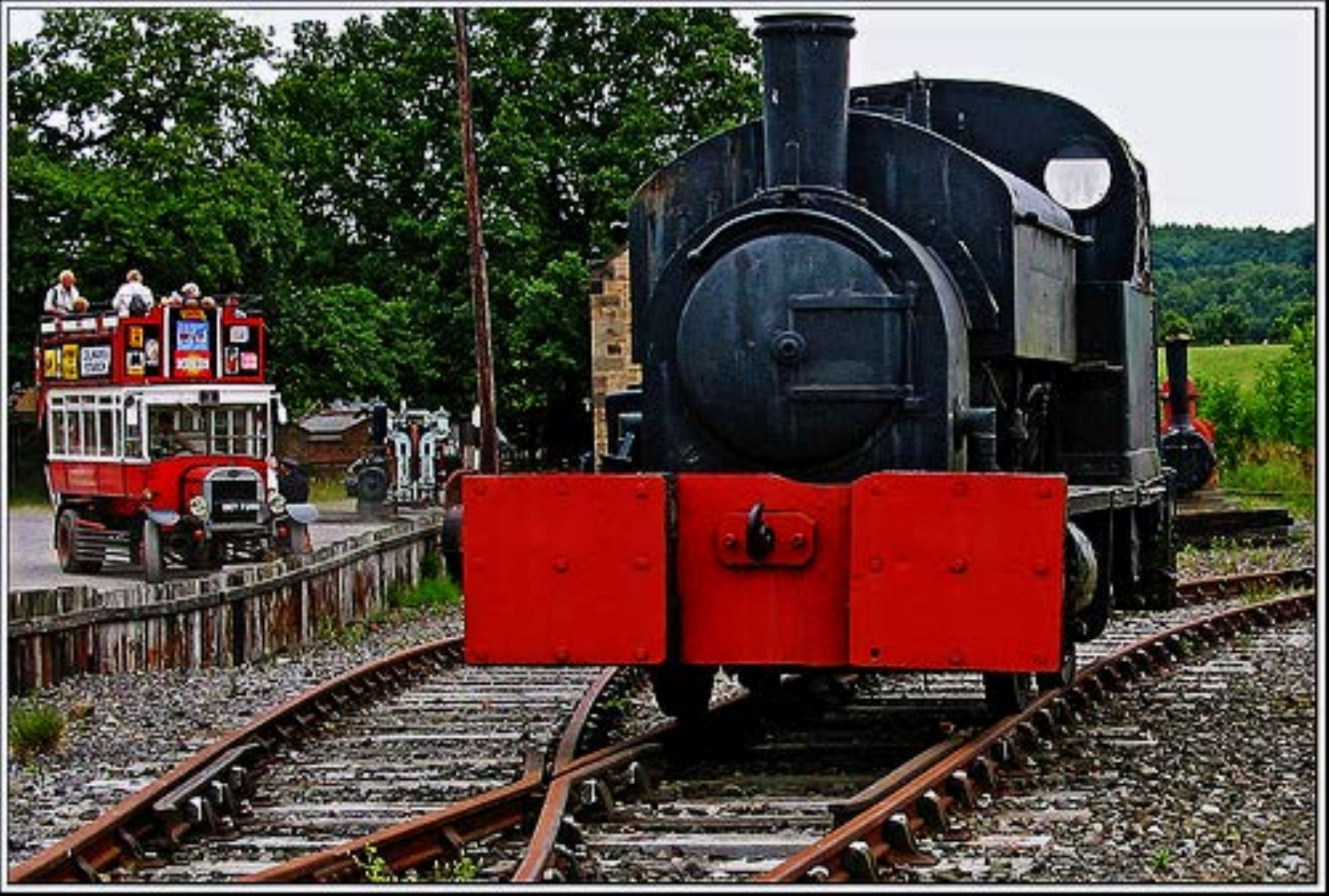}&
			\includegraphics[width=0.156\linewidth]{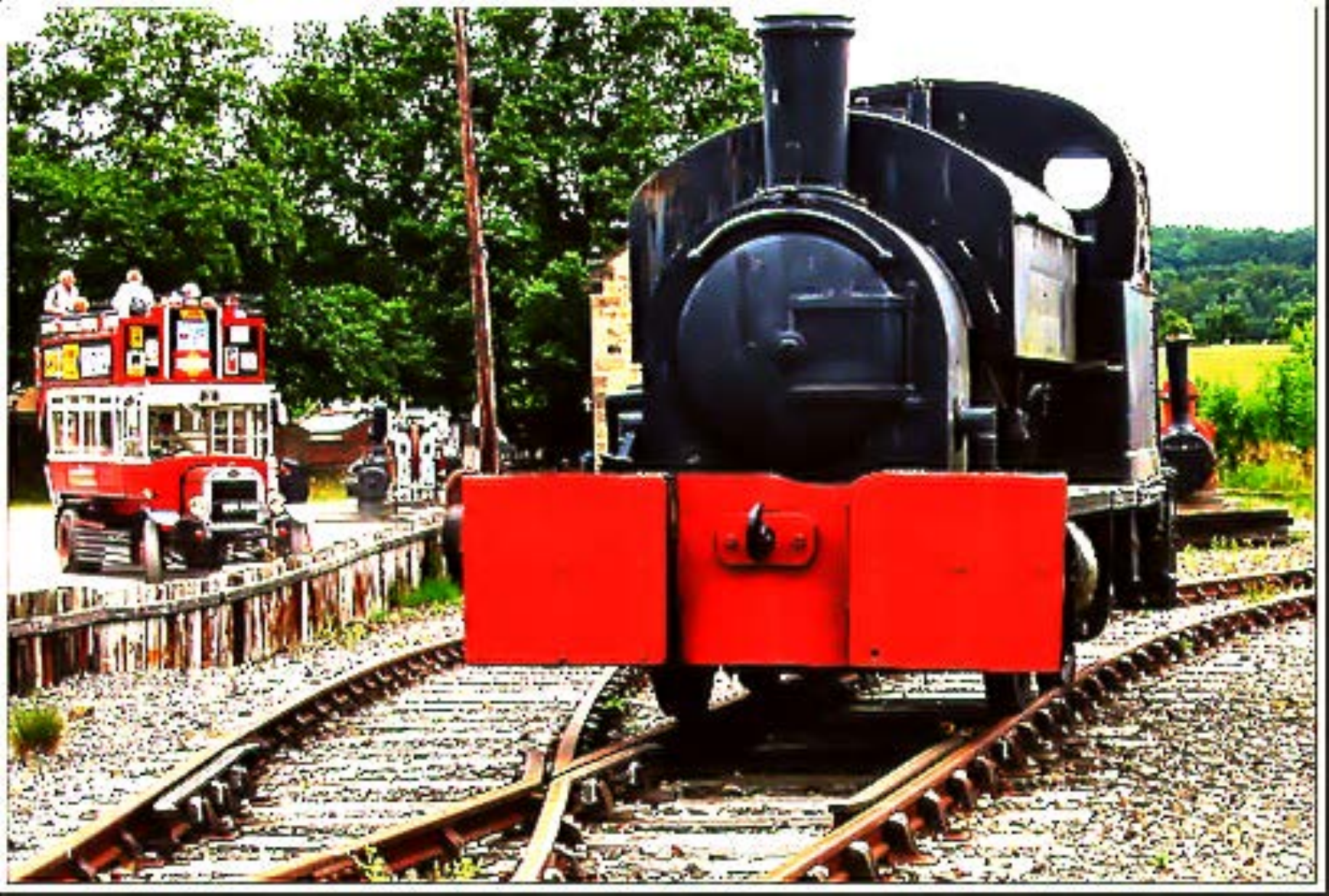}&
			\includegraphics[width=0.156\linewidth]{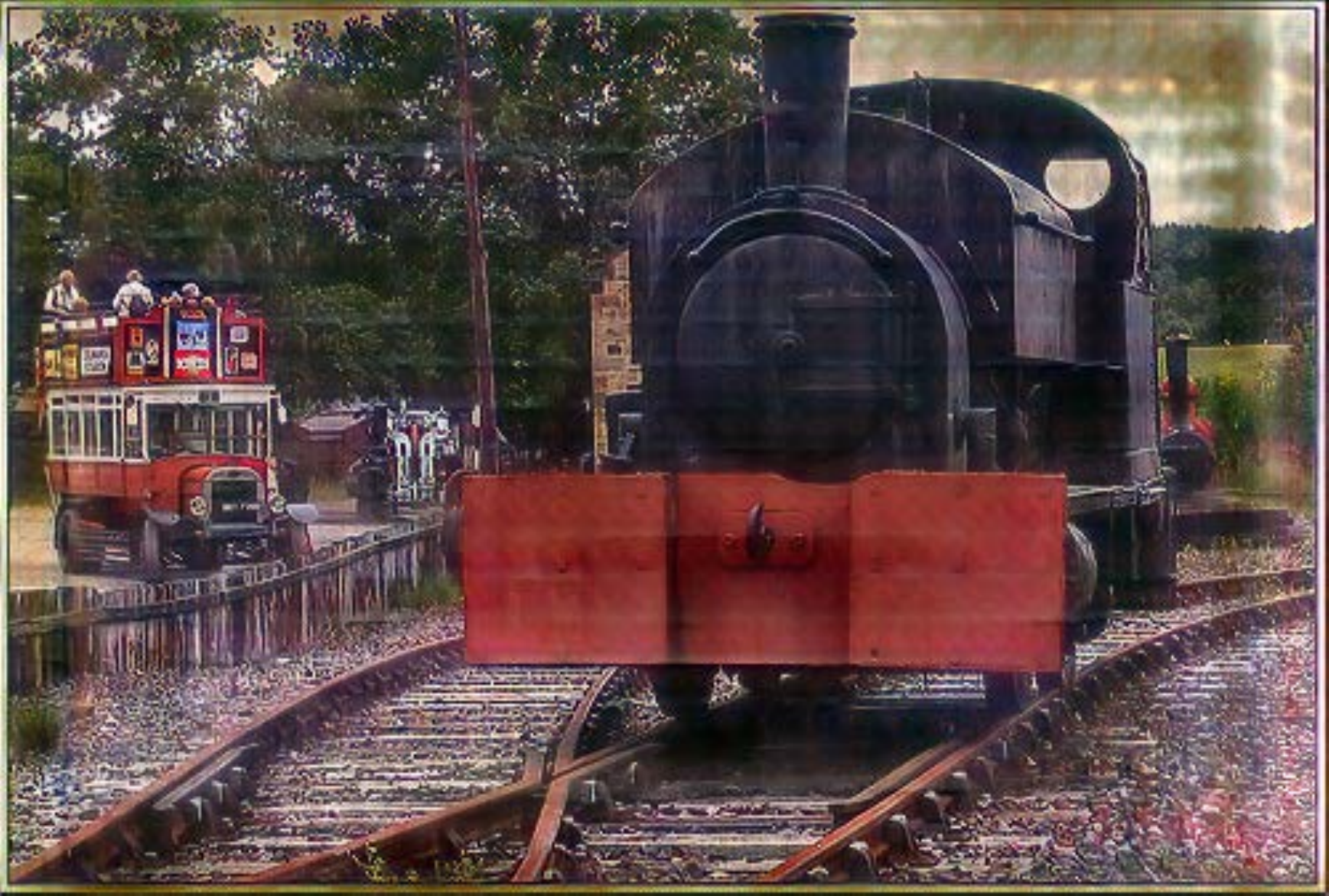}&
			\includegraphics[width=0.156\linewidth]{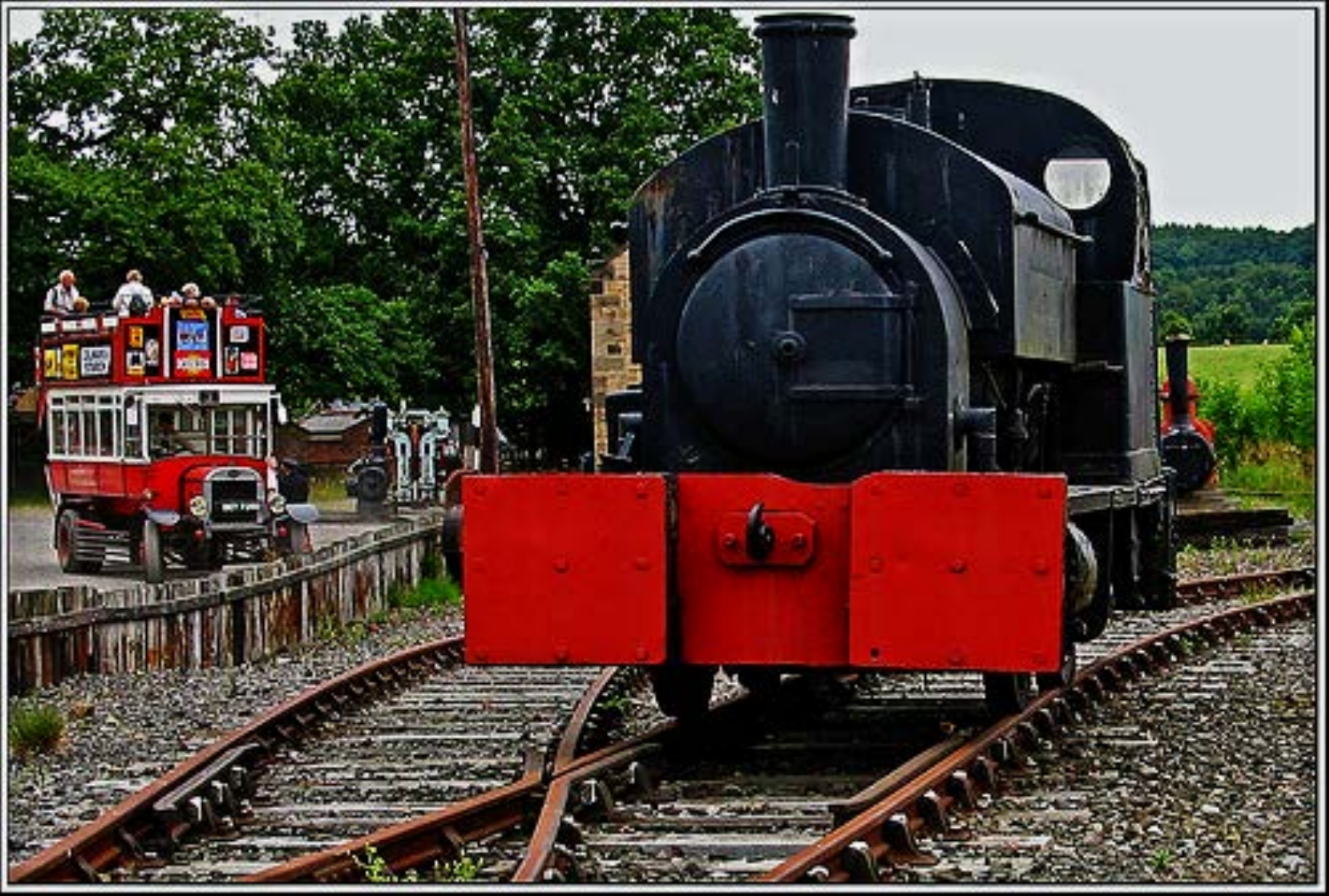}&
			\includegraphics[width=0.156\linewidth]{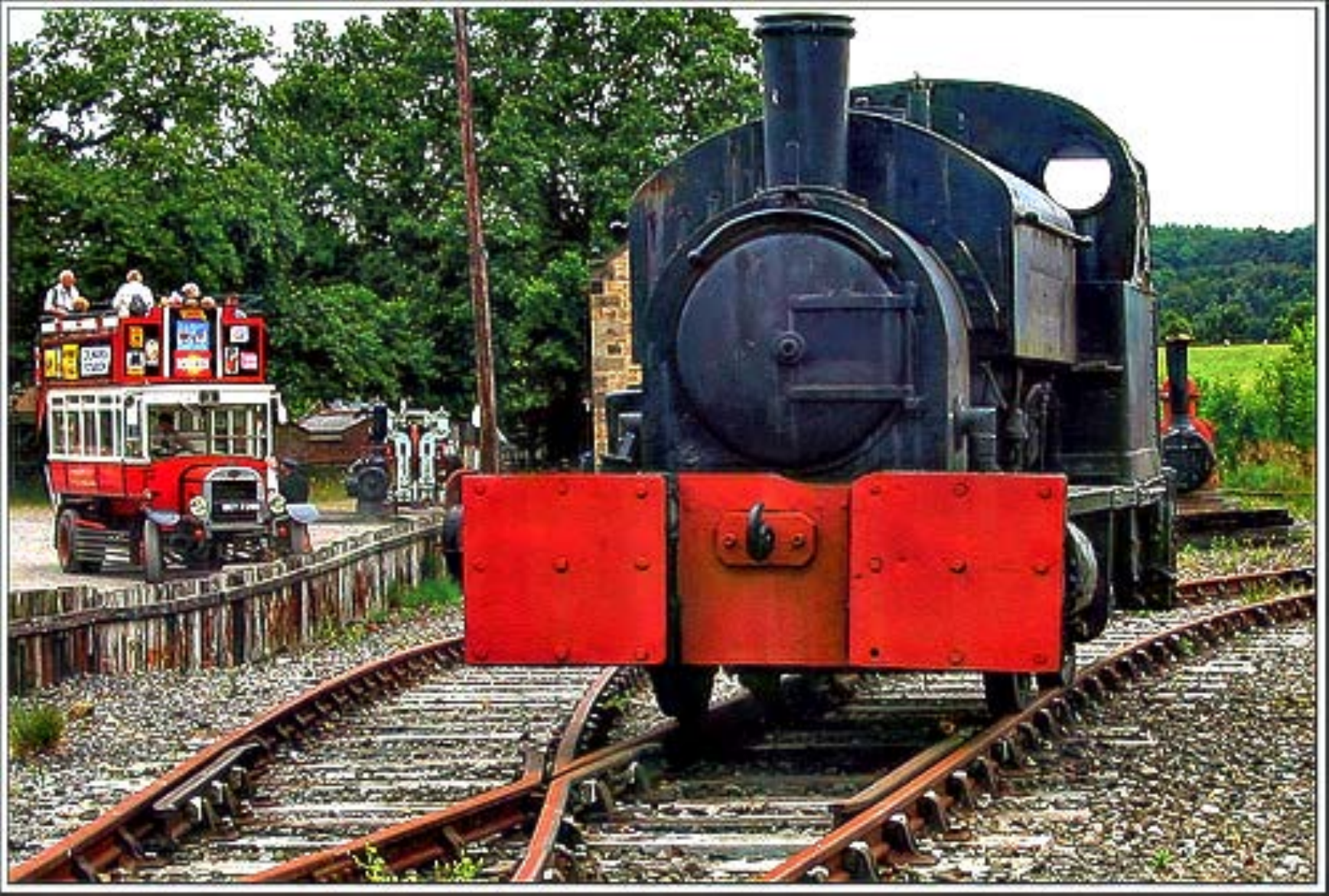}&
			\includegraphics[width=0.156\linewidth]{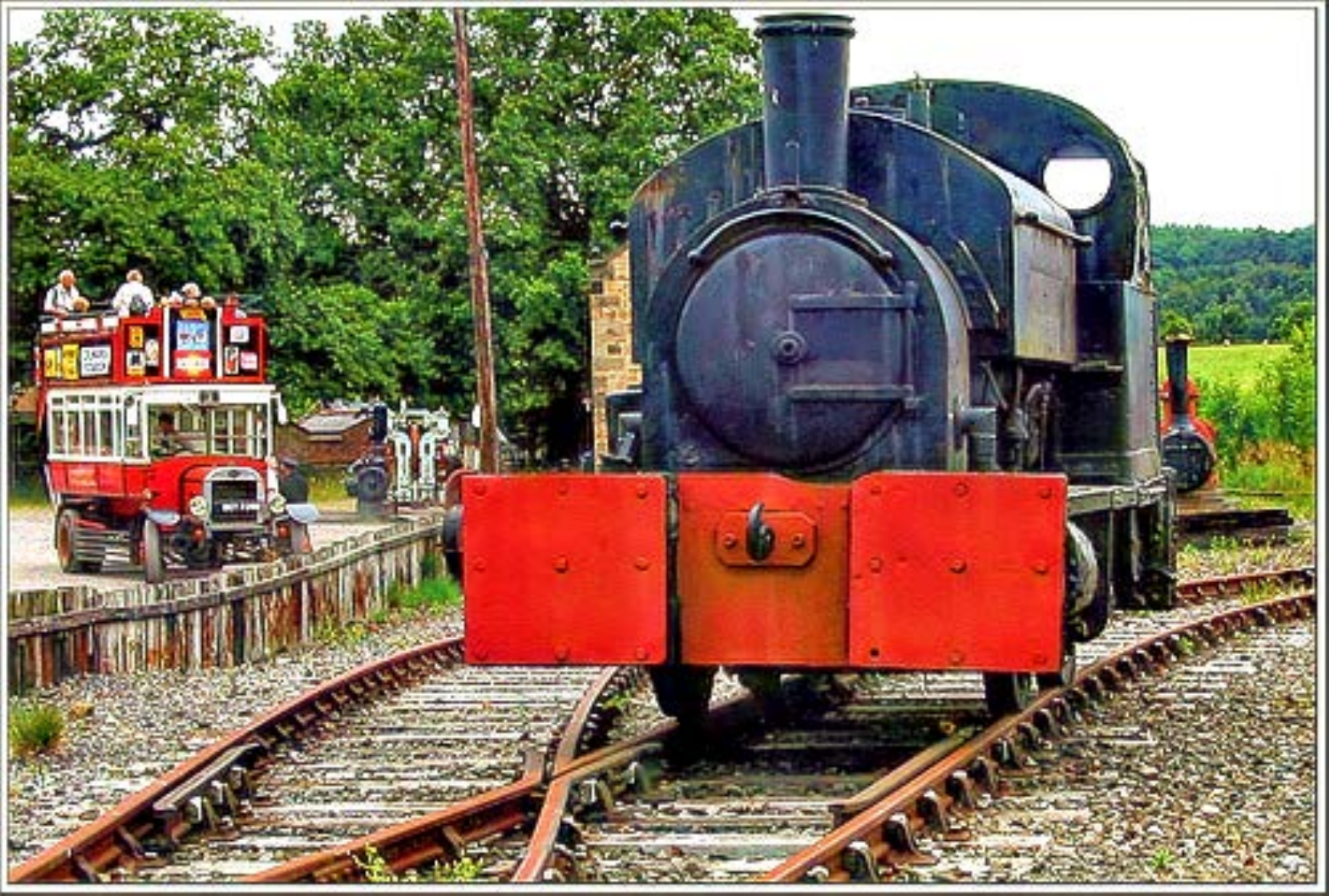}\\
			\footnotesize ZeroDCE&\footnotesize RUAS&\footnotesize UTVNet&\footnotesize SCL&\footnotesize \textbf{BL}&\footnotesize \textbf{RBL}\\
			\multicolumn{6}{c}{\footnotesize (b) Visual comparison among different methods on the VOC dataset~\citep{lv2021attention}}\\
		\end{tabular}
		\caption{Visual results of state-of-the-art methods and our two versions (BL and RBL) on different datasets.}
		\vspace{-1em}
		\label{fig: LSRW_VOC}
	\end{figure*}
	
	\begin{figure*}[t]
		\centering
		\begin{tabular}{c@{\extracolsep{0.3em}}c@{\extracolsep{0.3em}}c@{\extracolsep{0.3em}}c@{\extracolsep{0.3em}}c@{\extracolsep{0.3em}}c@{\extracolsep{0.3em}}c}			
			\includegraphics[width=0.1354\linewidth]{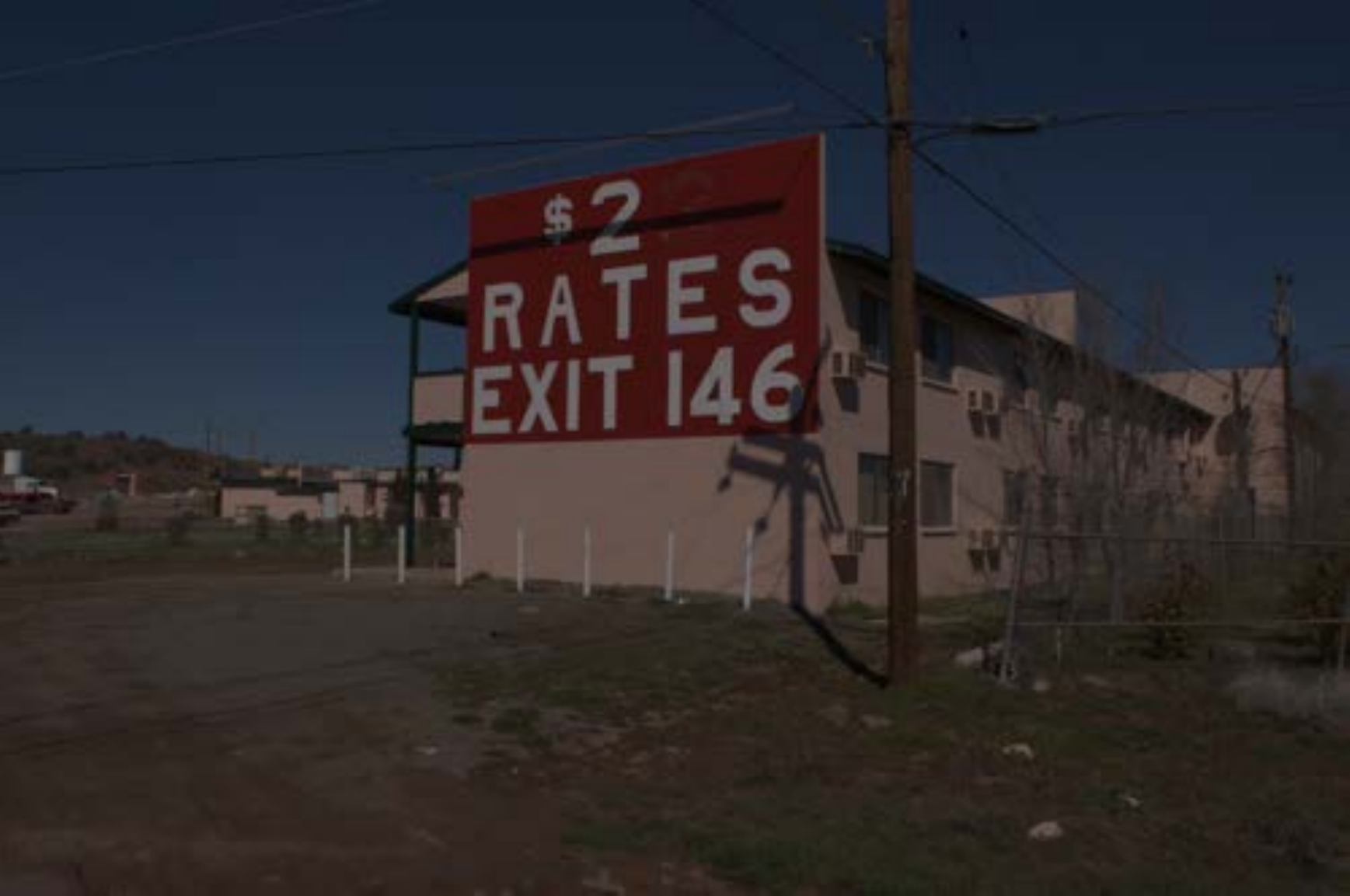}&
			\includegraphics[width=0.1354\linewidth]{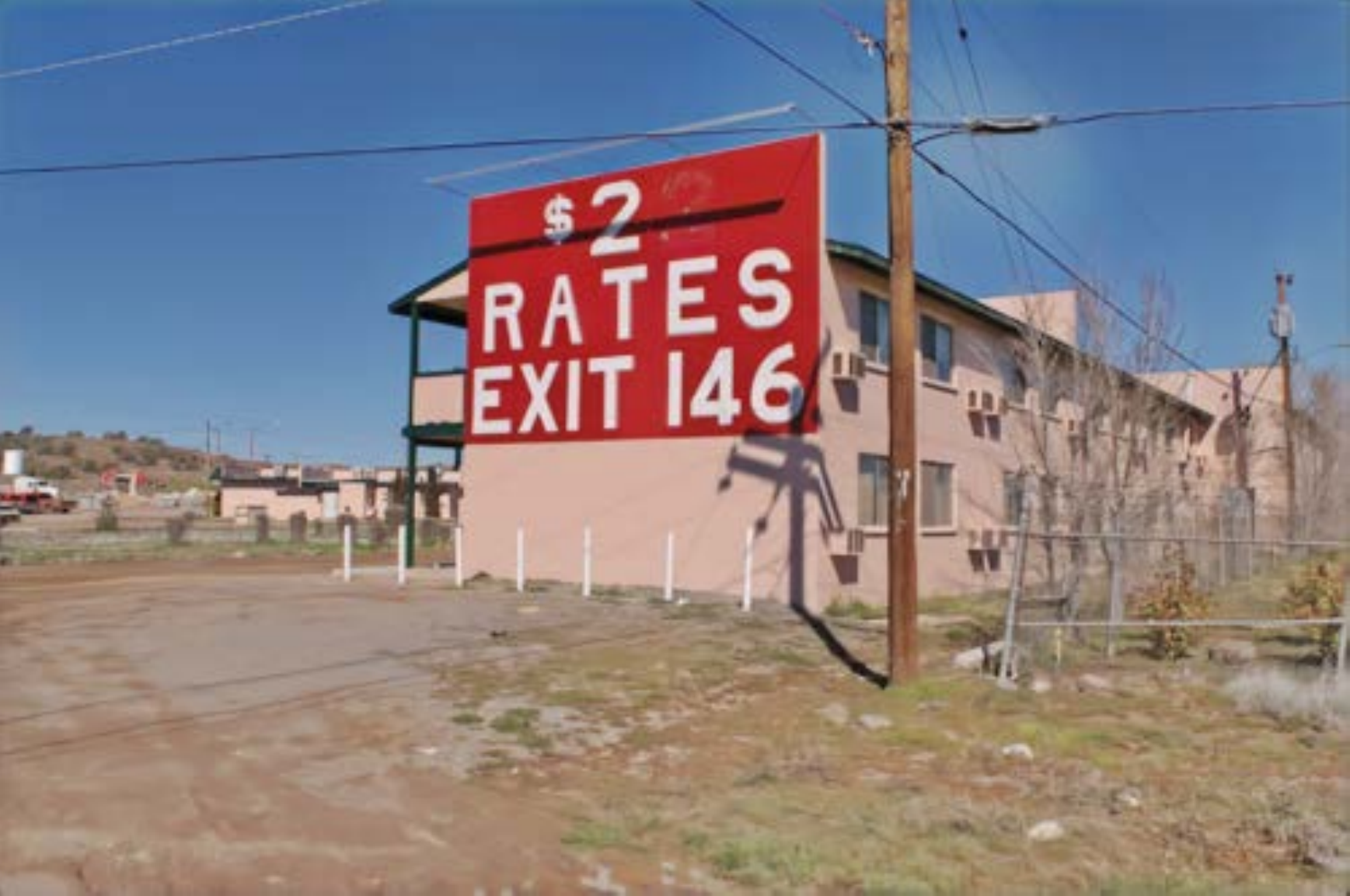}&
			\includegraphics[width=0.1354\linewidth]{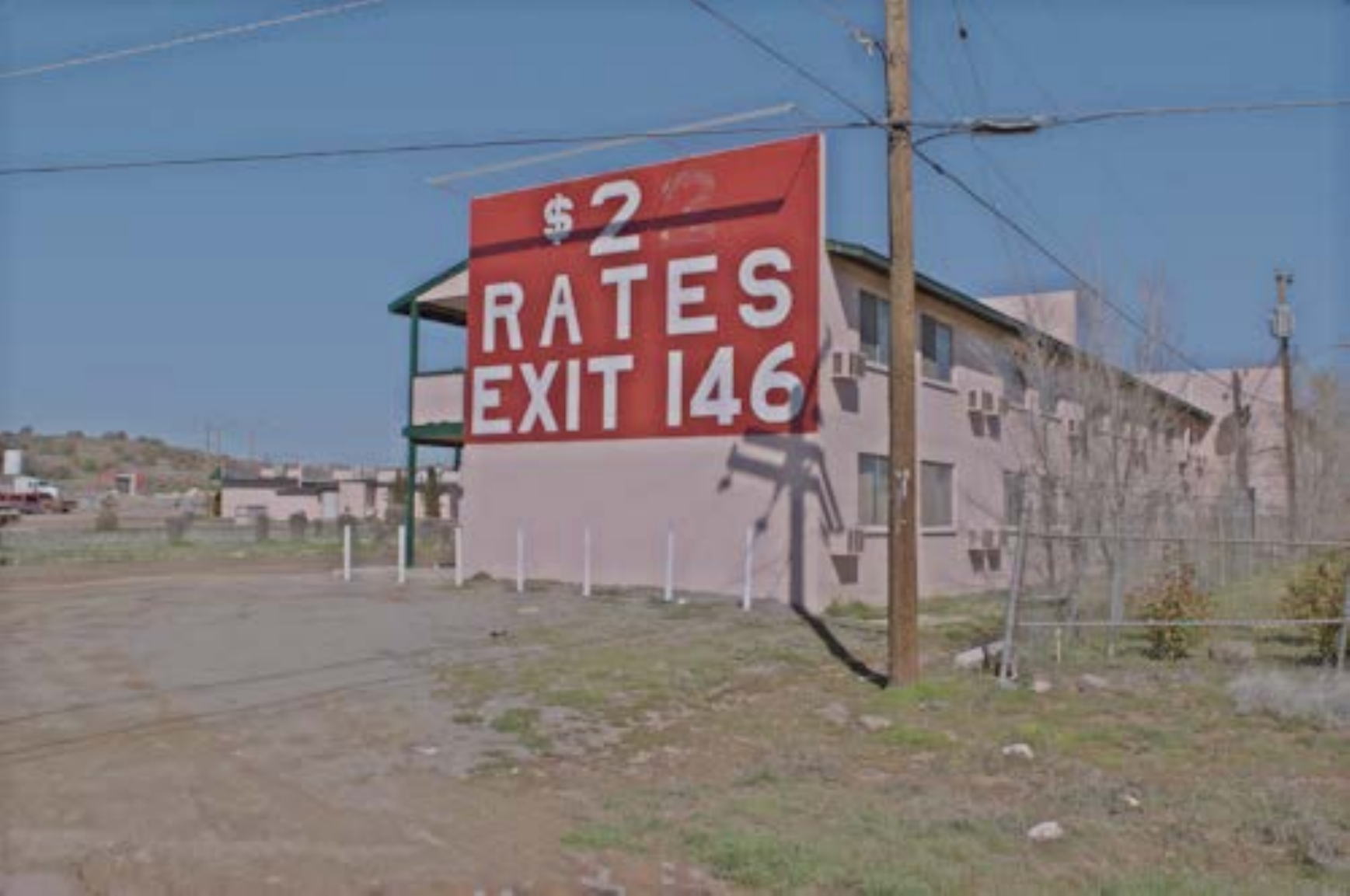}&
			\includegraphics[width=0.1354\linewidth]{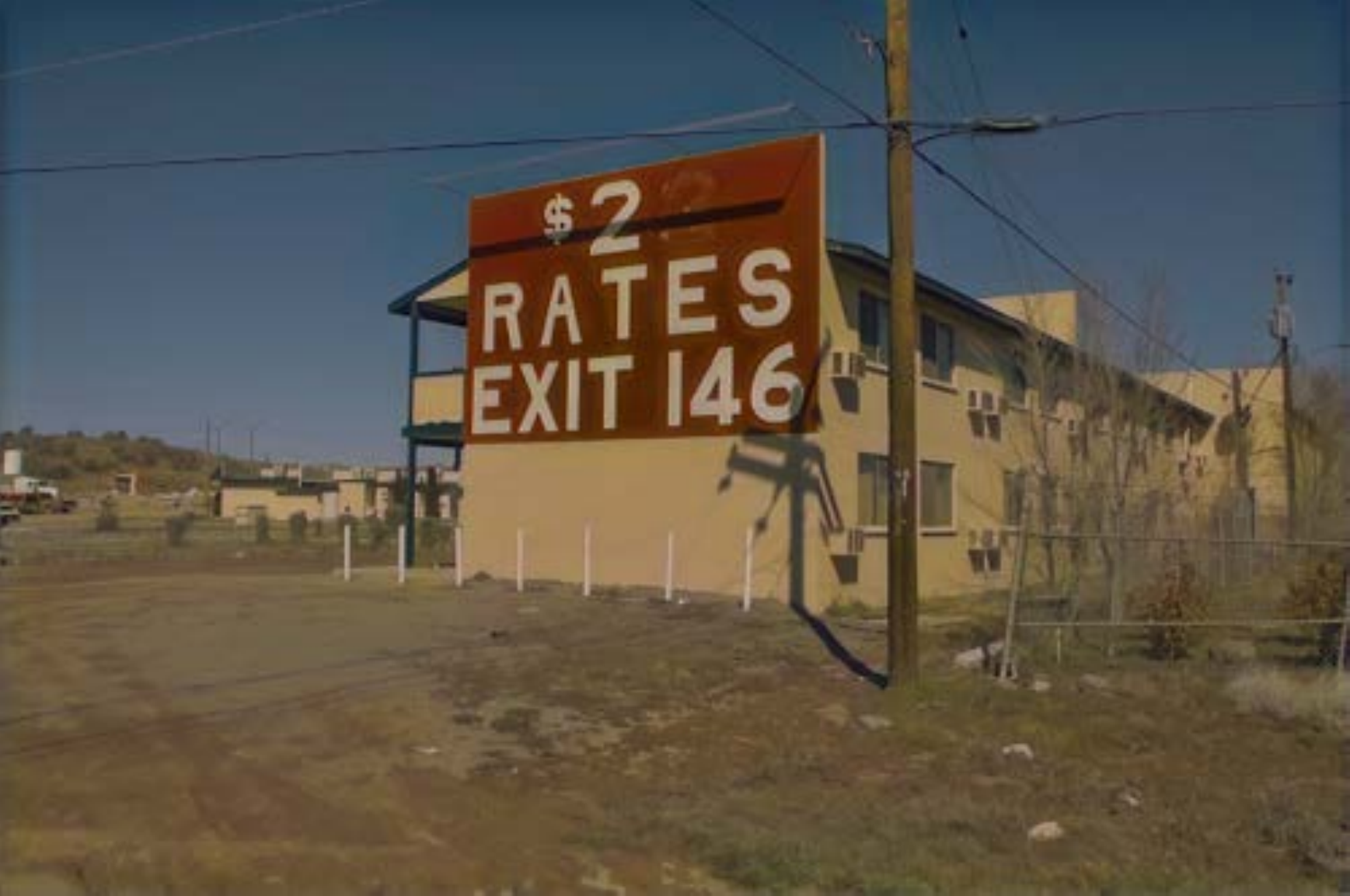}&
			\includegraphics[width=0.1354\linewidth]{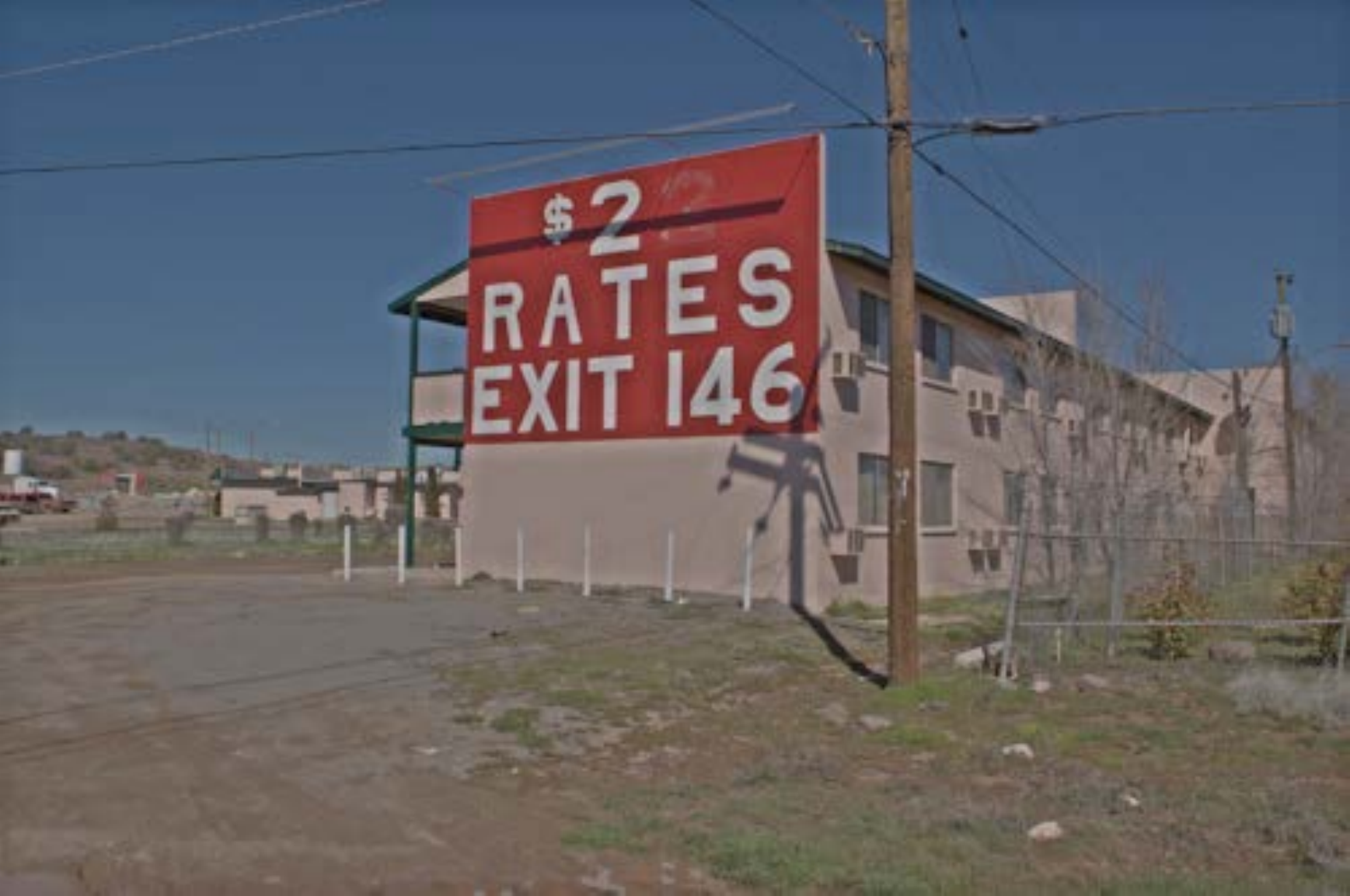}&
			\includegraphics[width=0.1354\linewidth]{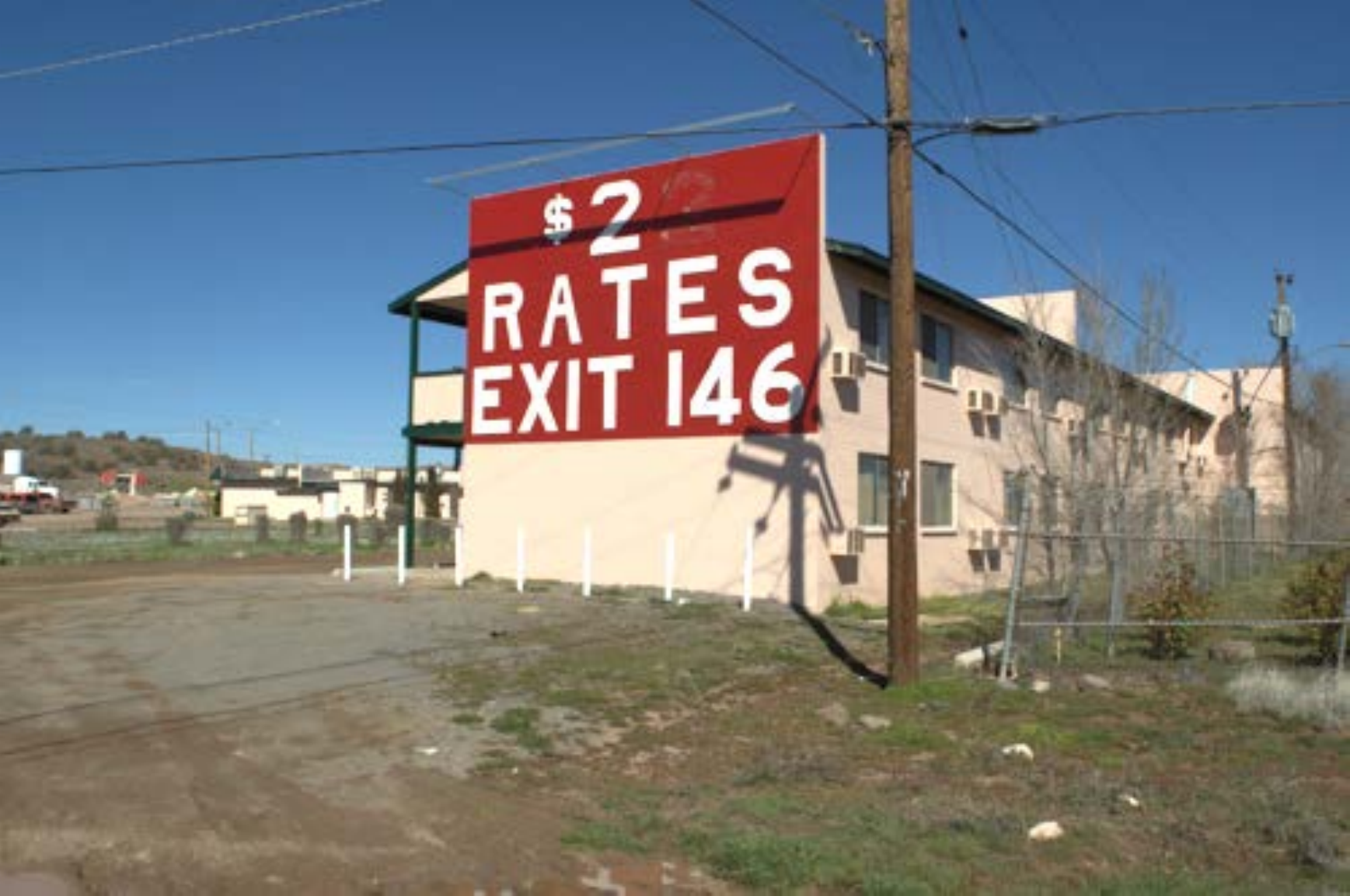}&
			\includegraphics[width=0.1354\linewidth]{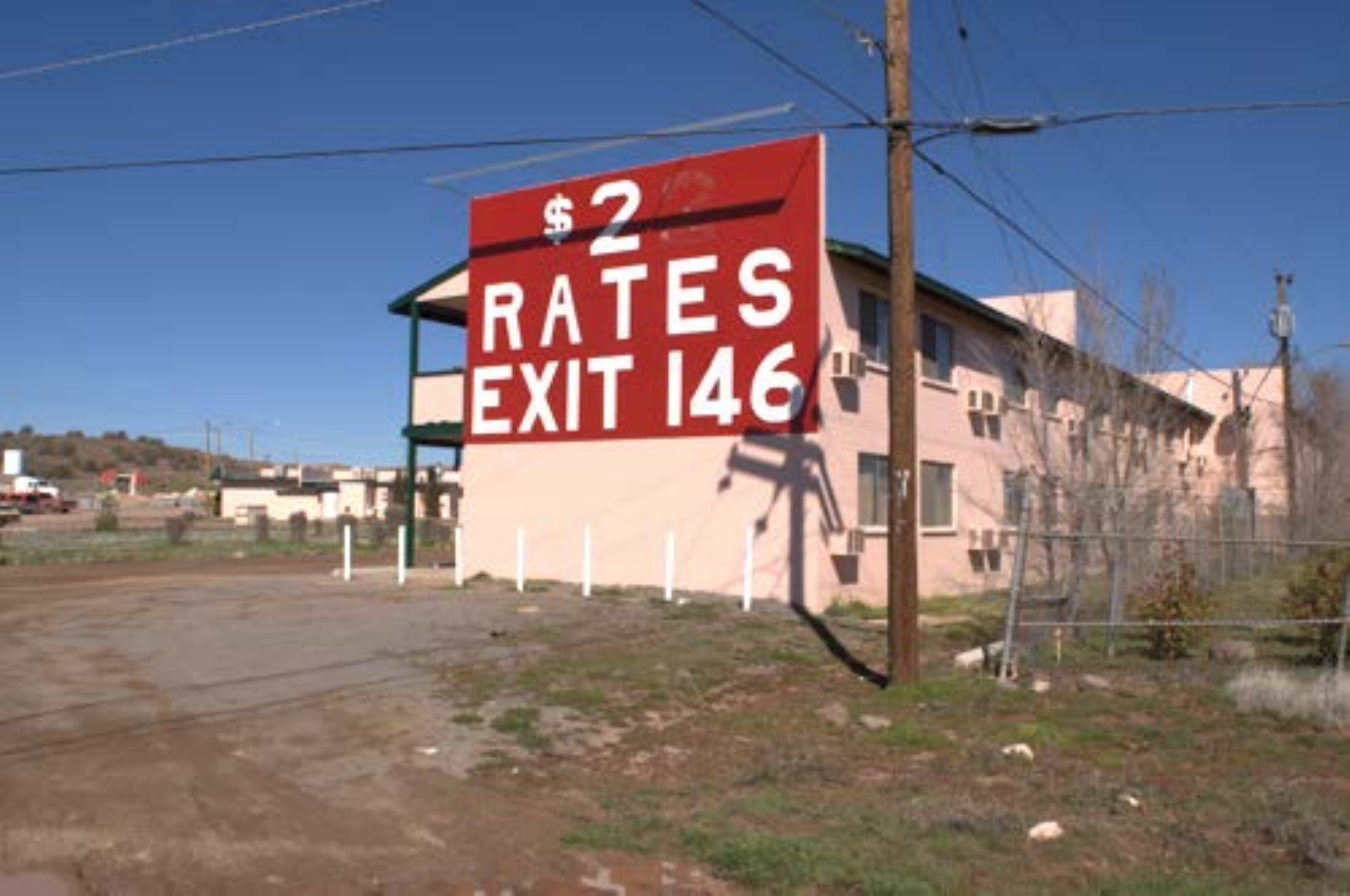}\\
			\includegraphics[width=0.1354\linewidth]{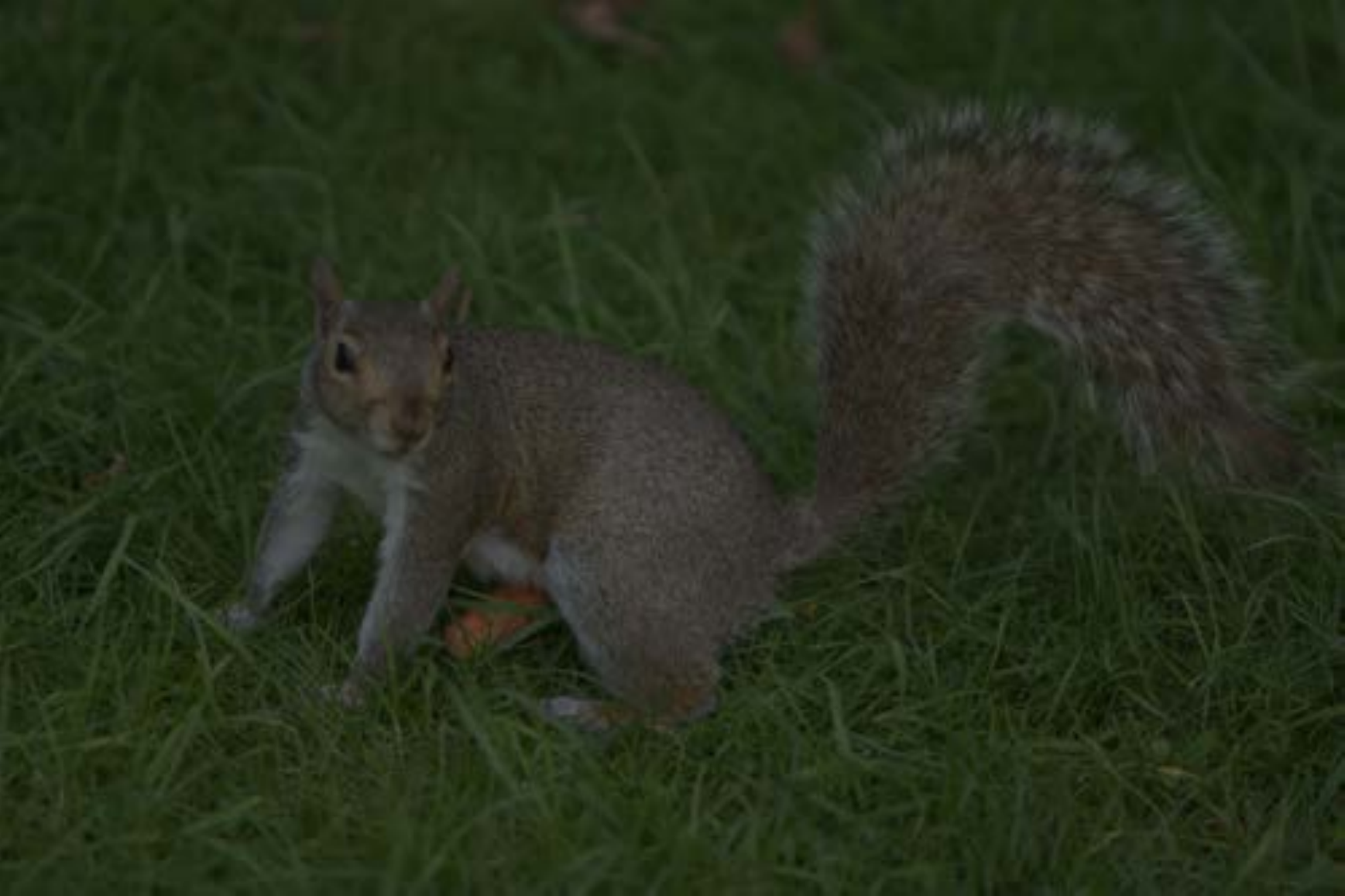}&
			\includegraphics[width=0.1354\linewidth]{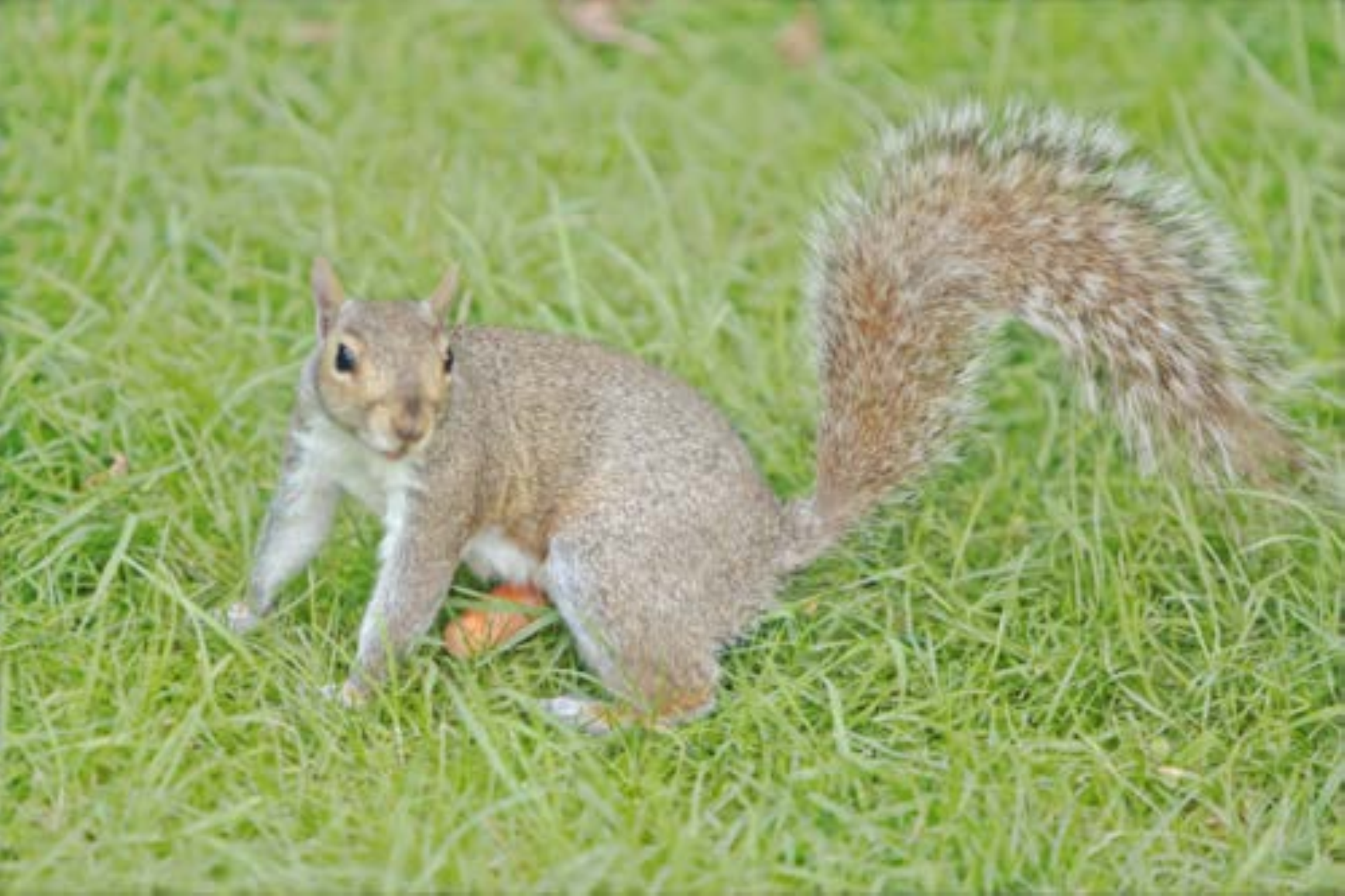}&
			\includegraphics[width=0.1354\linewidth]{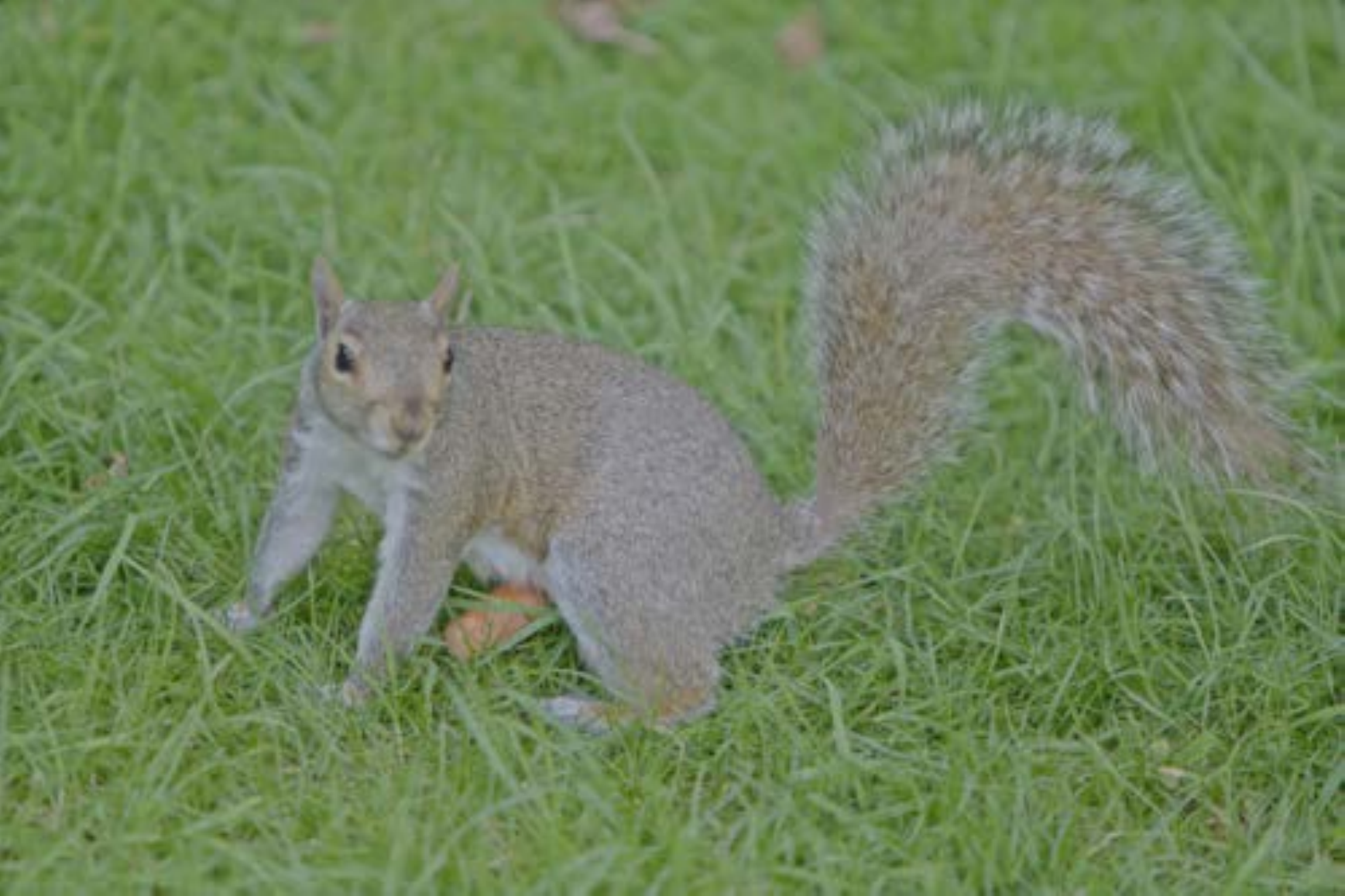}&
			\includegraphics[width=0.1354\linewidth]{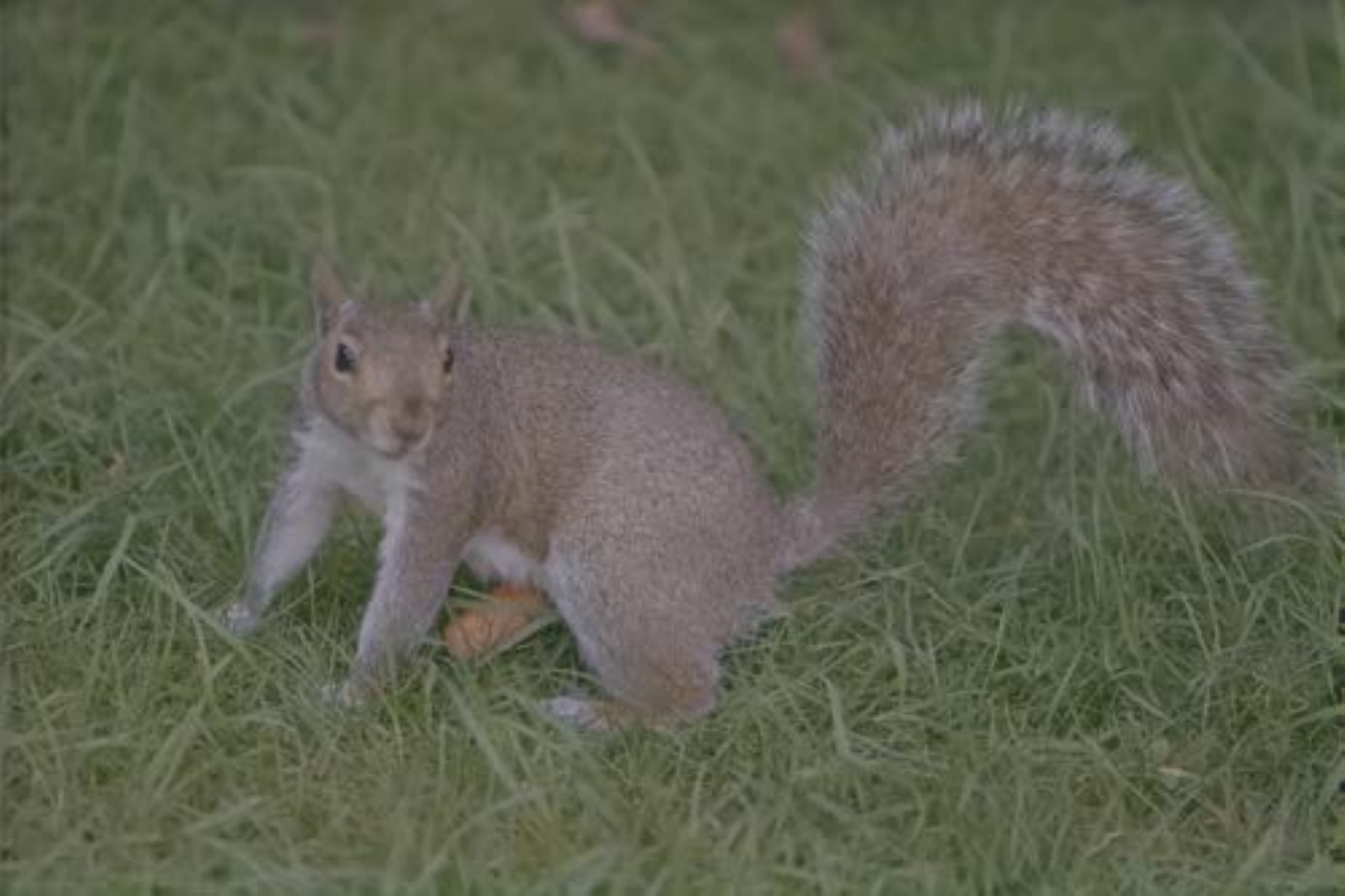}&
			\includegraphics[width=0.1354\linewidth]{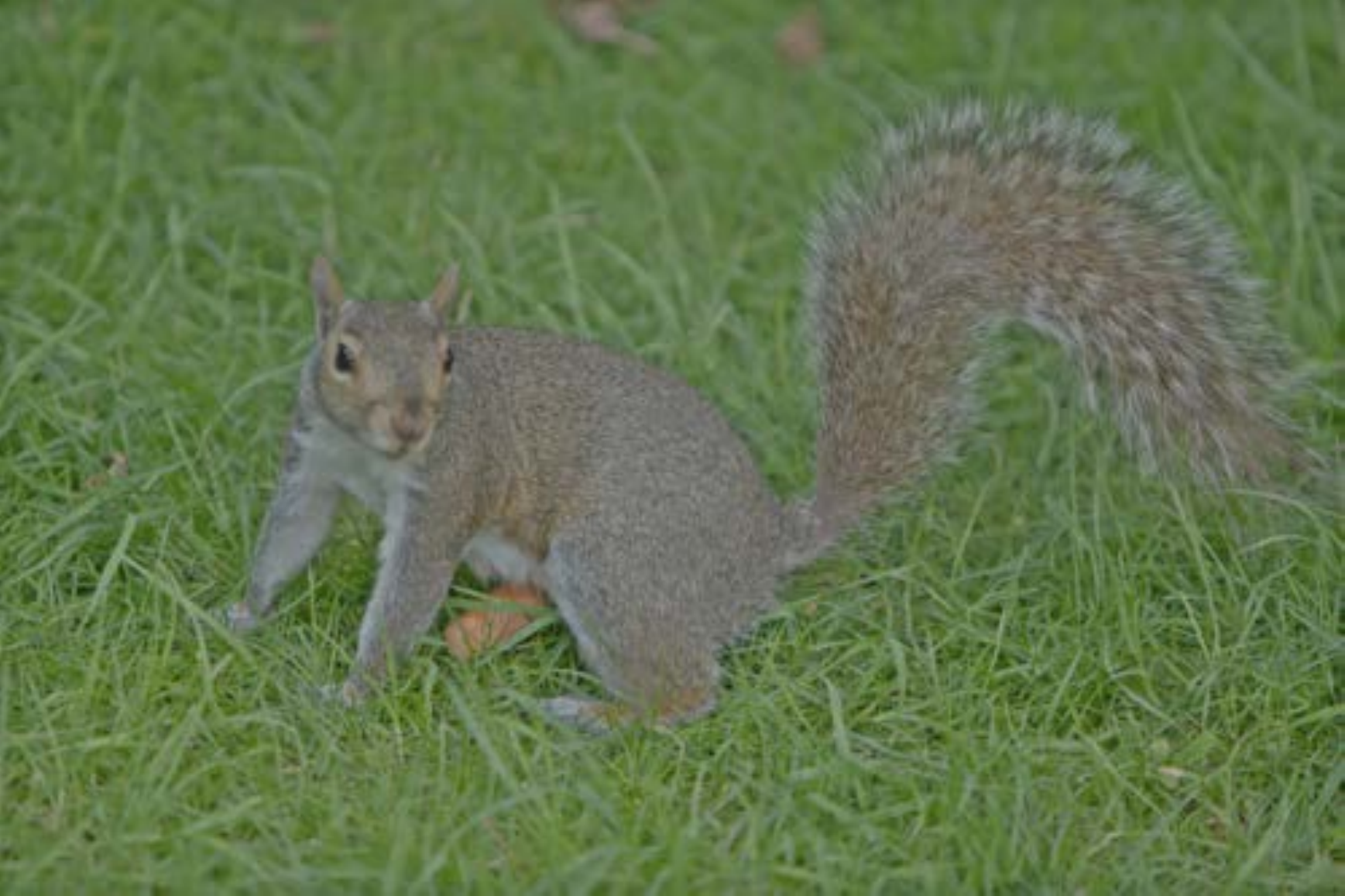}&
			\includegraphics[width=0.1354\linewidth]{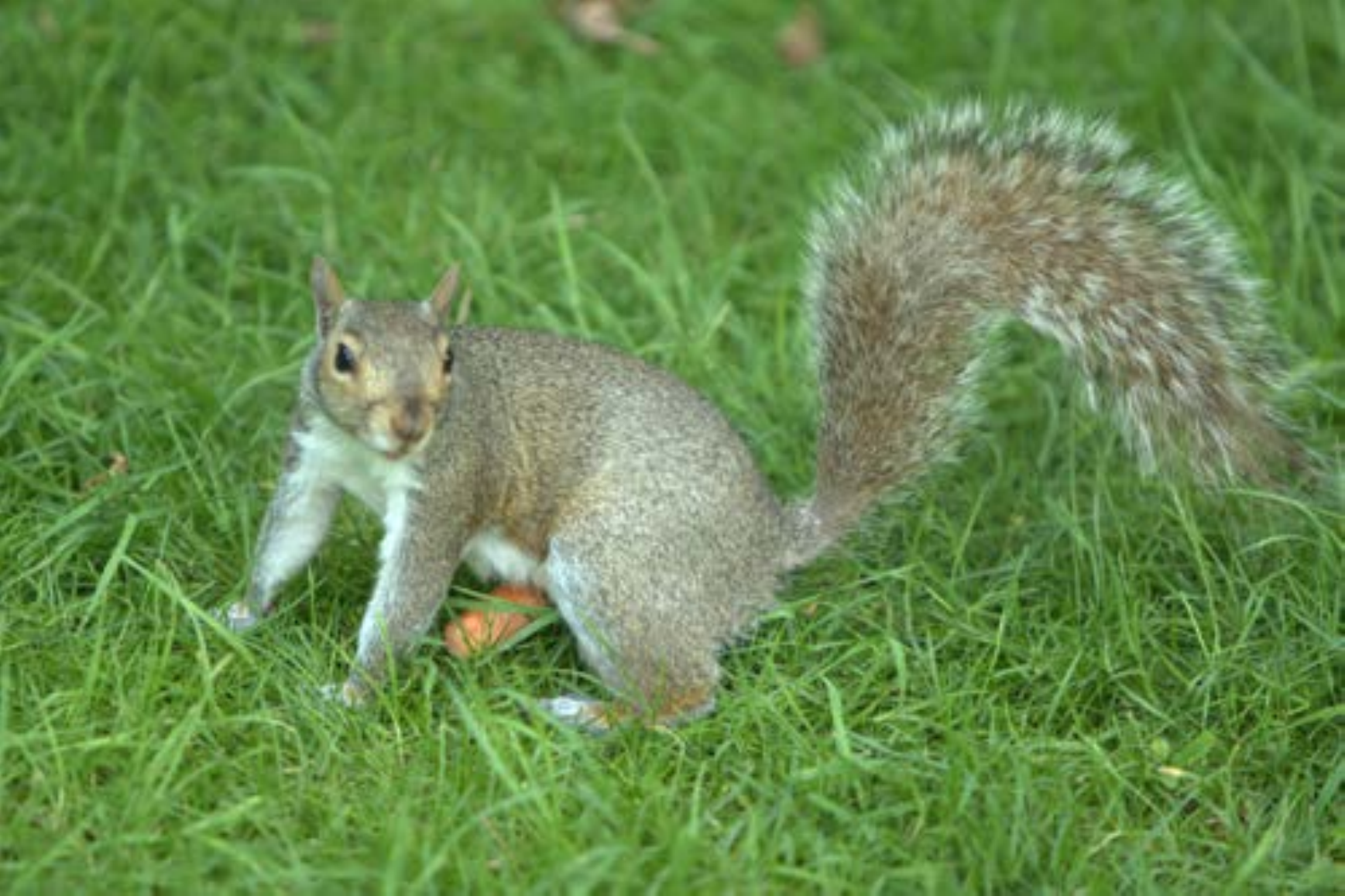}&
			\includegraphics[width=0.1354\linewidth]{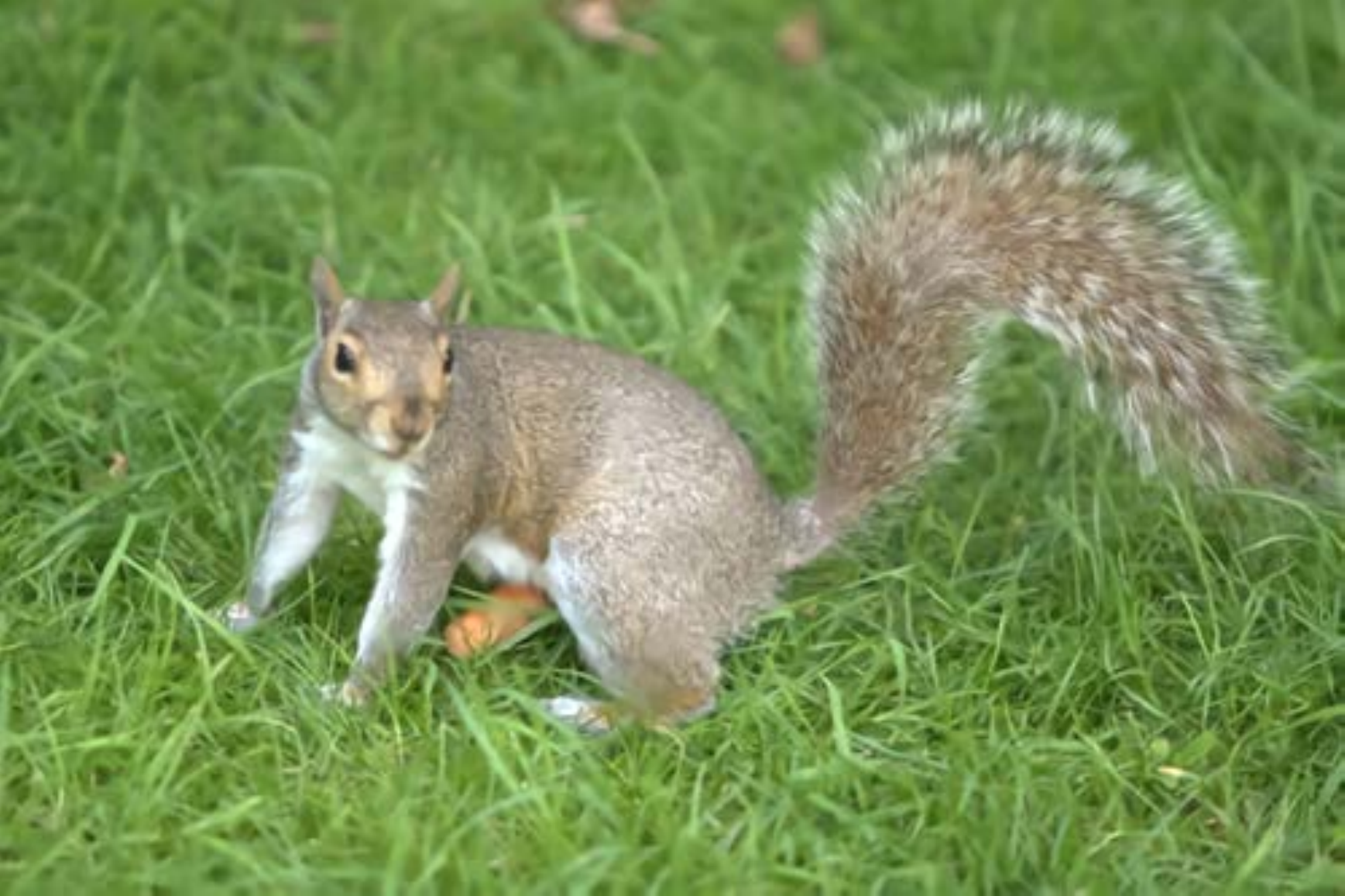}\\
			\includegraphics[width=0.1354\linewidth]{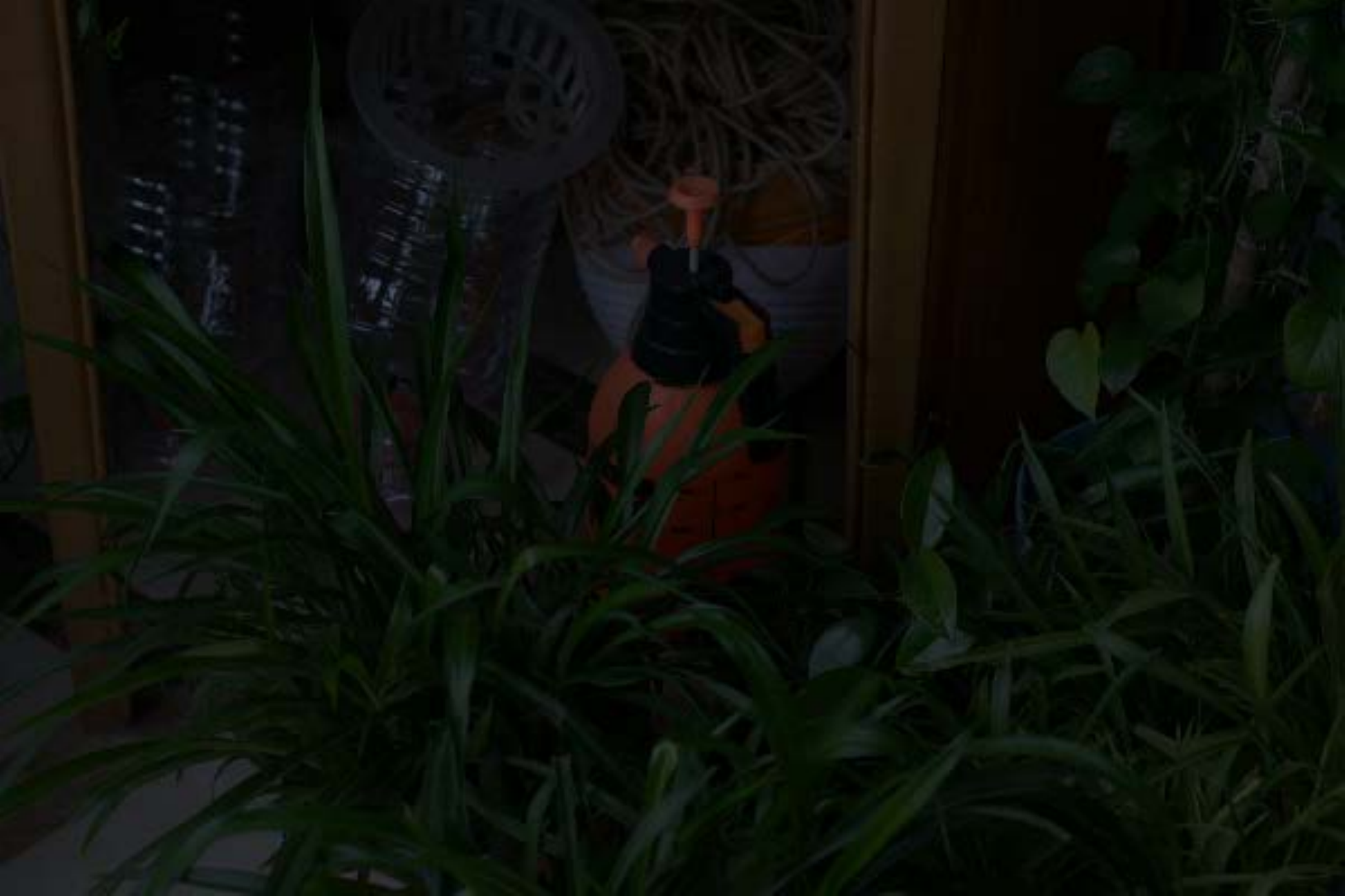}&
			\includegraphics[width=0.1354\linewidth]{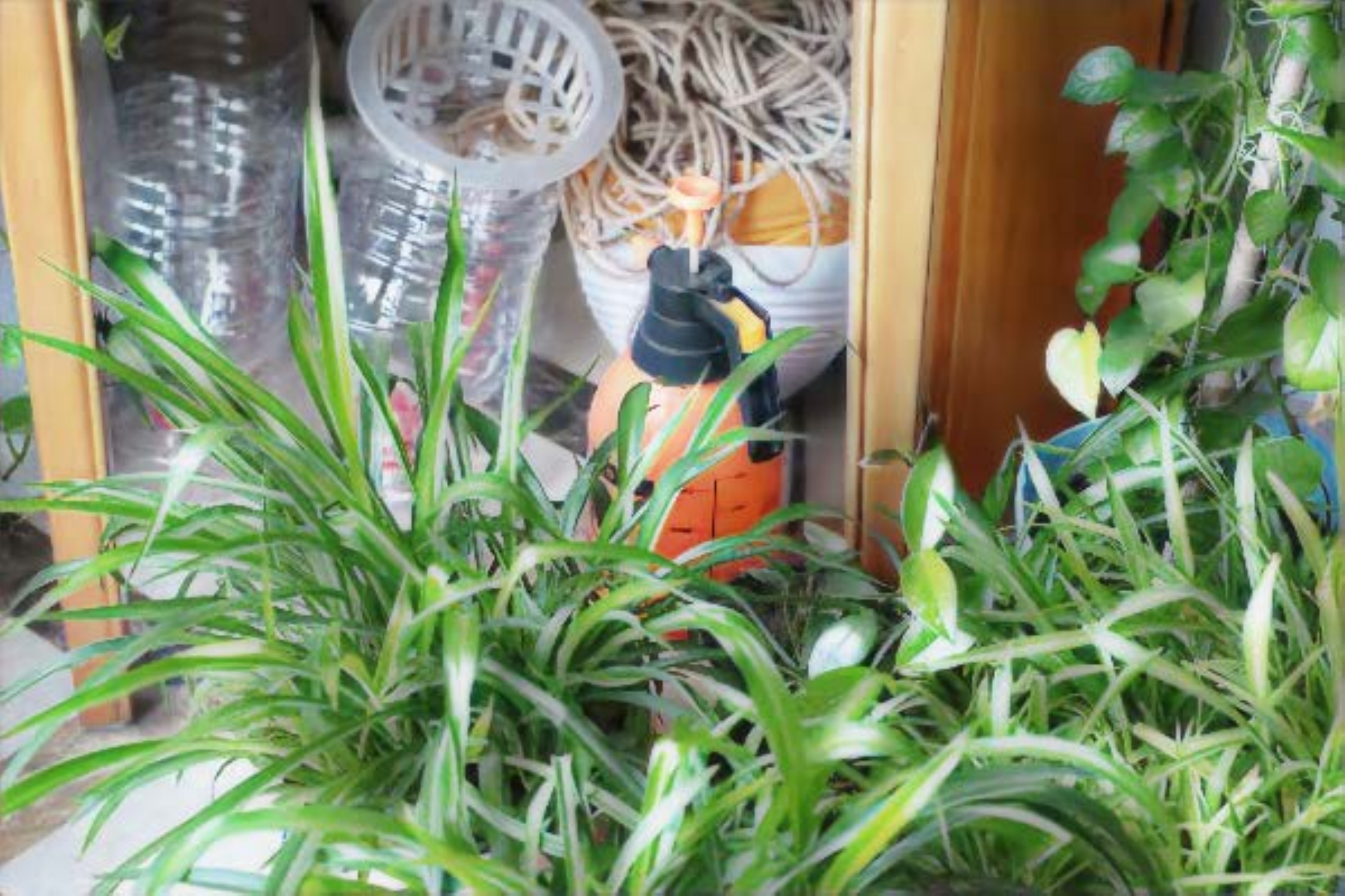}&
			\includegraphics[width=0.1354\linewidth]{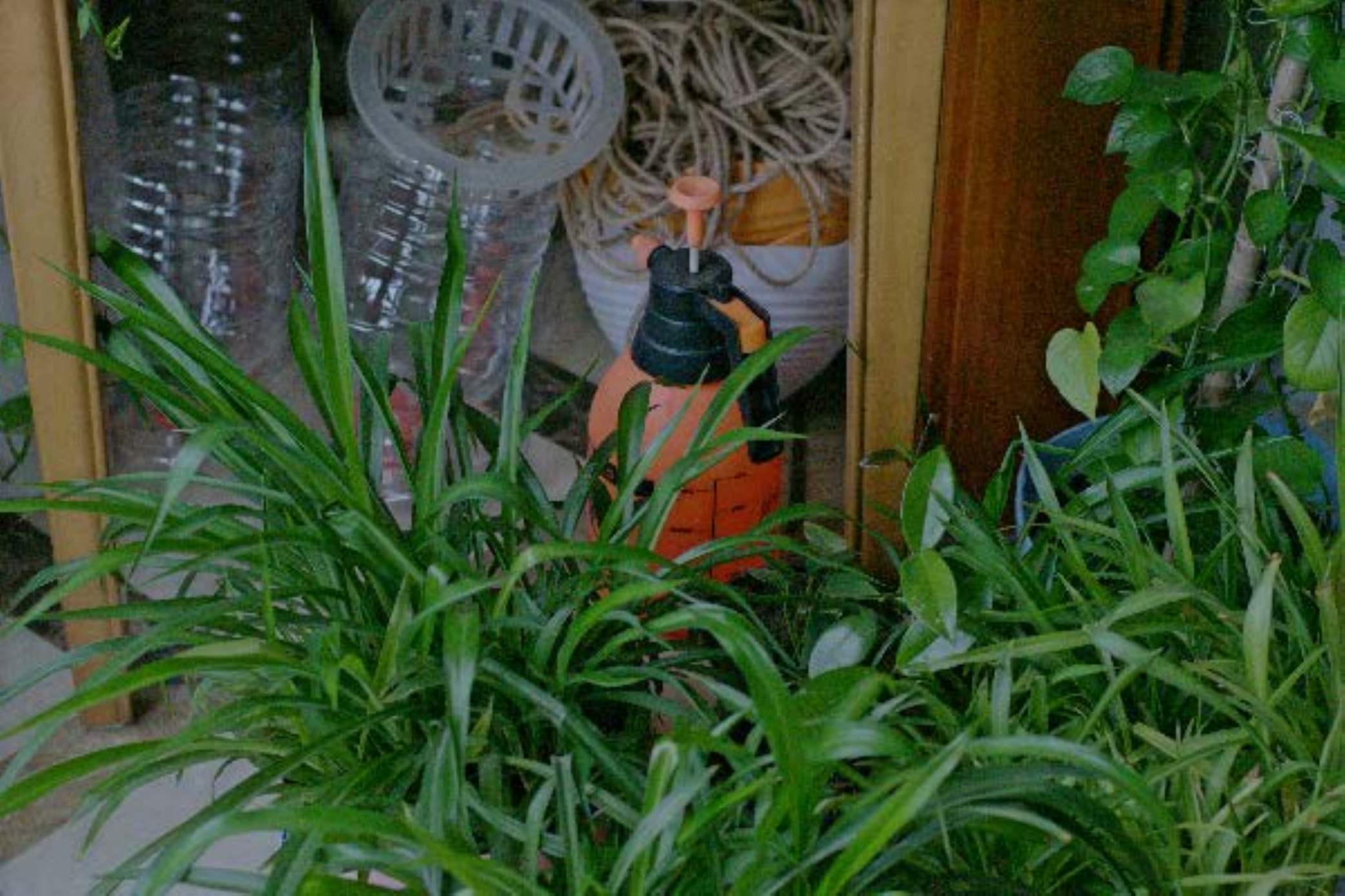}&
			\includegraphics[width=0.1354\linewidth]{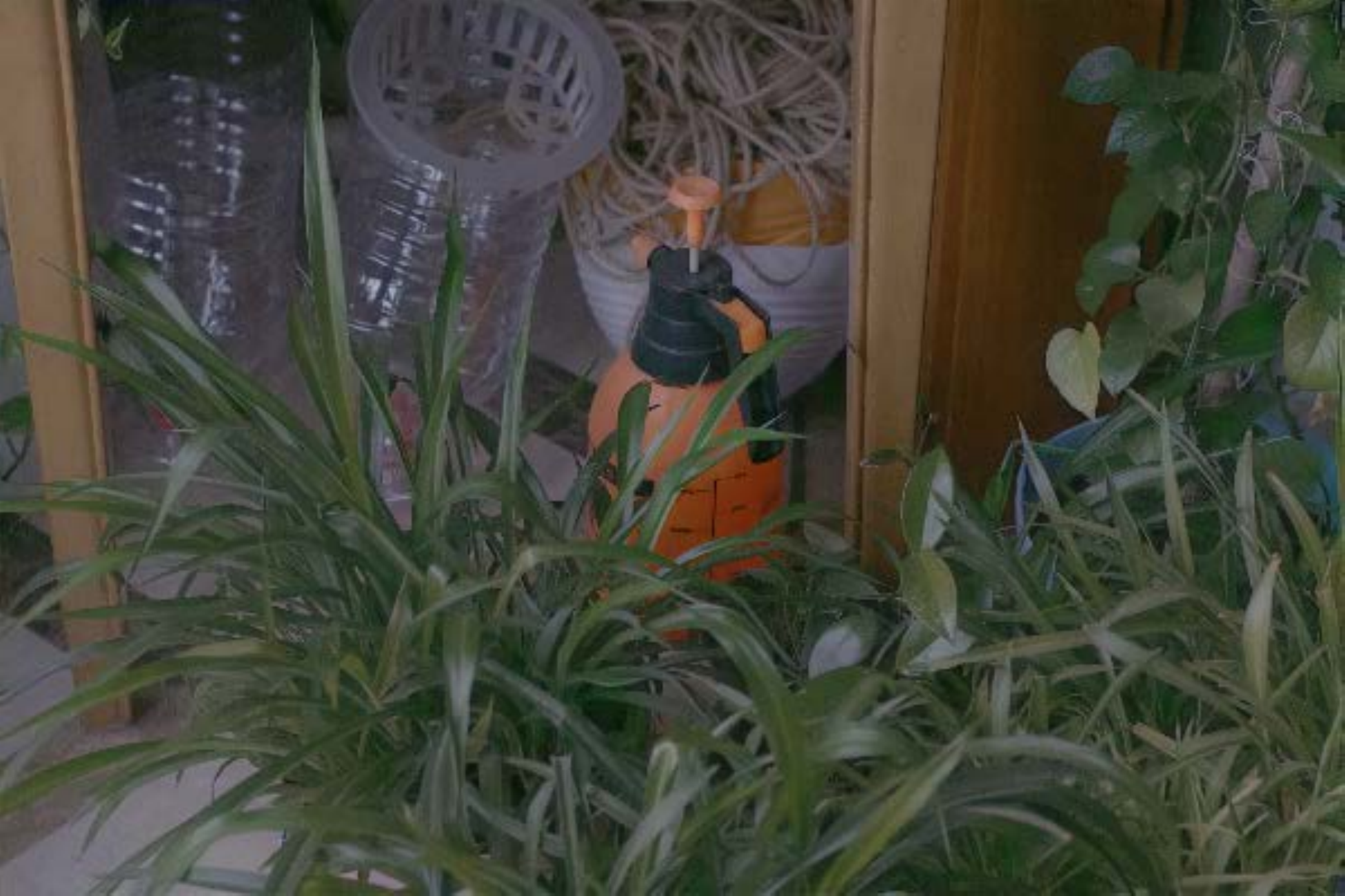}&
			\includegraphics[width=0.1354\linewidth]{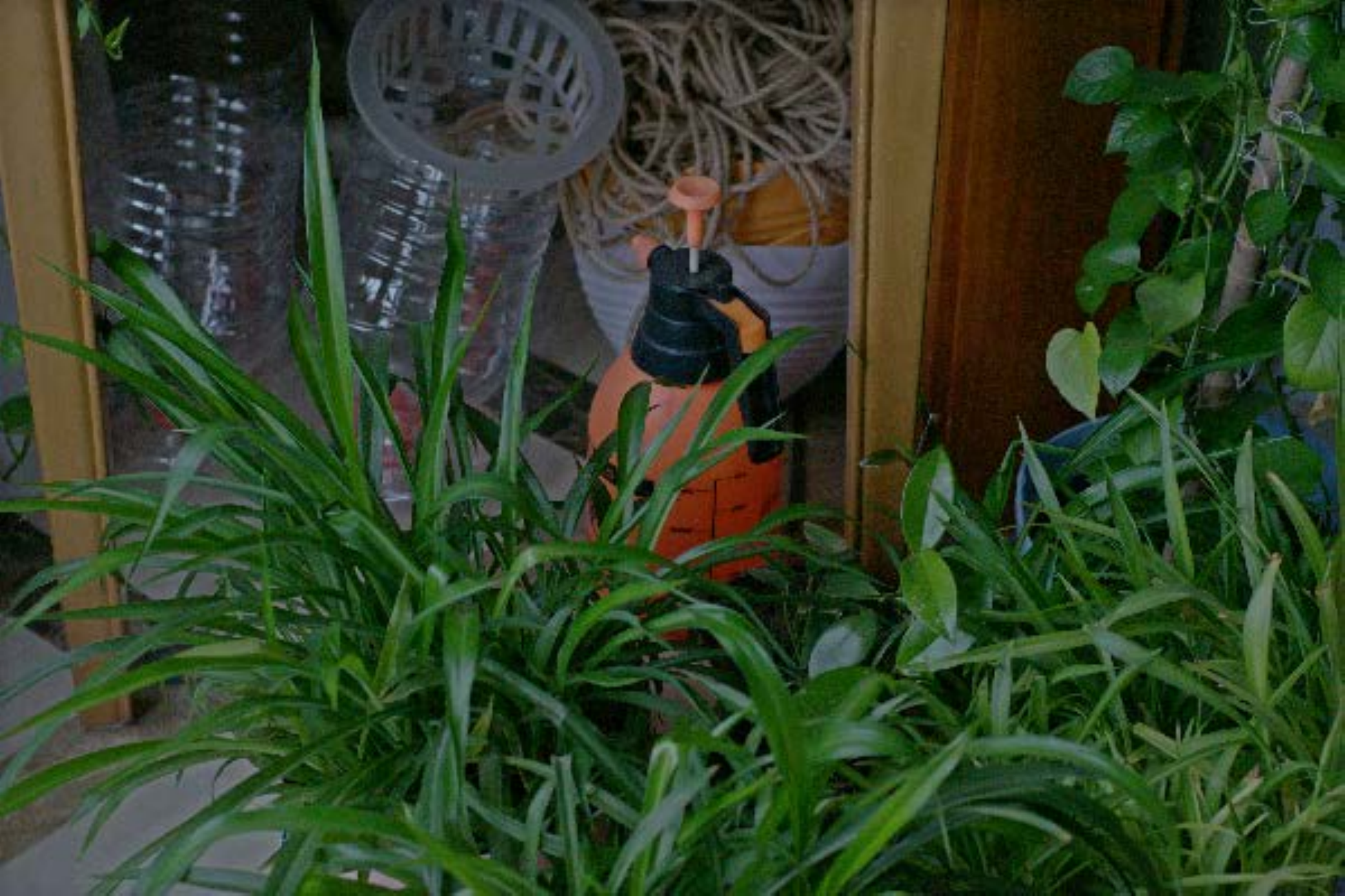}&
			\includegraphics[width=0.1354\linewidth]{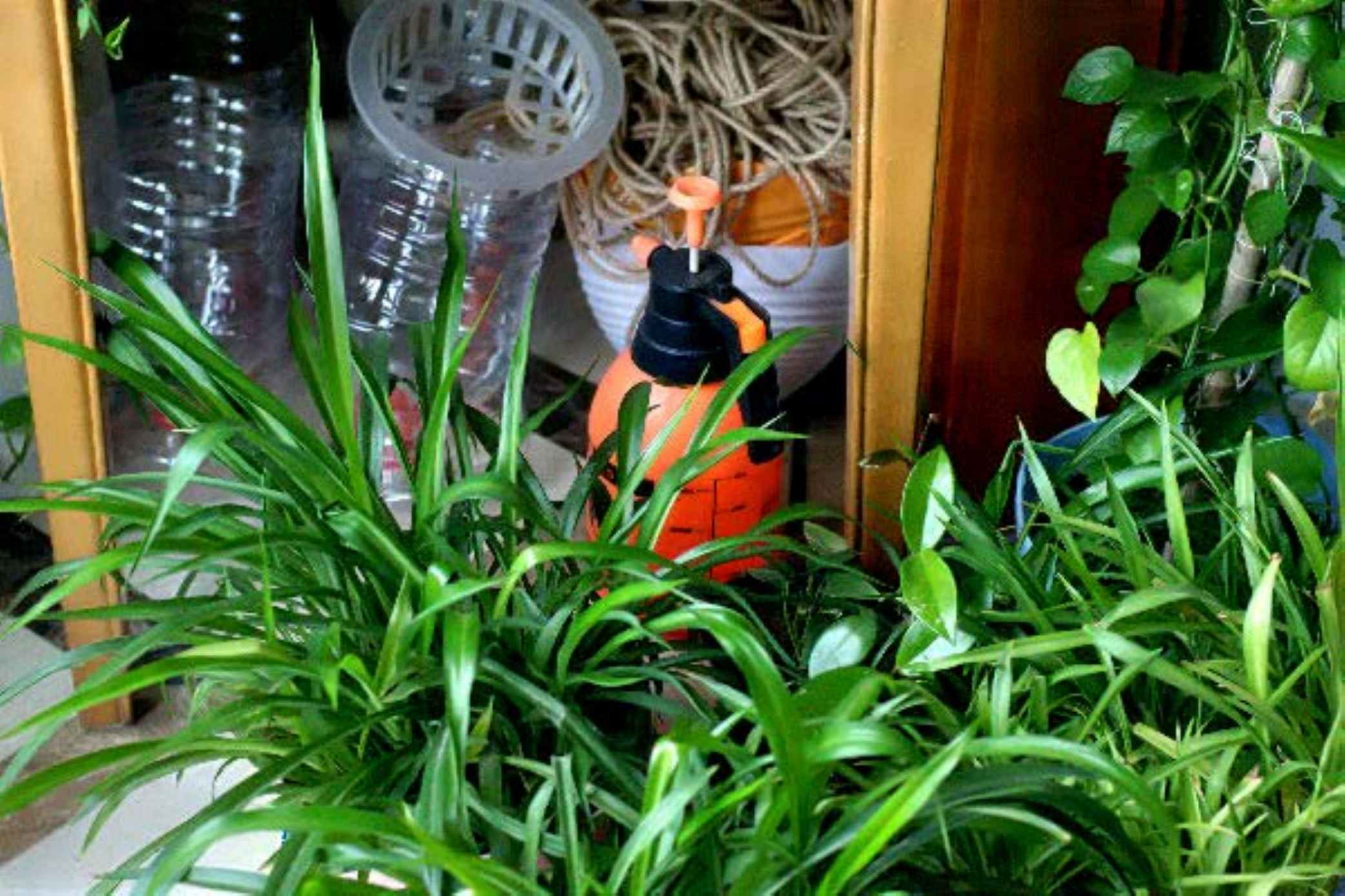}&
			\includegraphics[width=0.1354\linewidth]{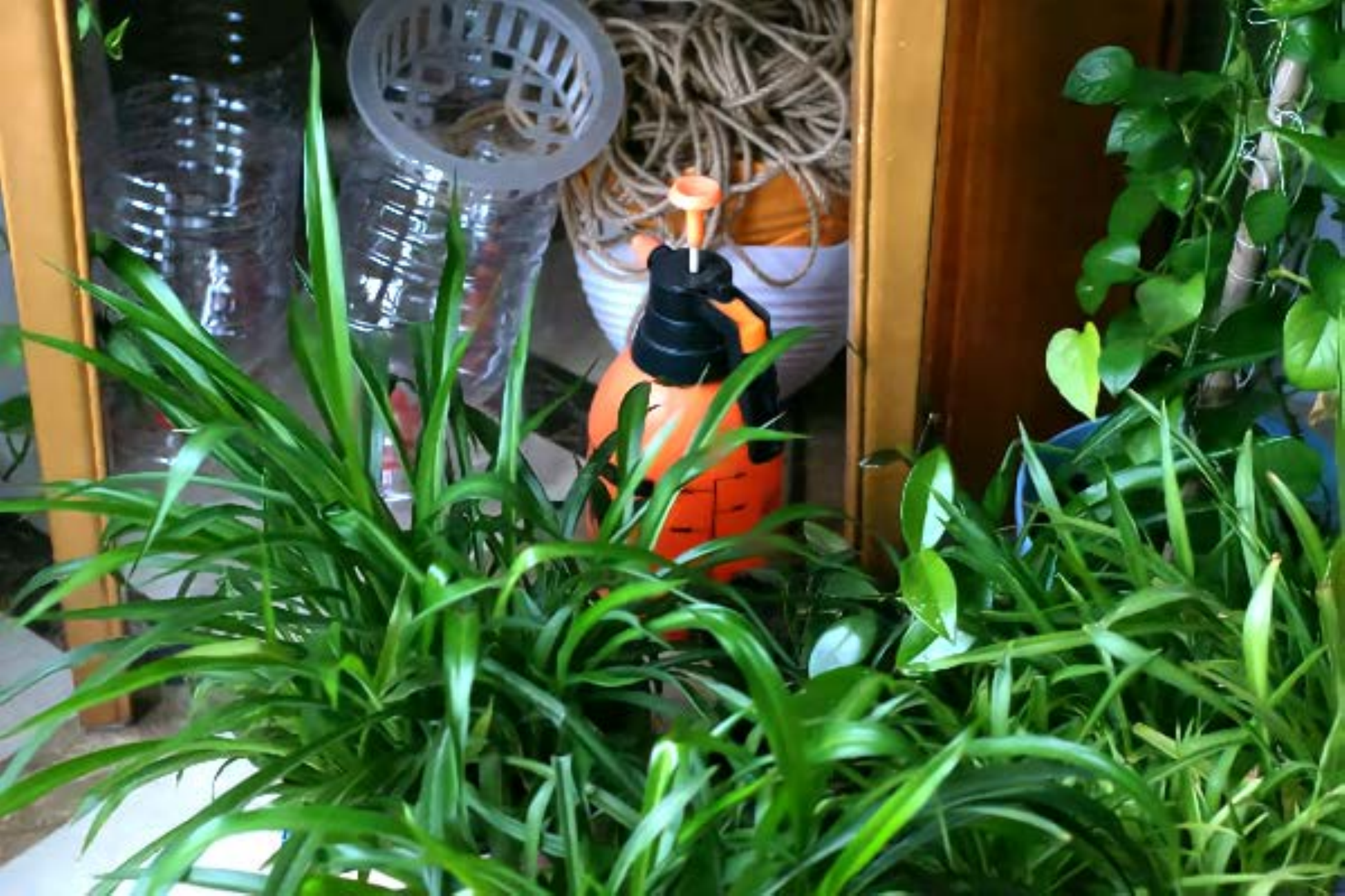}\\
			\includegraphics[width=0.1354\linewidth]{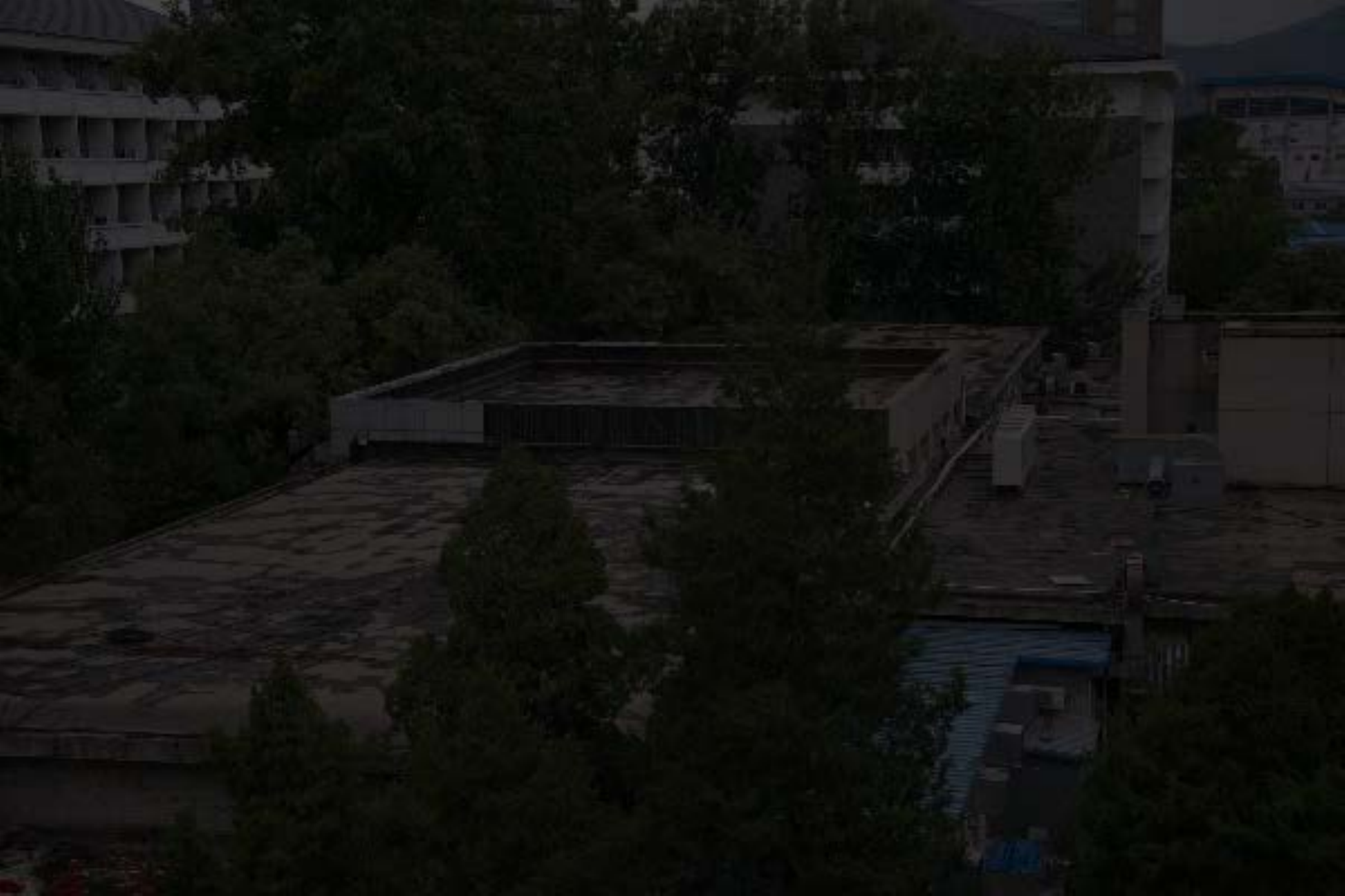}&
			\includegraphics[width=0.1354\linewidth]{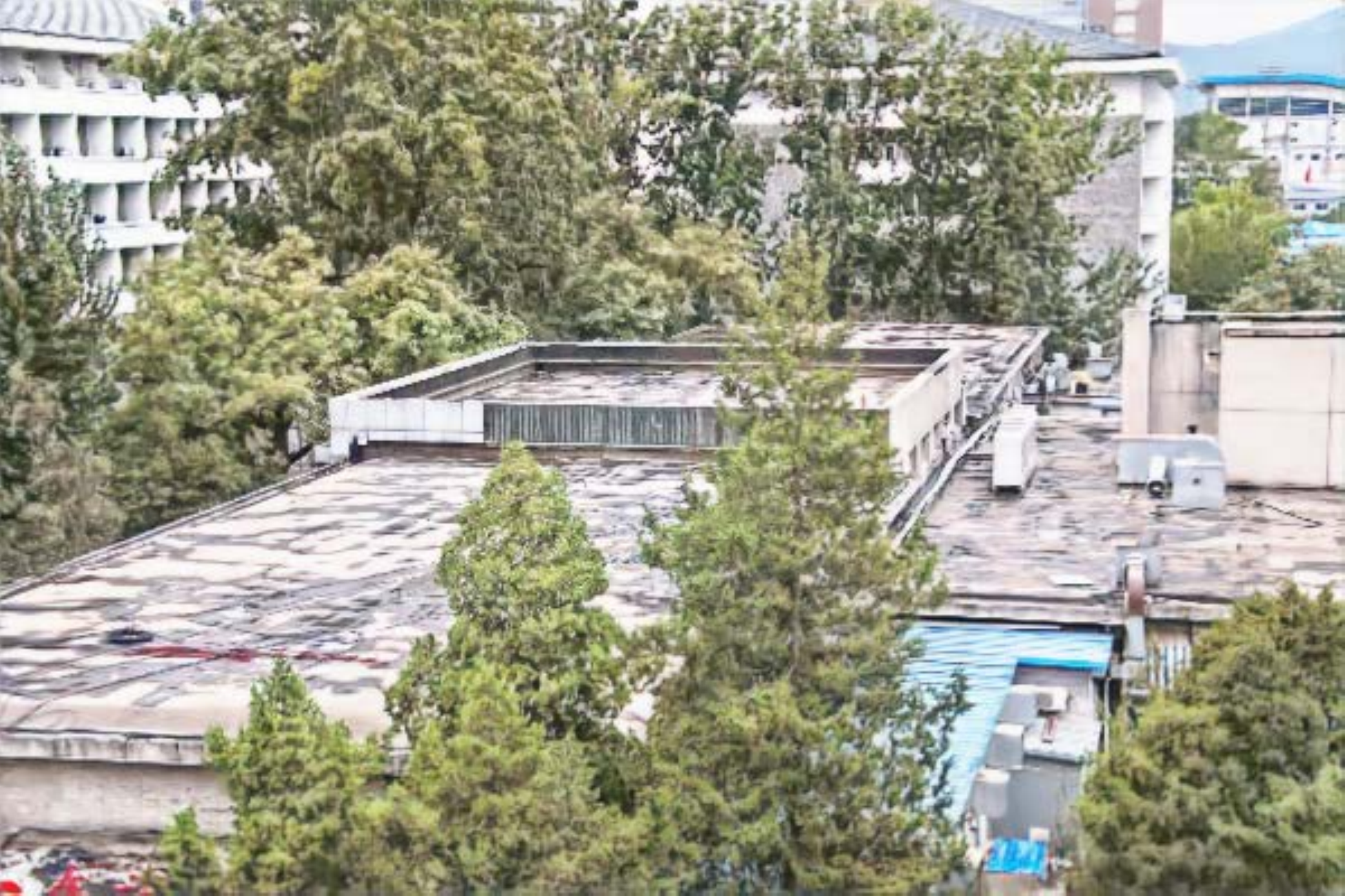}&
			\includegraphics[width=0.1354\linewidth]{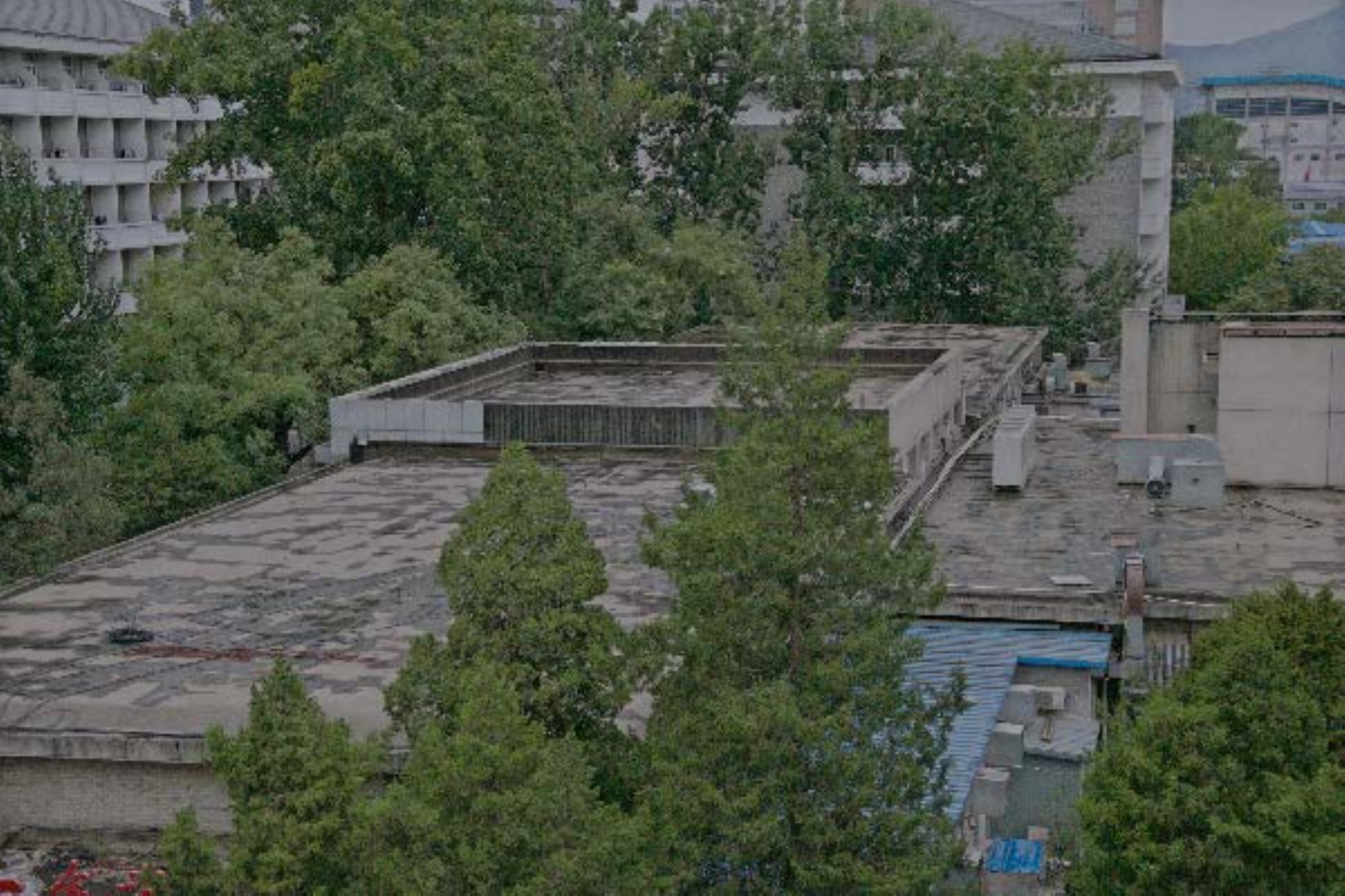}&
			\includegraphics[width=0.1354\linewidth]{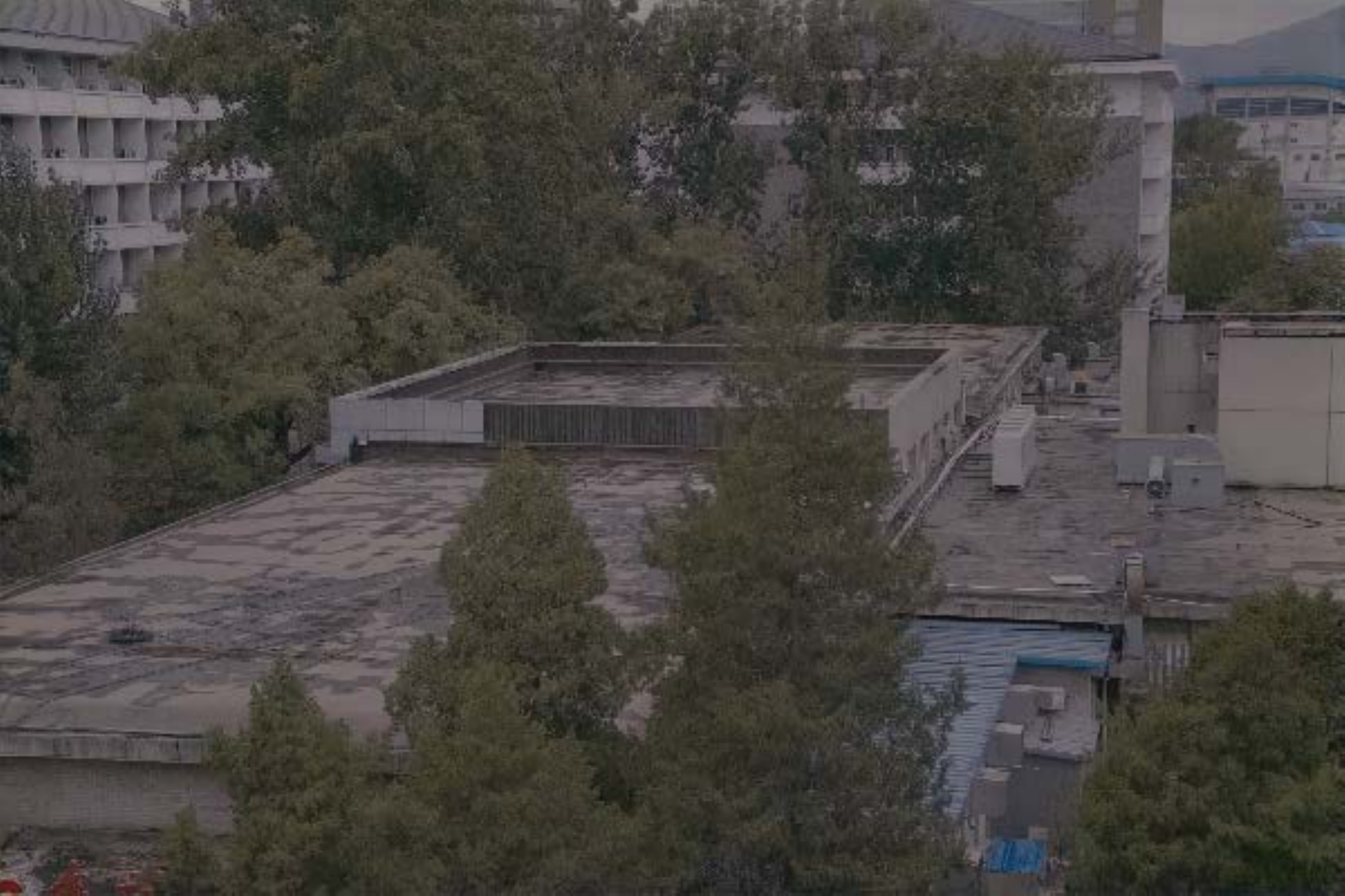}&
			\includegraphics[width=0.1354\linewidth]{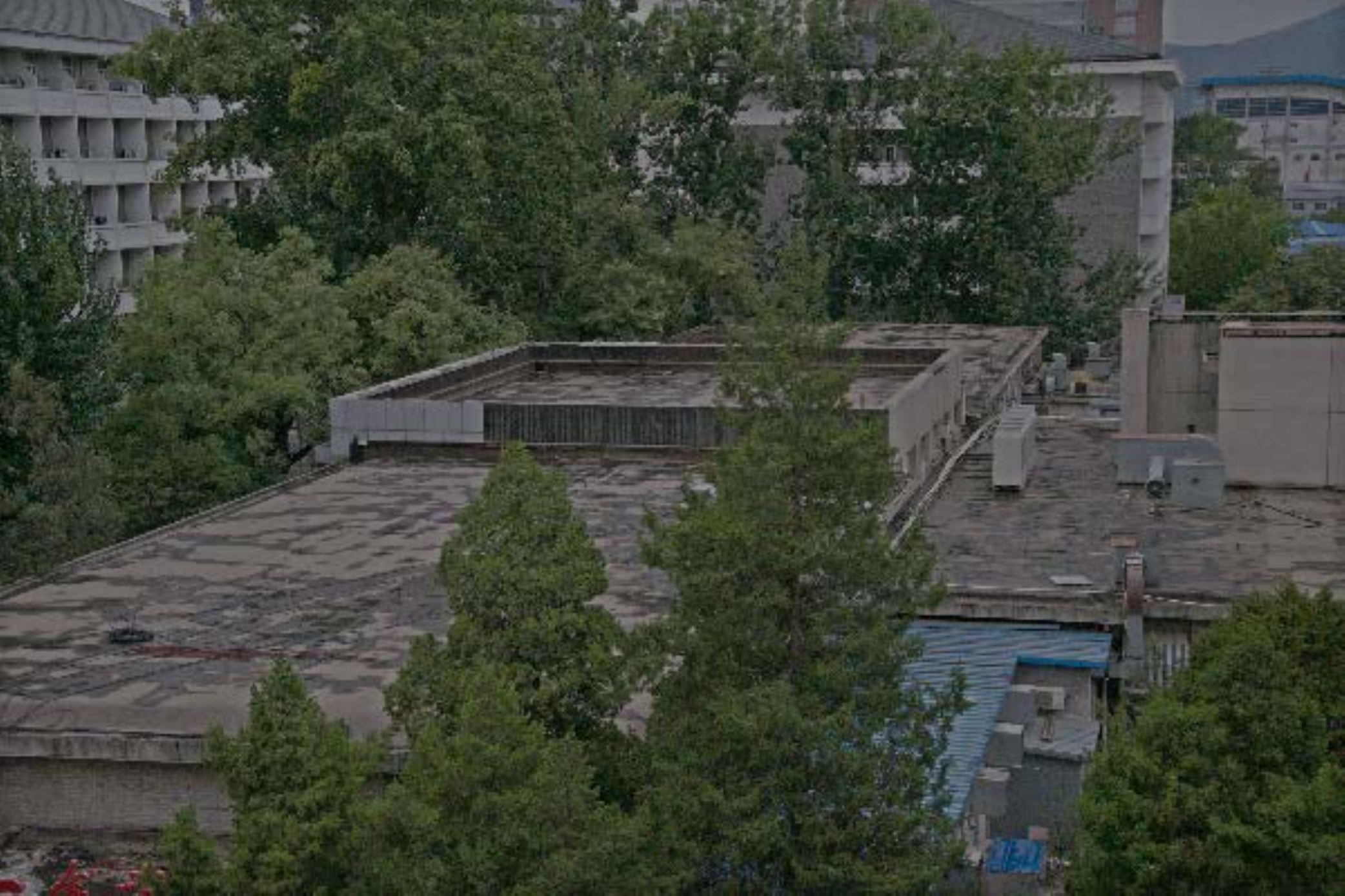}&
			\includegraphics[width=0.1354\linewidth]{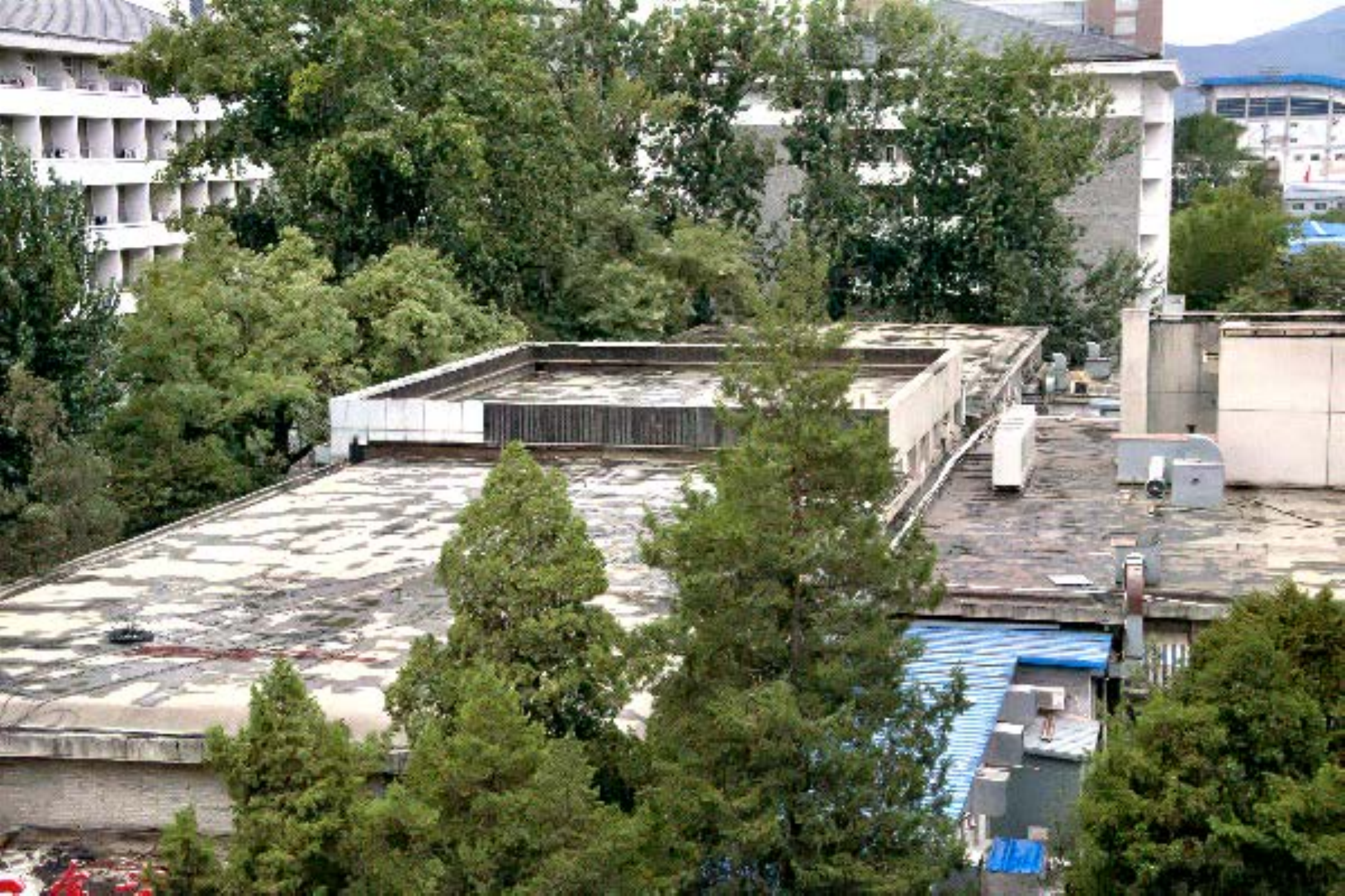}&
			\includegraphics[width=0.1354\linewidth]{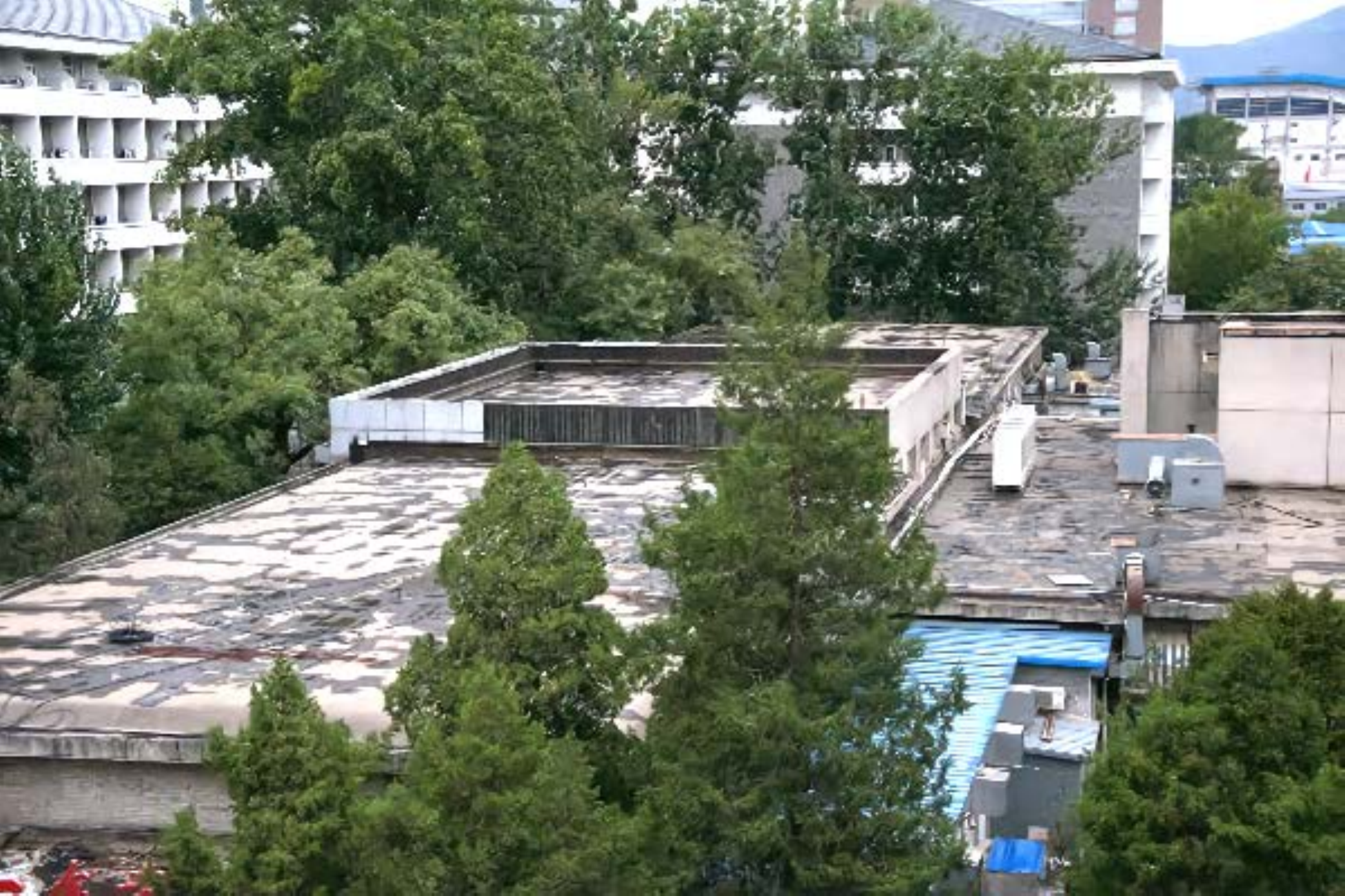}\\
			\includegraphics[width=0.1354\linewidth]{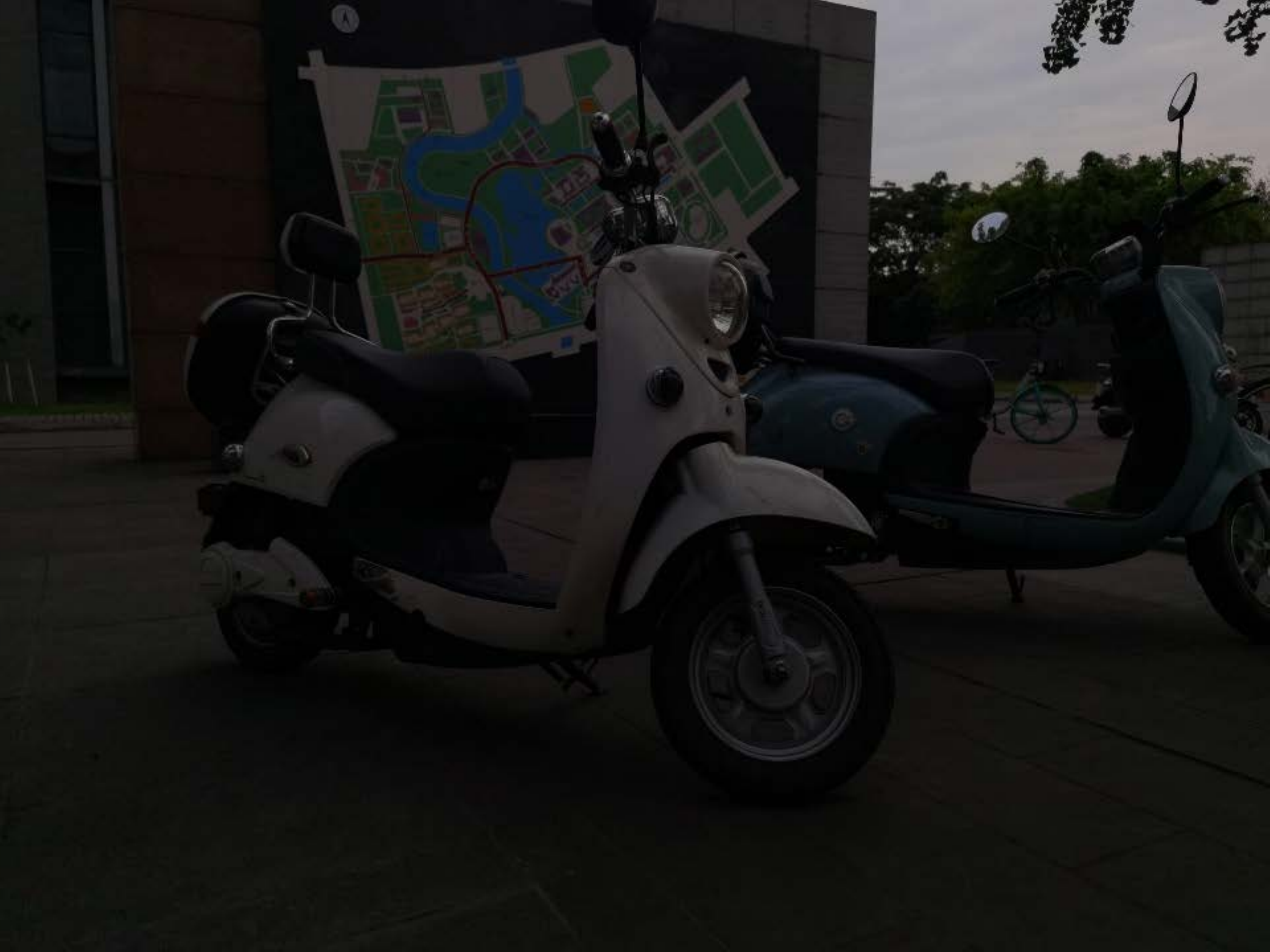}&
			\includegraphics[width=0.1354\linewidth]{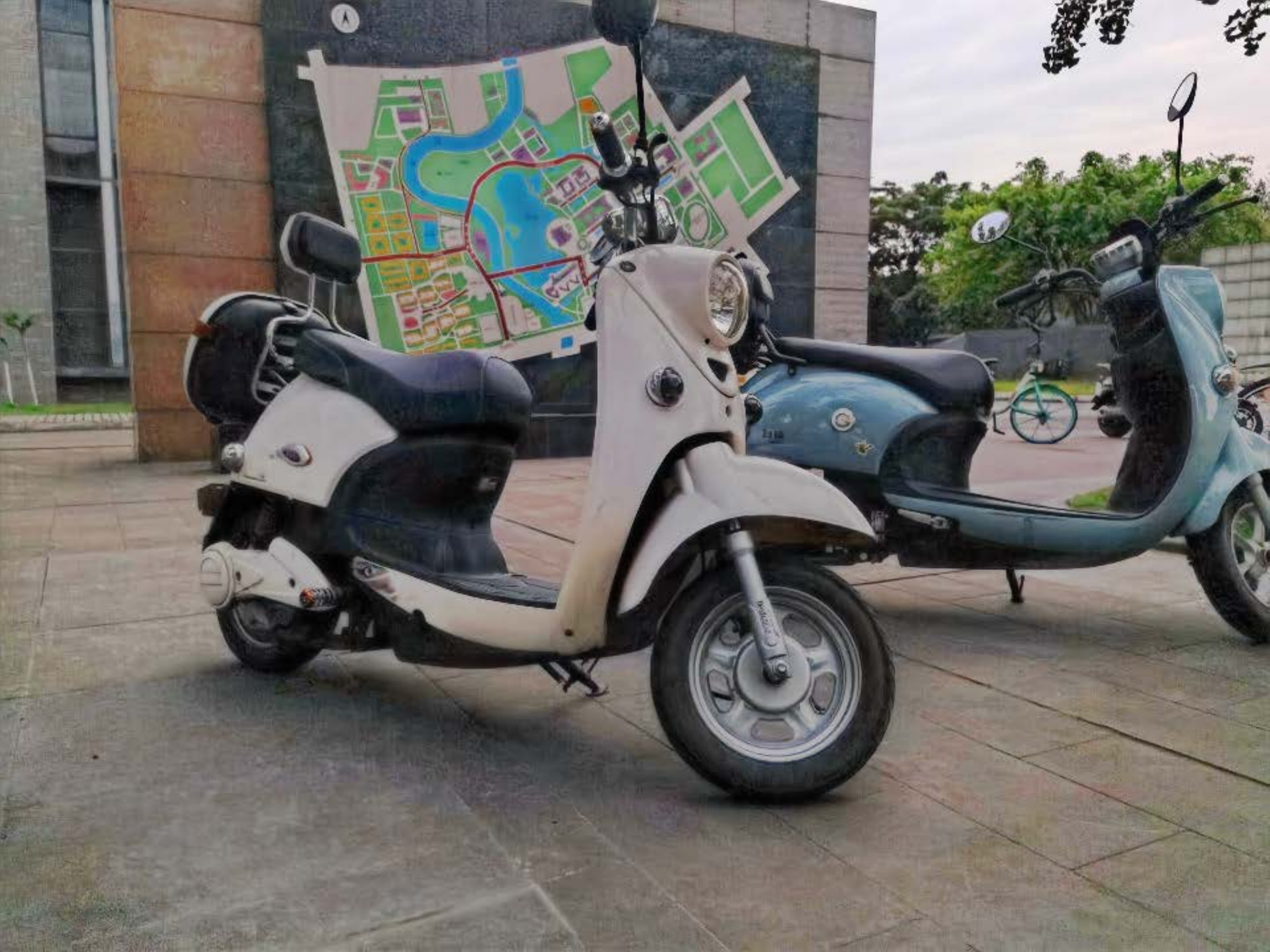}&
			\includegraphics[width=0.1354\linewidth]{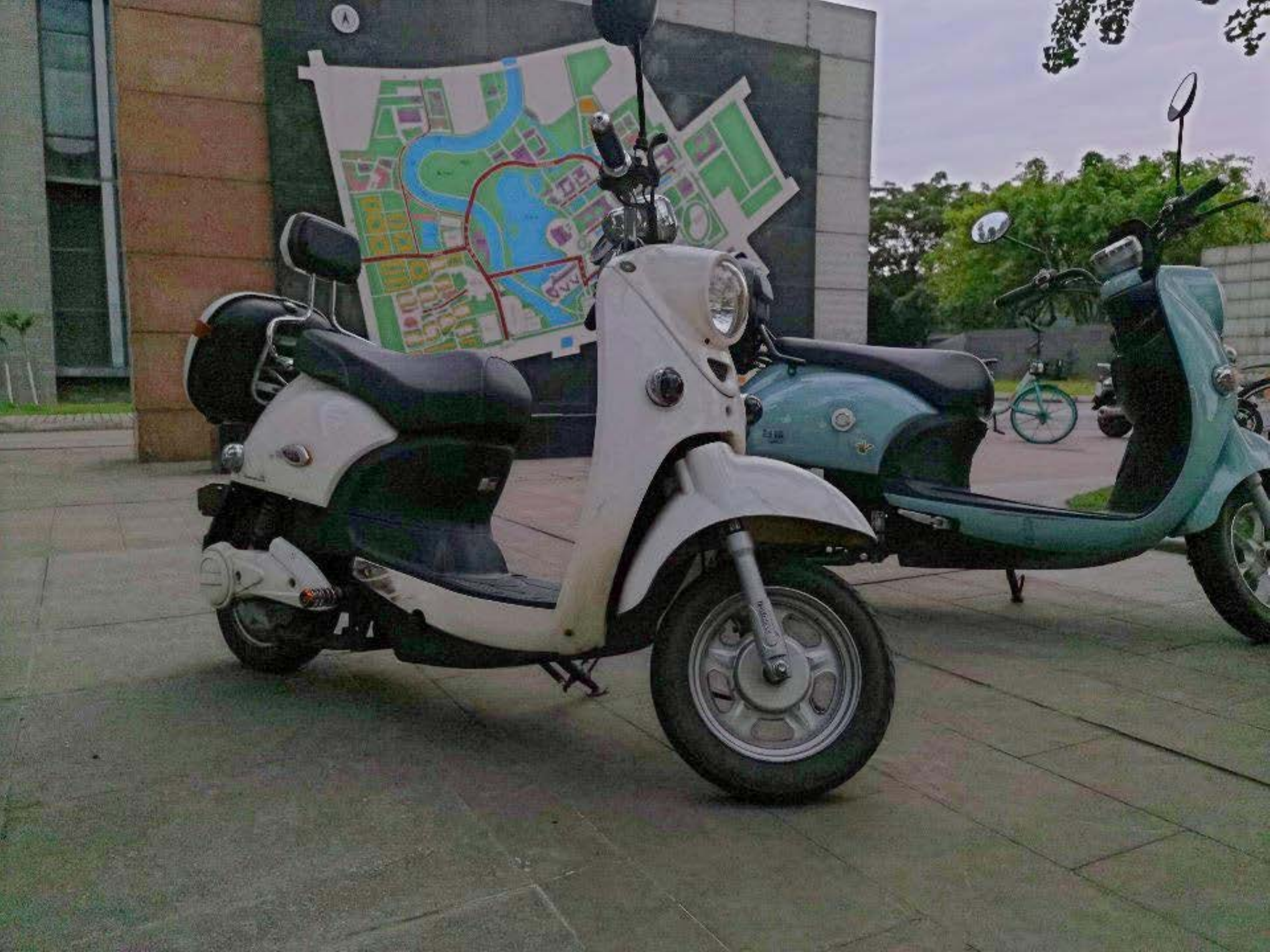}&
			\includegraphics[width=0.1354\linewidth]{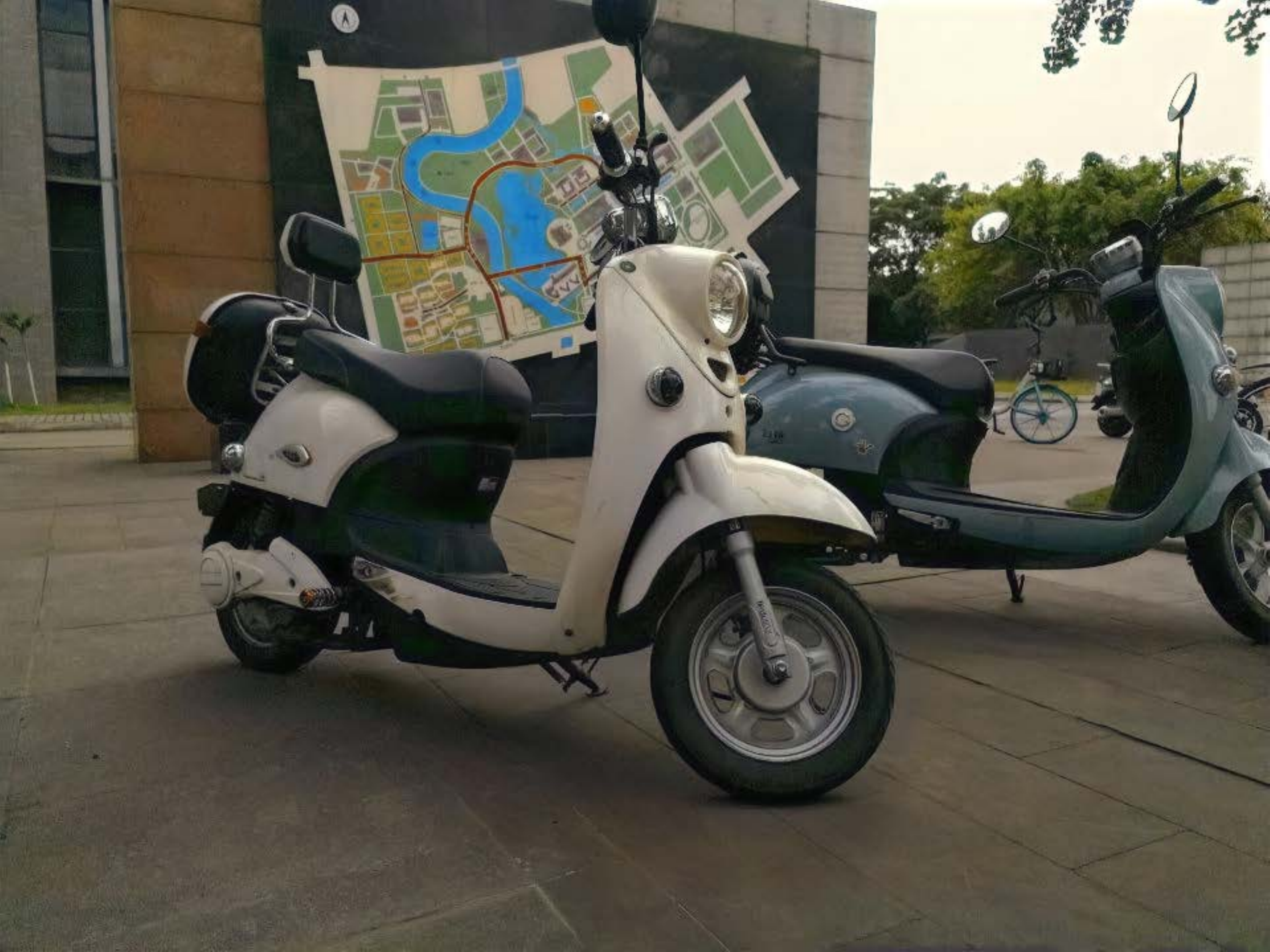}&
			\includegraphics[width=0.1354\linewidth]{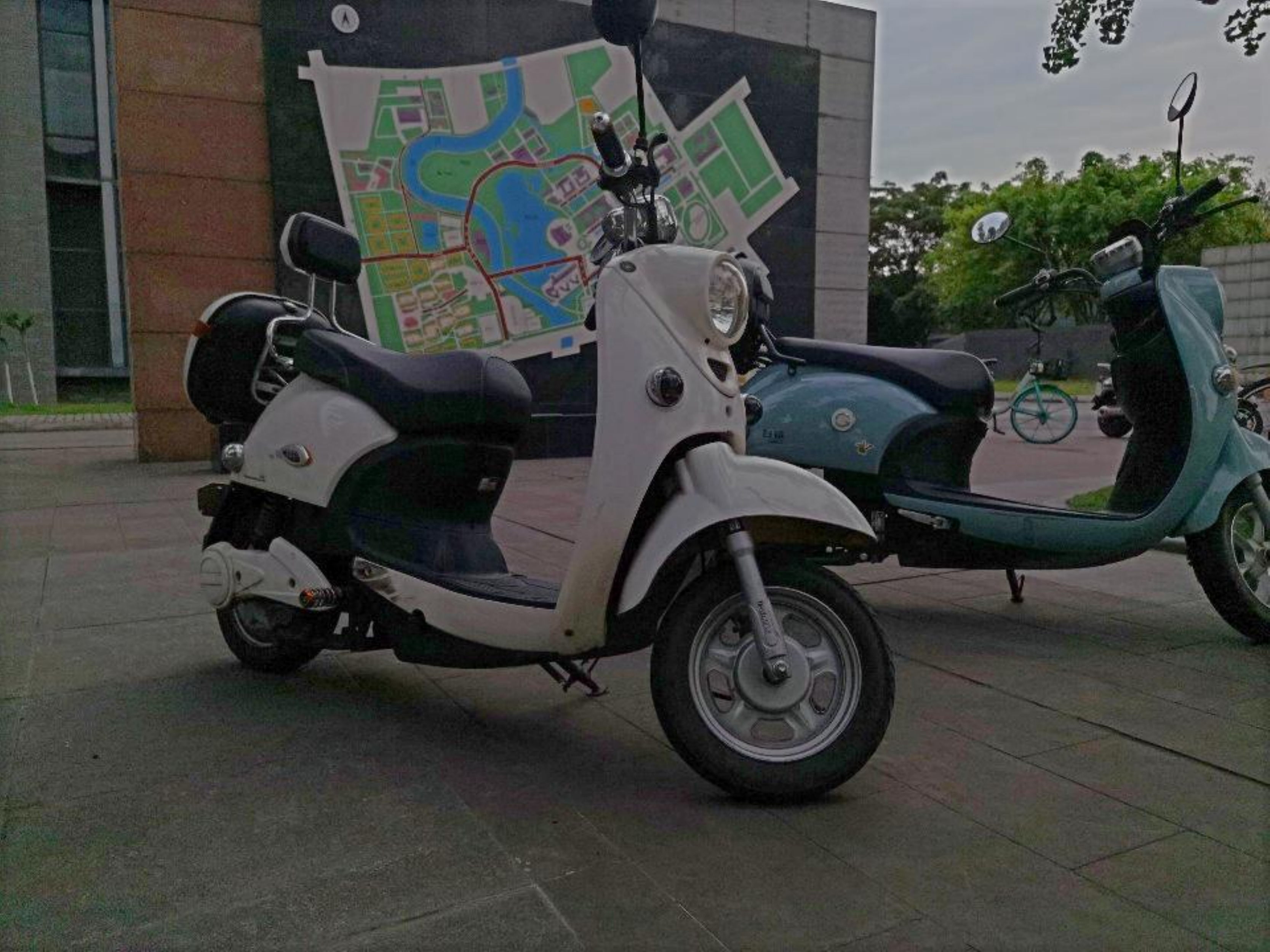}&
			\includegraphics[width=0.1354\linewidth]{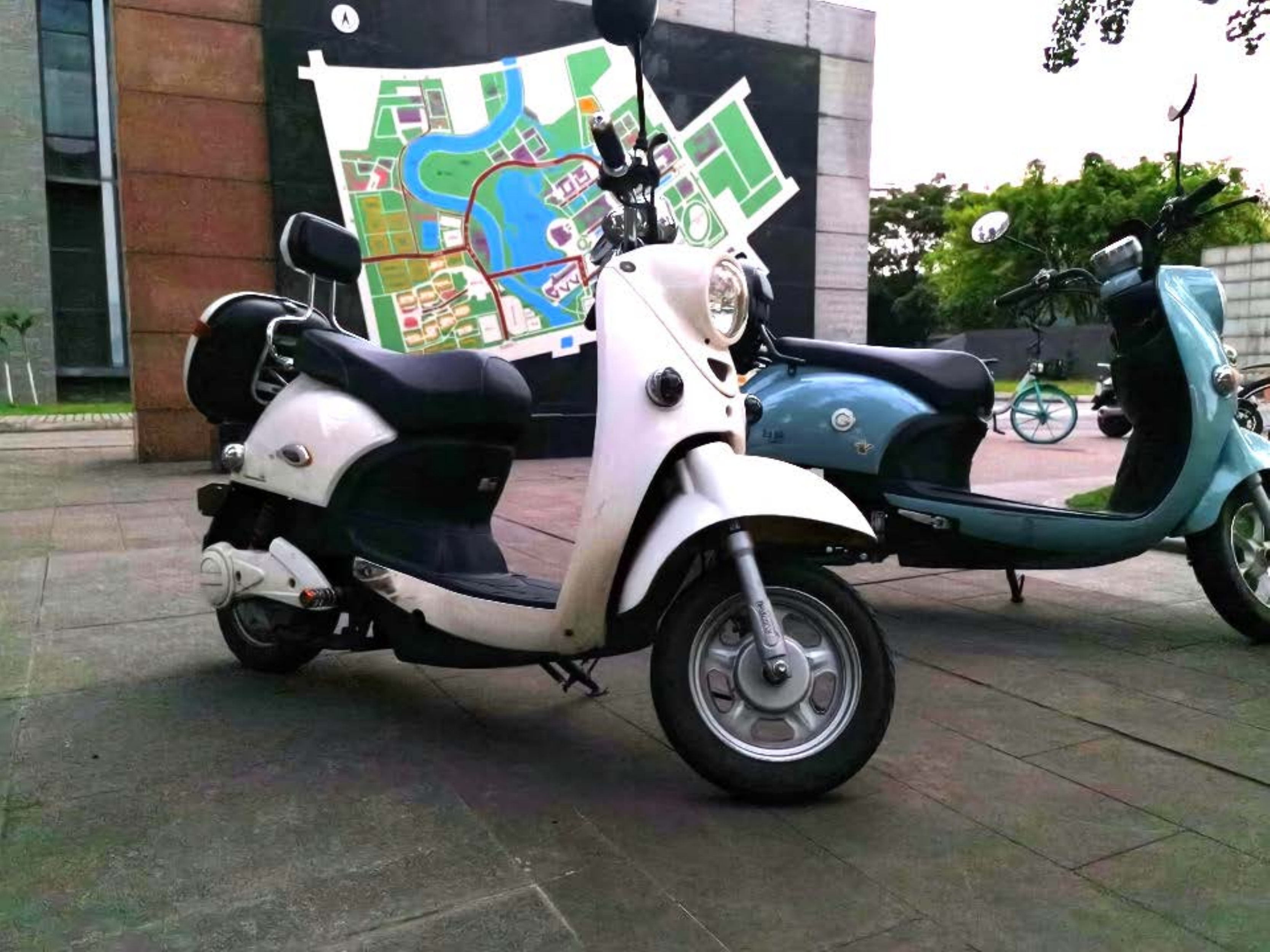}&
			\includegraphics[width=0.1354\linewidth]{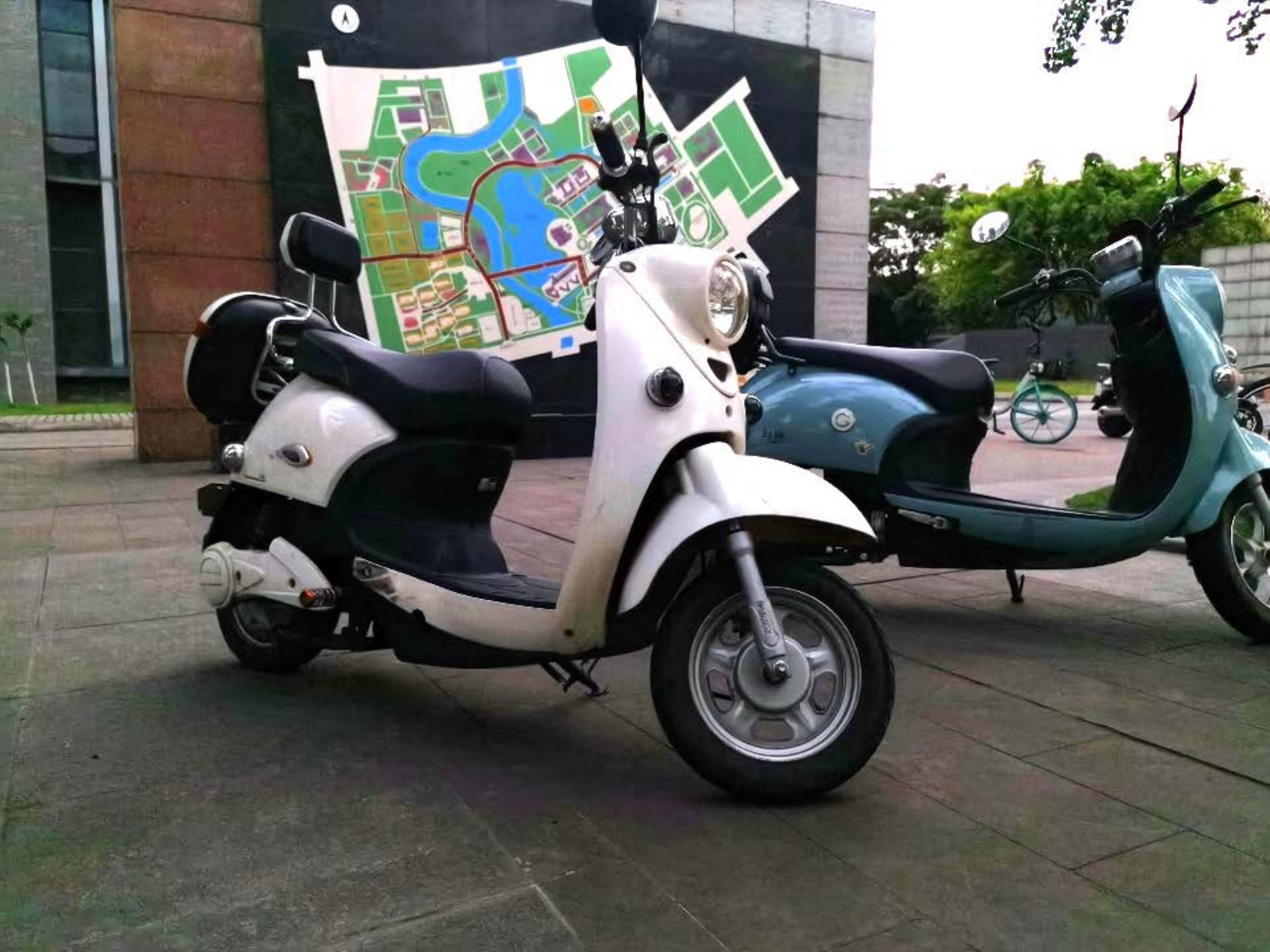}\\
			\includegraphics[width=0.1354\linewidth]{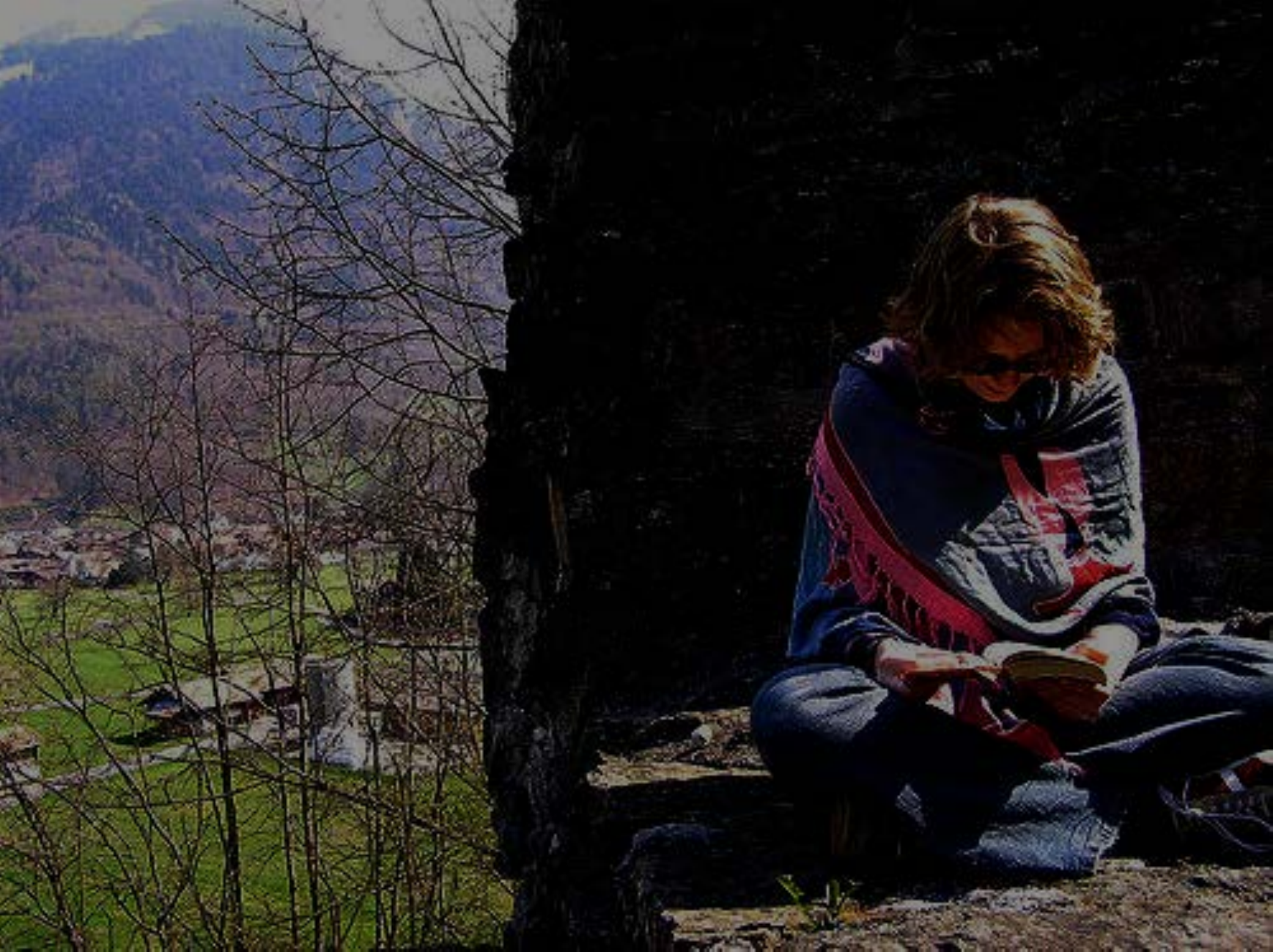}&
			\includegraphics[width=0.1354\linewidth]{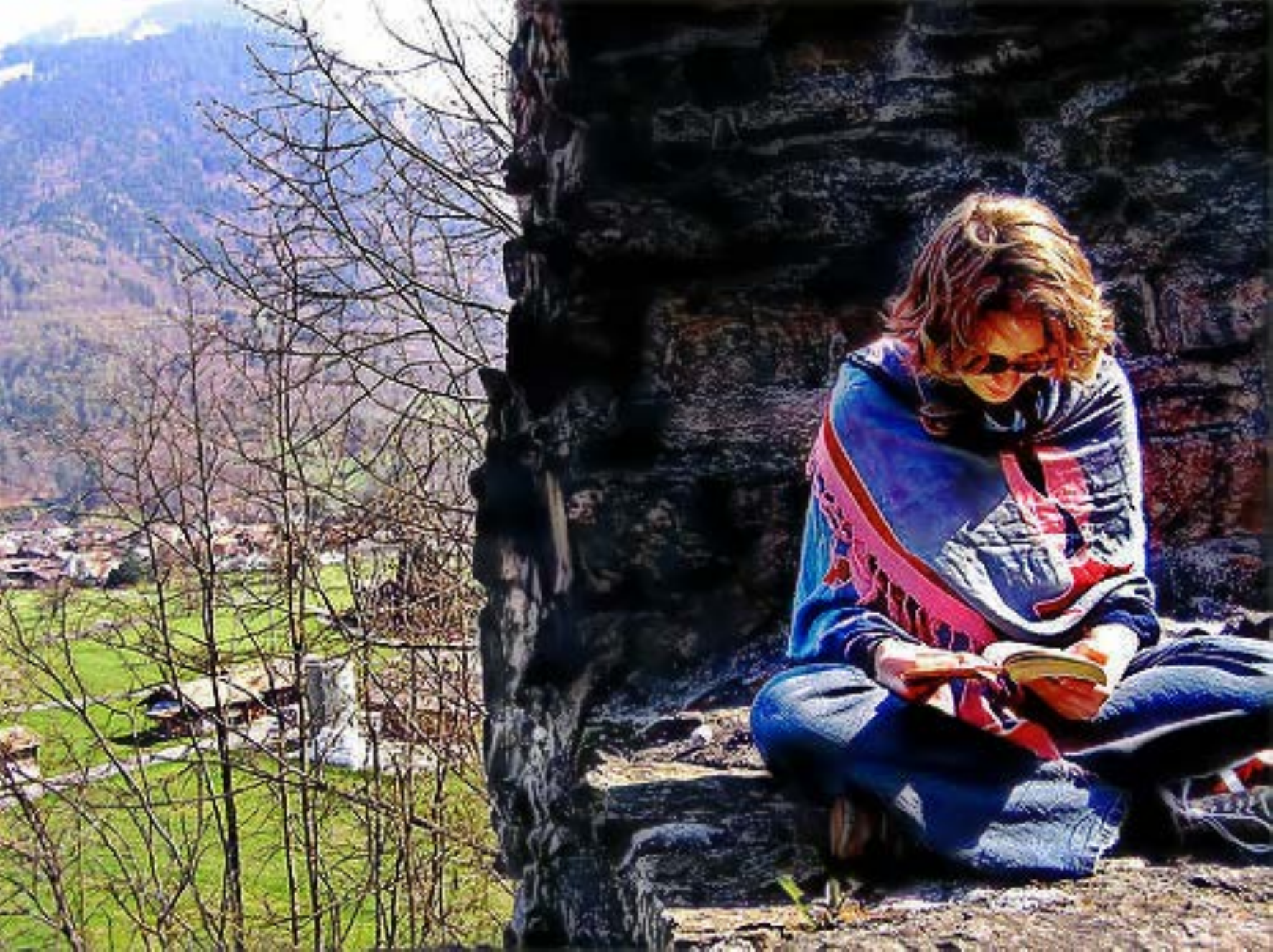}&
			\includegraphics[width=0.1354\linewidth]{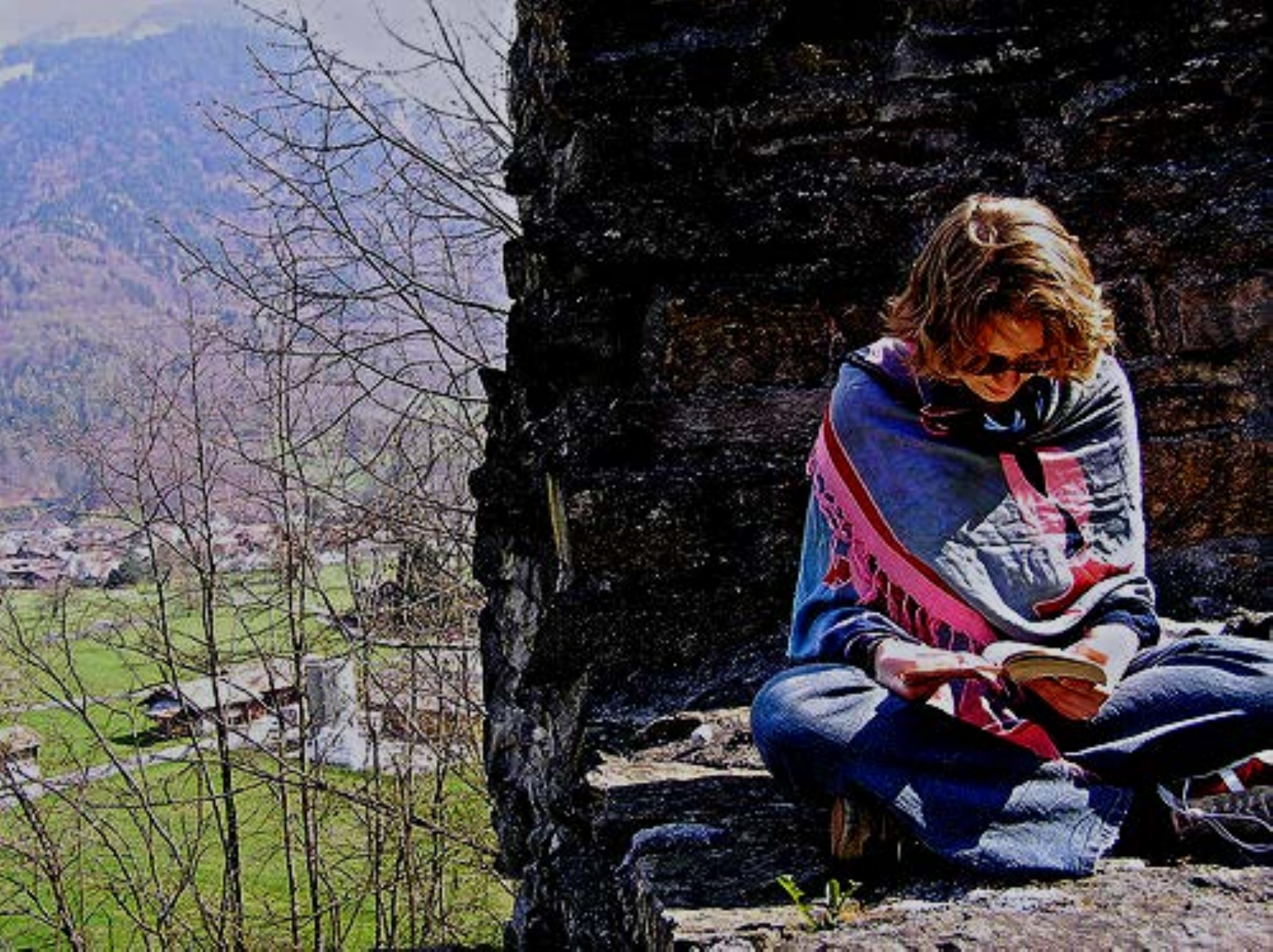}&
			\includegraphics[width=0.1354\linewidth]{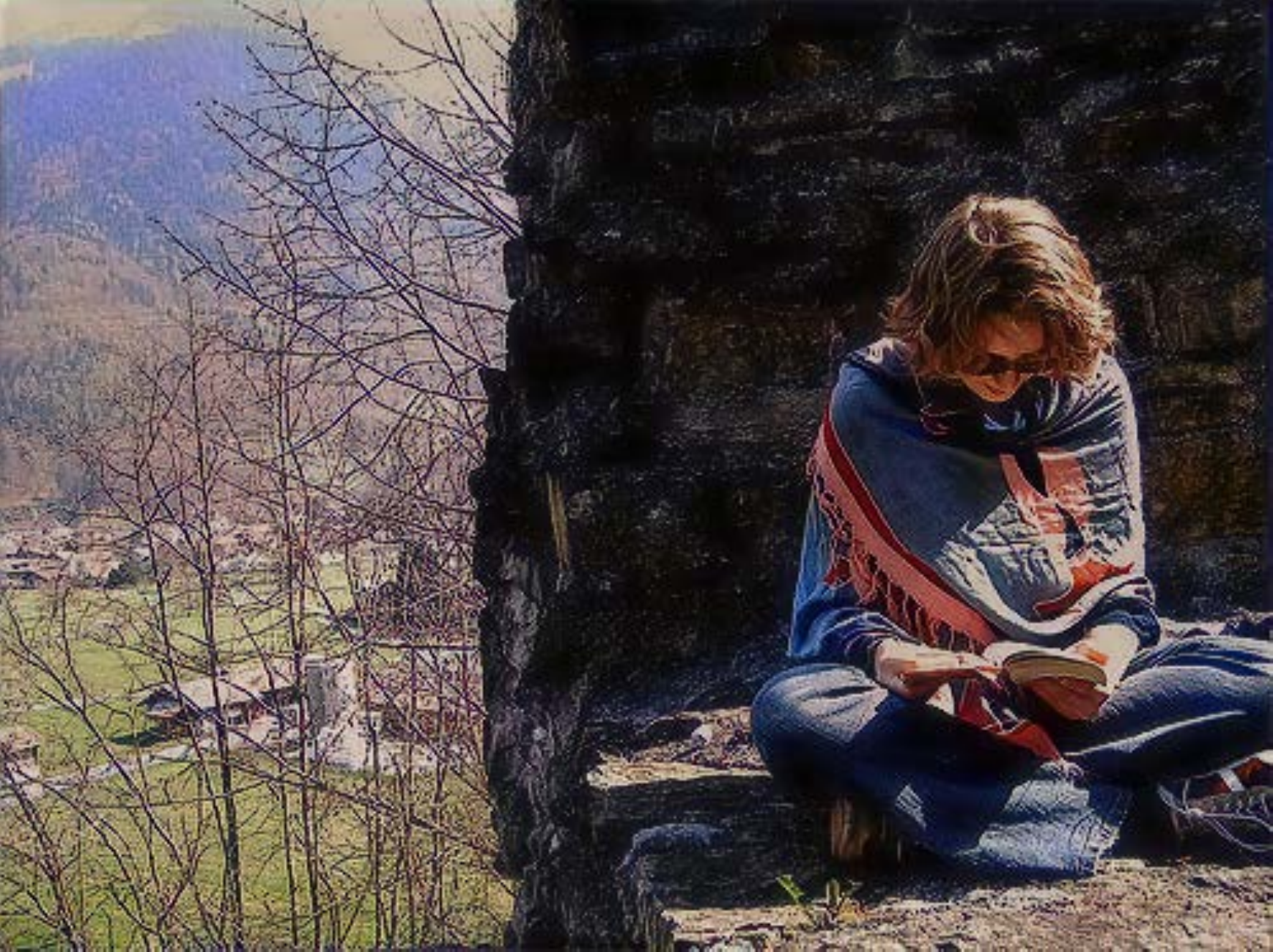}&
			\includegraphics[width=0.1354\linewidth]{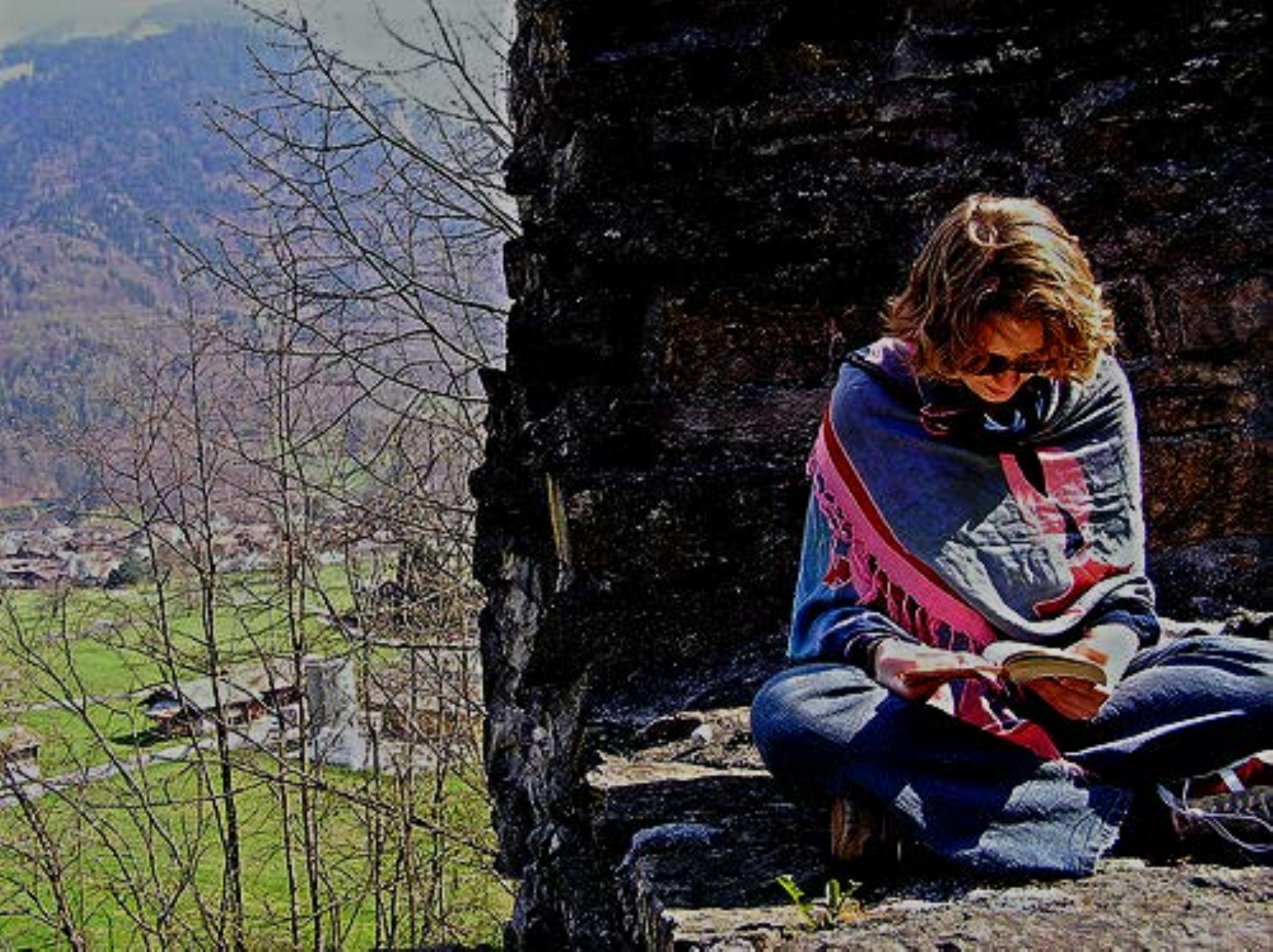}&
			\includegraphics[width=0.1354\linewidth]{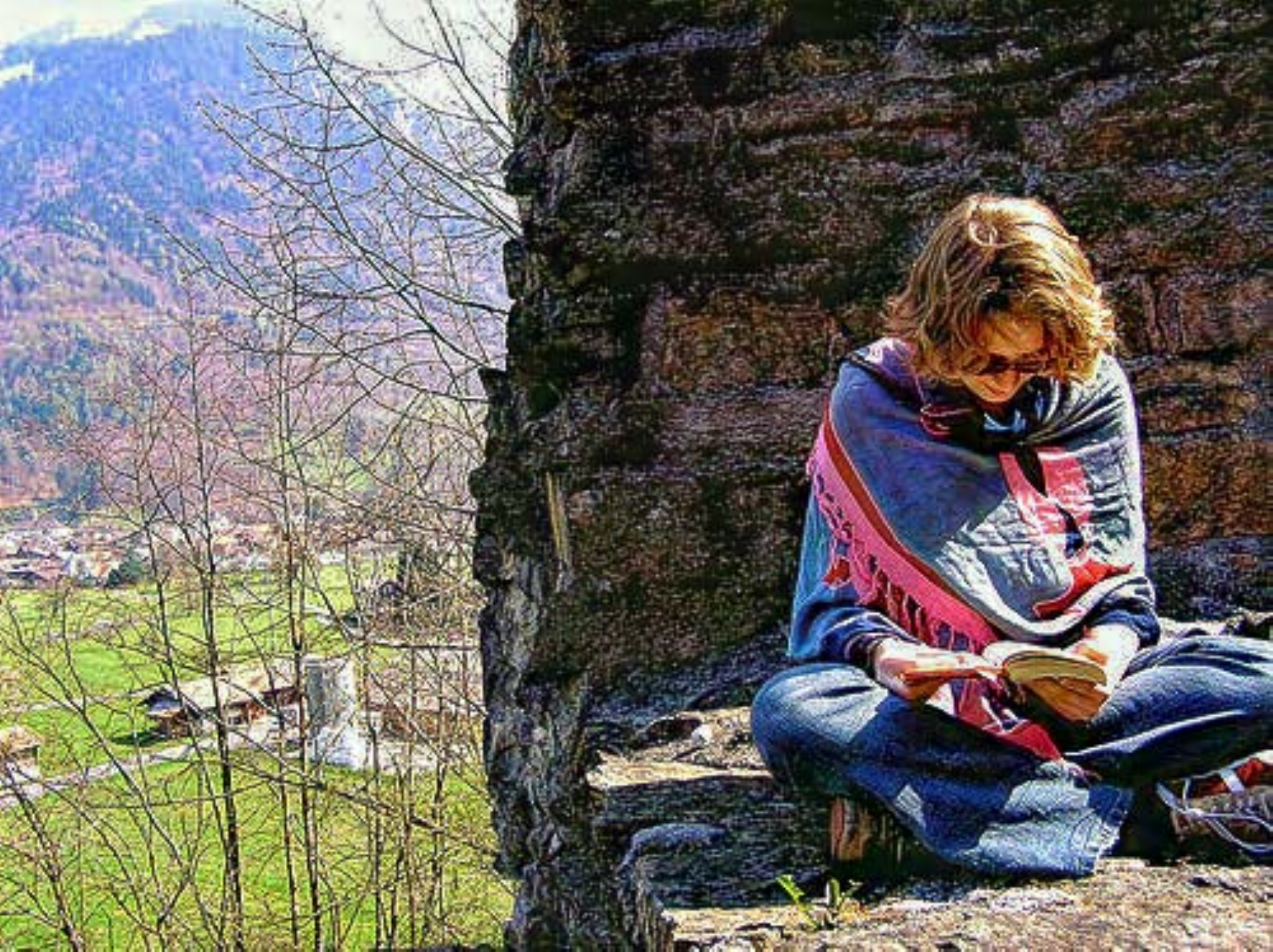}&
			\includegraphics[width=0.1354\linewidth]{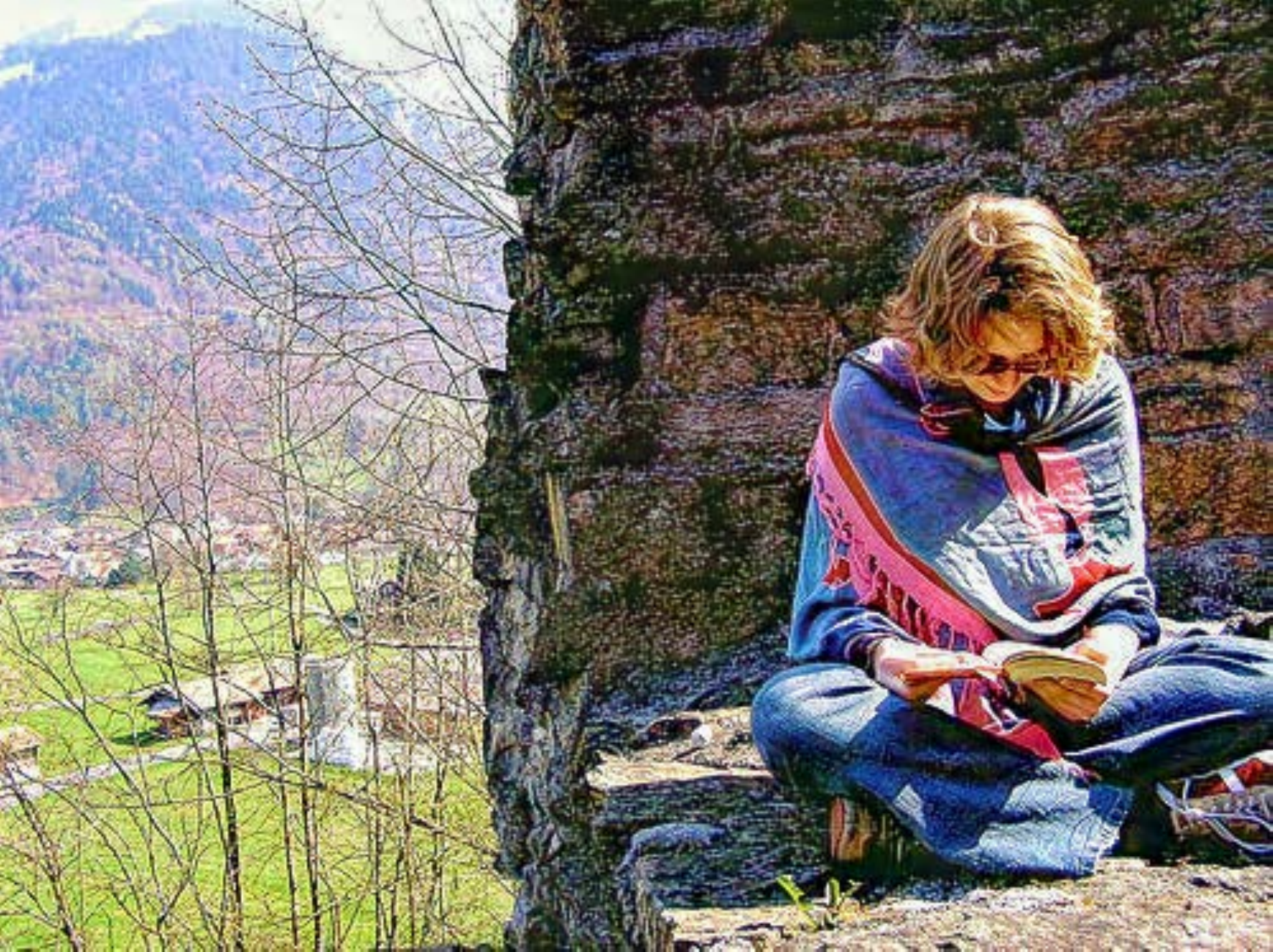}\\
			\footnotesize Input&\footnotesize KinD&\footnotesize ZeroDCE&\footnotesize UTVNet&\footnotesize SCL&\footnotesize \textbf{BL}&\footnotesize \textbf{RBL}\\
		\end{tabular}
		\caption{More visual comparisons on the standard datasets. The top two and middle two rows come from the MIT and LOL datasets, respectively. The last two rows come from the LSRW and VOC datasets, respectively.}
		\label{fig: Moredatasets}
	\end{figure*}

	\begin{figure*}[t]
		\centering
		\begin{tabular}{c}
			\includegraphics[width=1\linewidth]{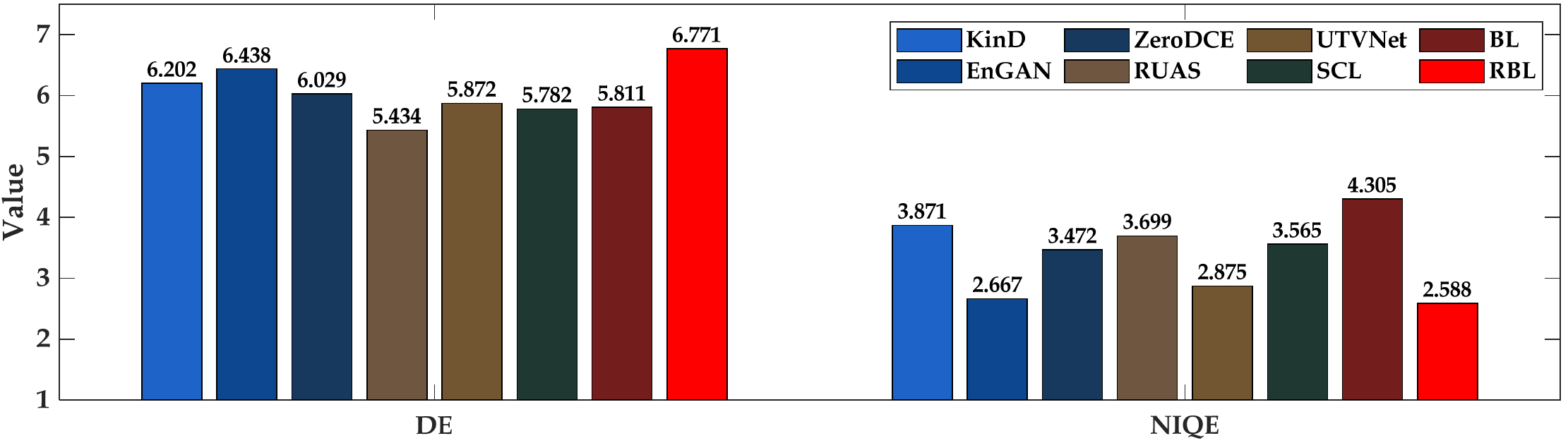}\\
		\end{tabular}
		%	\vspace{-0.4cm}
		\caption{Numerical scores (DE$\uparrow$ and NIQE$\downarrow$) among different state-of-the-art methods and our two versions on the DARK FACE dataset. }
		\label{fig: Hist}
	\end{figure*}
	
	\begin{figure*}[t]
		\centering
		\begin{tabular}{c@{\extracolsep{0.3em}}c@{\extracolsep{0.3em}}c@{\extracolsep{0.3em}}c@{\extracolsep{0.3em}}c}
			\includegraphics[width=0.192\linewidth]{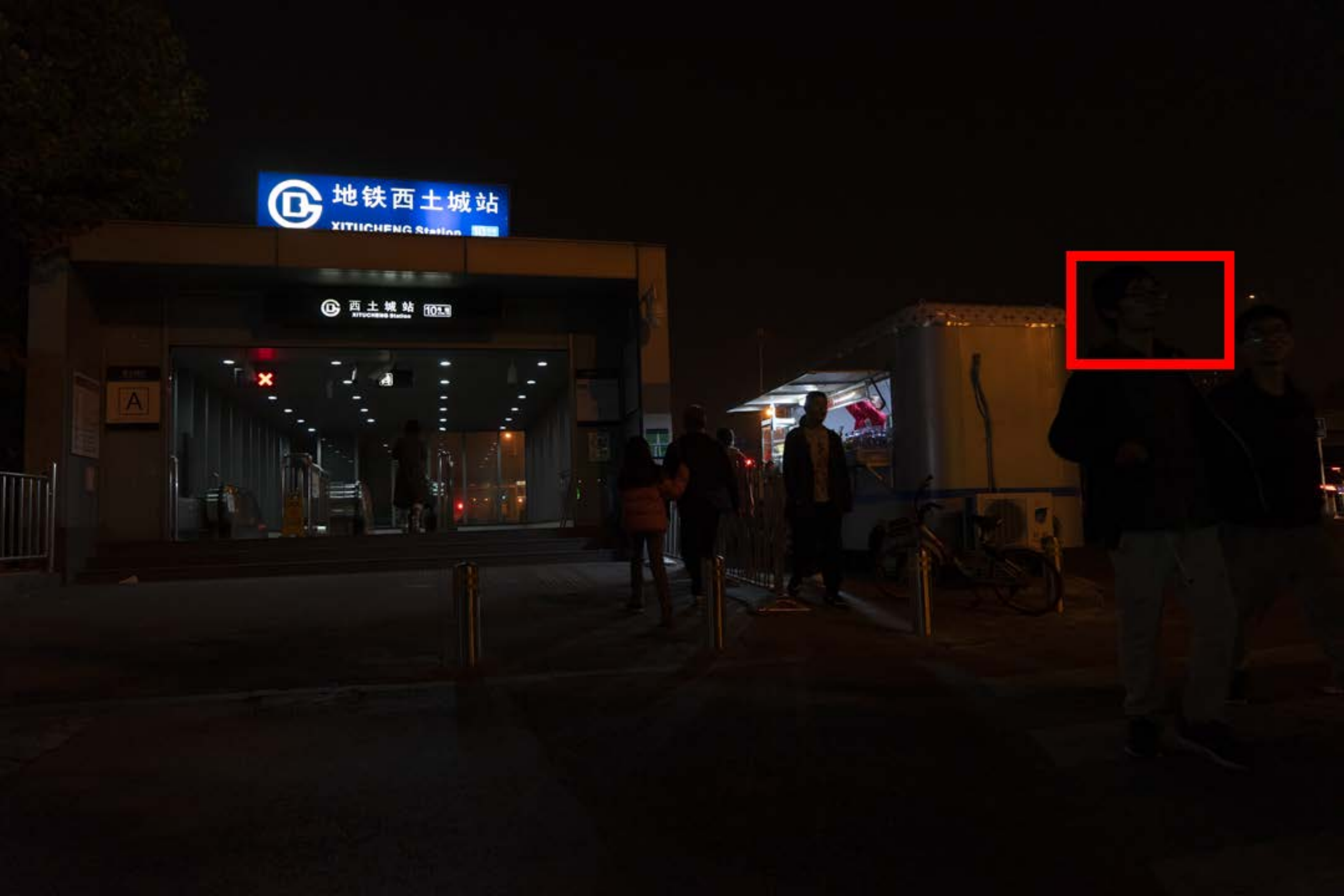}&
			\includegraphics[width=0.192\linewidth]{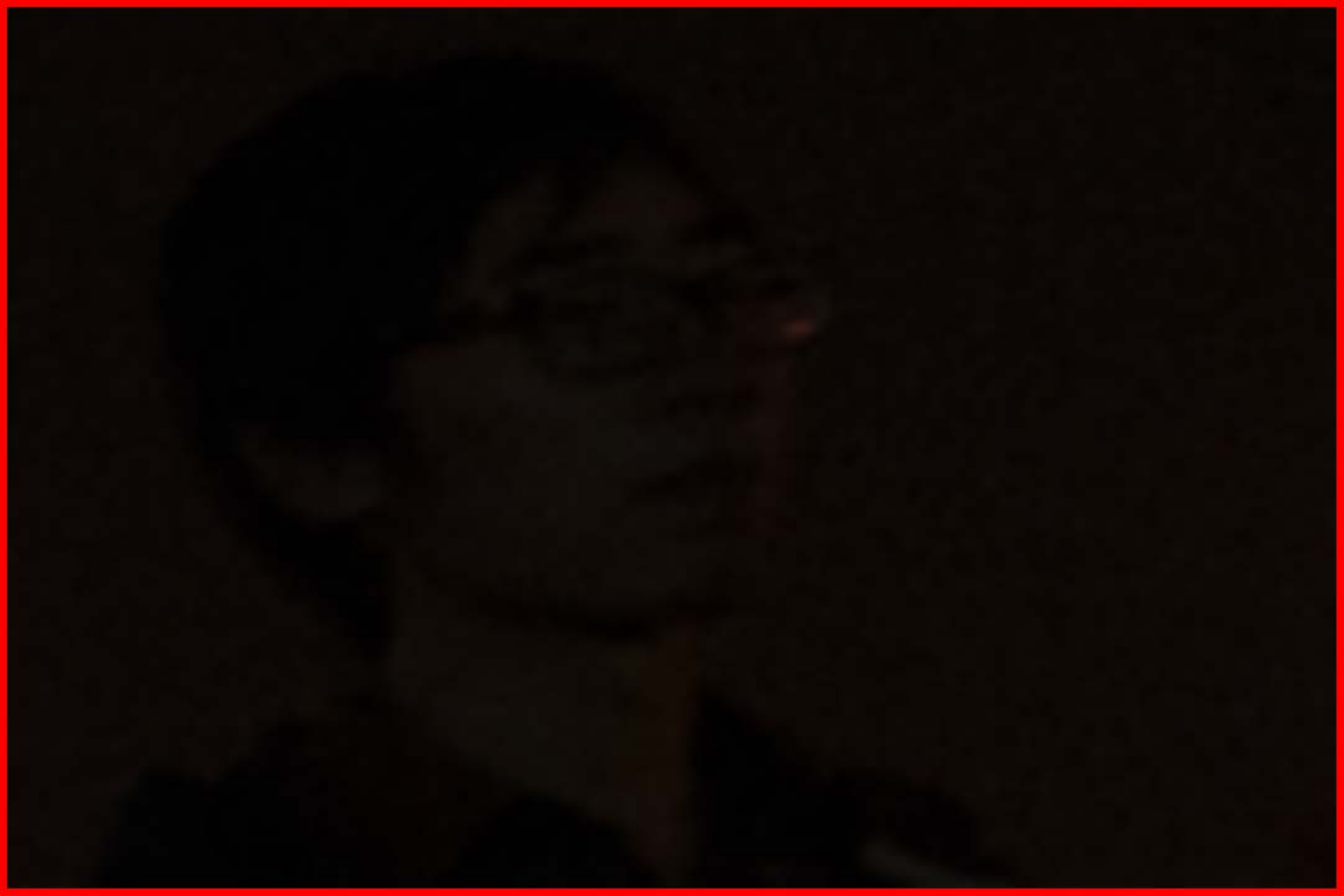}&
			\includegraphics[width=0.192\linewidth]{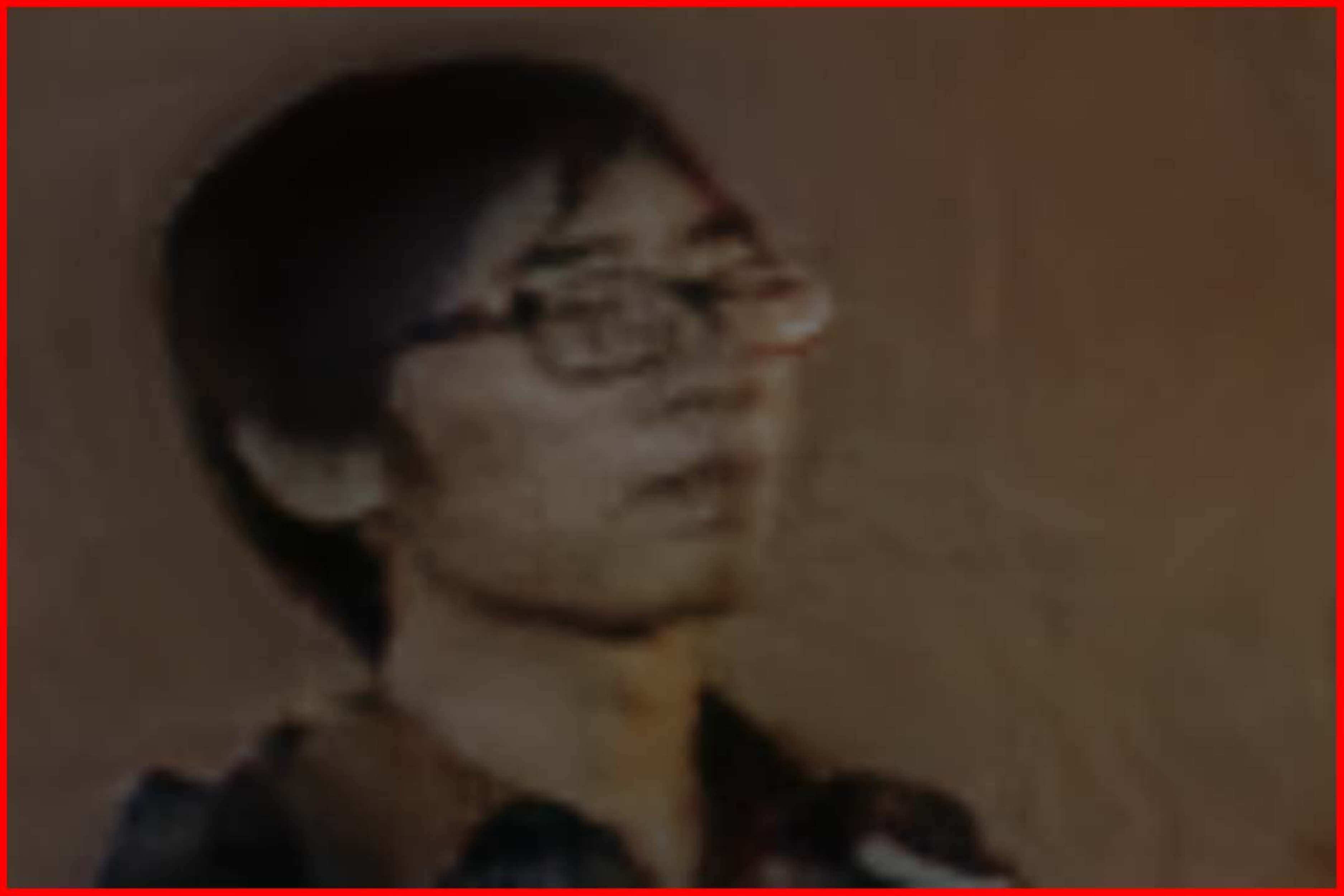}&
			\includegraphics[width=0.192\linewidth]{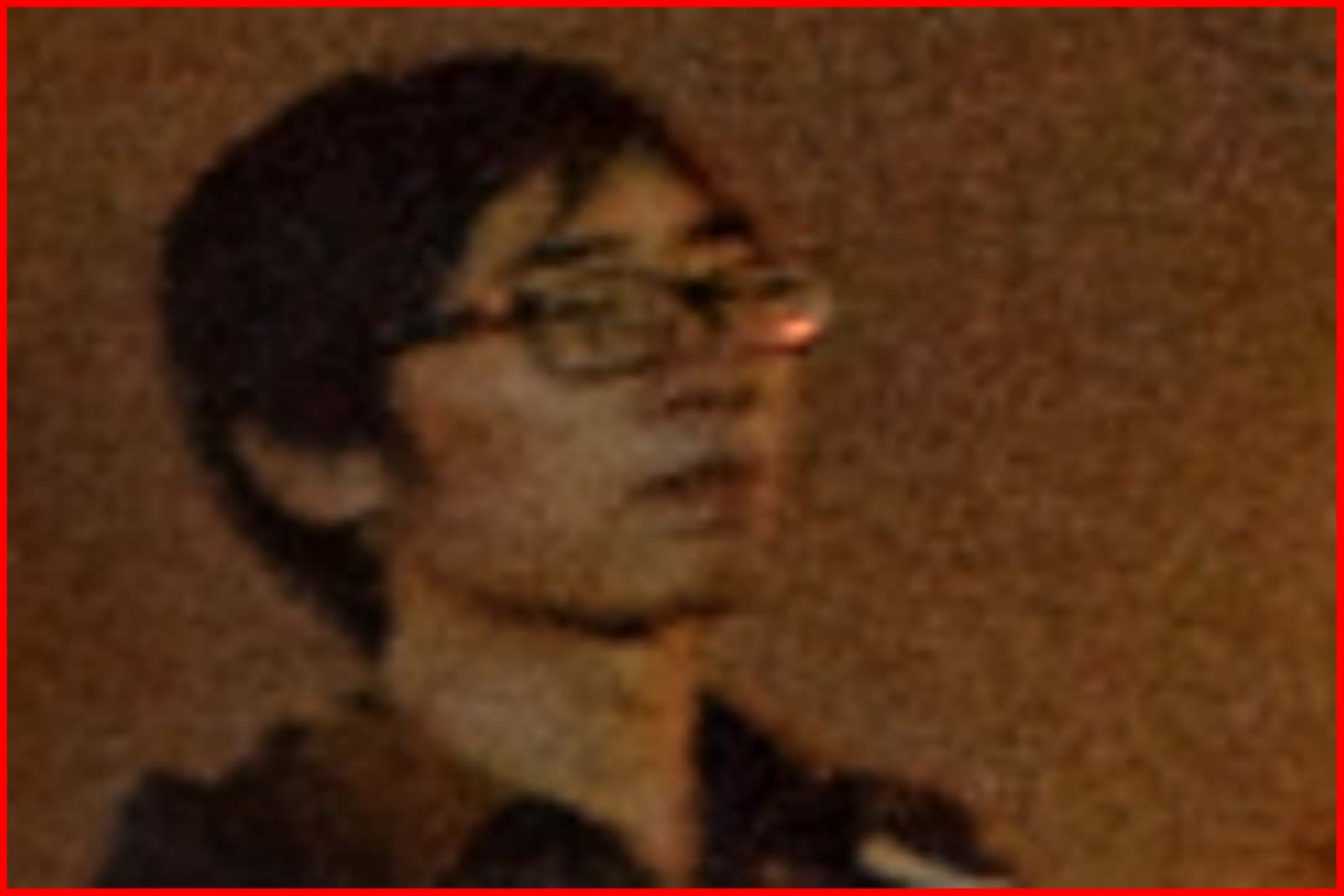}&
			\includegraphics[width=0.192\linewidth]{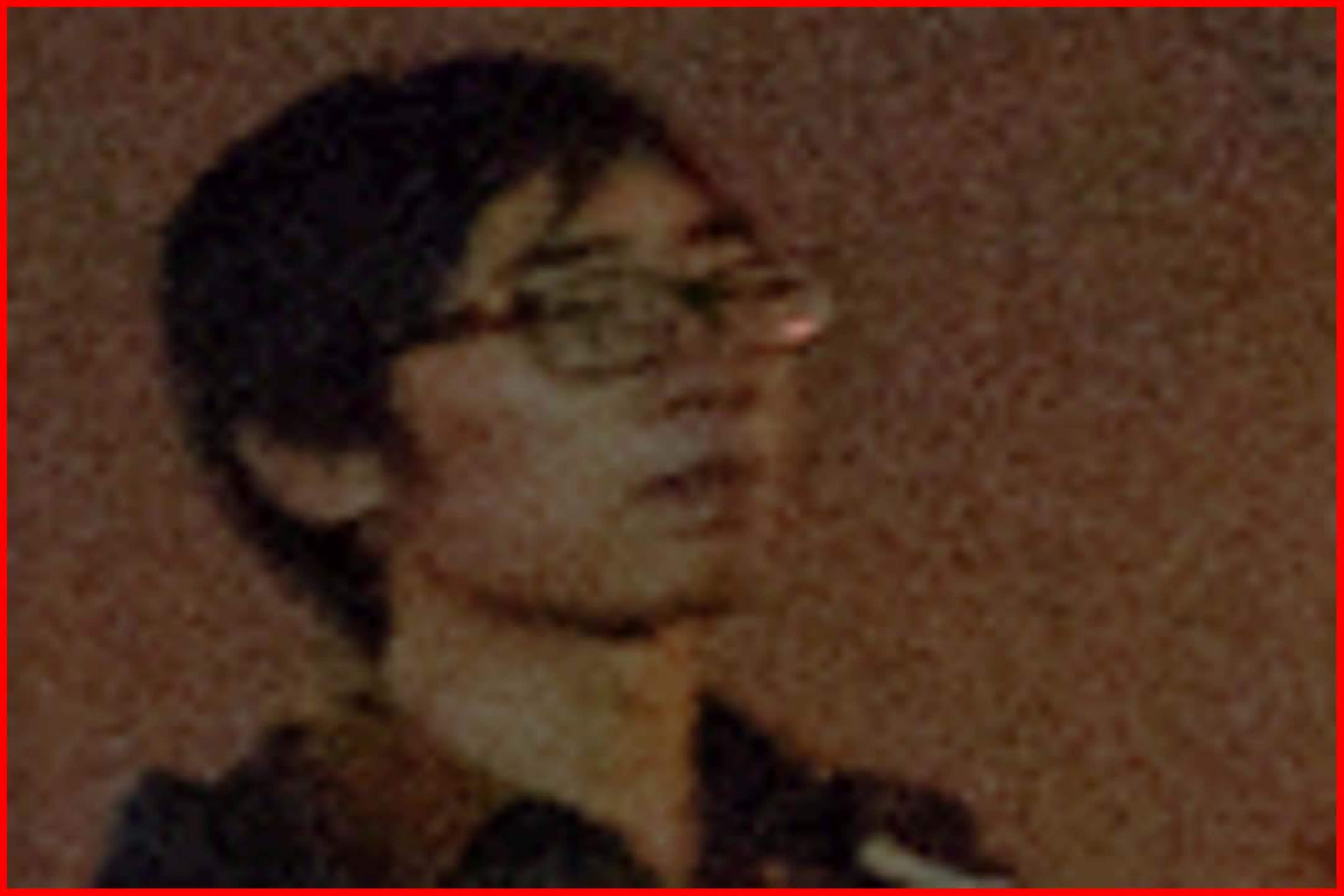}\\
			\footnotesize Input (Full Size)&\footnotesize Input&\footnotesize KinD&\footnotesize EnGAN&\footnotesize ZeroDCE\\
			\includegraphics[width=0.192\linewidth]{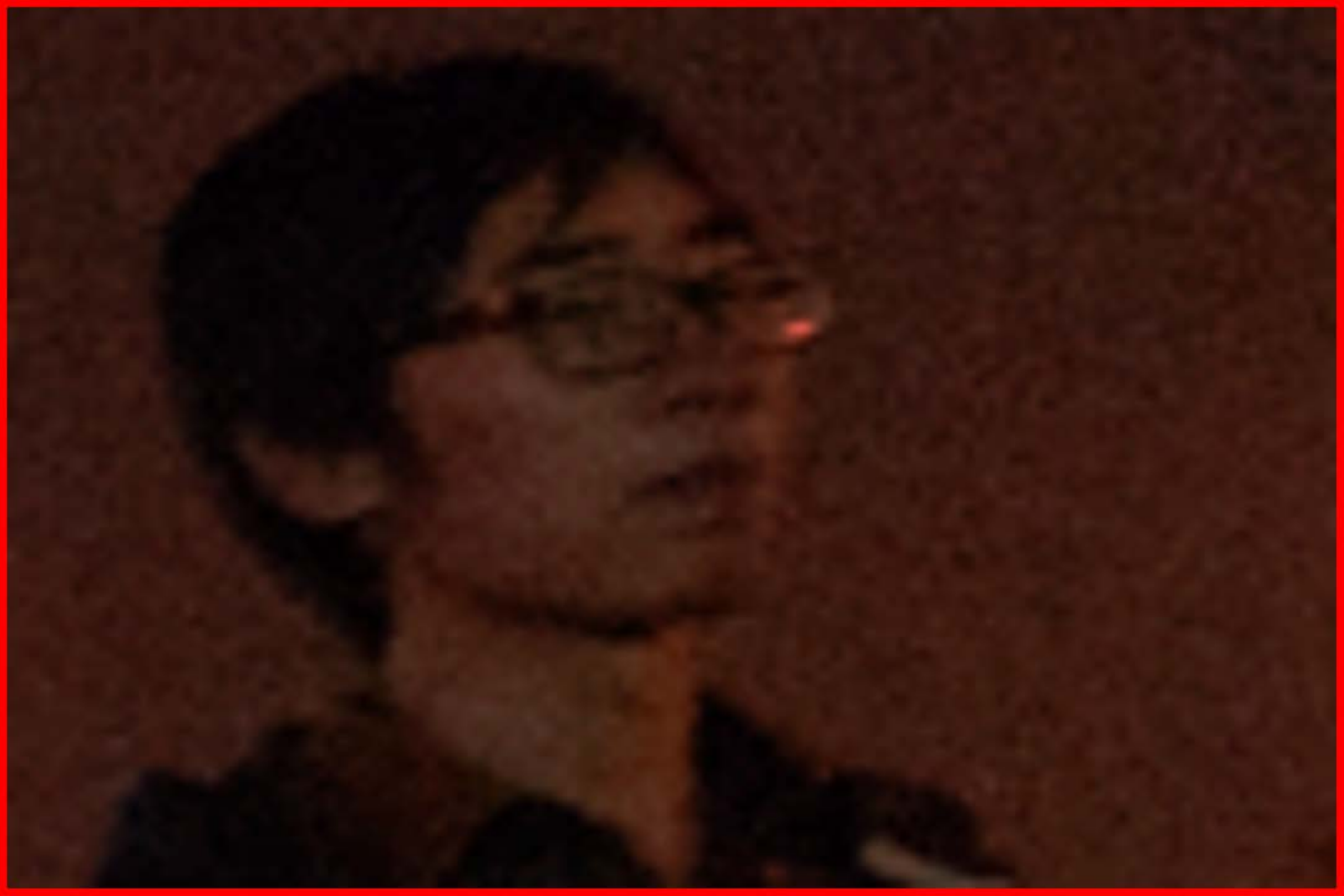}&
			\includegraphics[width=0.192\linewidth]{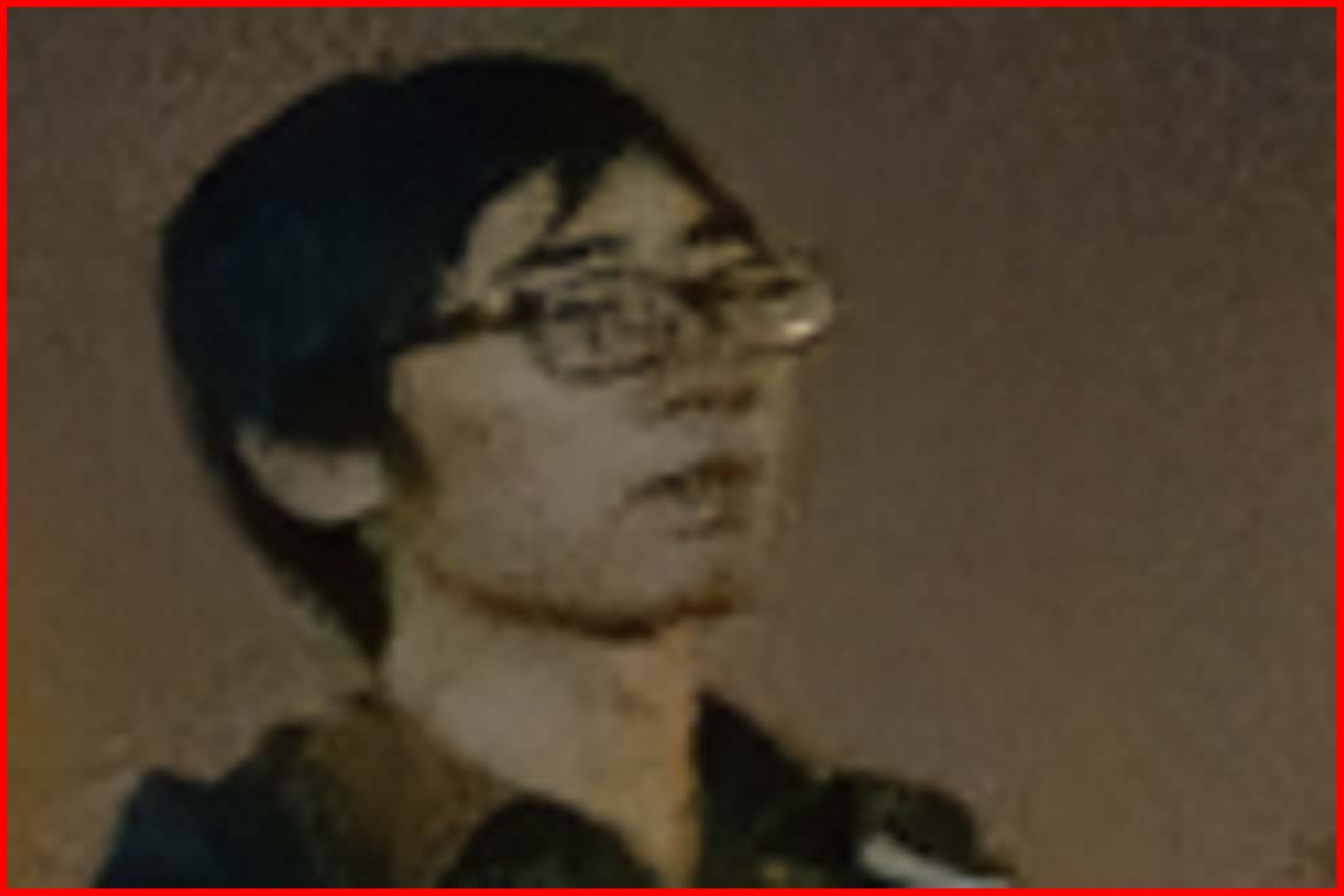}&
			\includegraphics[width=0.192\linewidth]{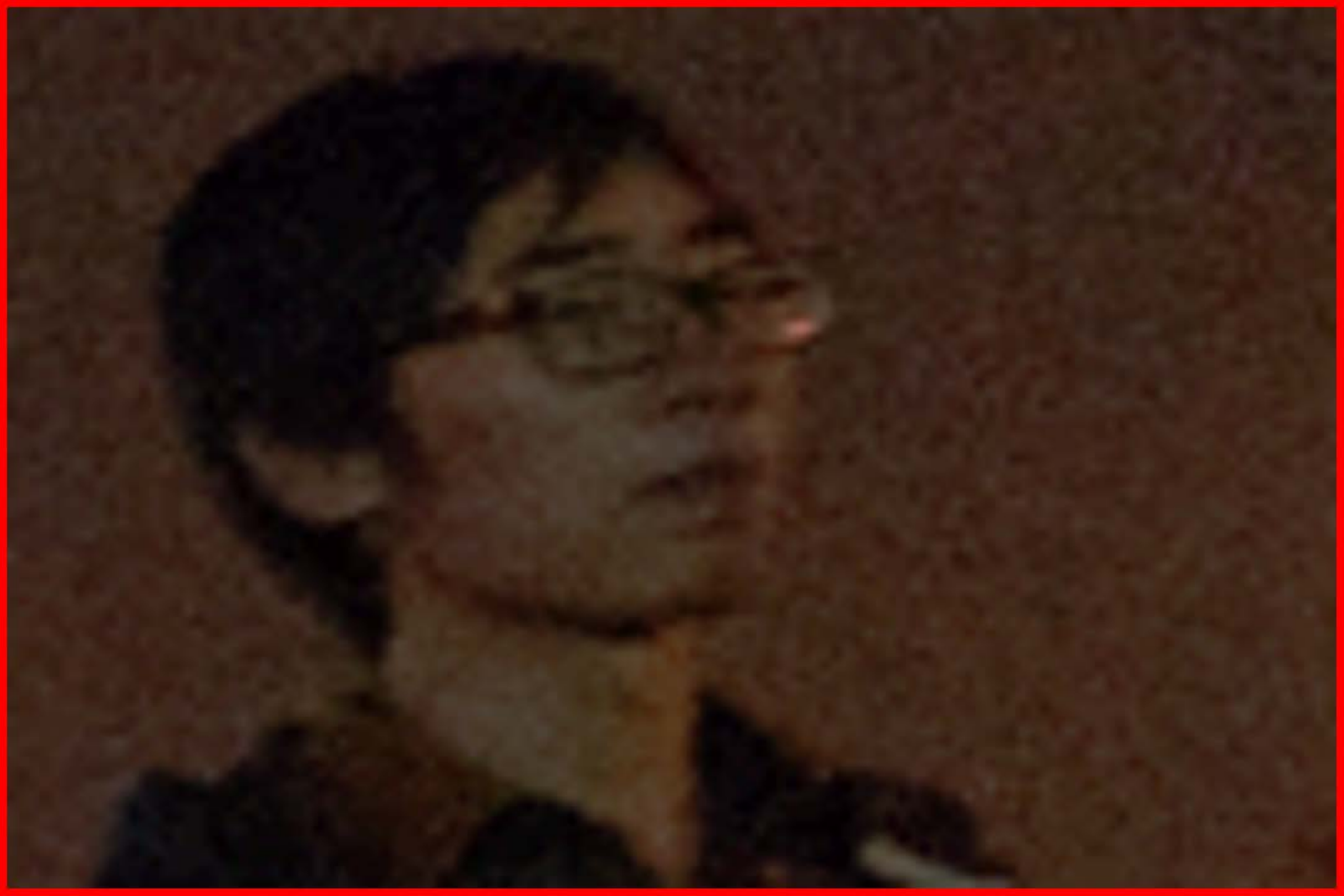}&
			\includegraphics[width=0.192\linewidth]{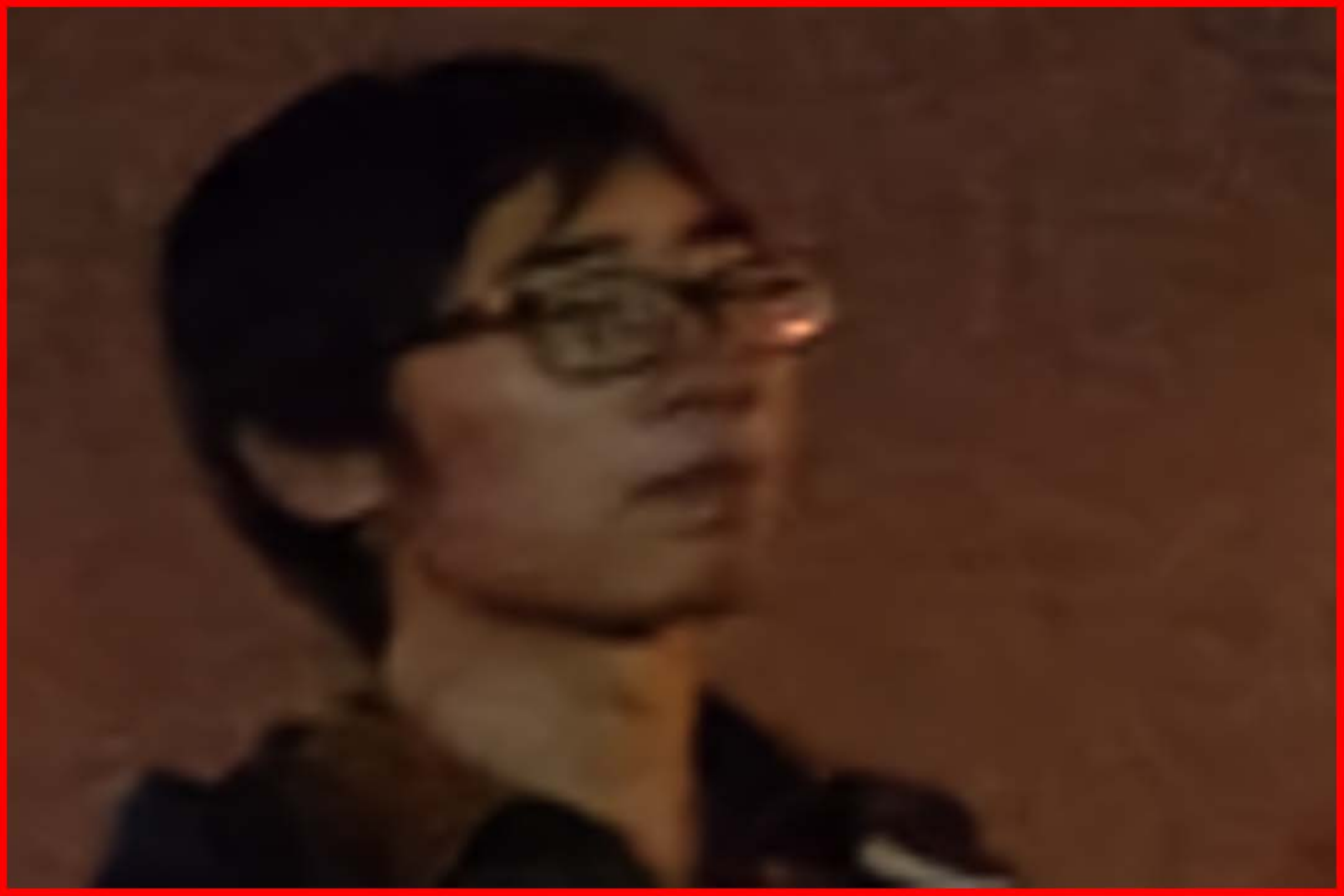}&
			\includegraphics[width=0.192\linewidth]{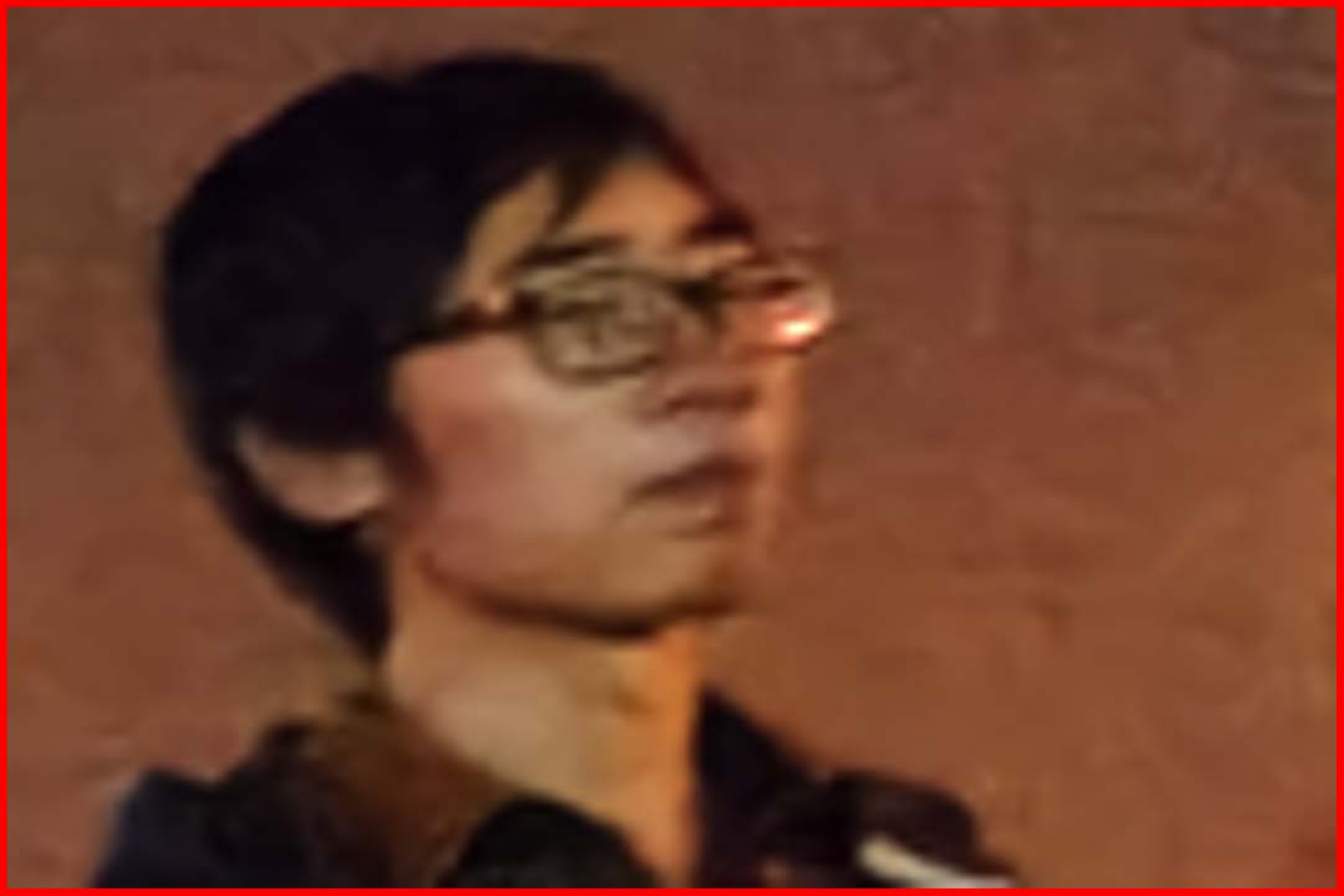}\\
			\footnotesize RUAS&\footnotesize UTVNet&\footnotesize SCL&\footnotesize \textbf{BL}&\footnotesize \textbf{RBL}\\
			\includegraphics[width=0.192\linewidth]{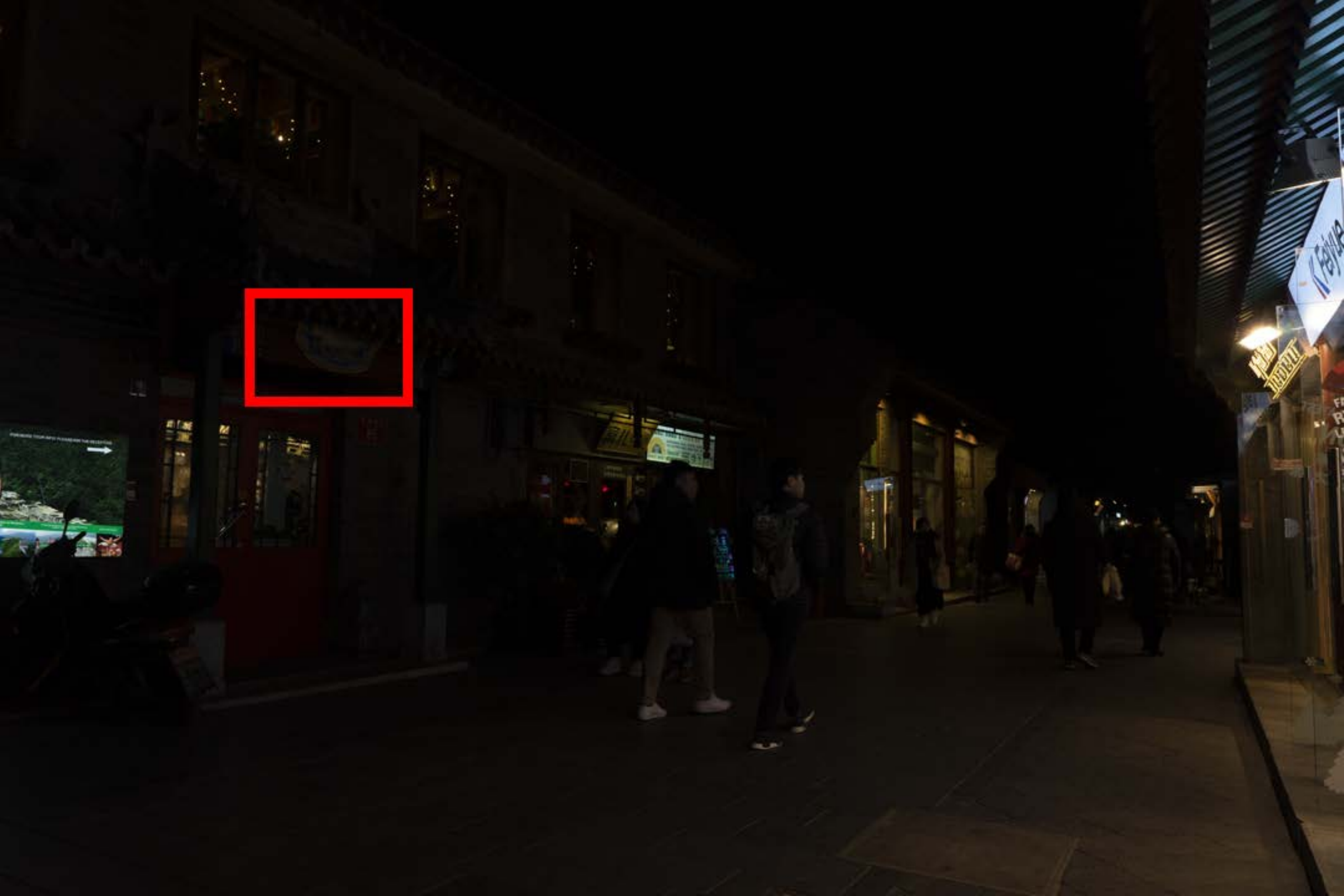}&
			\includegraphics[width=0.192\linewidth]{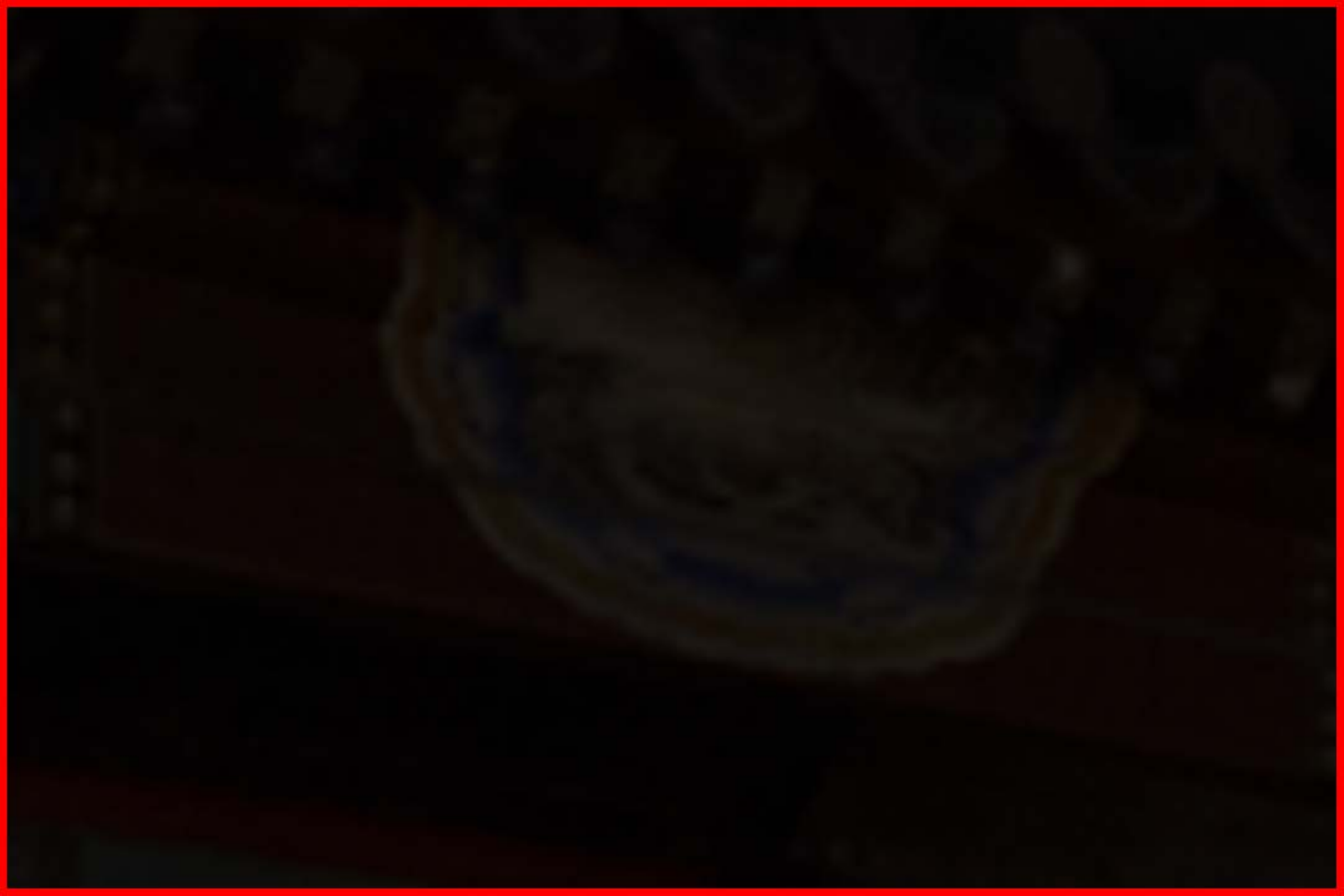}&
			\includegraphics[width=0.192\linewidth]{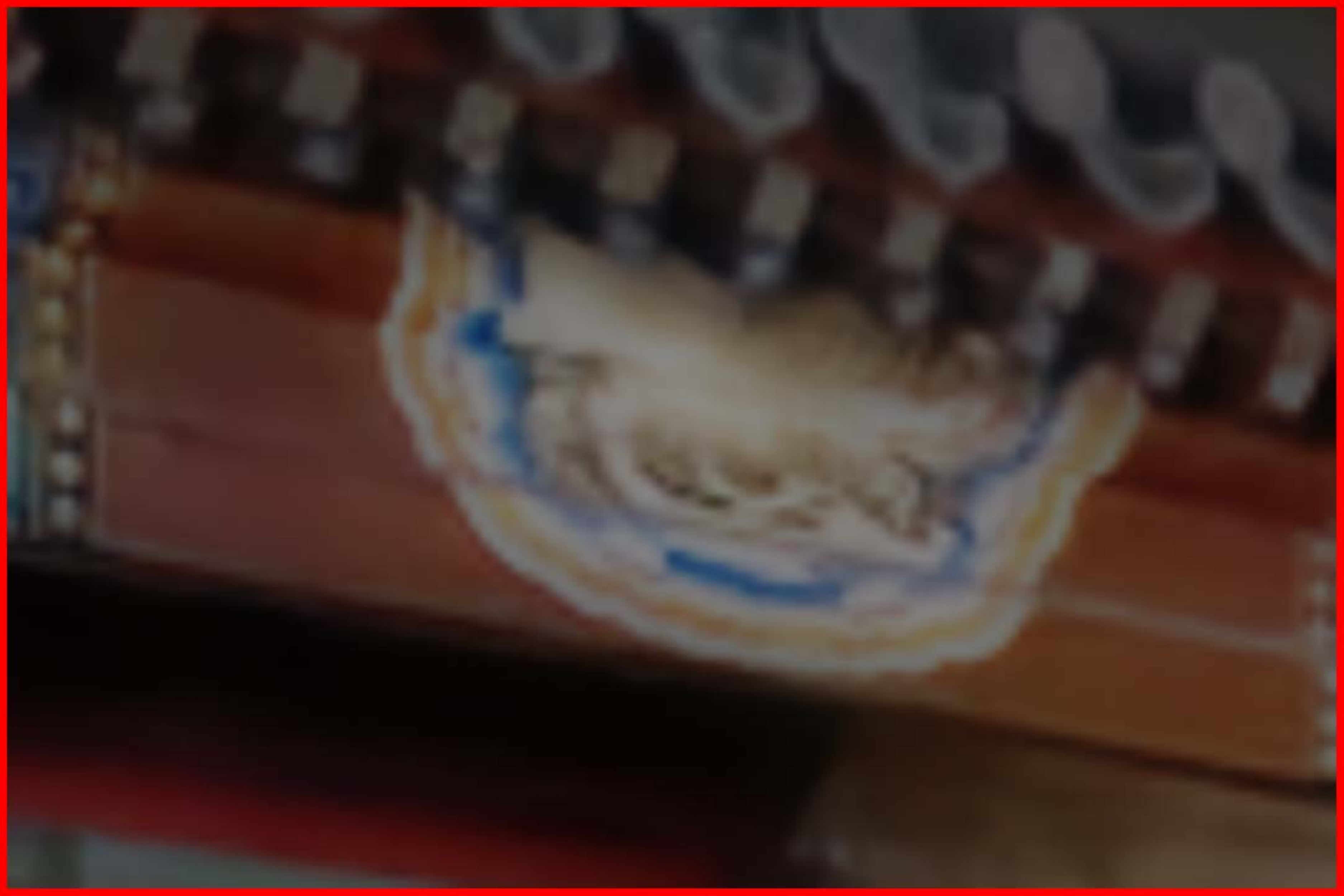}&
			\includegraphics[width=0.192\linewidth]{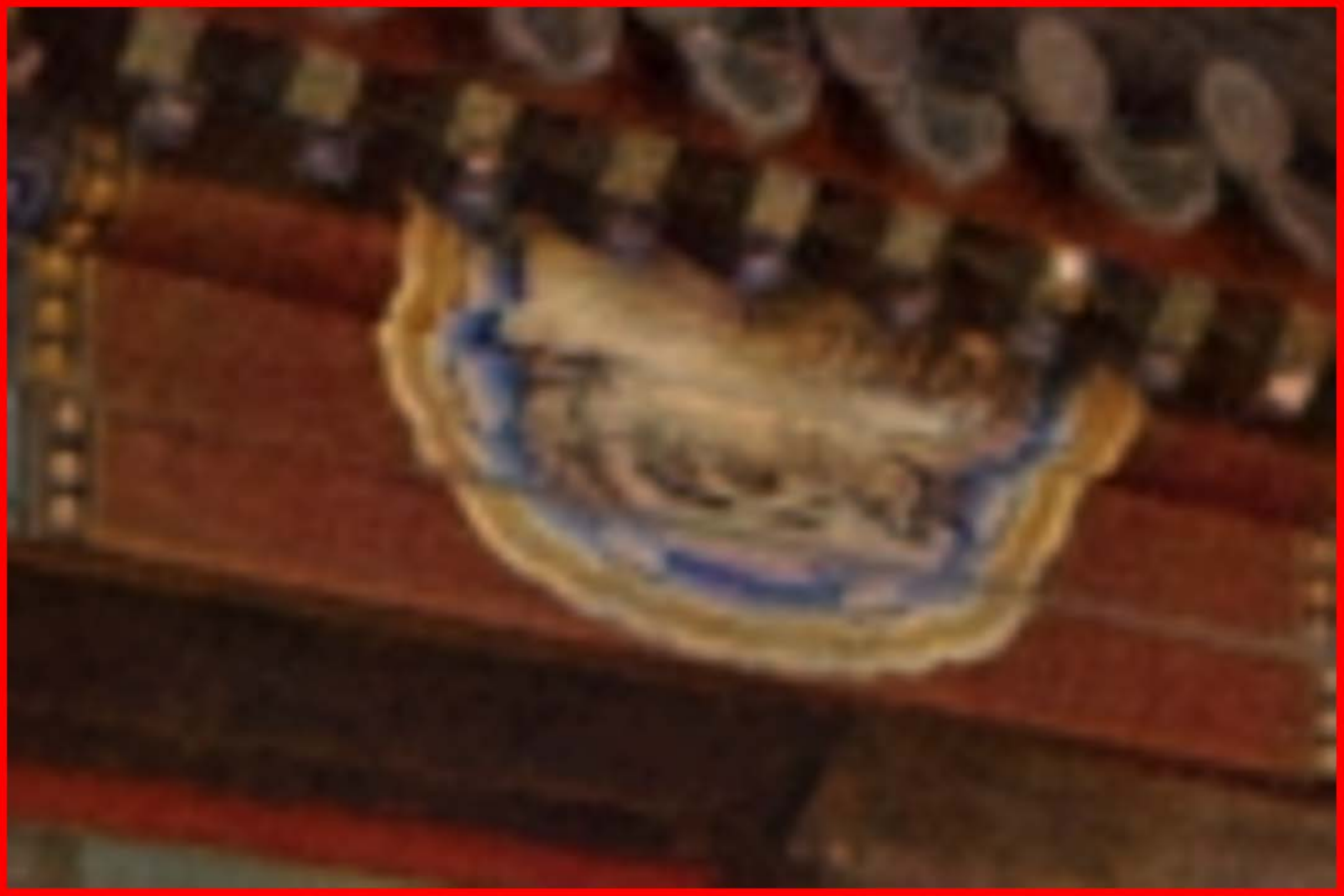}&
			\includegraphics[width=0.192\linewidth]{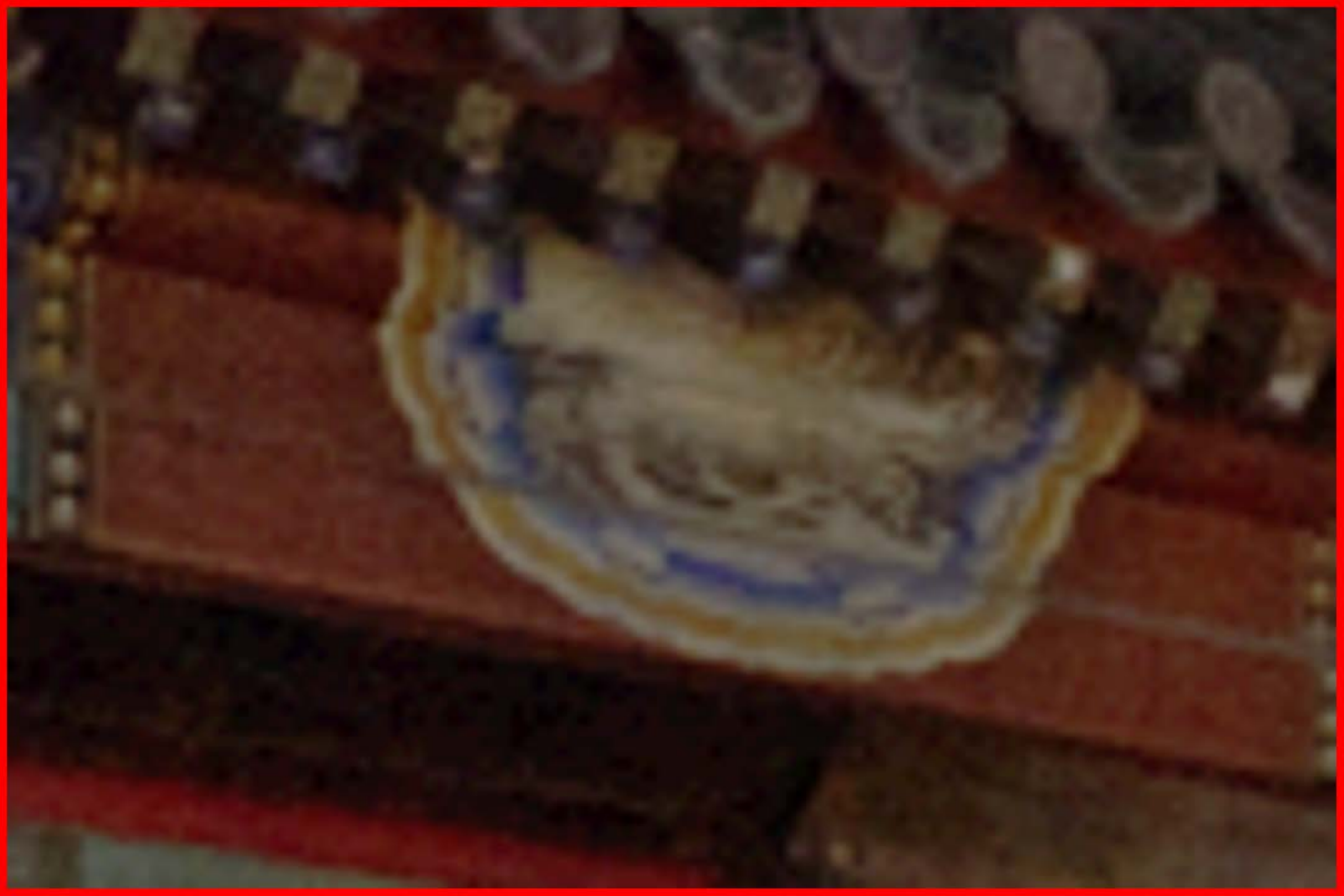}\\
			\footnotesize Input (Full Size)&\footnotesize Input&\footnotesize KinD&\footnotesize EnGAN&\footnotesize ZeroDCE\\
			\includegraphics[width=0.192\linewidth]{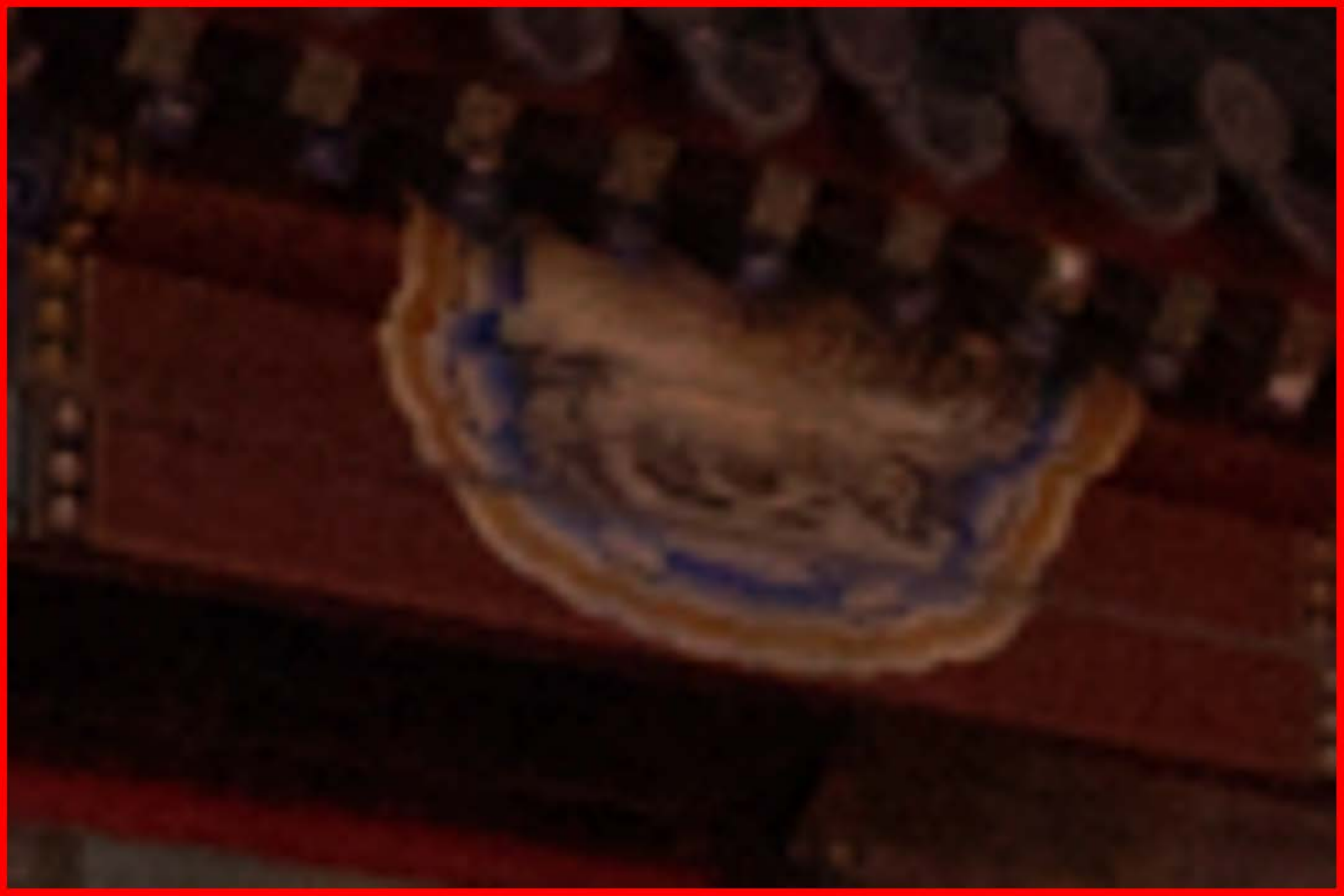}&
			\includegraphics[width=0.192\linewidth]{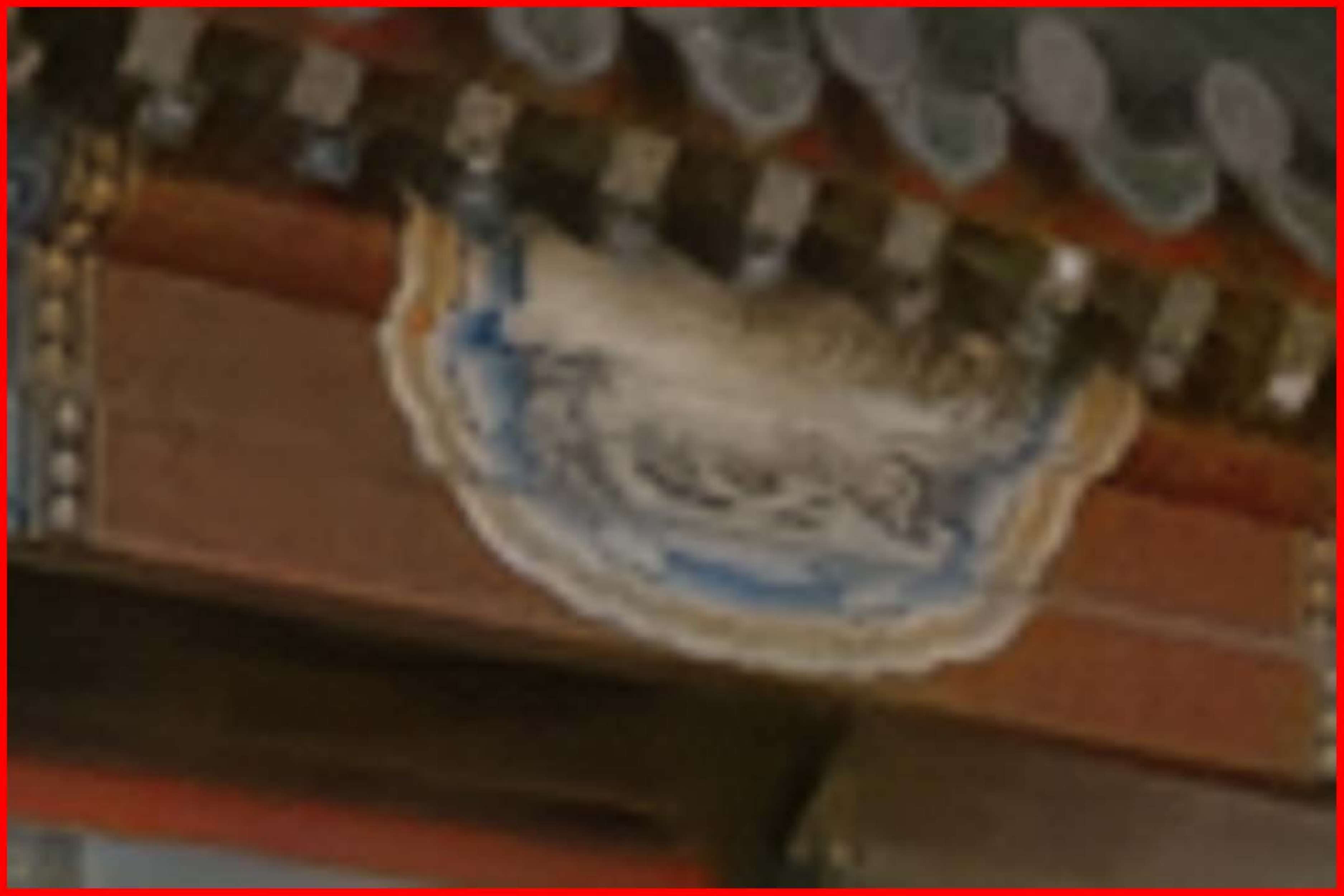}&
			\includegraphics[width=0.192\linewidth]{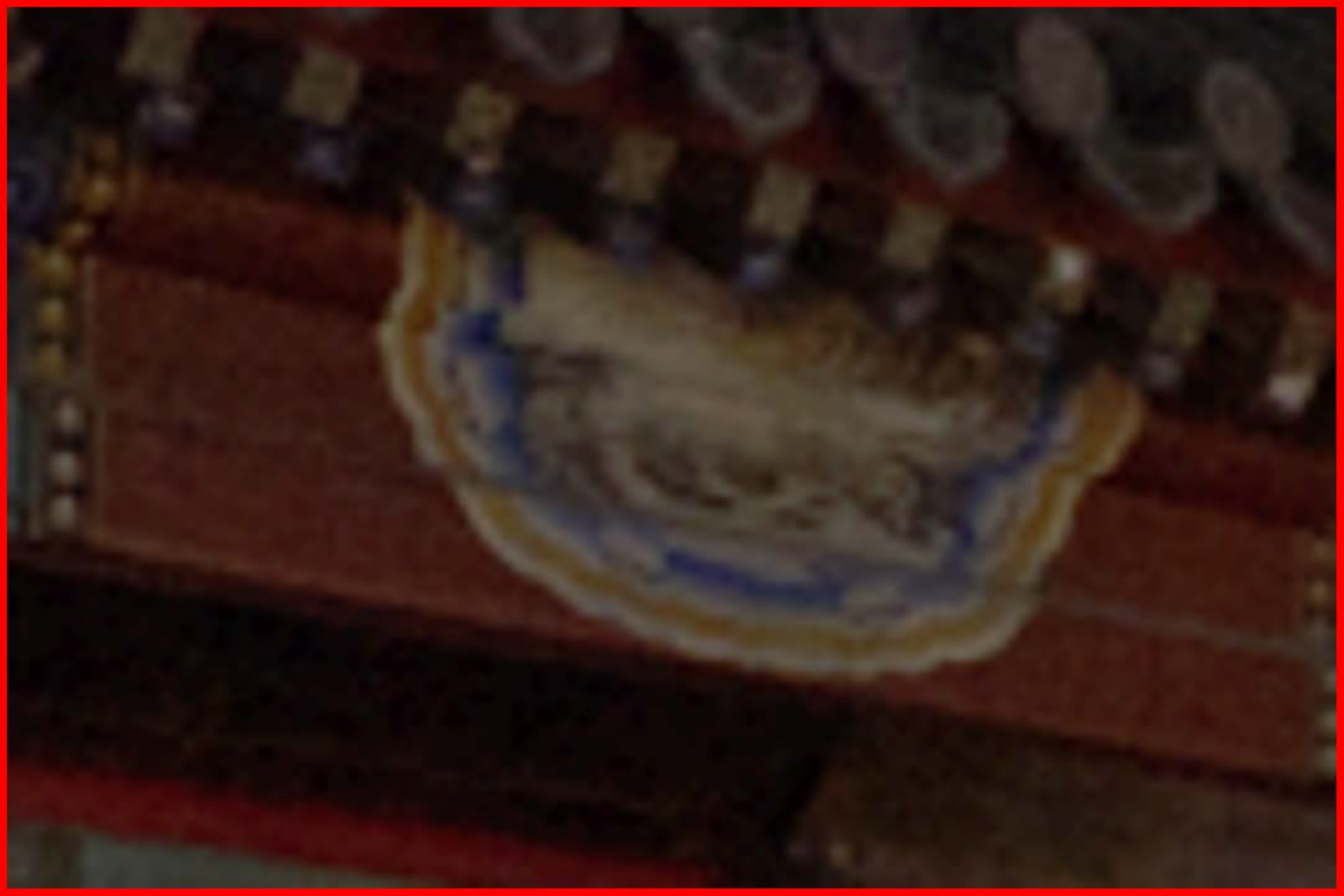}&
			\includegraphics[width=0.192\linewidth]{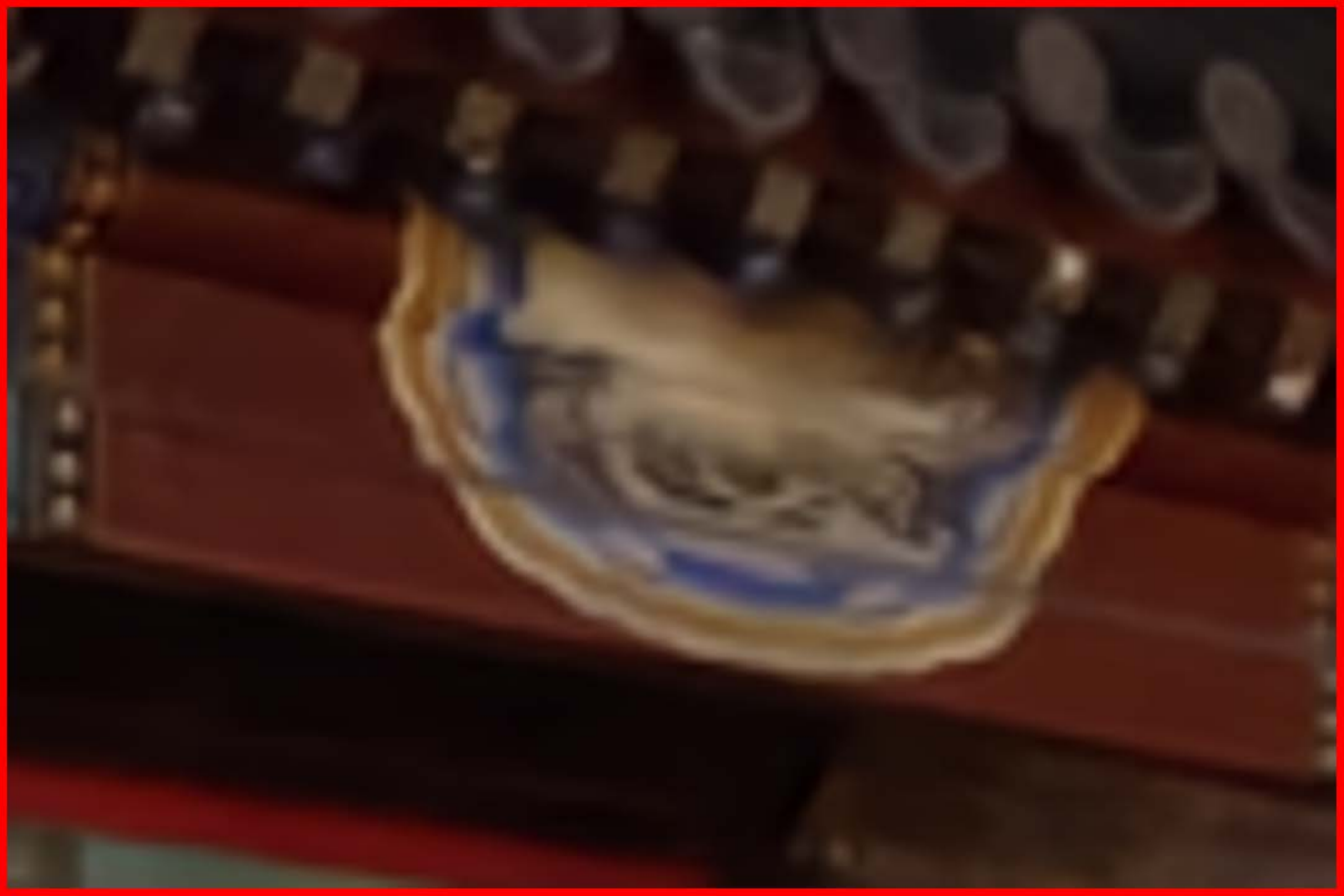}&
			\includegraphics[width=0.192\linewidth]{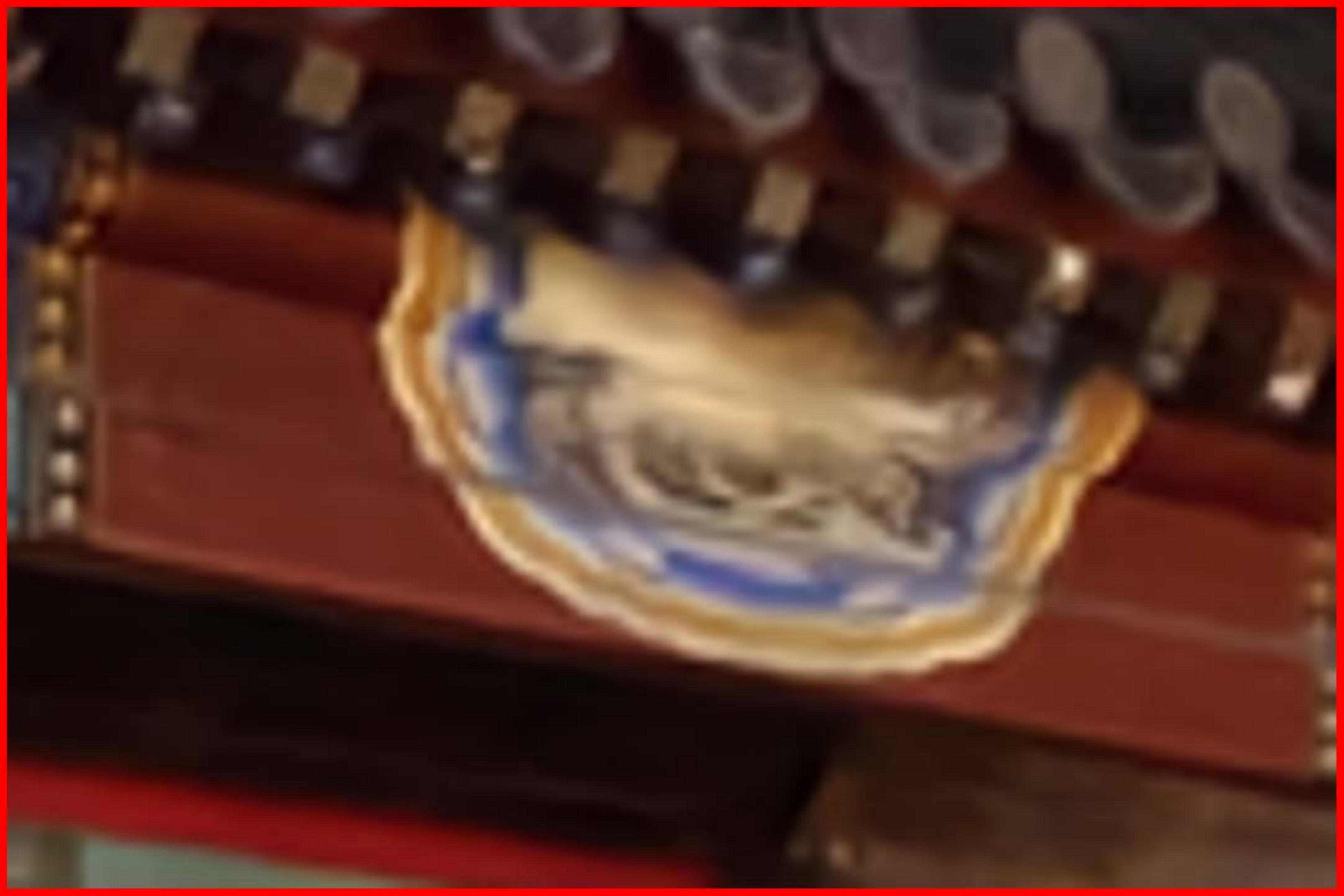}\\
			\footnotesize RUAS&\footnotesize UTVNet&\footnotesize SCL&\footnotesize \textbf{BL}&\footnotesize \textbf{RBL}\\
			\includegraphics[width=0.192\linewidth]{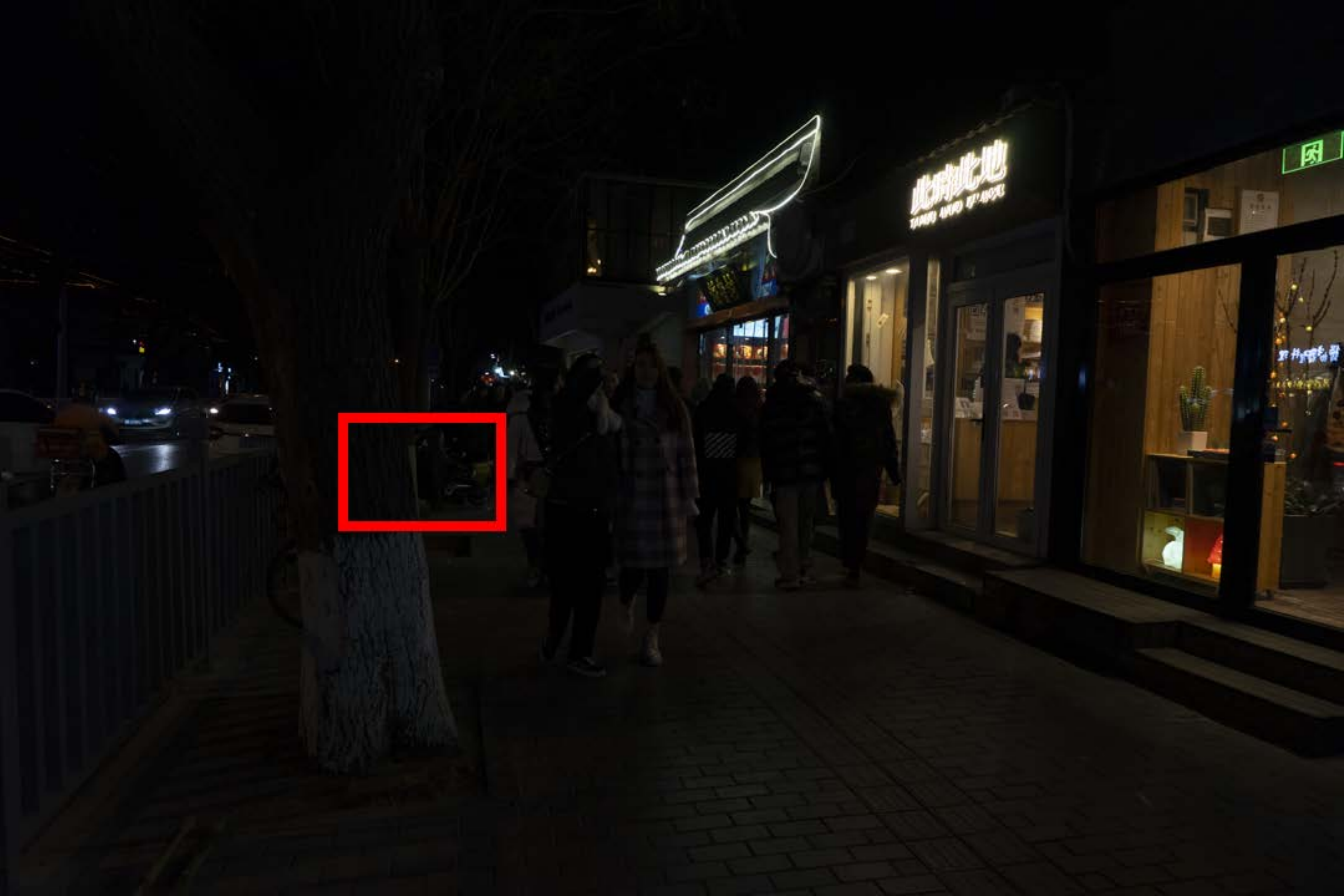}&
			\includegraphics[width=0.192\linewidth]{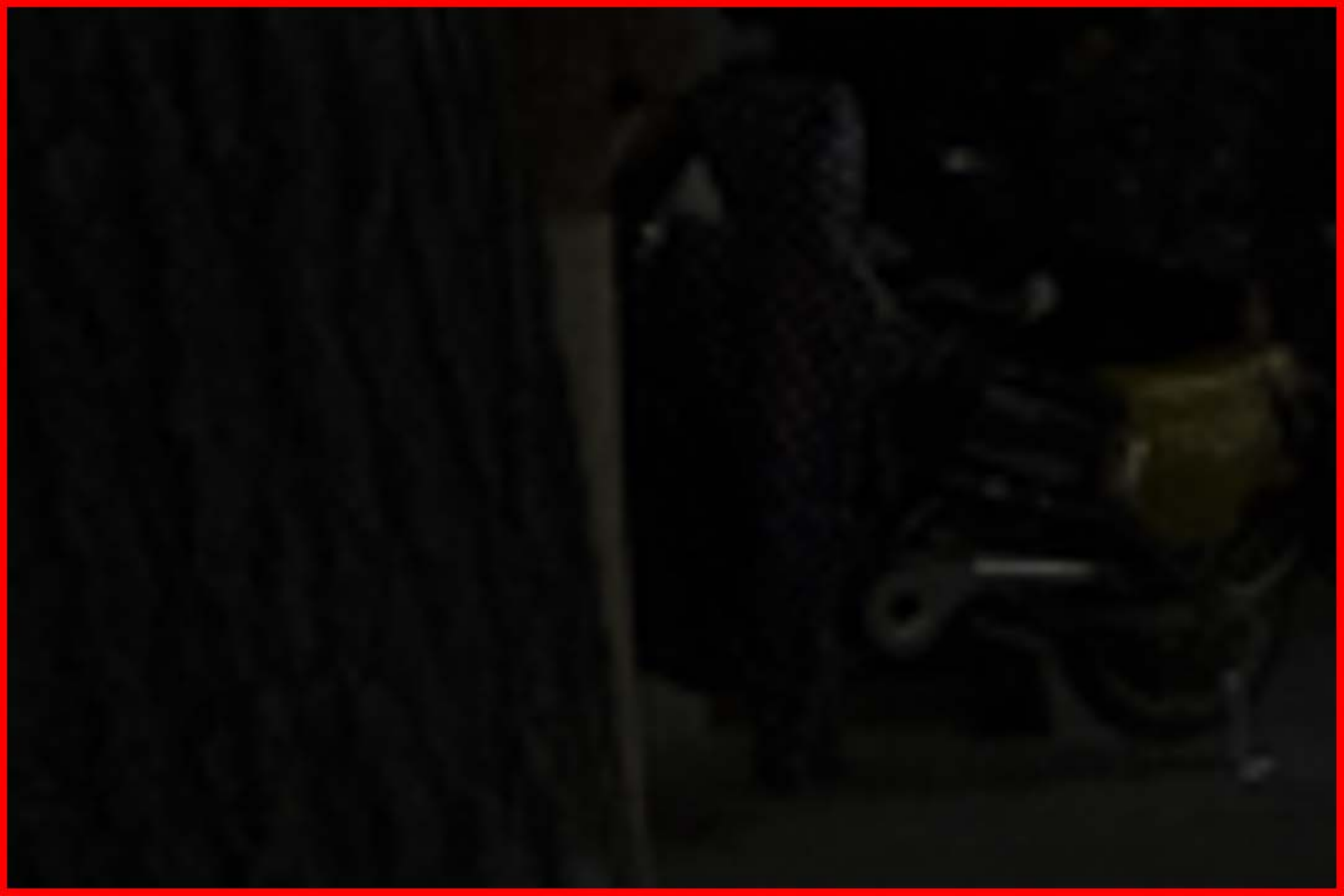}&
			\includegraphics[width=0.192\linewidth]{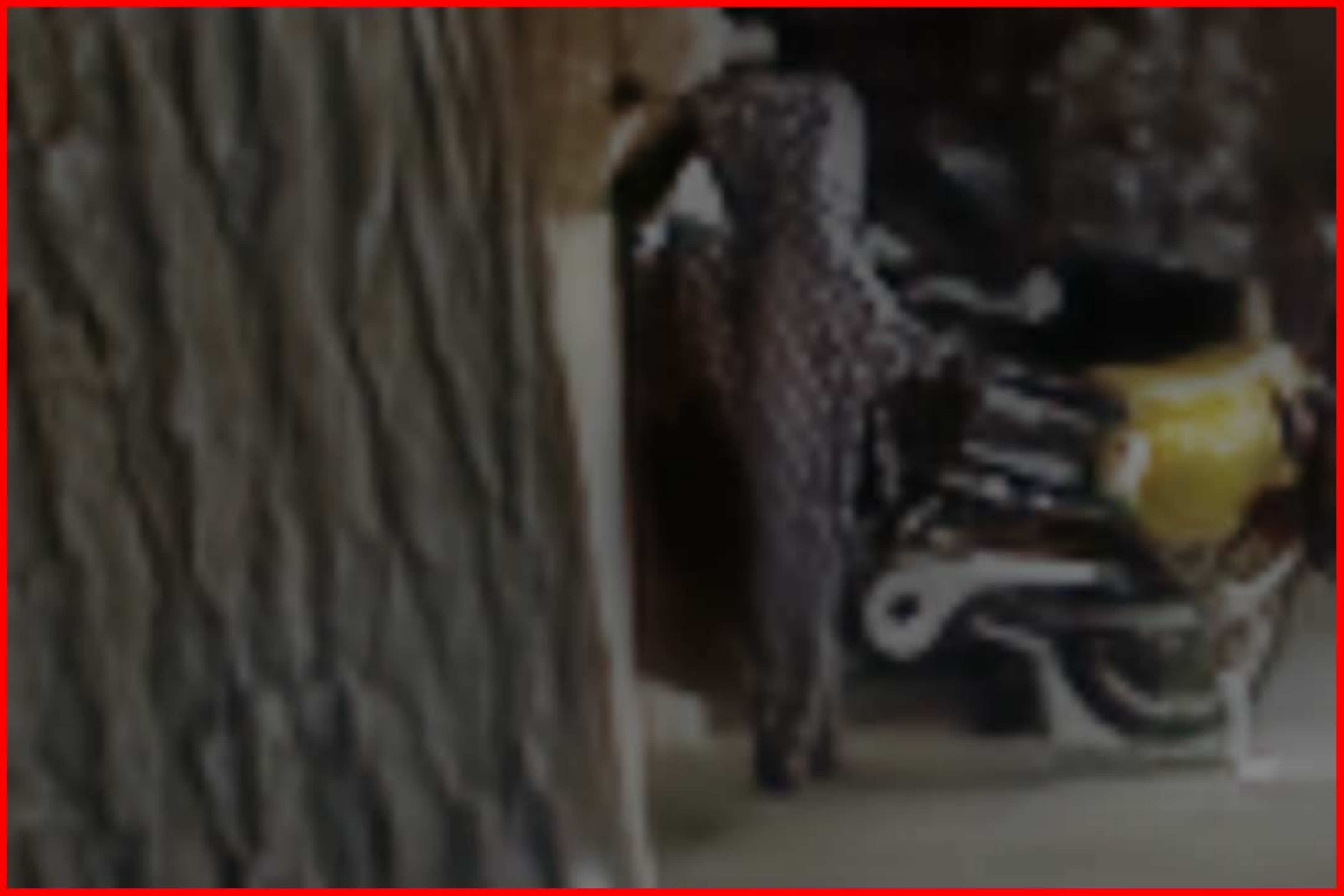}&
			\includegraphics[width=0.192\linewidth]{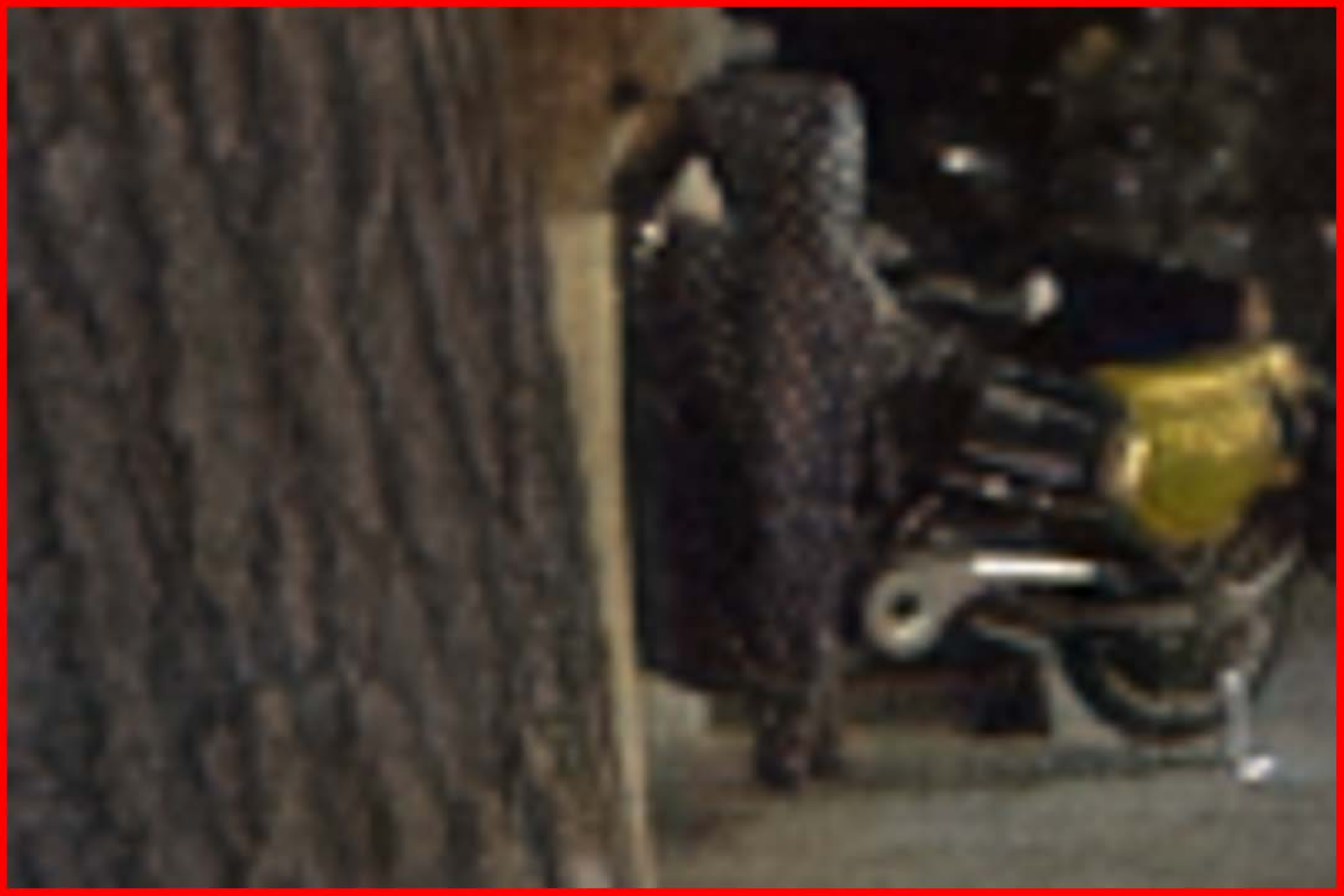}&
			\includegraphics[width=0.192\linewidth]{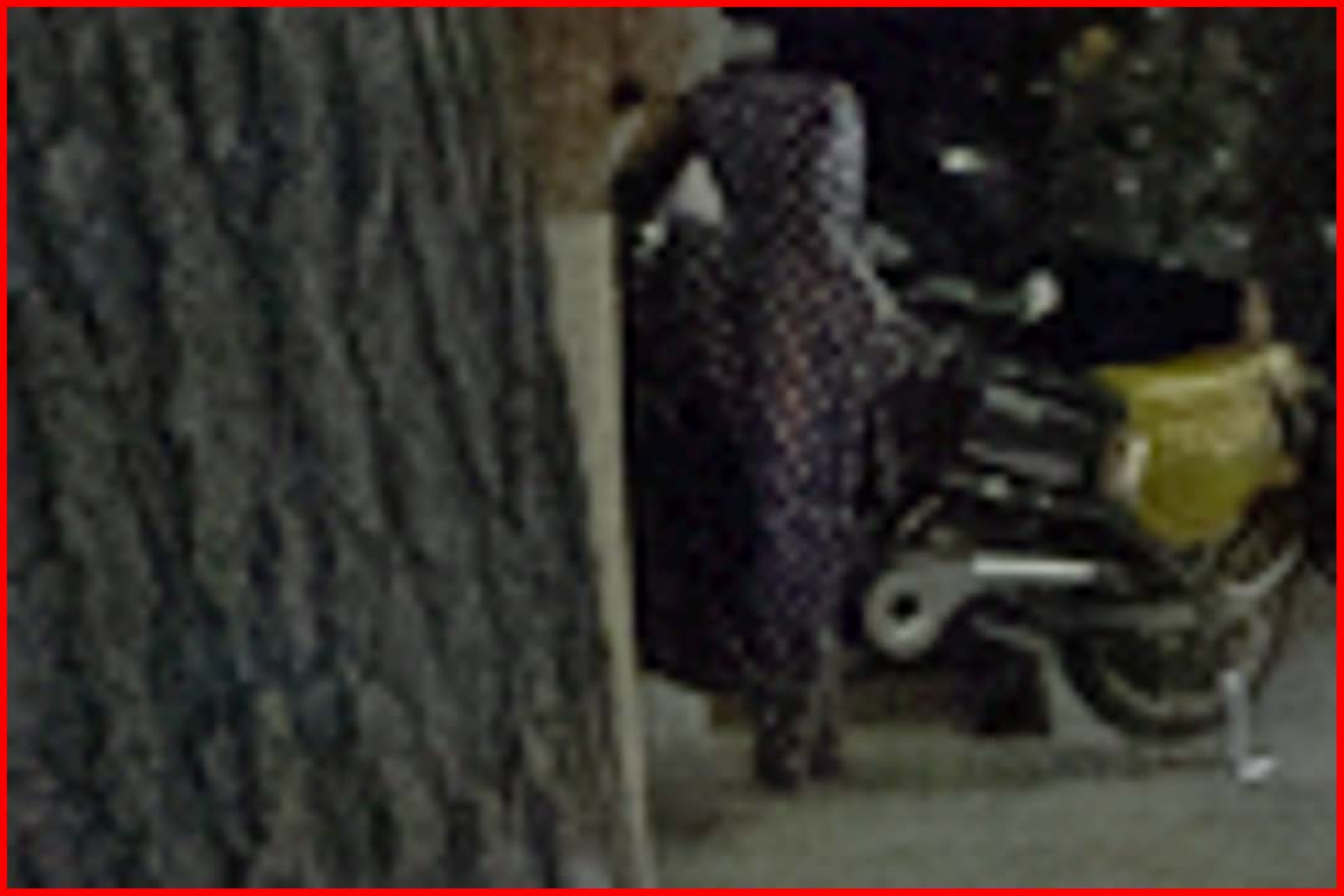}\\
			\footnotesize Input (Full Size)&\footnotesize Input&\footnotesize KinD&\footnotesize EnGAN&\footnotesize ZeroDCE\\
			\includegraphics[width=0.192\linewidth]{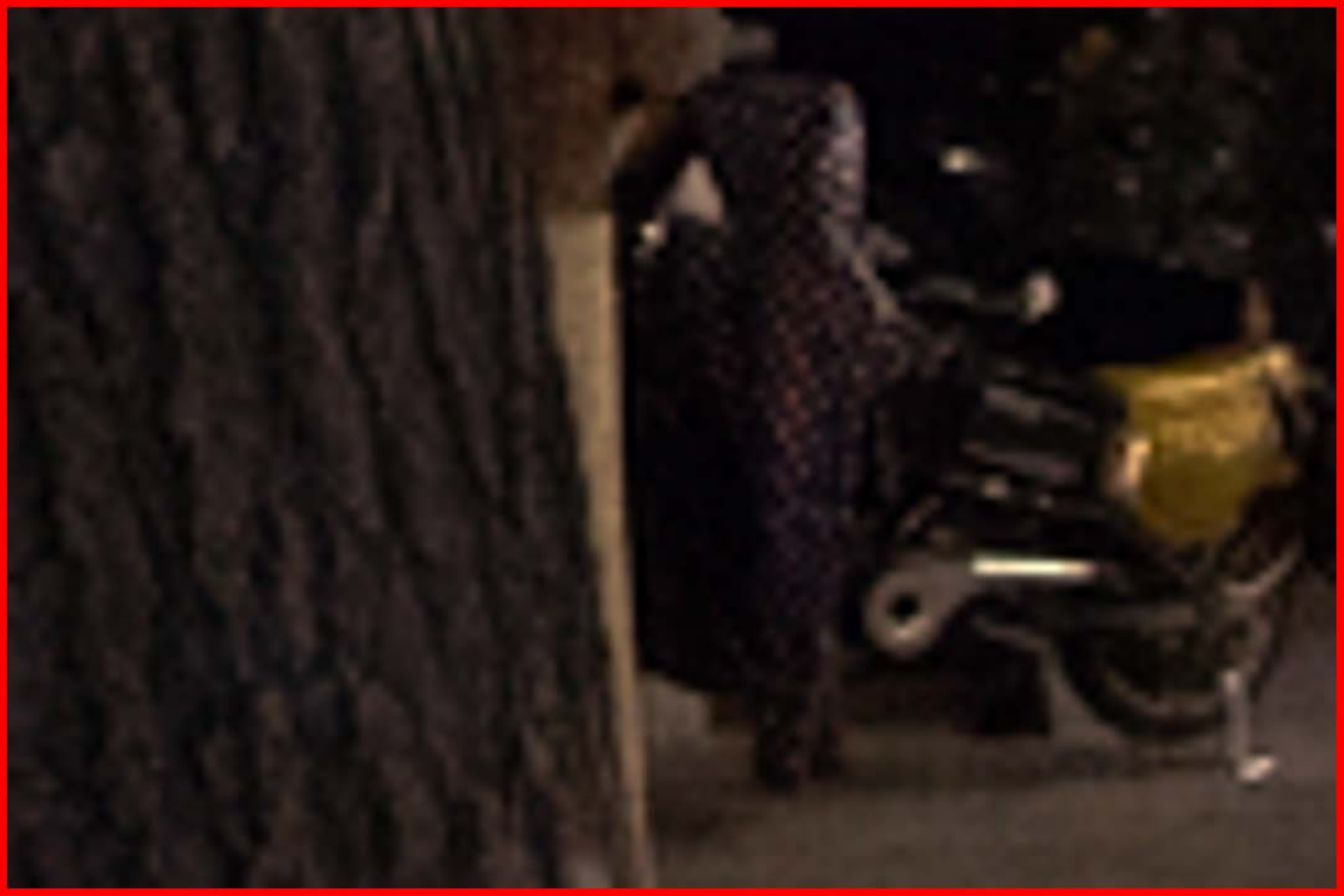}&
			\includegraphics[width=0.192\linewidth]{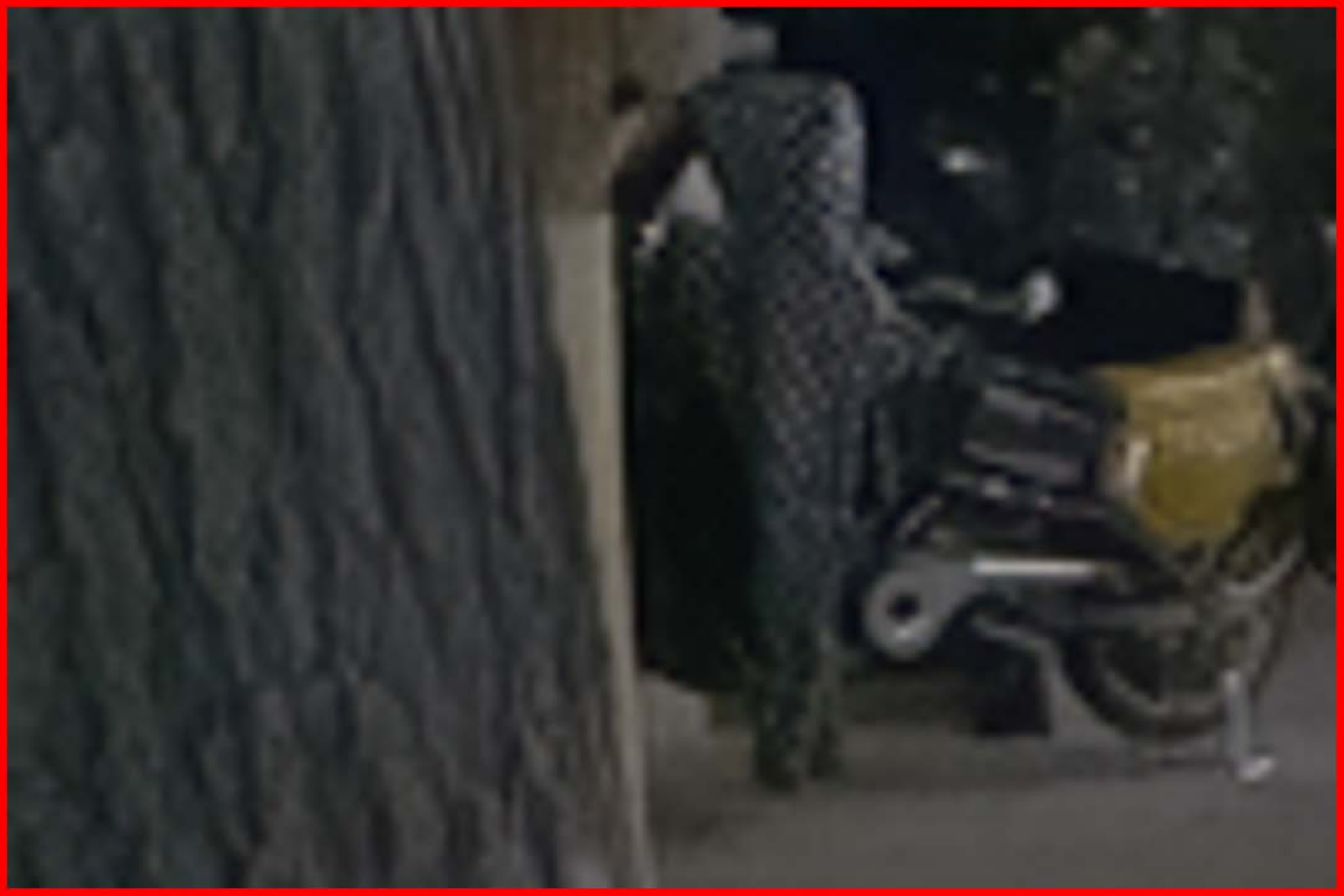}&
			\includegraphics[width=0.192\linewidth]{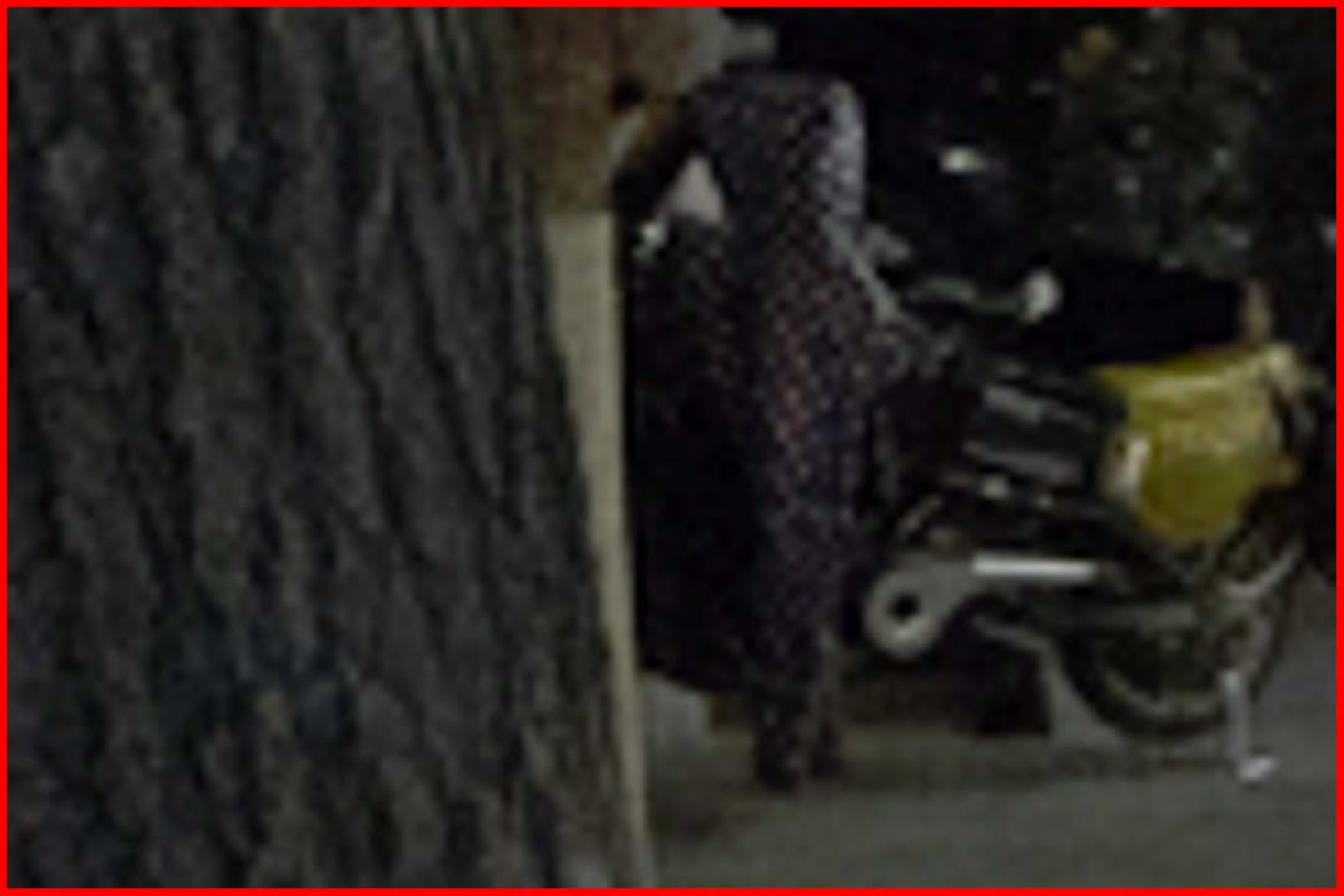}&
			\includegraphics[width=0.192\linewidth]{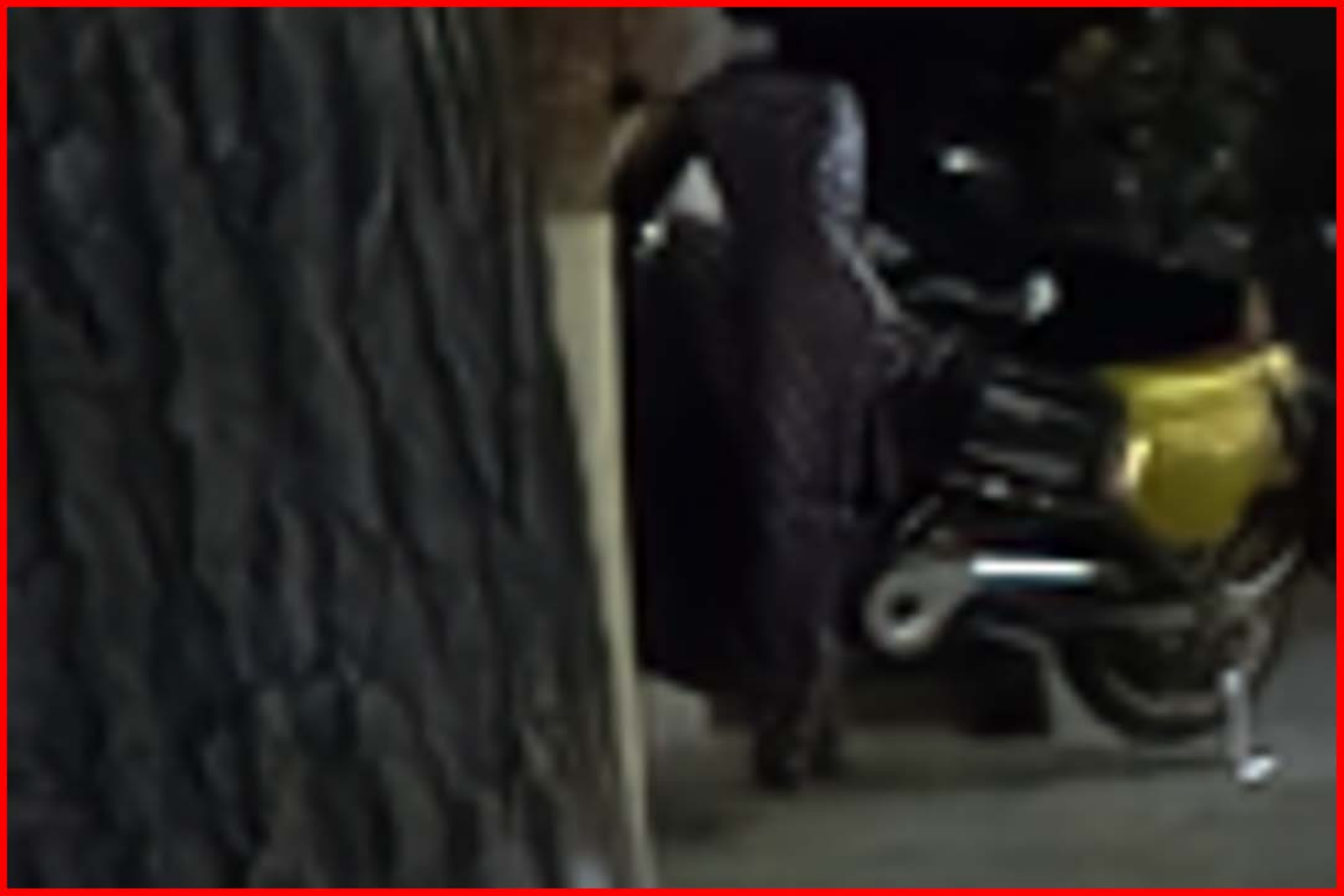}&
			\includegraphics[width=0.192\linewidth]{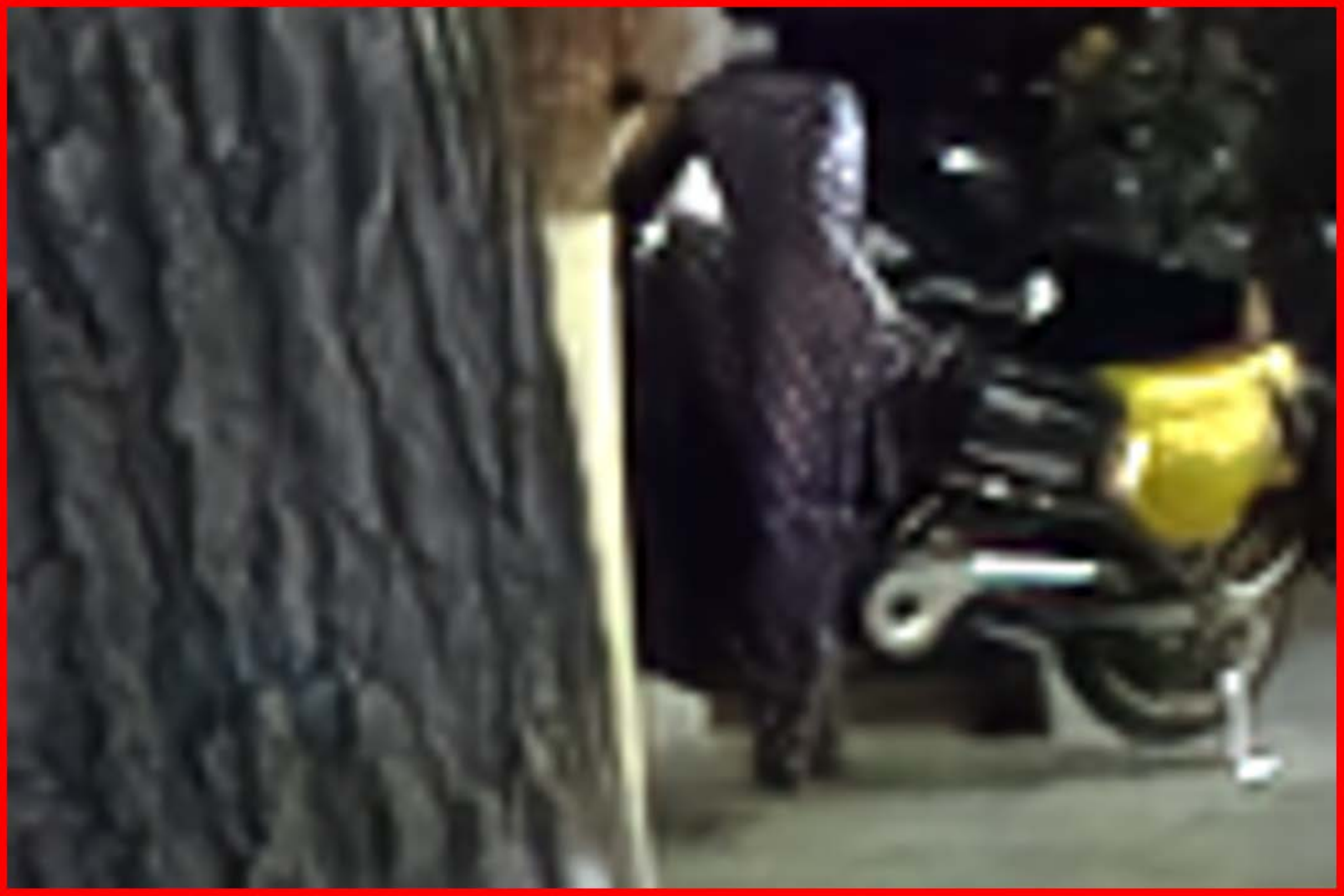}\\
			\footnotesize RUAS&\footnotesize UTVNet&\footnotesize SCL&\footnotesize \textbf{BL}&\footnotesize \textbf{RBL}\\
		\end{tabular}
		\caption{Visual results of state-of-the-art methods and our two versions (BL and RBL) on the DARKFACE dataset.}
		\label{fig: DARKFACE}
	\end{figure*}

	\begin{figure*}[t]
		\centering
		\begin{tabular}{c@{\extracolsep{0.3em}}c@{\extracolsep{0.3em}}c@{\extracolsep{0.3em}}c@{\extracolsep{0.3em}}c@{\extracolsep{0.3em}}c@{\extracolsep{0.3em}}c}
			\includegraphics[width=0.135\linewidth]{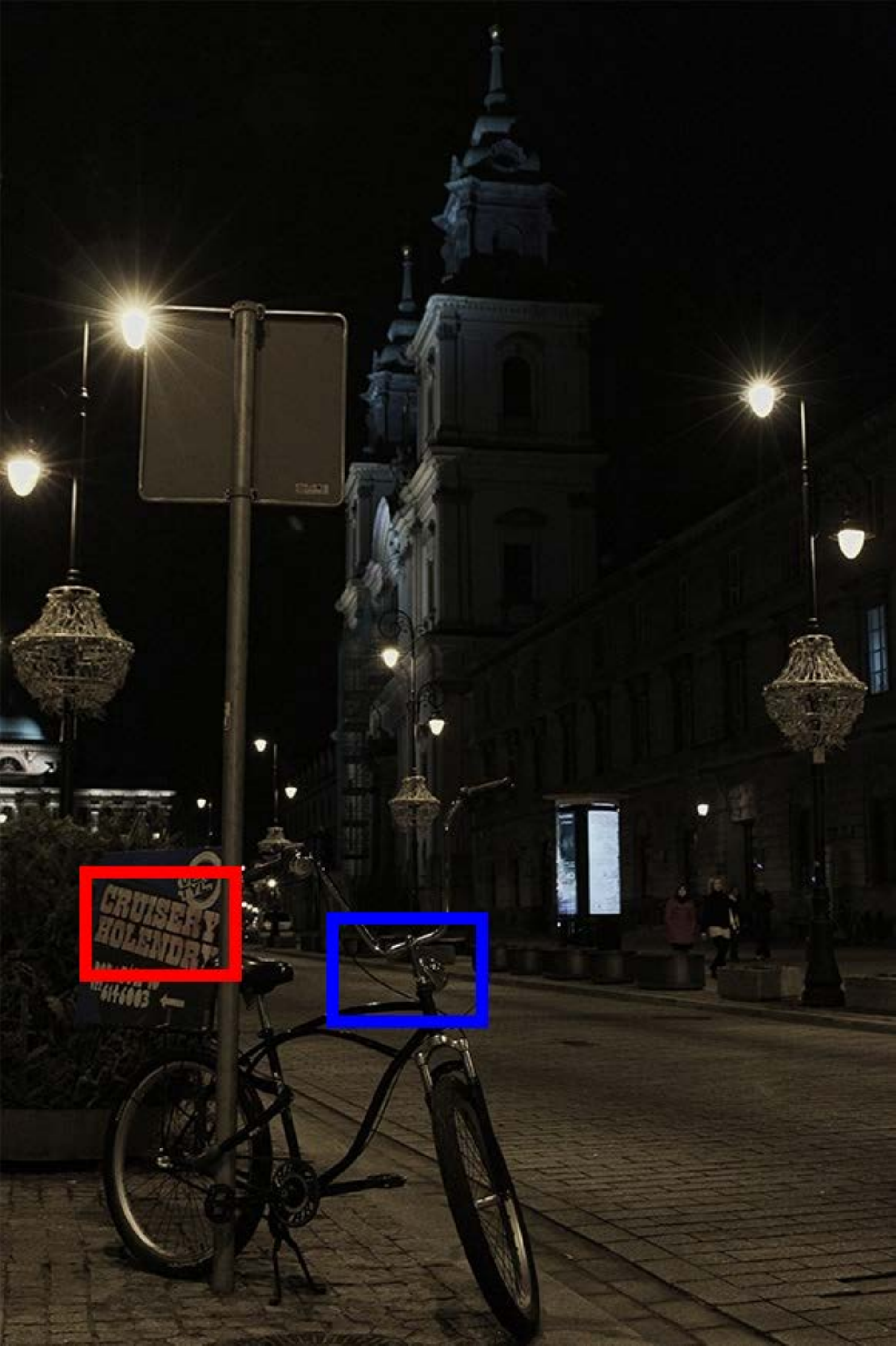}&
			\includegraphics[width=0.135\linewidth]{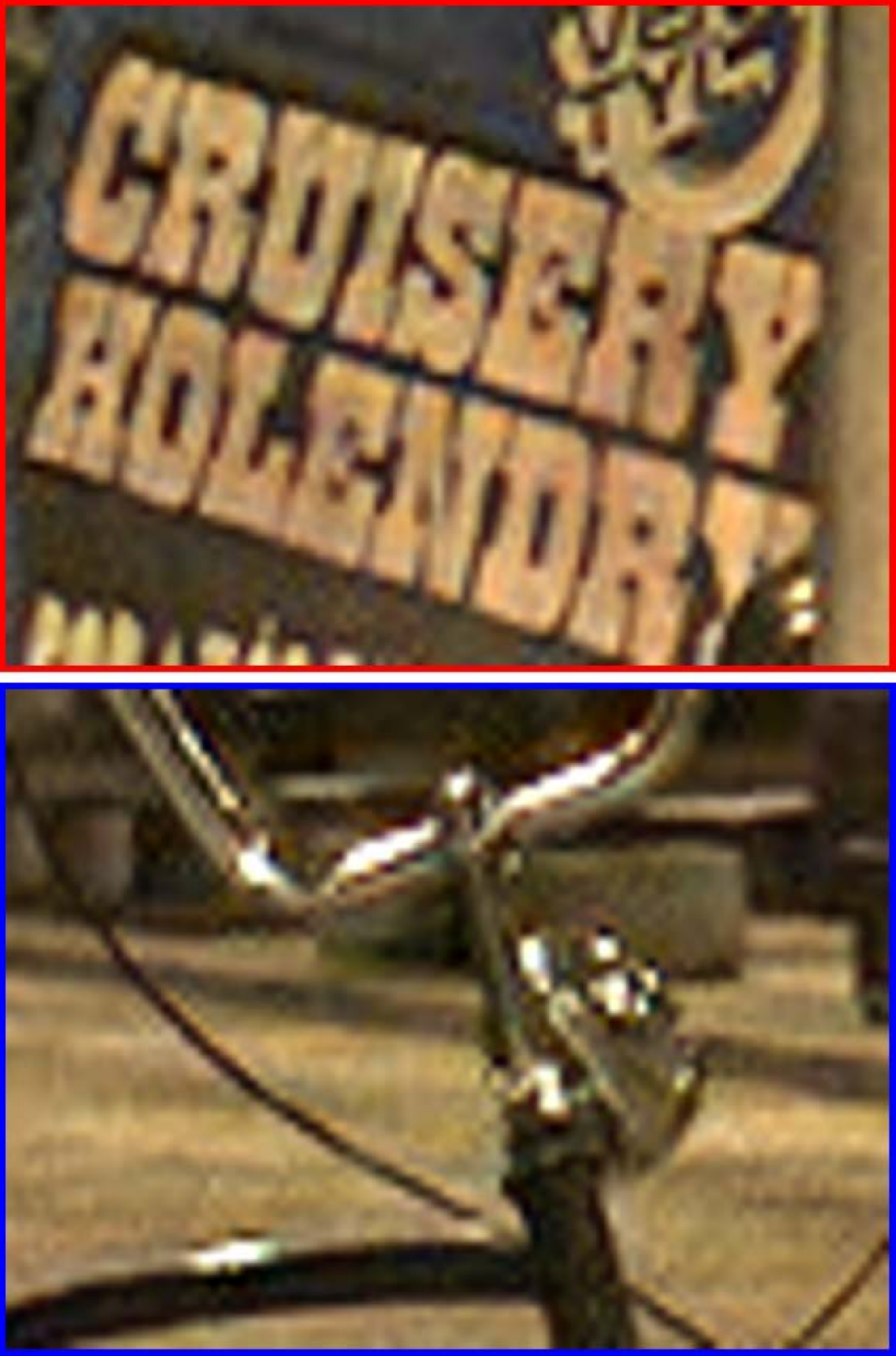}&
			\includegraphics[width=0.135\linewidth]{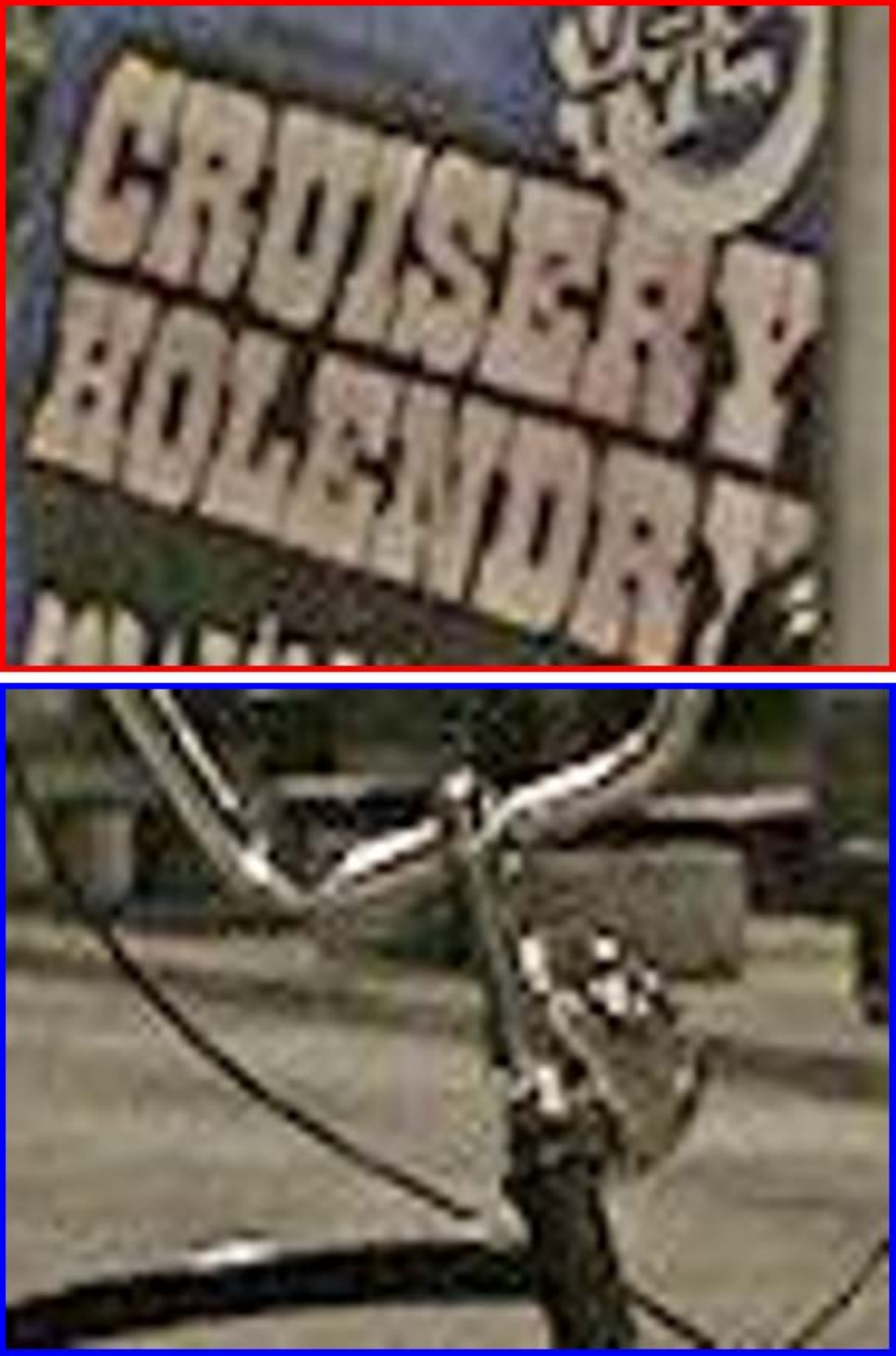}&
			\includegraphics[width=0.135\linewidth]{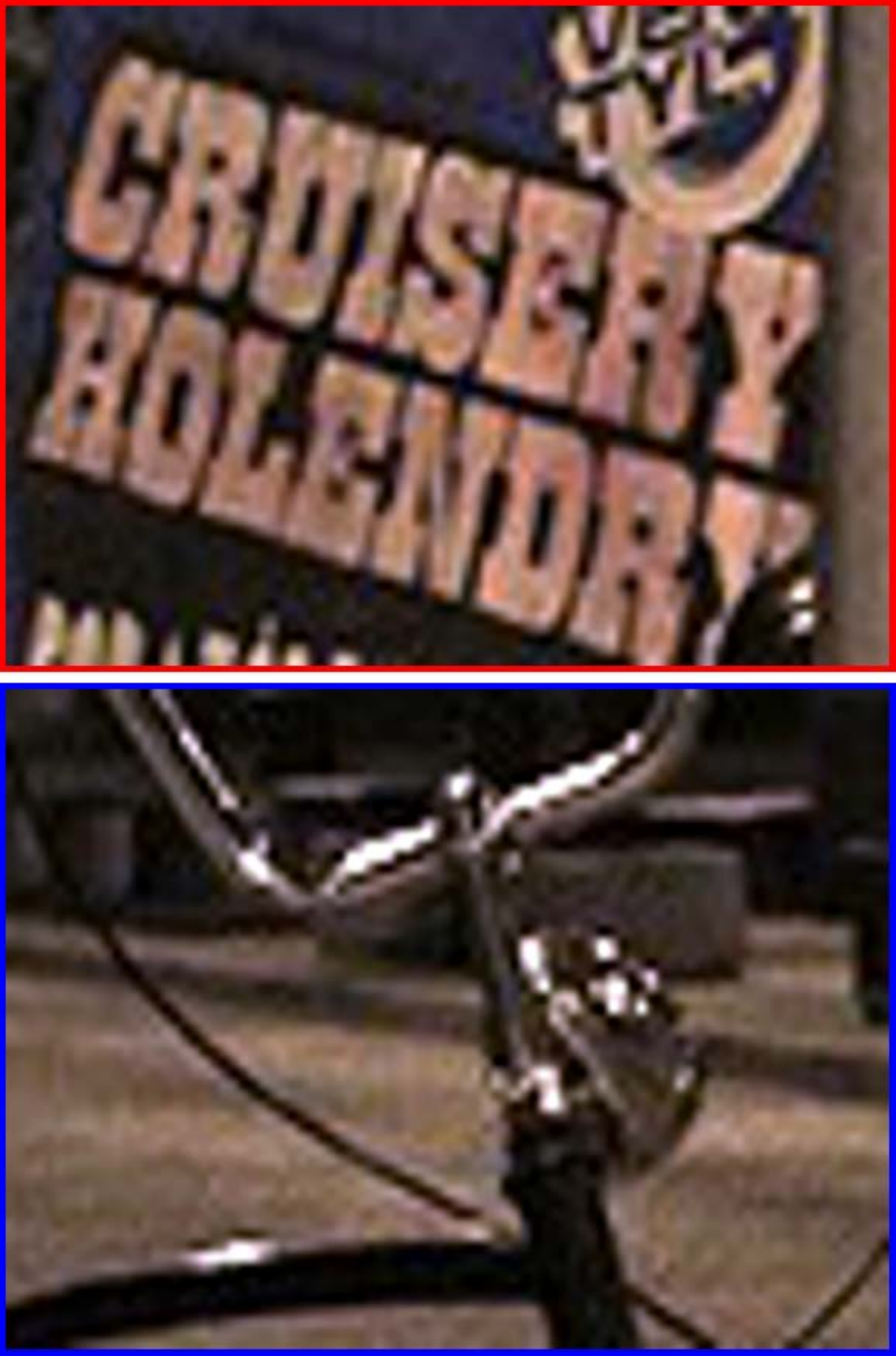}&
			\includegraphics[width=0.135\linewidth]{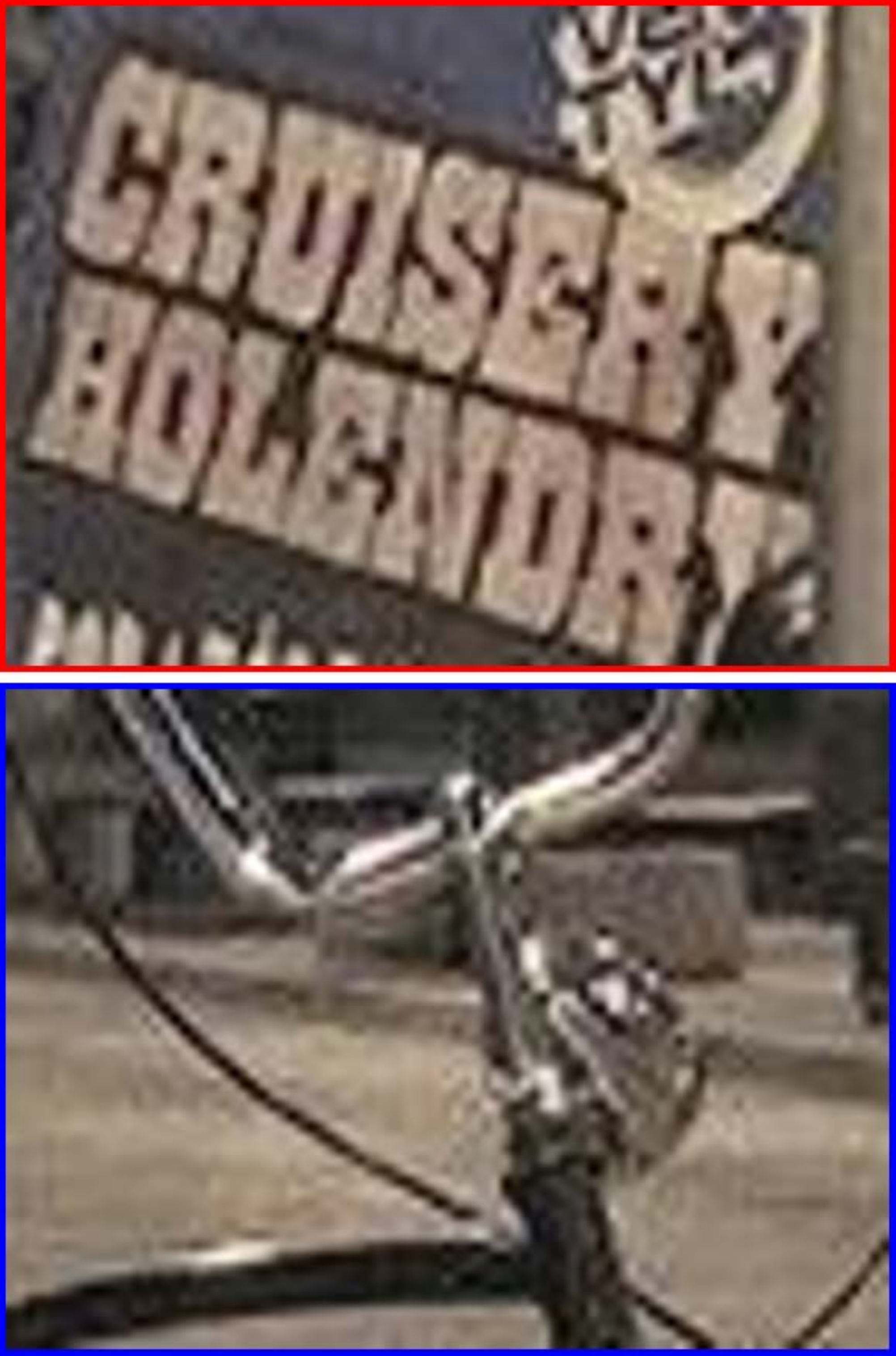}&
			\includegraphics[width=0.135\linewidth]{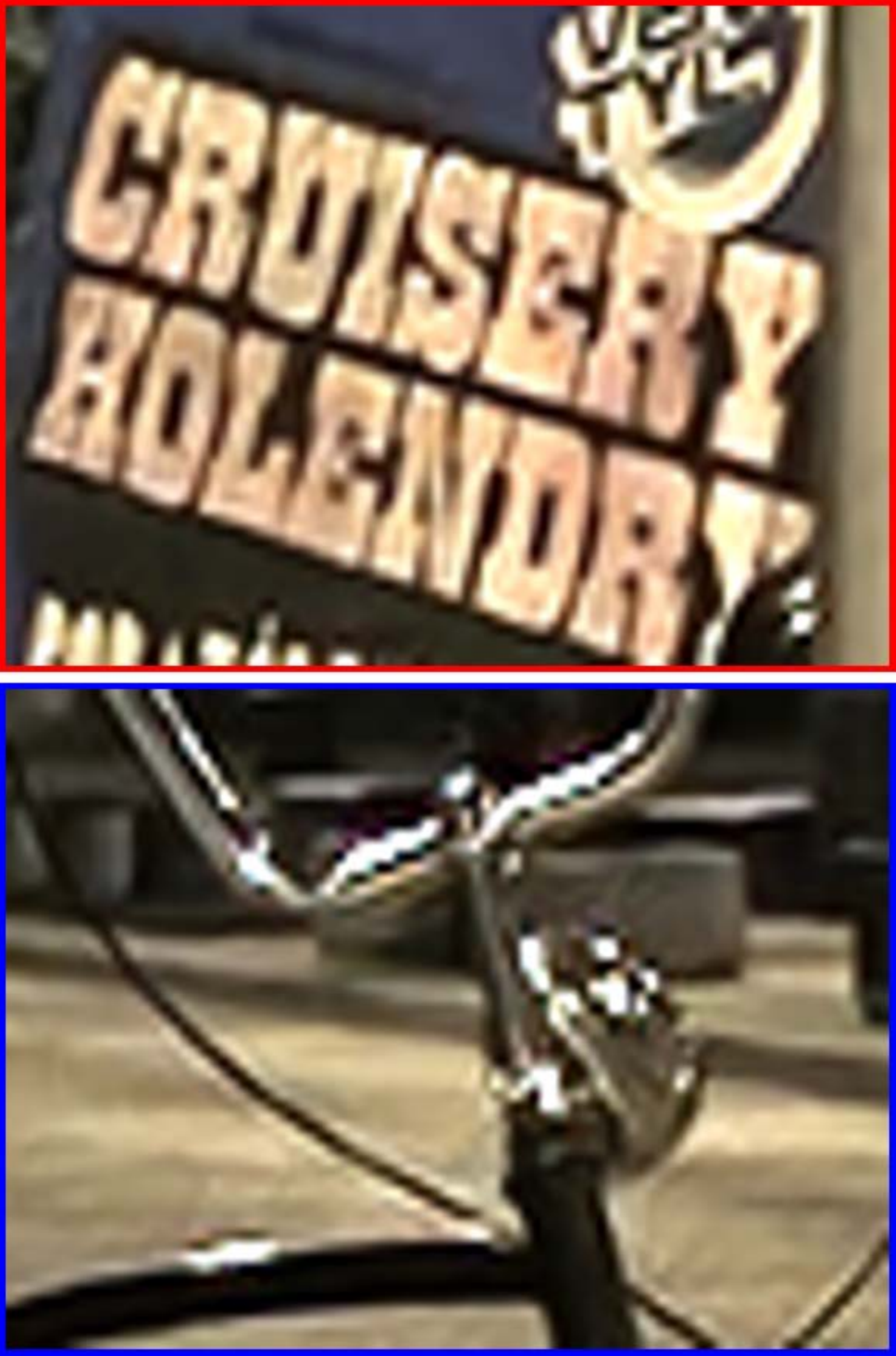}&
			\includegraphics[width=0.135\linewidth]{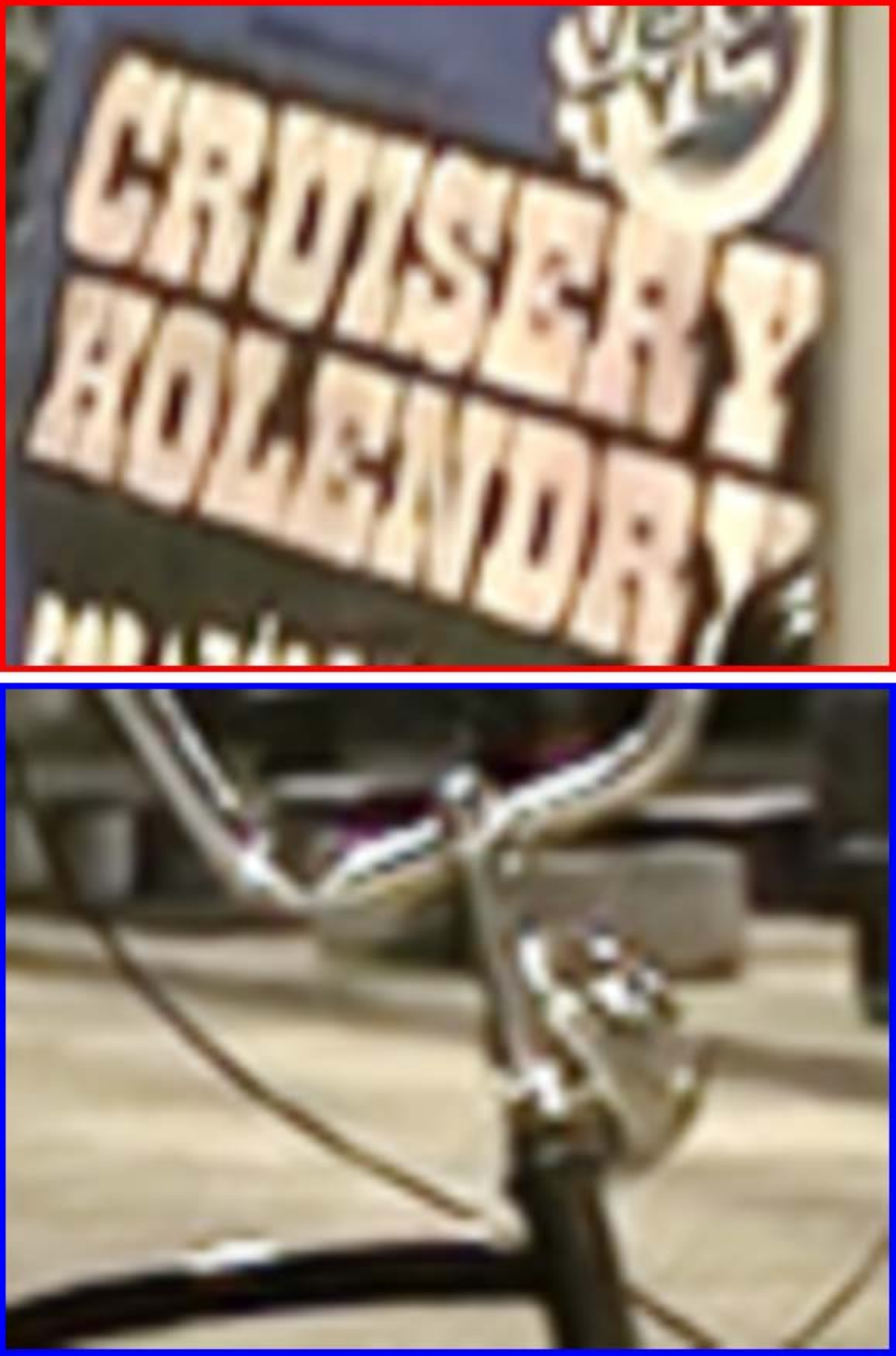}\\
			\footnotesize Input&\footnotesize EnGAN&\footnotesize ZeroDCE&\footnotesize RUAS&\footnotesize UTVNet&\footnotesize \textbf{BL}&\footnotesize \textbf{RBL}\\
		\end{tabular}
		\caption{Visual results of state-of-the-art methods and our two versions (BL and RBL) on the ExDARK dataset.}
		\label{fig: ExDARK}
	\end{figure*}

	\subsection{Evaluations on Unseen-Paired Benchmarks}
	To demonstrate the generalization capability of our method, we compared the proposed method with the state-of-the-art approaches on other datasets which contained the reference. It is noteworthy that the quantitative results provided in the first part differed from the conventional setting, as the testing phase utilized completely new data with a distinct distribution compared to the learning phase.
	
	\textbf{Quantitative comparison.}$\;$ 
	As shown in Table~\ref{table: LSRW_VOC}, although testing on unseen scenarios had somewhat affected the advantages of our method, it still maintained a competitive performance. Furthermore, regarding the results of our two proposed versions, the RBL outperformed the original scheme in the most of performance metrics. The aforementioned phenomenon provided evidence that by introducing optimization measures for the initial values of the decoder, we had indeed achieved a more beneficial initialization, enabling it to fast adapt to different scenarios.
	
	\textbf{Qualitative comparison.}$\;$ 
	The results of our method and other approaches on the LSRW and VOC datasets were presented in Fig.~\ref{fig: LSRW_VOC}. From Fig.~\ref{fig: LSRW_VOC} (a), it could be seen that other methods exhibited poor performance on the LSRW dataset. RUAS suffered from overexposure, while EnGAN and FIDE exhibited noticeable underexposure. Additionally, RetinexNet disrupted the details and textures, and UTVNet not only had underexposure but also exhibited color shifts. From the results in Fig.~\ref{fig: LSRW_VOC} (b), it is also obvious that UTVNet exhibited noticeable artifacts. In contrast, the enhancement results of our BL and RBL not only achieved optimal lighting intensity but also outperformed other methods significantly in terms of both color and texture details, highlighting the advantages of the proposed approach. The enhanced visual appeal of RBL compared to BL further validated the effectiveness of RBL.
	Besides, more subjective results on four datasets in Fig.~\ref{fig: Moredatasets} further illustrated our advantages.
	The first four rows of results demonstrated that our method produced more suitable lighting and vivid colors while effectively handling scenarios with significant noise. The latter two rows of results indicated that our method exhibited good generalization capability by adapting well to unseen scenes.

	\begin{figure}[t] 
		\centering
		\begin{tabular}{c@{\extracolsep{0.3em}}c}
			\includegraphics[width=0.46\linewidth]{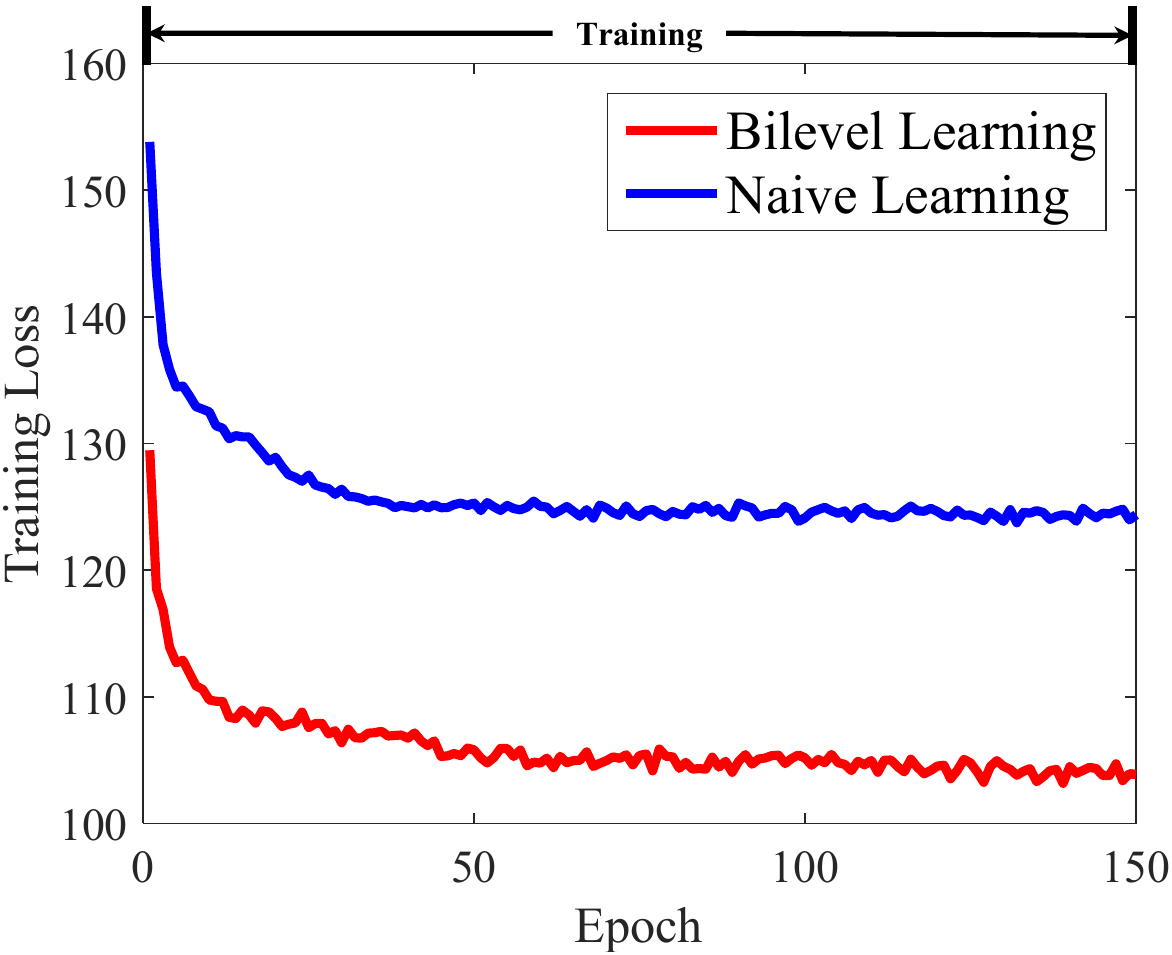}&
			\includegraphics[width=0.47\linewidth]{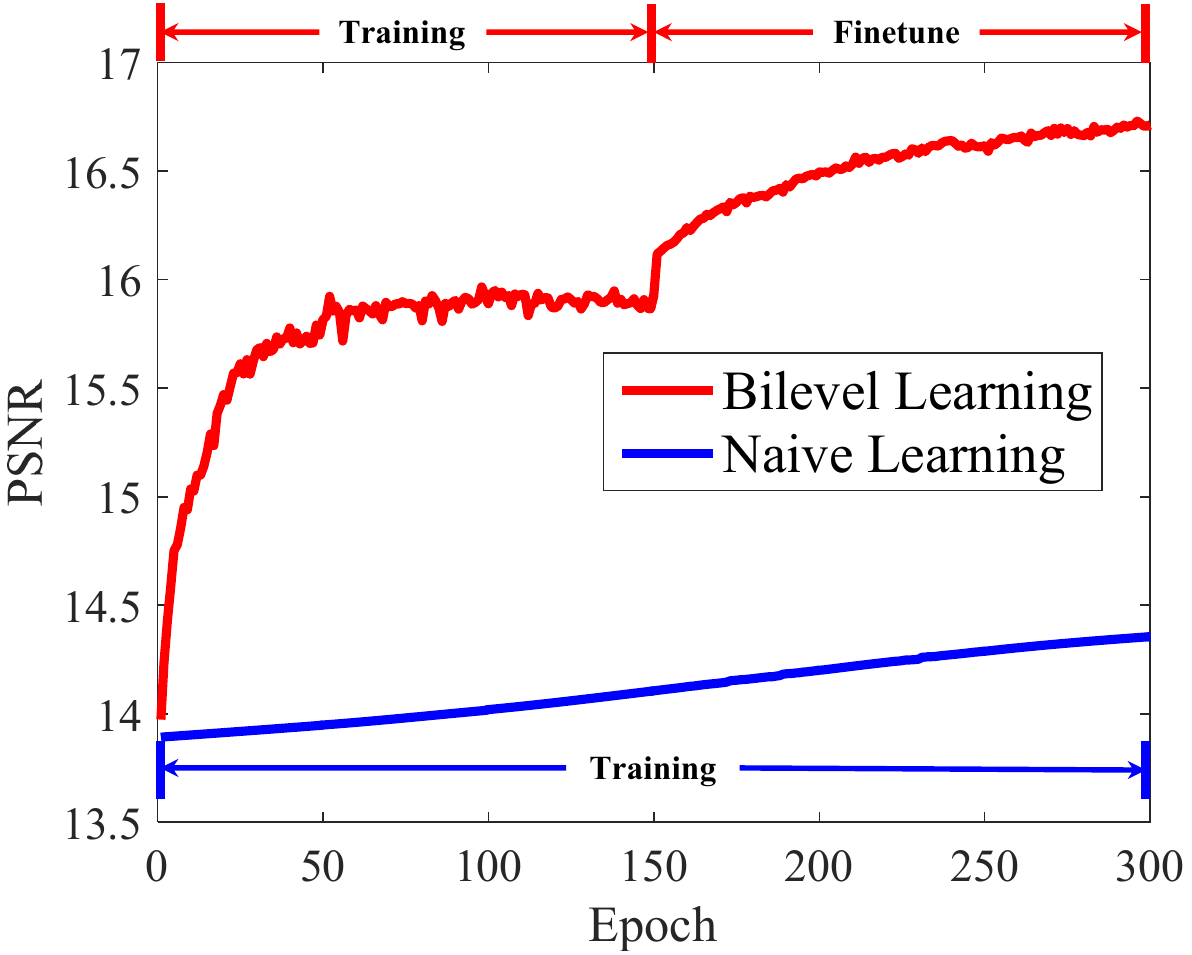}\\
			\footnotesize (a) &\footnotesize (b)\\
		\end{tabular}
		\caption{Convergence behaviors of naive learning and our method. (a) is the loss comparison when both of them are trained with mixed datasets of MIT and LOL. (b) is the PSNR comparison of our training strategy and naive approach (training with a single dataset (VOC) only). As for the second graph, we execute the bilevel training for 150 epochs and finetuning for another 150 epochs, and train the naive approach for 300 epochs to ensure fairness.}
		\label{linegraphs}
		\vspace{-3mm}
	\end{figure}
	
	\begin{figure}[t]
		\centering
		\begin{tabular}{c@{\extracolsep{0.3em}}c@{\extracolsep{0.3em}}c}
			\includegraphics[width=0.31\linewidth]{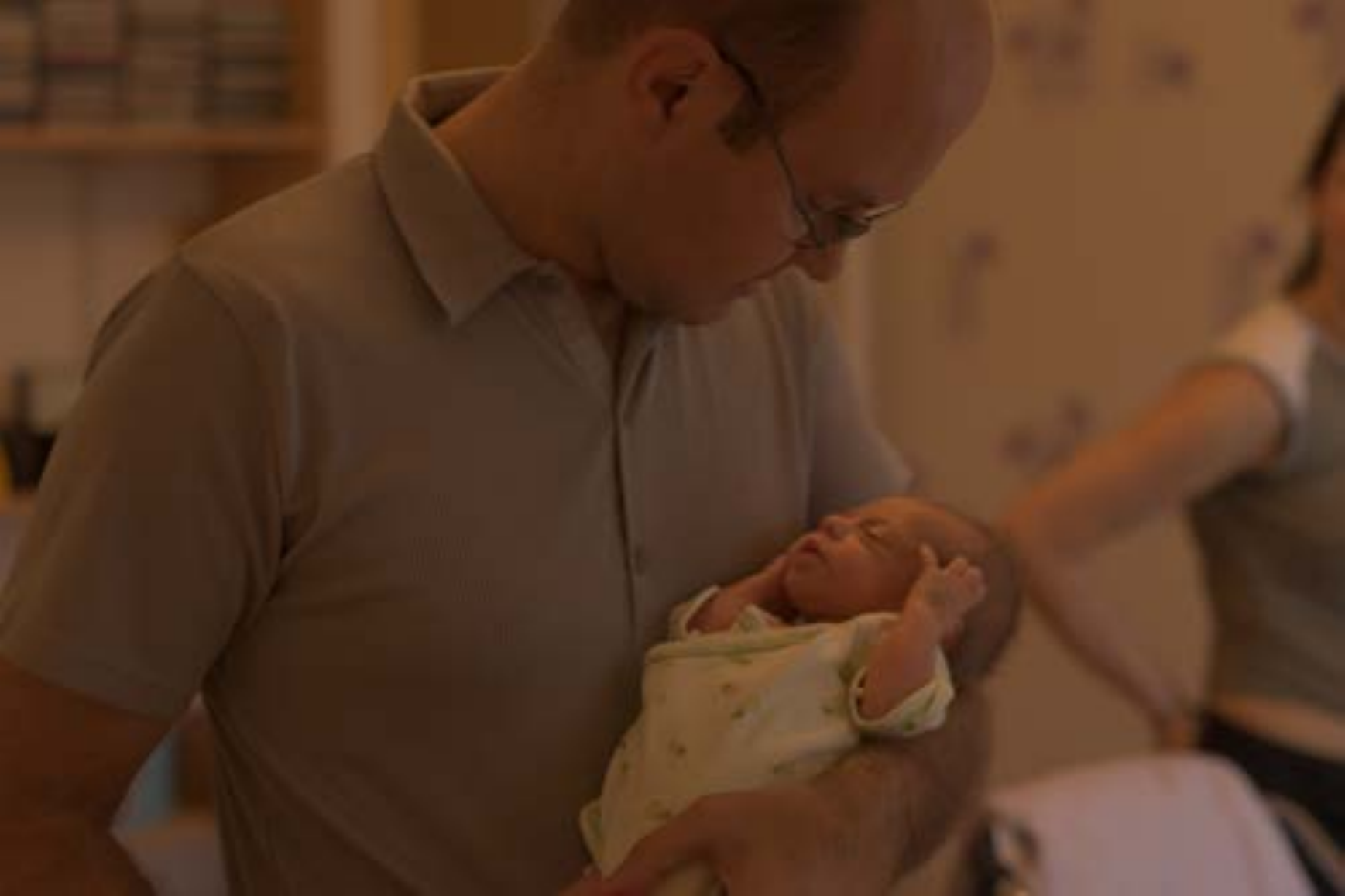}&
			\includegraphics[width=0.31\linewidth]{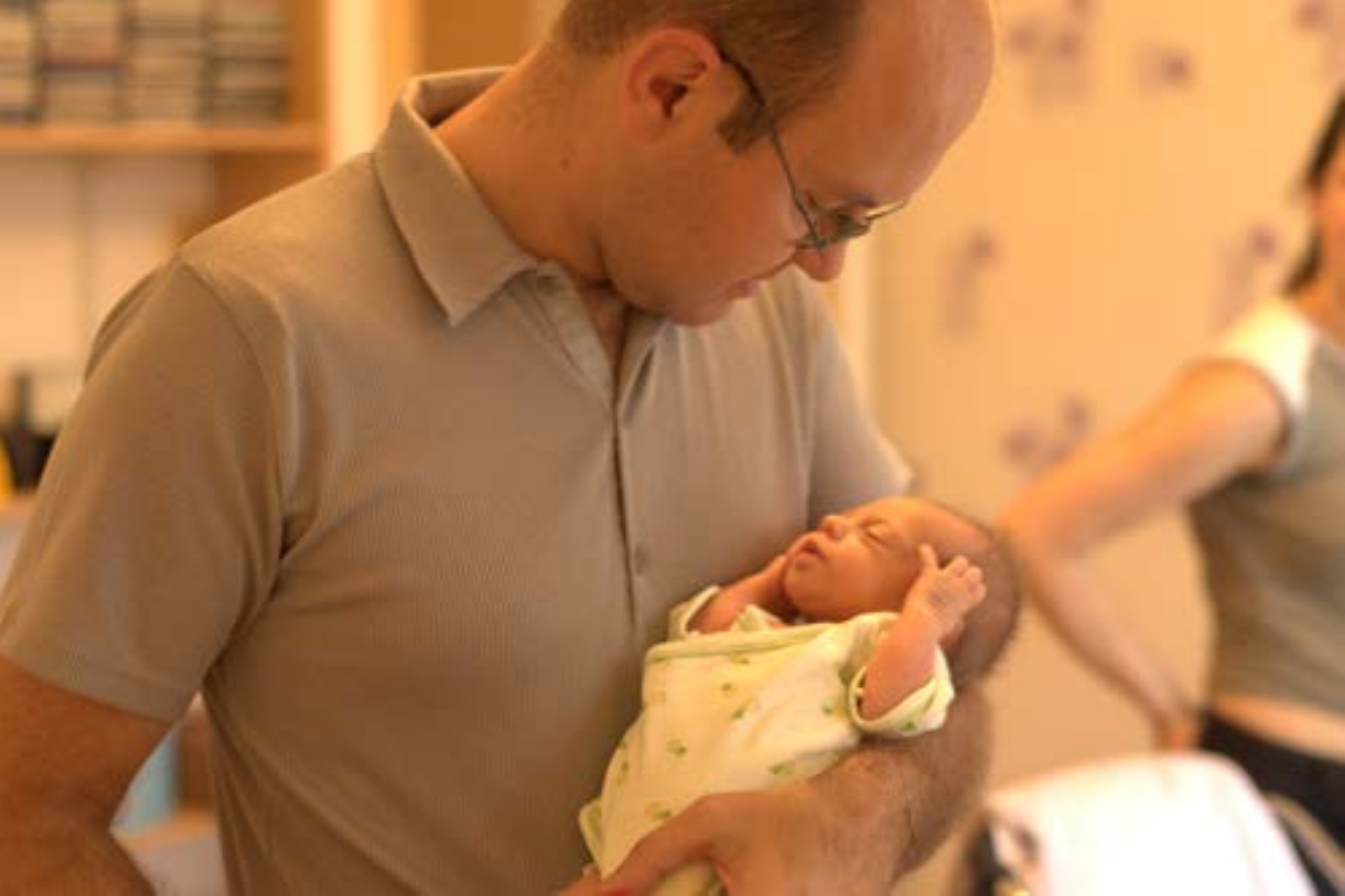}&
			\includegraphics[width=0.31\linewidth]{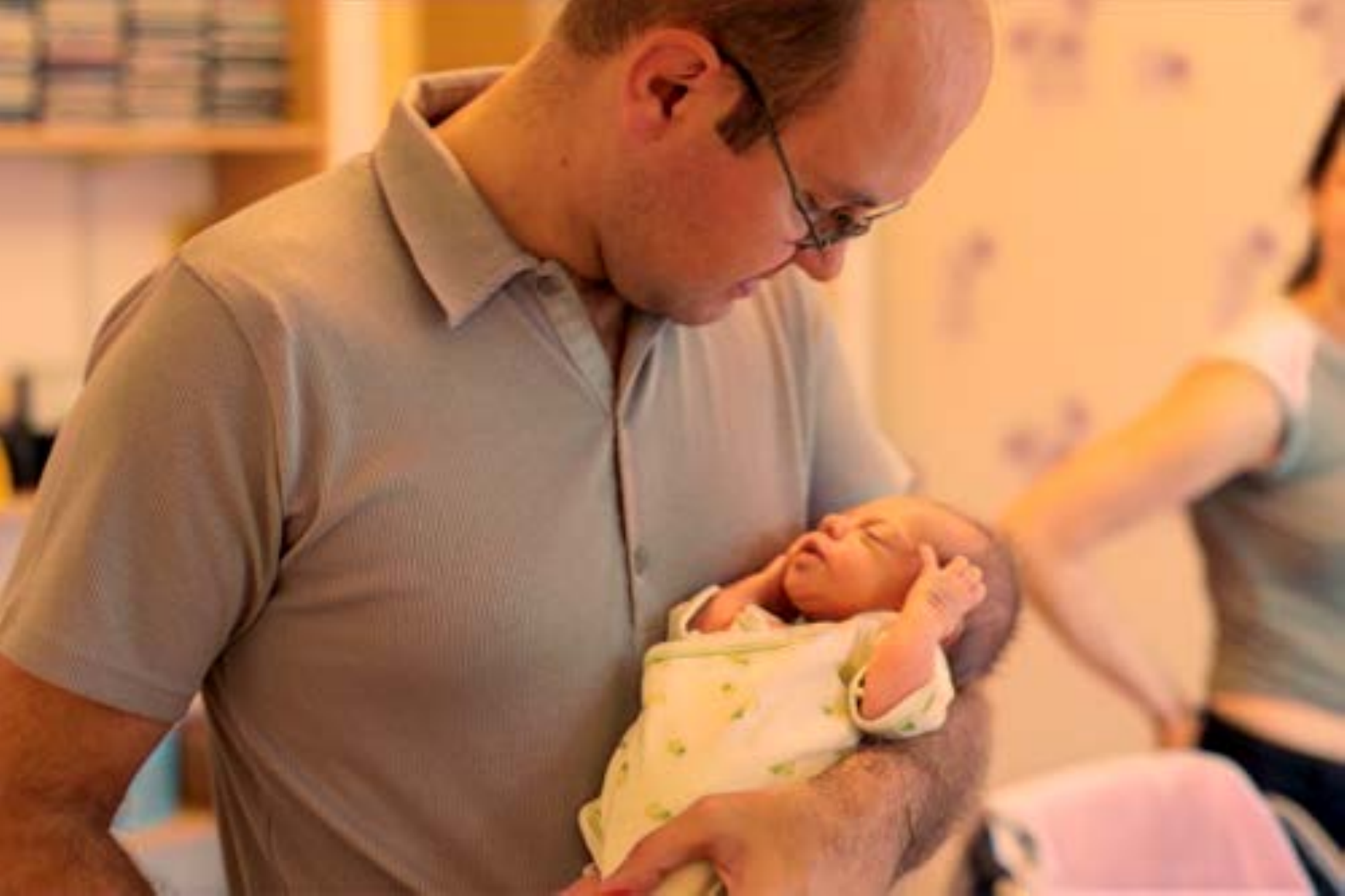}\\
			\includegraphics[width=0.31\linewidth]{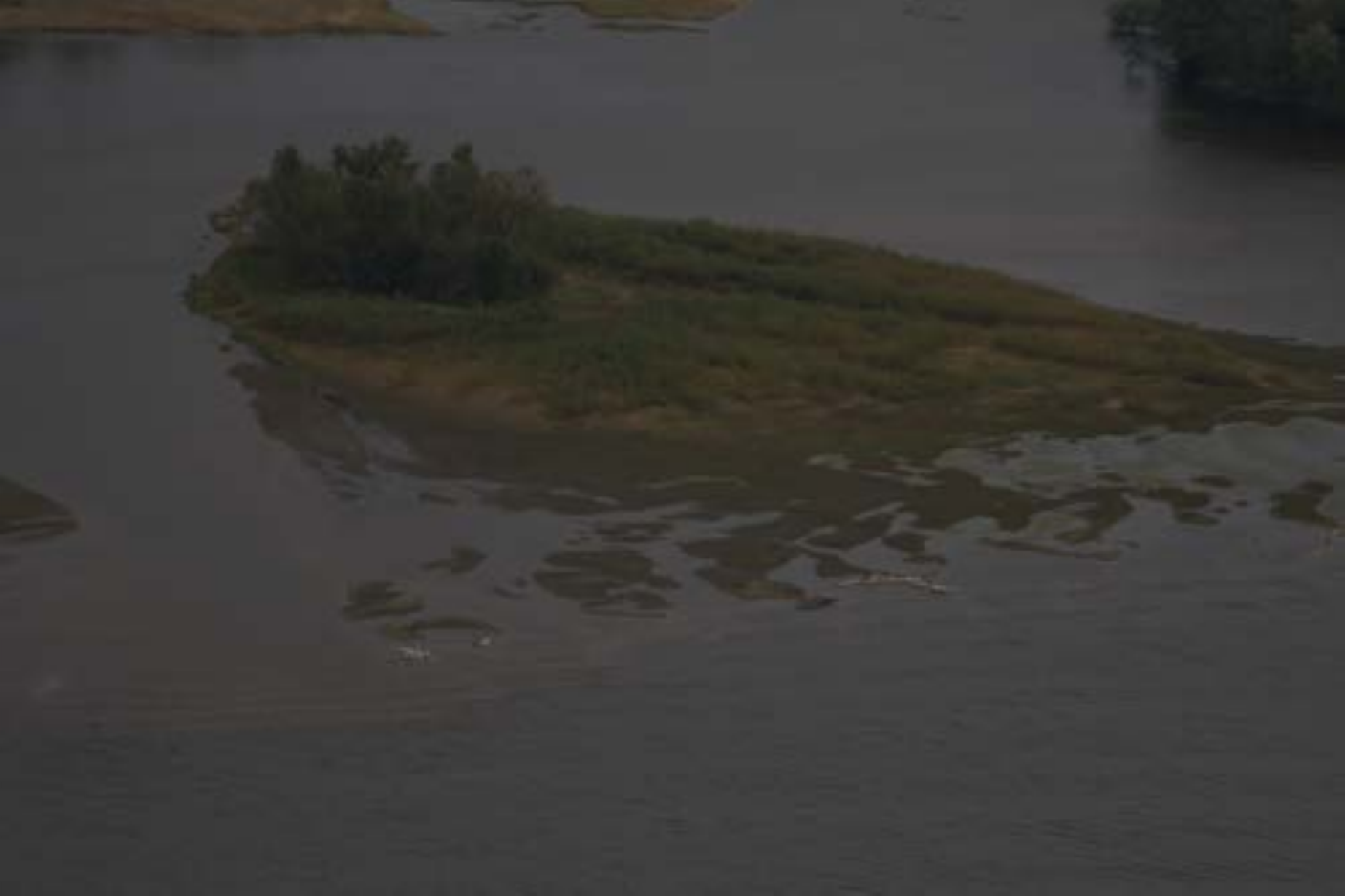}&
			\includegraphics[width=0.31\linewidth]{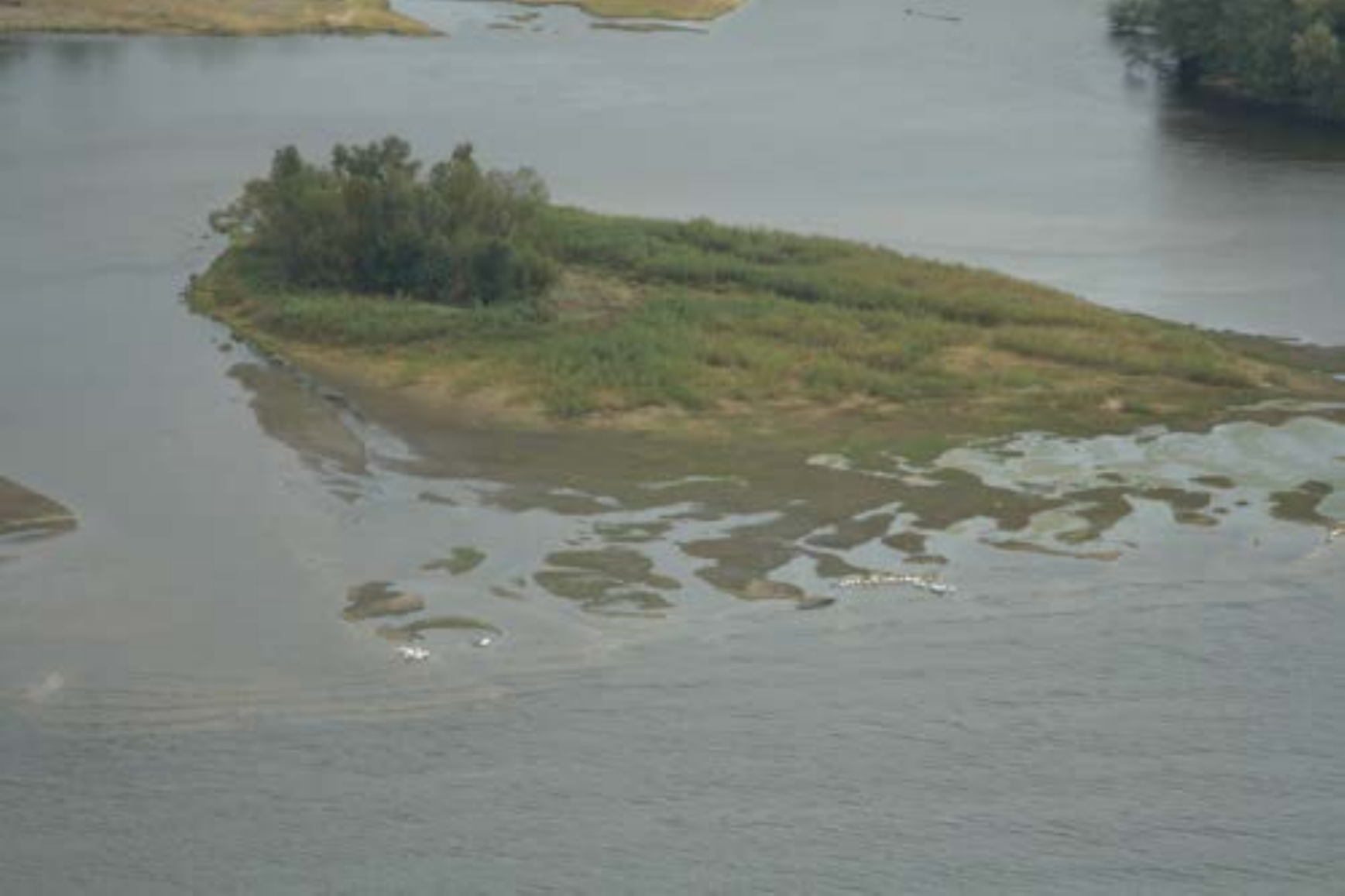}&
			\includegraphics[width=0.31\linewidth]{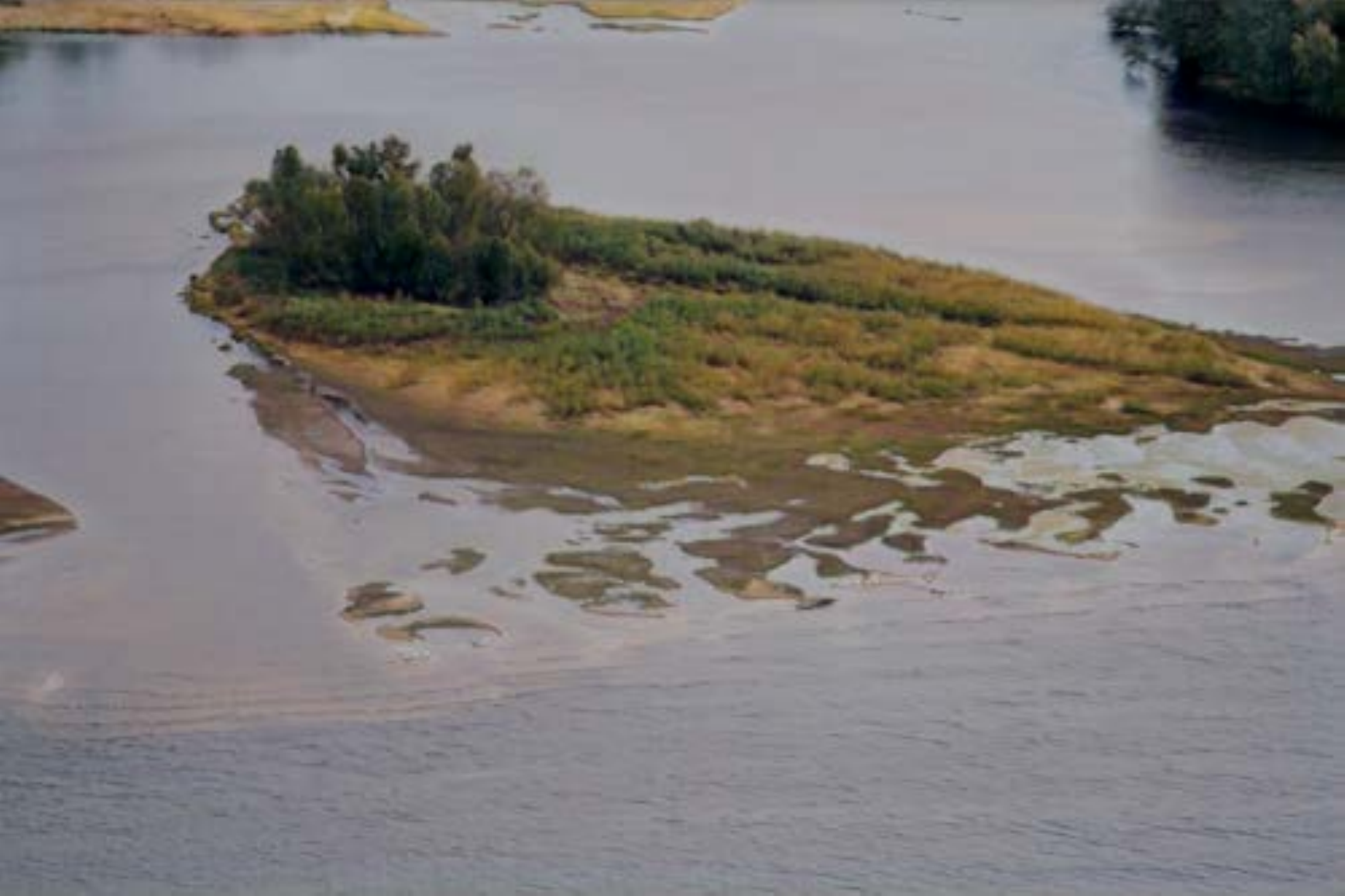}\\
			\footnotesize Input&\footnotesize Naive Learning&\footnotesize Ours\\
		\end{tabular}
		\caption{Visual comparison among different training strategy on the MIT dataset. Ours are better than naive both on color and exposure.}
		\label{fig: Analysis_MIT}
	\end{figure}
	\subsection{Evaluations on Unseen-Unpaired Benchmarks}
	We conducted a series of comparative experiments on two challenging unpaired datasets to further demonstrate the fast adaptability of our method in extreme low-light conditions. The testing protocol used in this section remains consistent with Section 6.3.
	
	\textbf{Quantitative comparison.}$\;$ 
	As illustrated by the numerical results shown in Fig.~\ref{fig: Hist}, the proposed RBL achieved significant improvements over BL in both DE and NIQE, outperforming other methods and demonstrating that our method could obtain the enhancement results whose texture detail and color were more in line with human visual habits.
	
	\begin{figure}[t] 
		\centering
		\begin{tabular}{c@{\extracolsep{0.3em}}c}
			\includegraphics[width=0.48\linewidth]{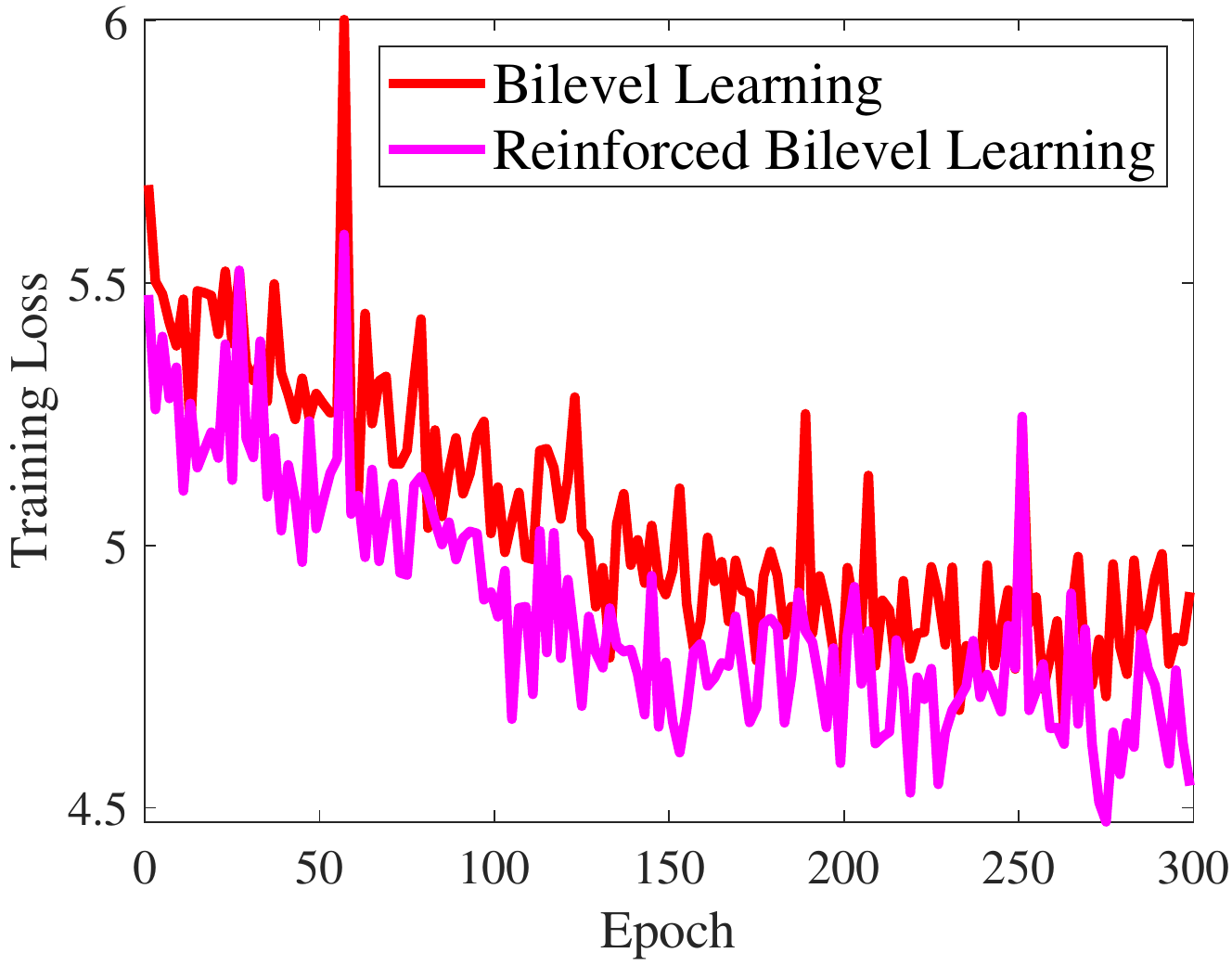}&
			\includegraphics[width=0.48\linewidth]{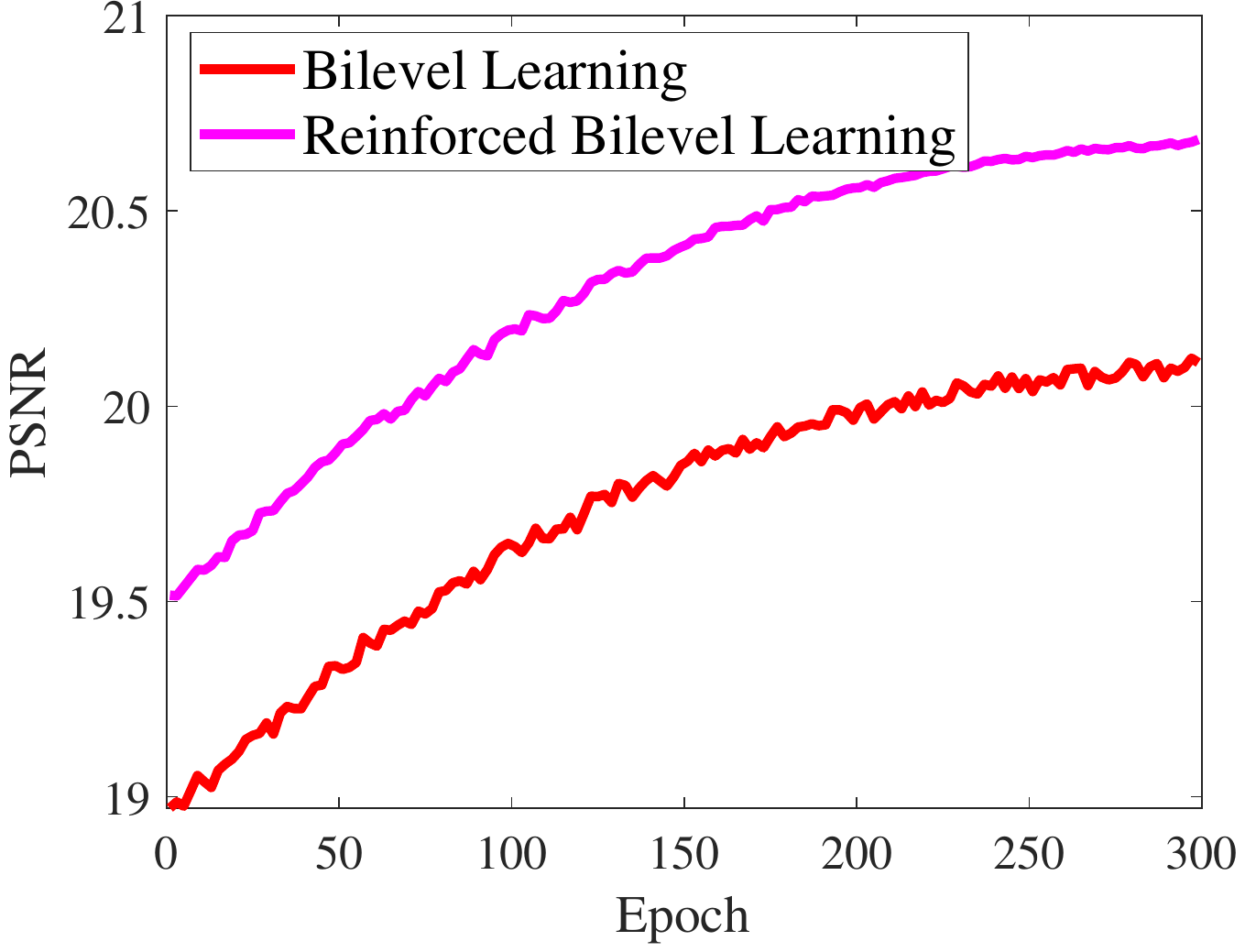}\\
			\footnotesize (a) &\footnotesize (b)\\
		\end{tabular}
		\caption{Convergence behaviors of our two versions in the adaptation stage on the MIT dataset. (a) is the loss comparison. (b) is the PSNR value comparison.}
		\label{linegraph2}
	\end{figure}
	
	\begin{figure}[t]
		\centering
		\begin{tabular}{c@{\extracolsep{0.3em}}c@{\extracolsep{0.3em}}c}
			\includegraphics[width=0.31\linewidth]{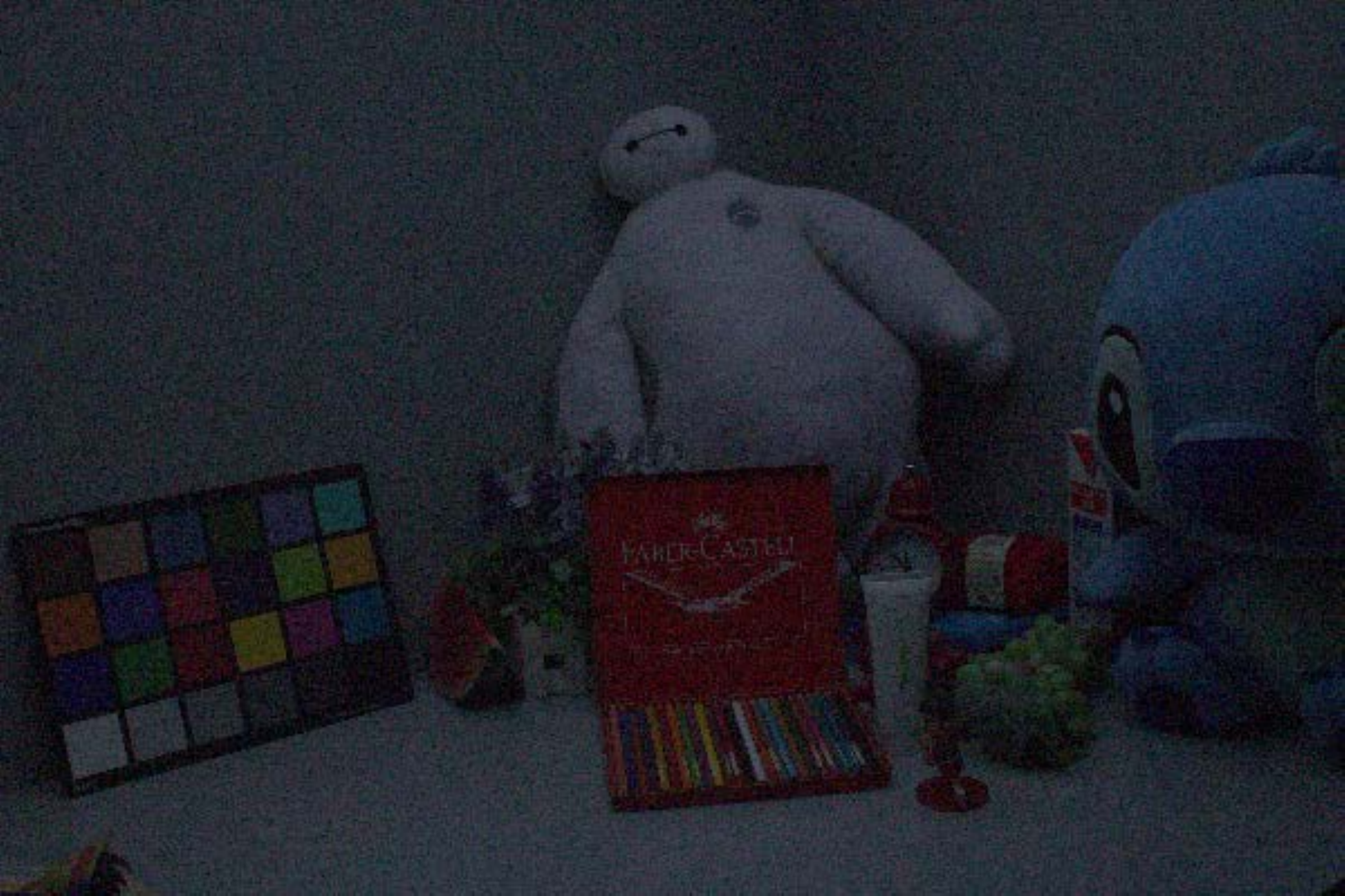}&
			\includegraphics[width=0.31\linewidth]{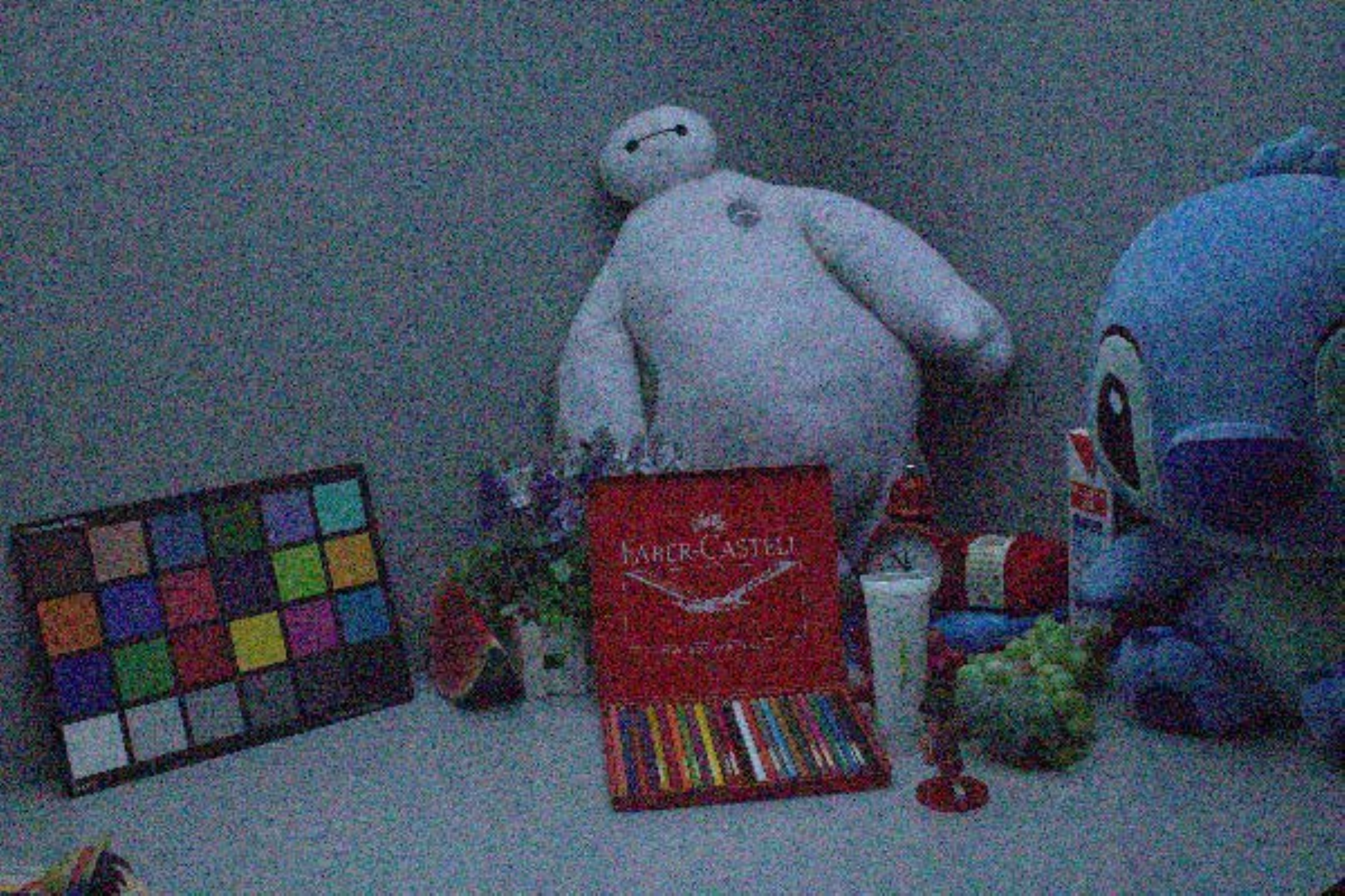}&
			\includegraphics[width=0.31\linewidth]{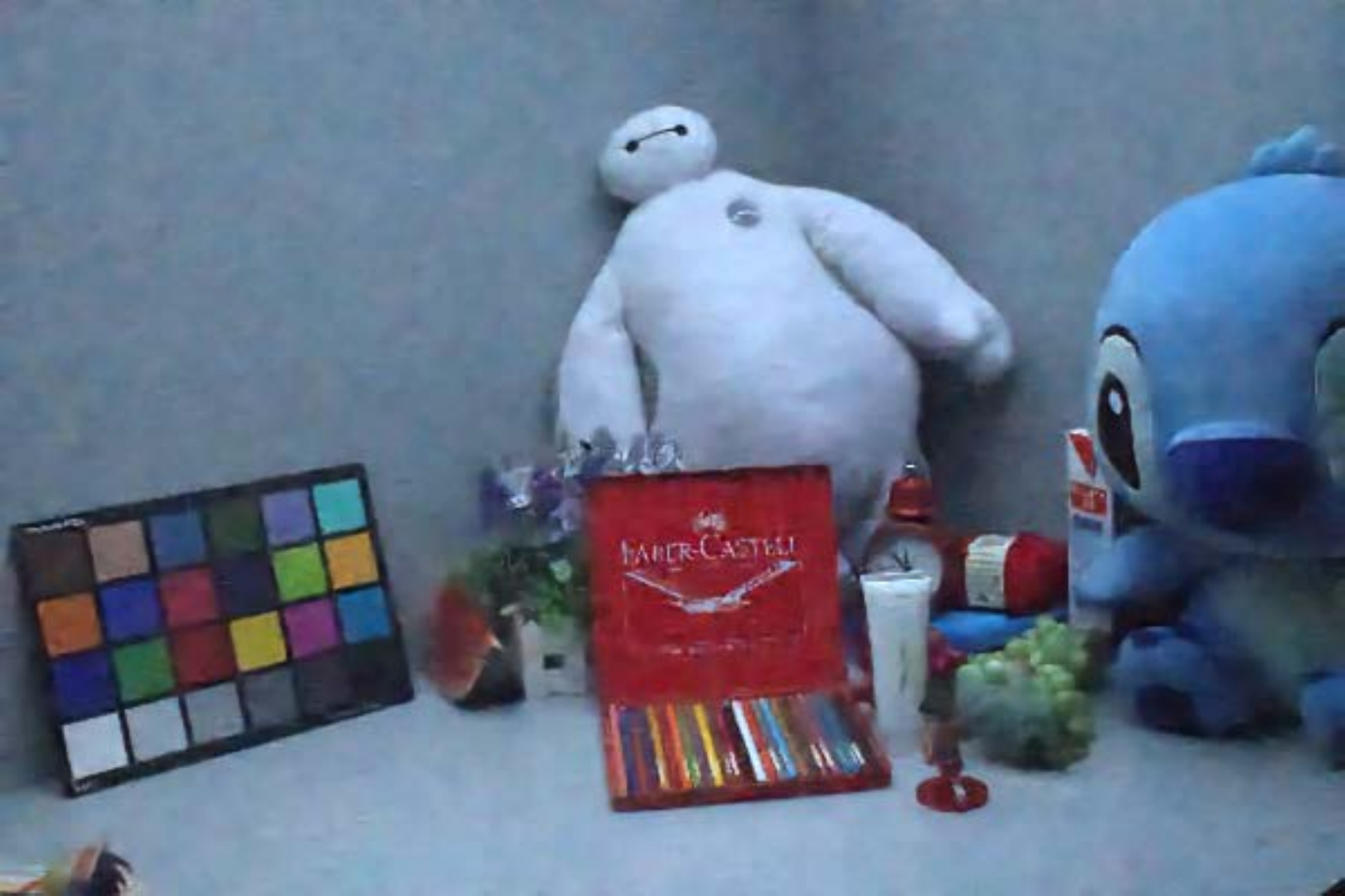}\\
			\includegraphics[width=0.31\linewidth]{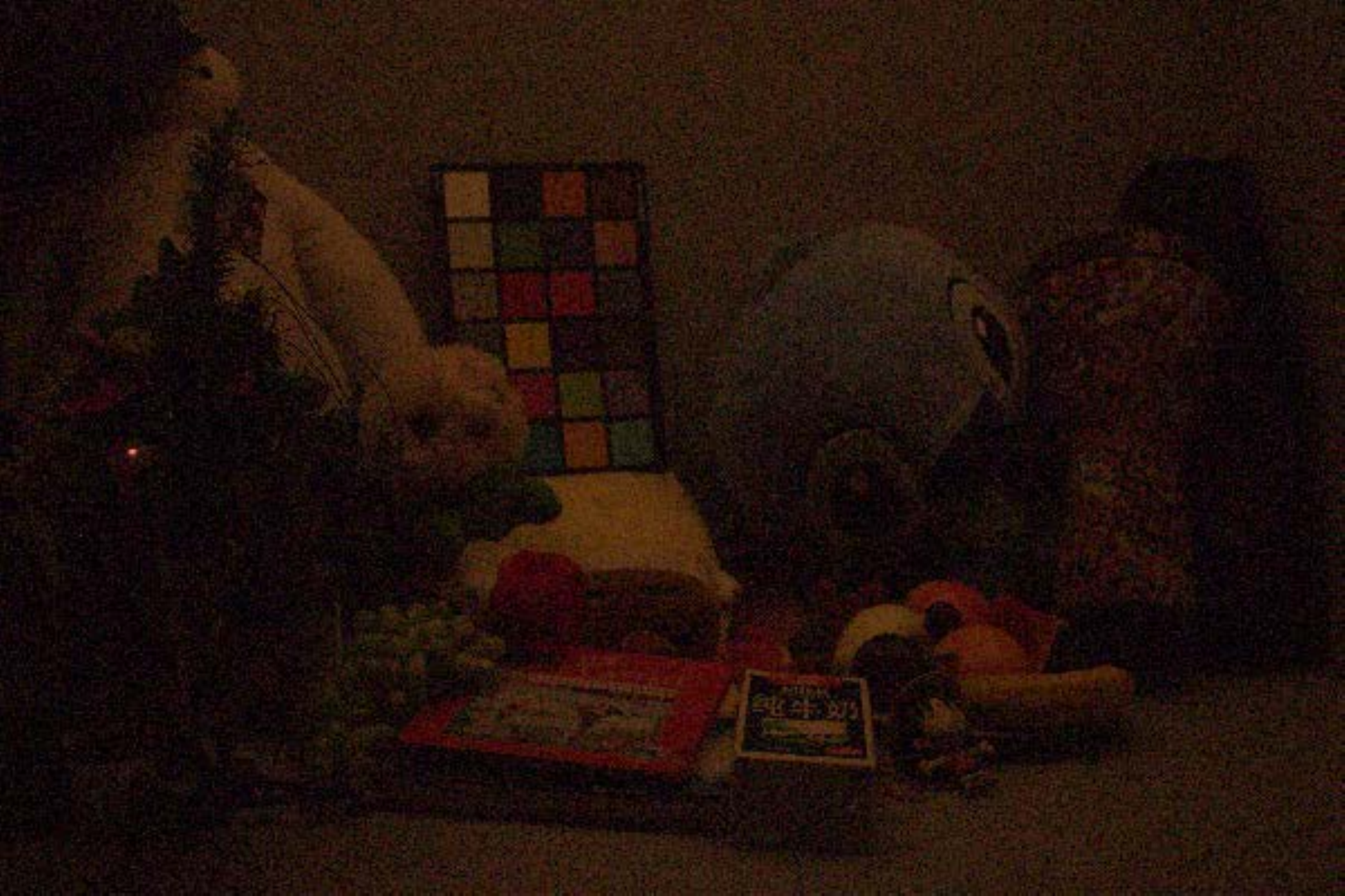}&
			\includegraphics[width=0.31\linewidth]{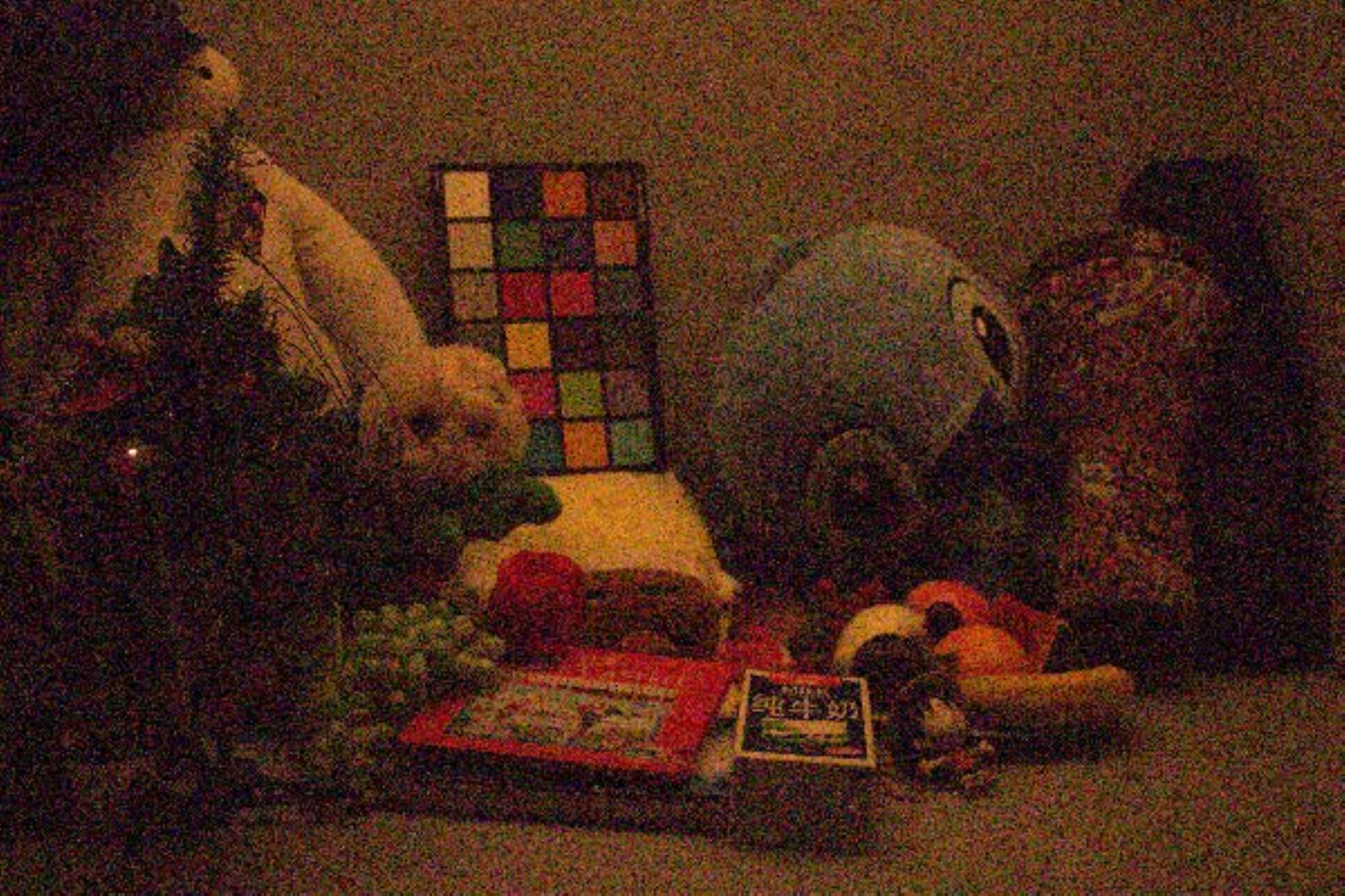}&
			\includegraphics[width=0.31\linewidth]{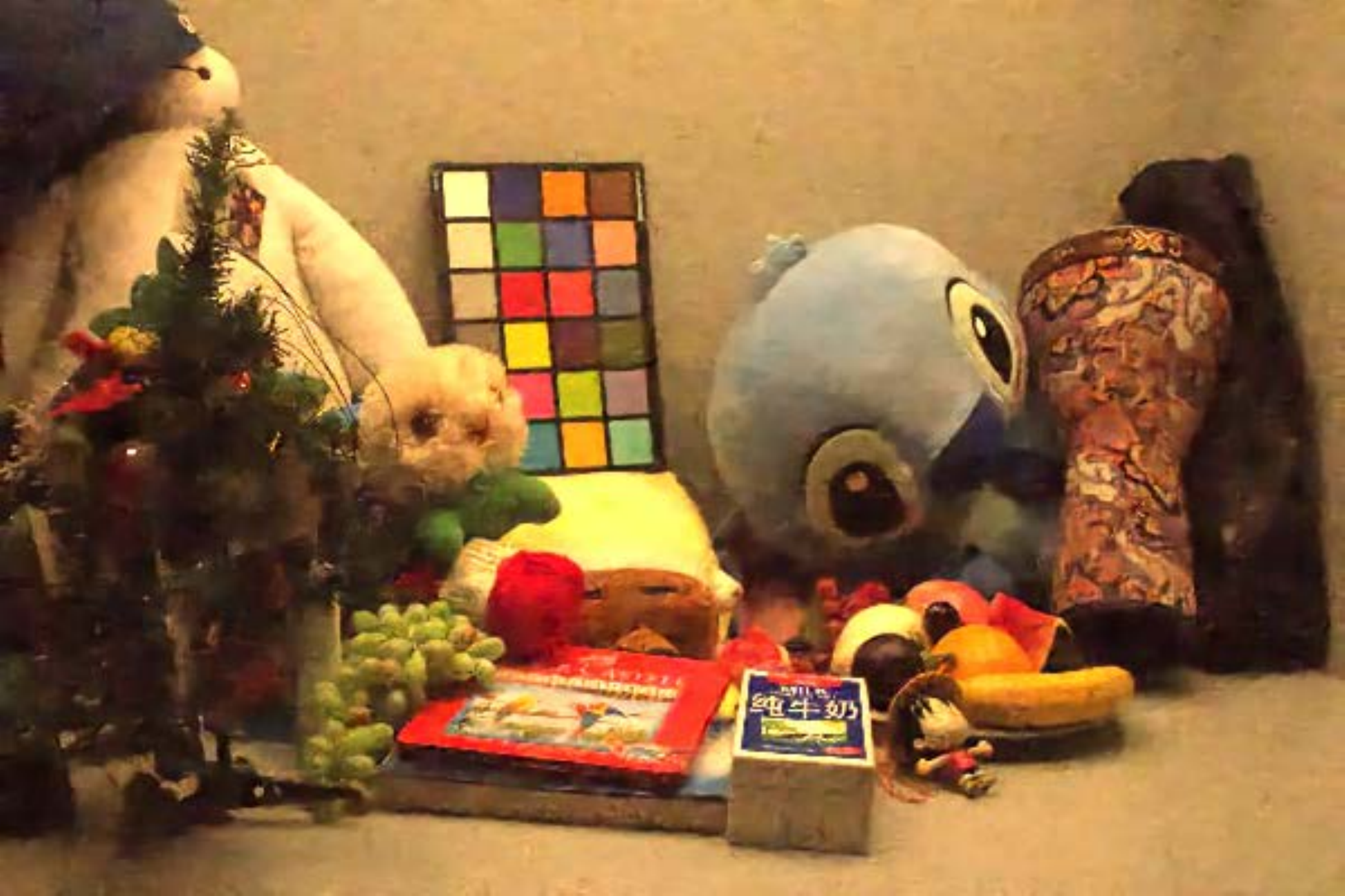}\\			
			\footnotesize Input&\footnotesize w/o Denoising&\footnotesize Ours\\
		\end{tabular}
		\caption{Ablation study of denoise module. The result with denoising module is not only brighter but also clearer, which proves the effectiveness of our denoising module. }
		\label{fig: ablation_denoise}
		%	\vspace{-0.2cm}
	\end{figure}
	
	\textbf{Qualitative comparison.}$\;$ 
	The three sets of qualitative results presented in Fig.~\ref{fig: DARKFACE} revealed a common issue of underexposure in the compared methods. In contrast, our method significantly improved image brightness and better preserved texture details. Further, we tested our algorithm on the Exclusively Dark (ExDARK)~\citep{Exdark_dataset} dataset, an unpaired challenging dataset constructed by collecting low-light images from different datasets with object class annotation. As Fig.~\ref{fig: ExDARK} illustrated, although the other methods achieved brightness enhancement, their results still exhibited a significant amount of noticeable noise. In contrast, our method could effectively brighten the image while ensuring that the noise in the image was removed to a certain extent. It proved that our method could not only adapt to different scenes in illumination estimation, but also achieve good results in denoising at the same time.

	%\begin{table}[t]
	%    \renewcommand\arraystretch{1.3}
	%    \caption{PSNR, SSIM and LPIPS results on images randomly selected from VOC dataset.}
	%	\centering
	%		\begin{tabular}{|c|c|c|c|}
		%			\hline  
		%			Model & PSNR & SSIM&LPIPS\\
		%			\hline  
		%			w/o finetune & 15.7692 & 0.6181& 0.2503 \\
		%			\hline 
		%			w/ finetune & {\textbf{16.7280}} &{\textbf{0.6325}} & {\textbf{0.1987}}\\
		%			\hline
		%		\end{tabular}
	%	
	%	
	%	\label{tab:finetune}
	%\end{table}

	\begin{figure}[t]
		\centering
		\begin{tabular}{c@{\extracolsep{0.3em}}c@{\extracolsep{0.3em}}c}
			\includegraphics[width=0.3\linewidth]{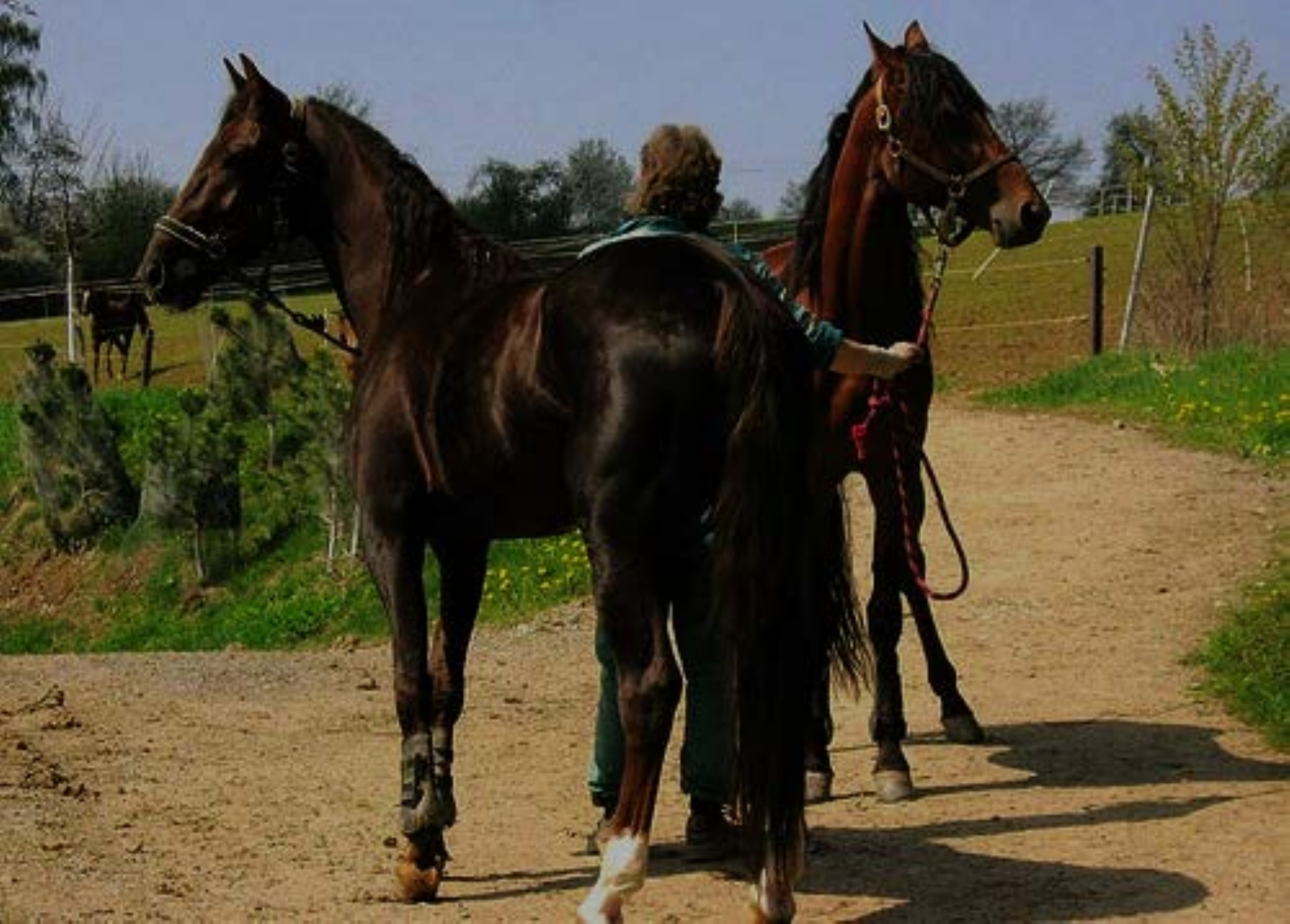}&
			\includegraphics[width=0.3\linewidth]{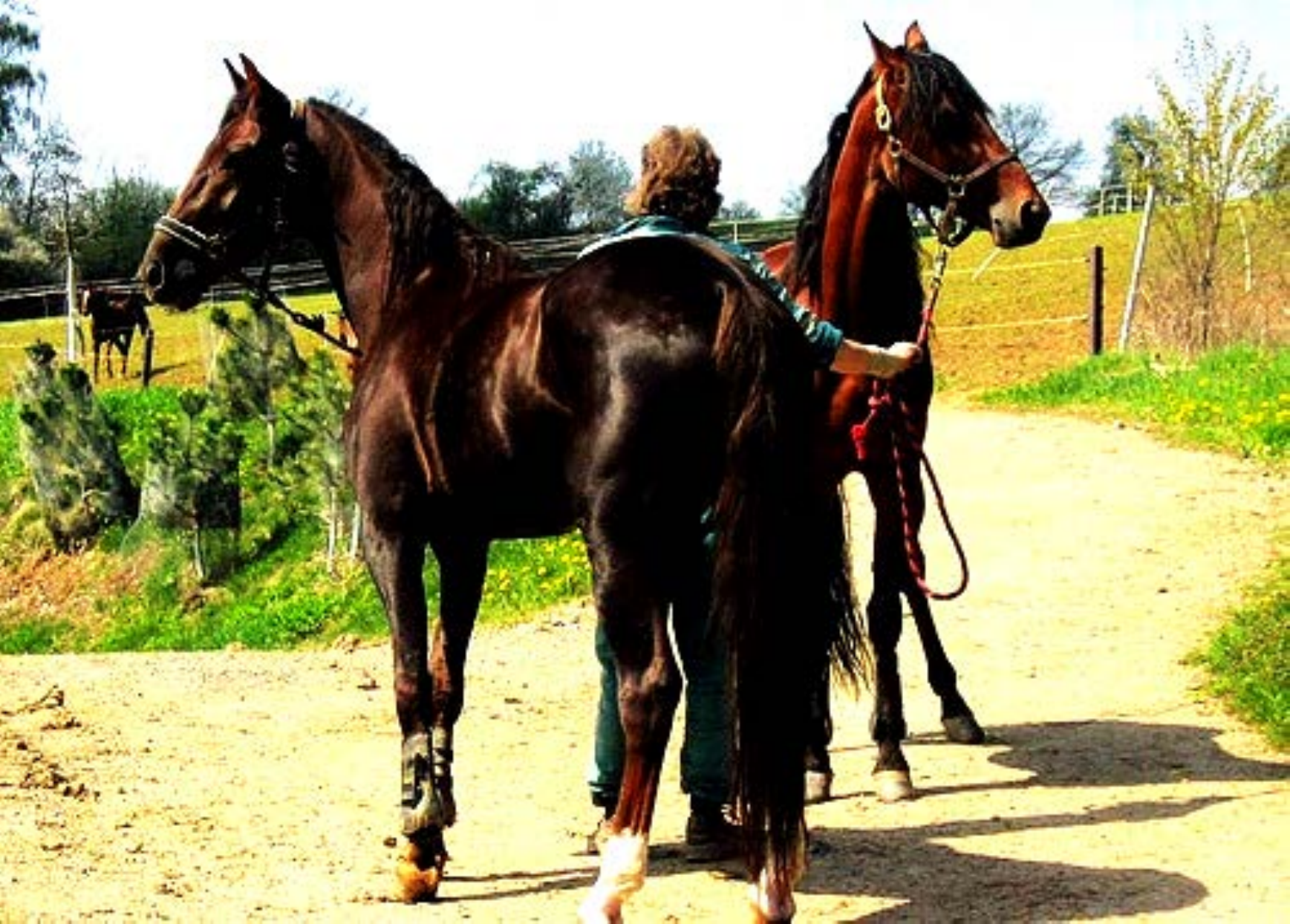}&
			\includegraphics[width=0.3\linewidth]{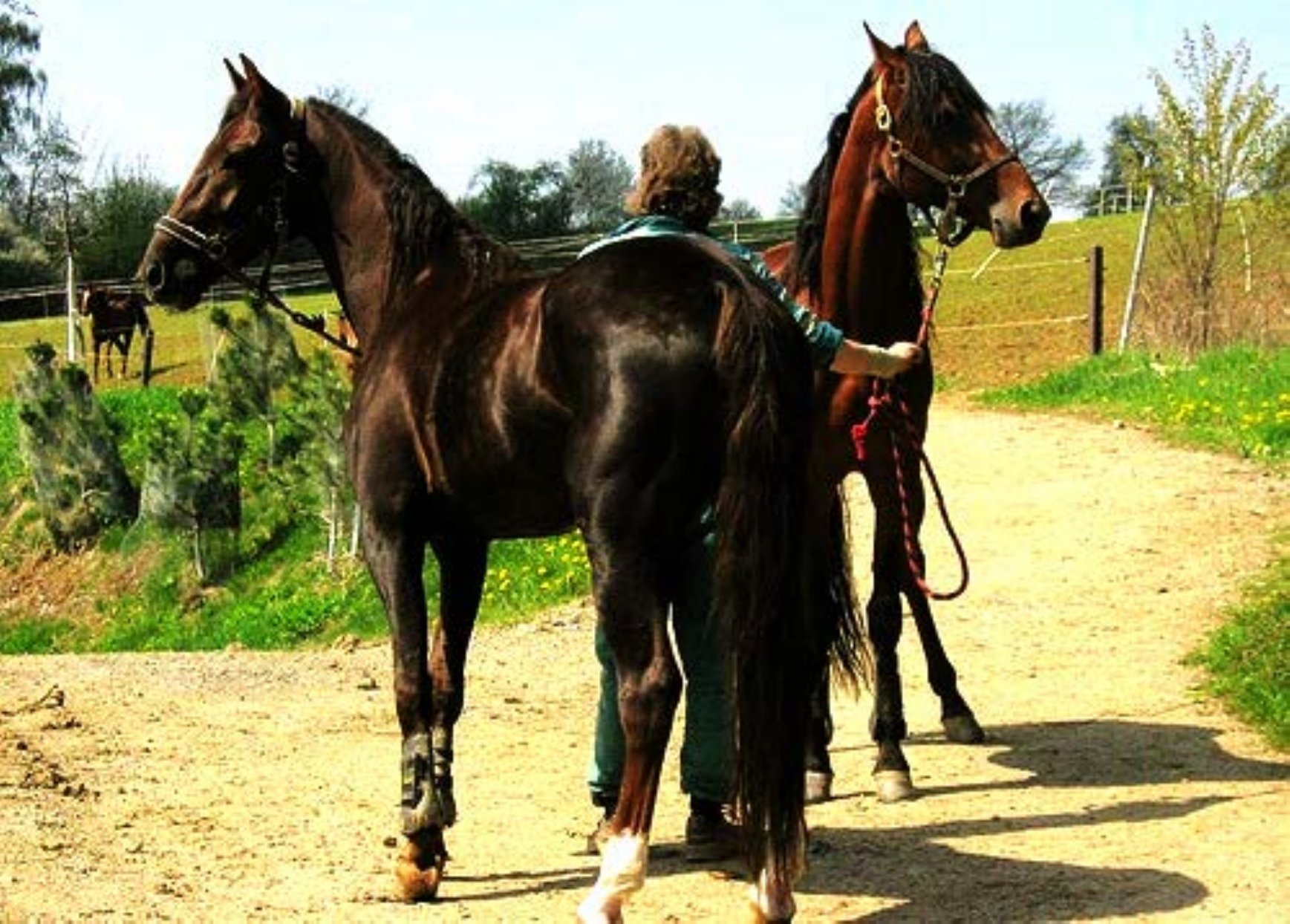}\\
			\includegraphics[width=0.3\linewidth]{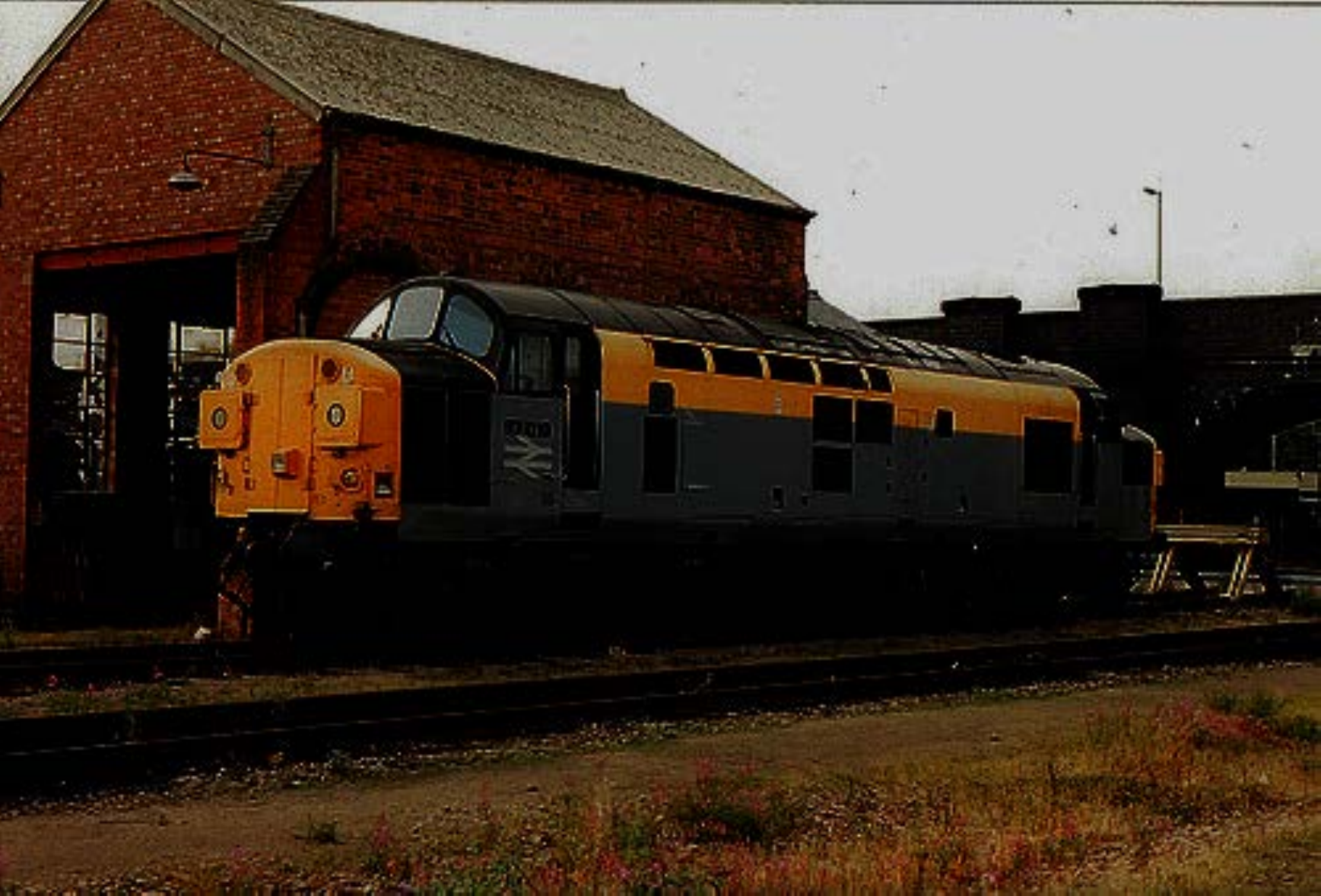}&
			\includegraphics[width=0.3\linewidth]{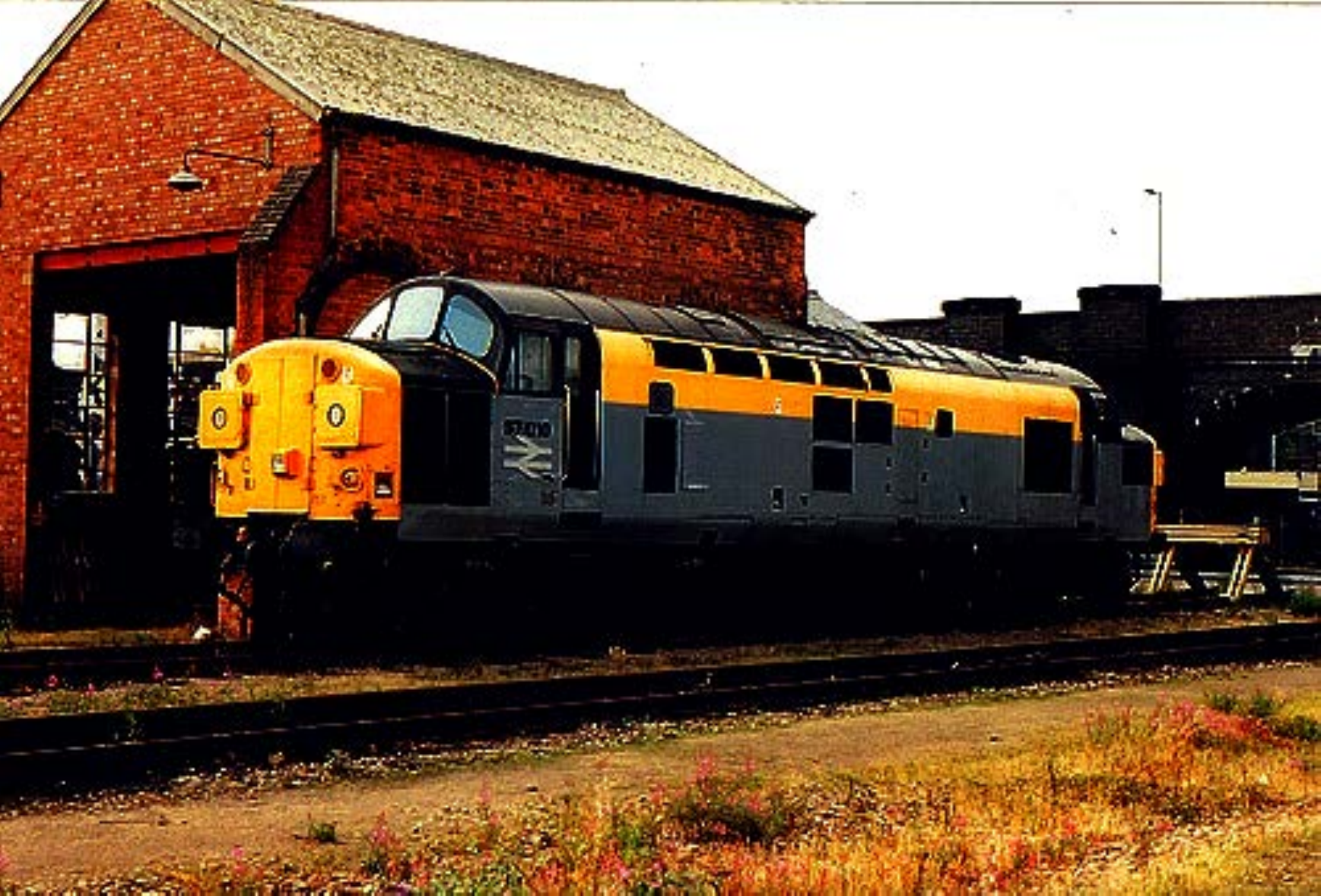}&
			\includegraphics[width=0.3\linewidth]{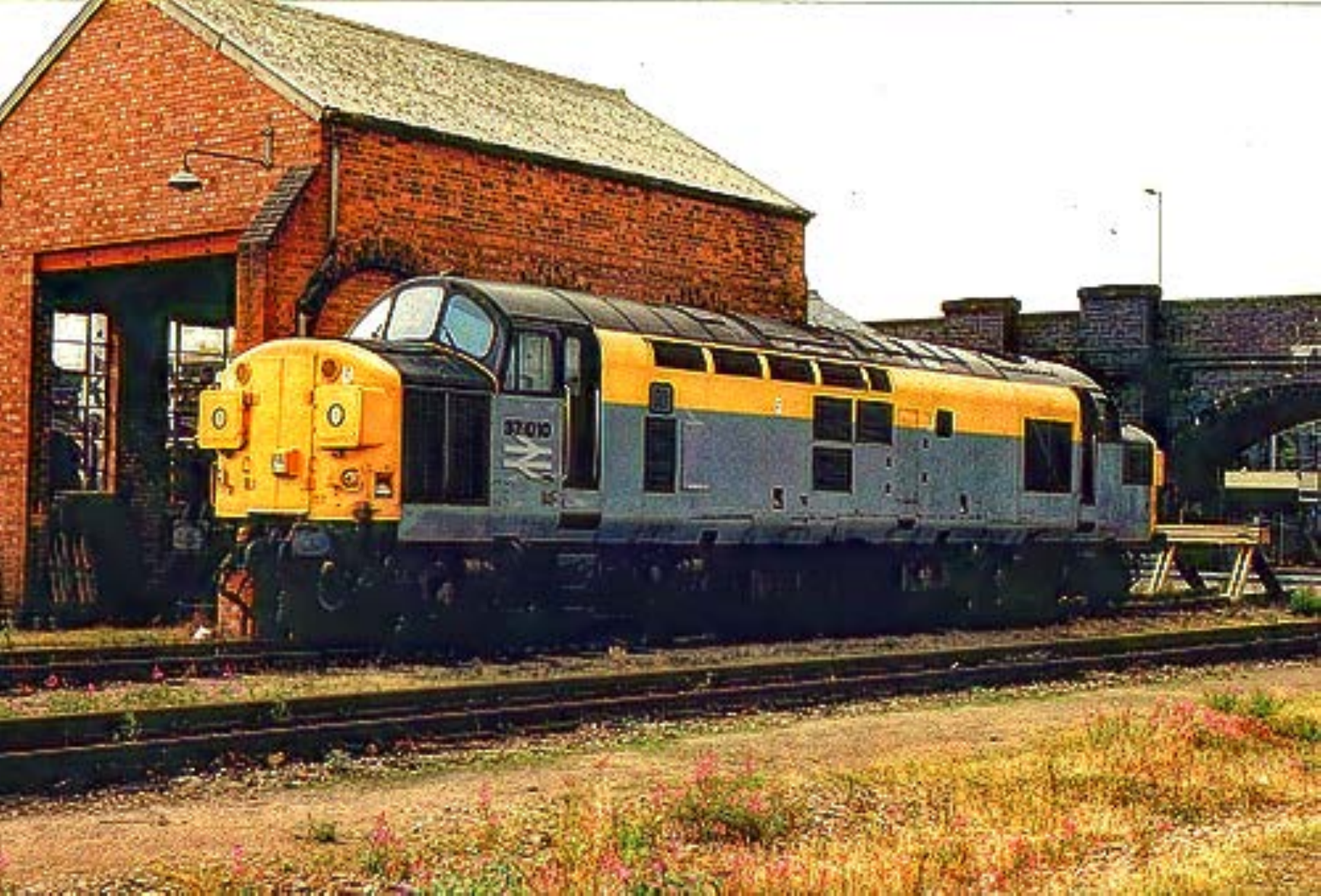}\\			
			\footnotesize Input&\footnotesize w/o Finetune&\footnotesize w/ Finetune\\
		\end{tabular}
		%	\vspace{-0.2cm}
		\caption{Ablation study of finetune process. }
		\label{fig: Ablation_finetune}
		%	\vspace{-0.2cm}
	\end{figure}

	\section{Algorithmic Analyses}
	In this section, we provided a series of algorithm analyses to validate the effectiveness of the proposed bilevel optimization framework. We first analyze the effect of different learning strategies by comparing the performance of the naive scheme and our method in terms of convergence. Subsequently, we analyze the advantages of the reinforced bilevel learning framework. Last but not least, we provide relevant results from ablation experiments on the employed finetuning strategy and the role of the constructed adaptive denoising module.
	%\subsection{Experimental Settings}
	%Our training process can be divided into two parts. To begin with, using the MIT and the LOL dataset for training, we trained the whole network with a learning rate of $1\times10^{-3}$. Thanks to the noise estimation network, our model was able to fit without/with noise scenarios. Next, after fixing parameters of encoder and denoising module, we finetune the scene-specific decoder by using Adam optimizer. 

	\subsection{Bilevel Learning \textit{vs} Naive Learning}
	
	Fig.~\ref {linegraphs} displayed the results of our training strategies compared to naive strategies. Both strategies were trained on a mixed dataset of the MIT and LOL datasets and tested on 100 randomly selected images from the VOC dataset. In each experiment, 300 epochs were trained in naive steps to ensure fairness. The results showed that the training loss, PSNR values of naive method were not as outstanding as our strategy. It is proved that the model generated by ours strategy possessed faster convergence speed and stronger generalization ability. As is shown in Fig.~\ref{fig: Analysis_MIT}, we tested our algorithm and naive methods on MIT dataset, and is is obvious that the proposed method took both illumination and color information into account.

	\subsection{Bilevel Learning \textit{vs} Reinforced Bilevel Learning}
	The comparative results regarding the convergence of the newly proposed RBL and BL were presented in Fig.~\ref{linegraph2}. The experimental setup for this analysis was largely consistent with Section 7.2, with the difference that we only presented the results from the adaptation phase to analyze the advantages of RBL. It is evident that RBL exhibited significant advantages over the original version in terms of both loss convergence and performance improvement. It indicated that with the reinforced bilevel learning strategy, we could further improve the convergence rate and obtain initializations, resulting in faster adaptation.
	
	%This indicates that by employing the reinforced dual-layer learning strategy, we can further enhance the convergence speed, resulting in faster adaptation.

	\begin{table}[t]
		\footnotesize
		\renewcommand\arraystretch{1.4}
		\caption{Effects of denoising module and finetune mechanism.}
		%	\vspace{-0.4cm}
		\centering
		\begin{tabular}{|c|c|c|c|}
			\hline
			\multicolumn{4}{|c|}{Effects of denoising on the LOL dataset}\\
			\hline Model & PSNR & SSIM & LPIPS \\
			\hline w/o Adaptive Denoising & 11.1692 & 0.2750 & 1.1439  \\
			\hline w/ Adaptive Denoising & {\textbf{15.4120}} & {\textbf{0.4205}}&{\textbf{0.5444}} \\
			\hline\hline
			\multicolumn{4}{|c|}{Effects of finetune on the VOC dataset}\\
			\hline Model & PSNR & SSIM & LPIPS \\
			\hline w/o Finetune & 15.7692 & 0.6181& 0.2503 \\
			\hline w/ Finetune & {\textbf{16.7280}} &{\textbf{0.6325}} & {\textbf{0.1987}}\\
			\hline
		\end{tabular}
		
		\label{tab:denoise_finetune}
	\end{table}
	
	\subsection{Effects of Adaptive Denoising}
	We presented the visual results of the ablation experiment on the adaptive denoising module in Fig.~\ref {fig: ablation_denoise}. It is evident that when the adaptive denoising module is removed, significant noise artifacts appear in the enhanced results, indicating that the network loses its ability to remove noise. This is because the network we constructed can accurately estimate the noise map. the relevant numerical results presented in Table \ref{tab:denoise_finetune} also demonstrated that the removal of the adaptive denoising module had a significant negative impact on the practical metric, which further verified its effectiveness.
	%The relevant numerical results presented in Table \ref{tab:denoise_finetune} also demonstrated that the removal of the denoising module had a significant negative impact on the practical performance.
	
	\subsection{Effects of Finetune Process}
	As for the finetune period shown in Fig.~\ref {fig: Ablation_finetune}, we only utilized 250 images from VOC to train 30 epochs. It could be observed that although the images without finetune could still recovered part of the original image information, the illumination estimation after finetune seemed to be more accurate. The quantitative comparison of the finetuning strategy in Table \ref{tab:denoise_finetune} can also show its positive effect. It is proved that a small amount of finetune could significantly improve effects.

	\subsection{Intermediate Results}
\begin{figure}[t]
		\centering
		\begin{tabular}{c@{\extracolsep{0.3em}}c@{\extracolsep{0.3em}}c@{\extracolsep{0.3em}}c@{\extracolsep{0.3em}}c}
			\includegraphics[width=0.18\linewidth]{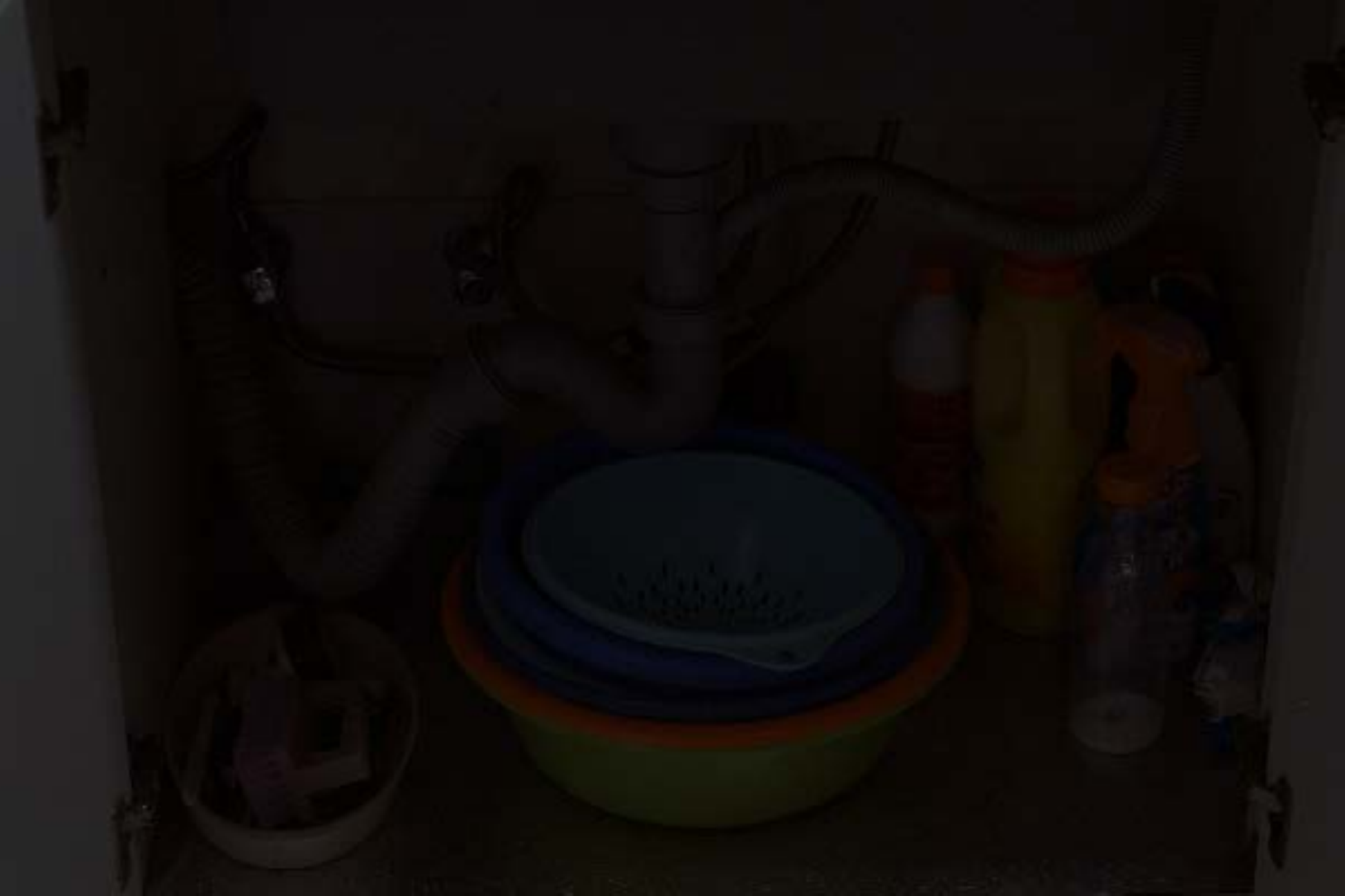}&
			\includegraphics[width=0.18\linewidth]{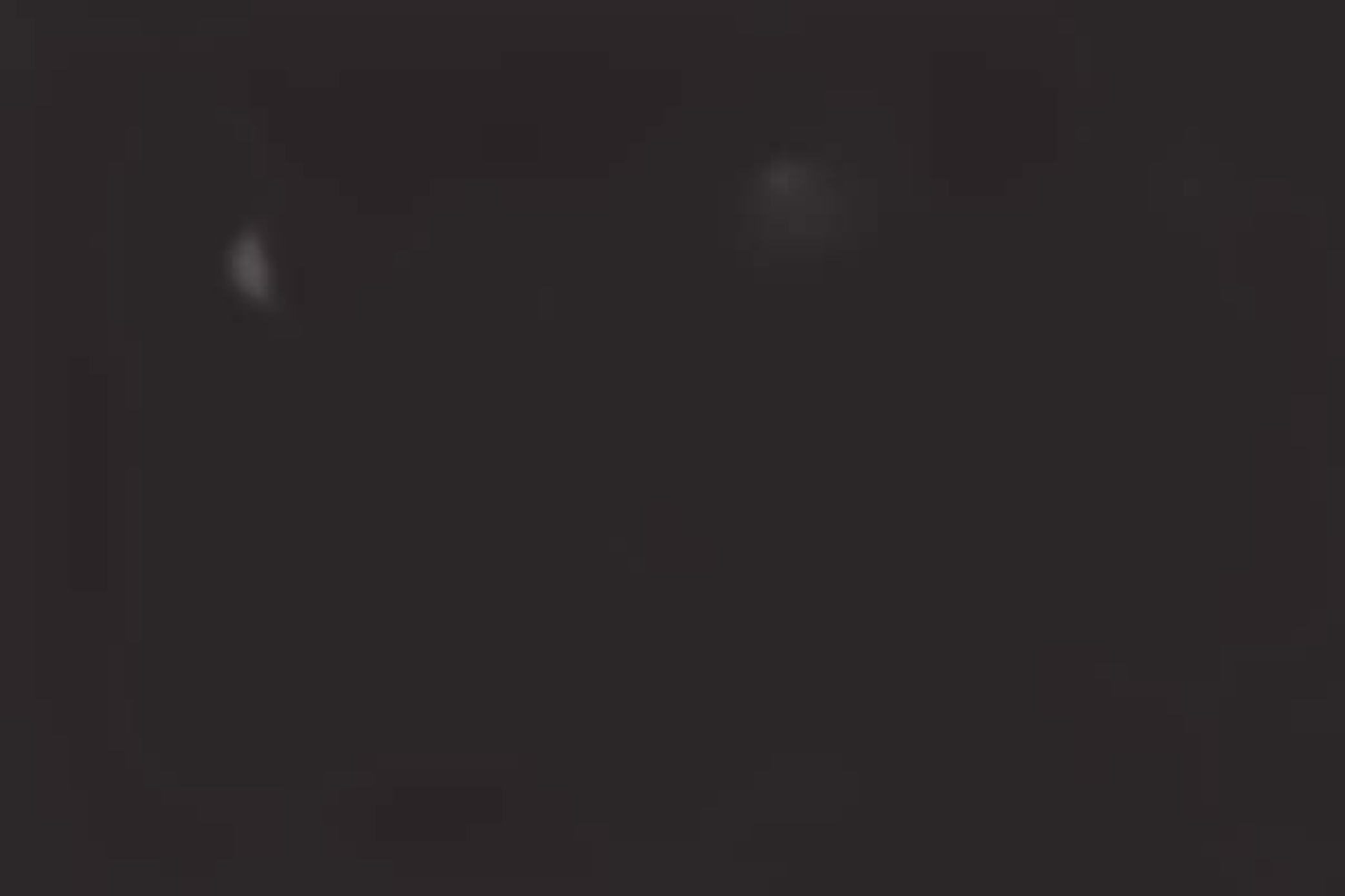}&
			\includegraphics[width=0.18\linewidth]{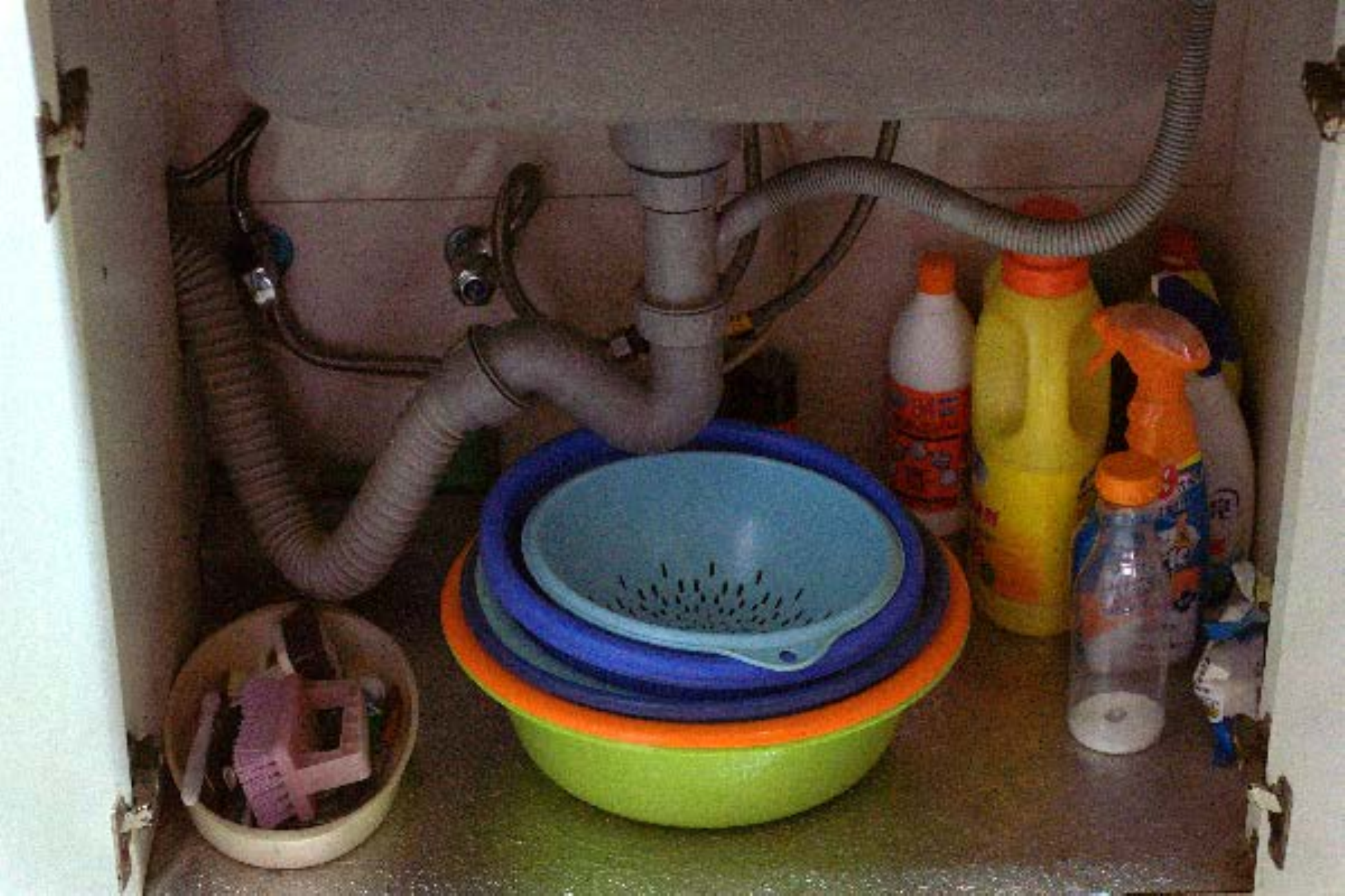}&
			\includegraphics[width=0.18\linewidth]{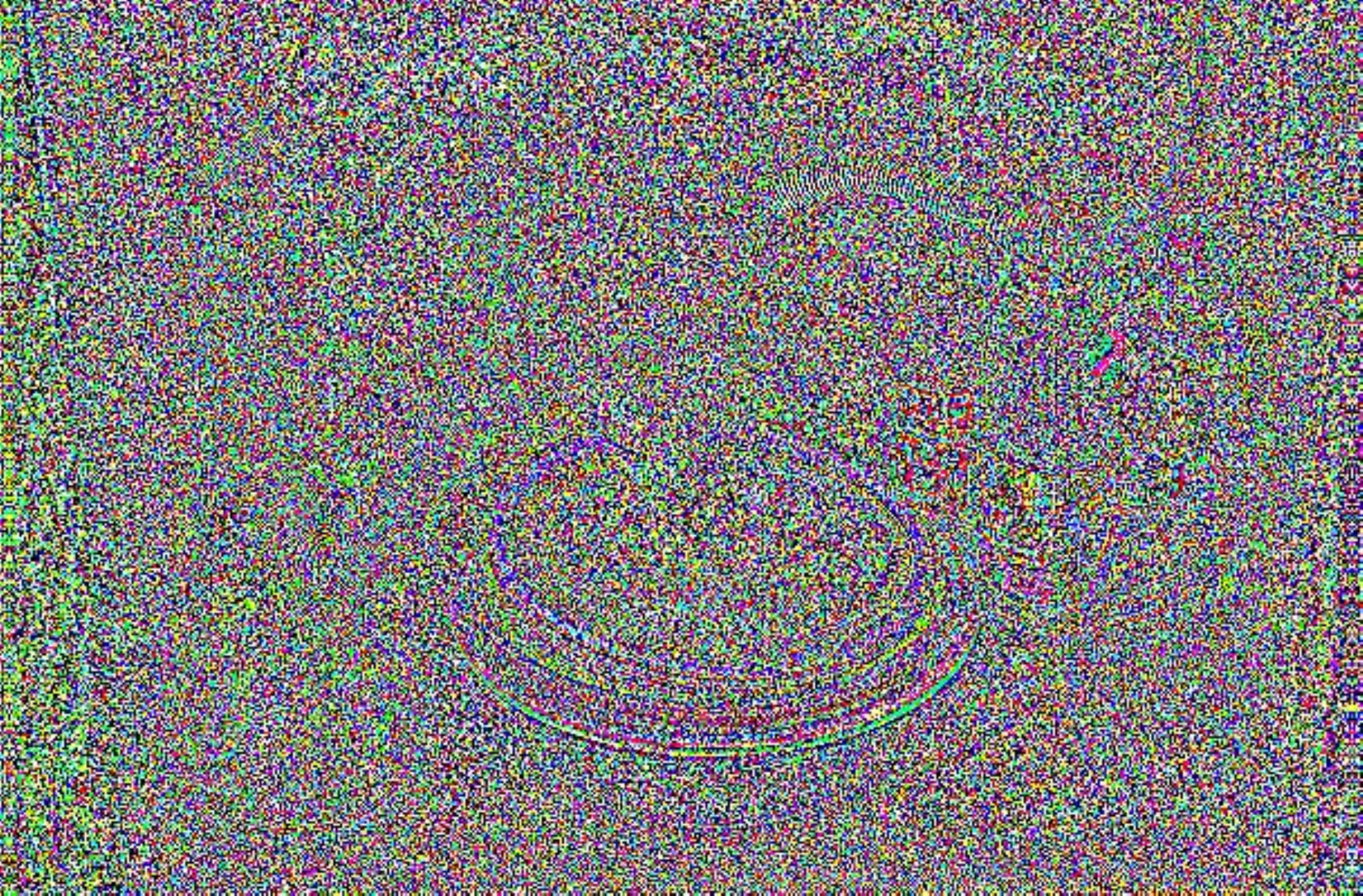}&
			\includegraphics[width=0.18\linewidth]{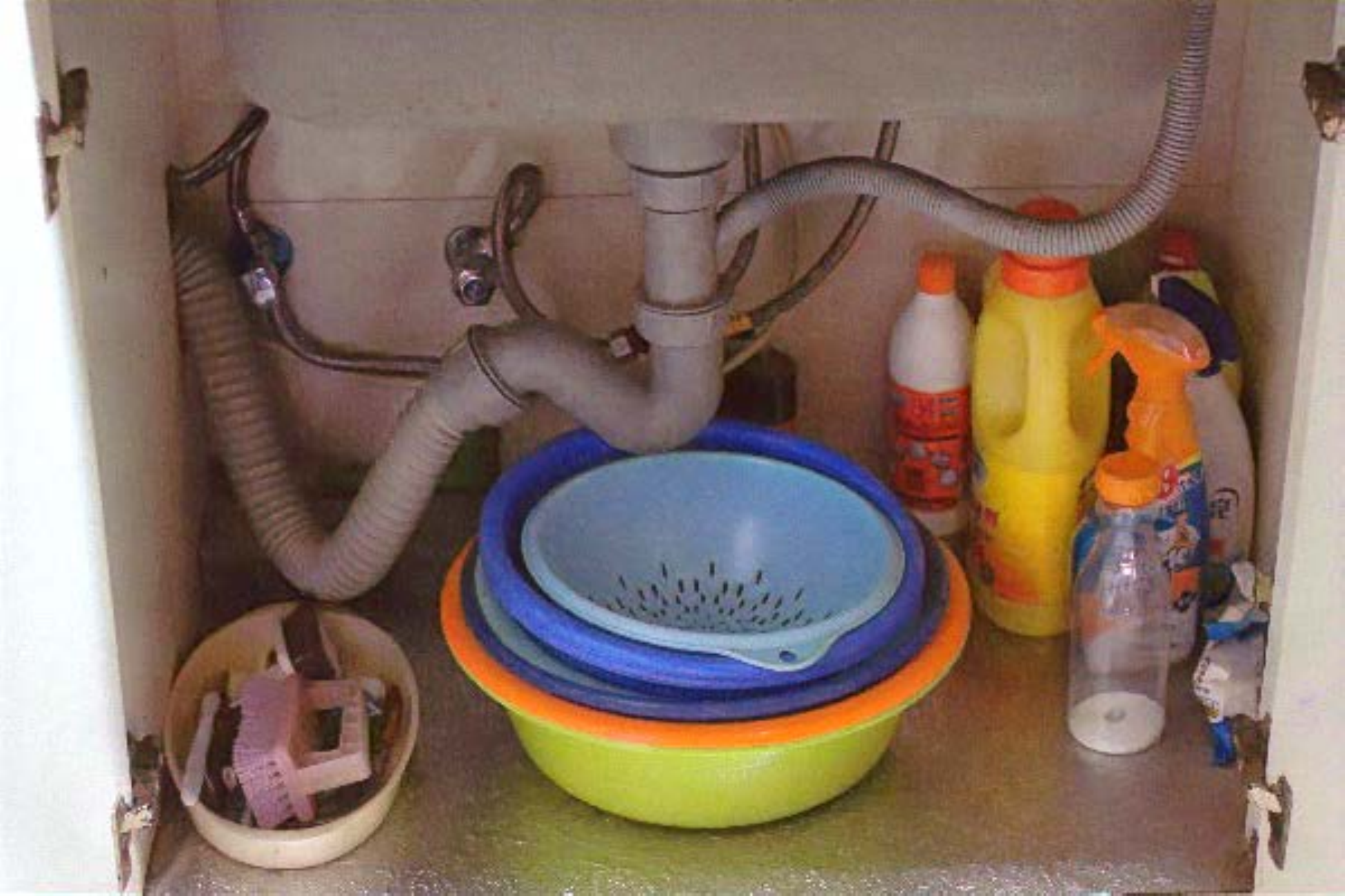}\\
			\includegraphics[width=0.18\linewidth]{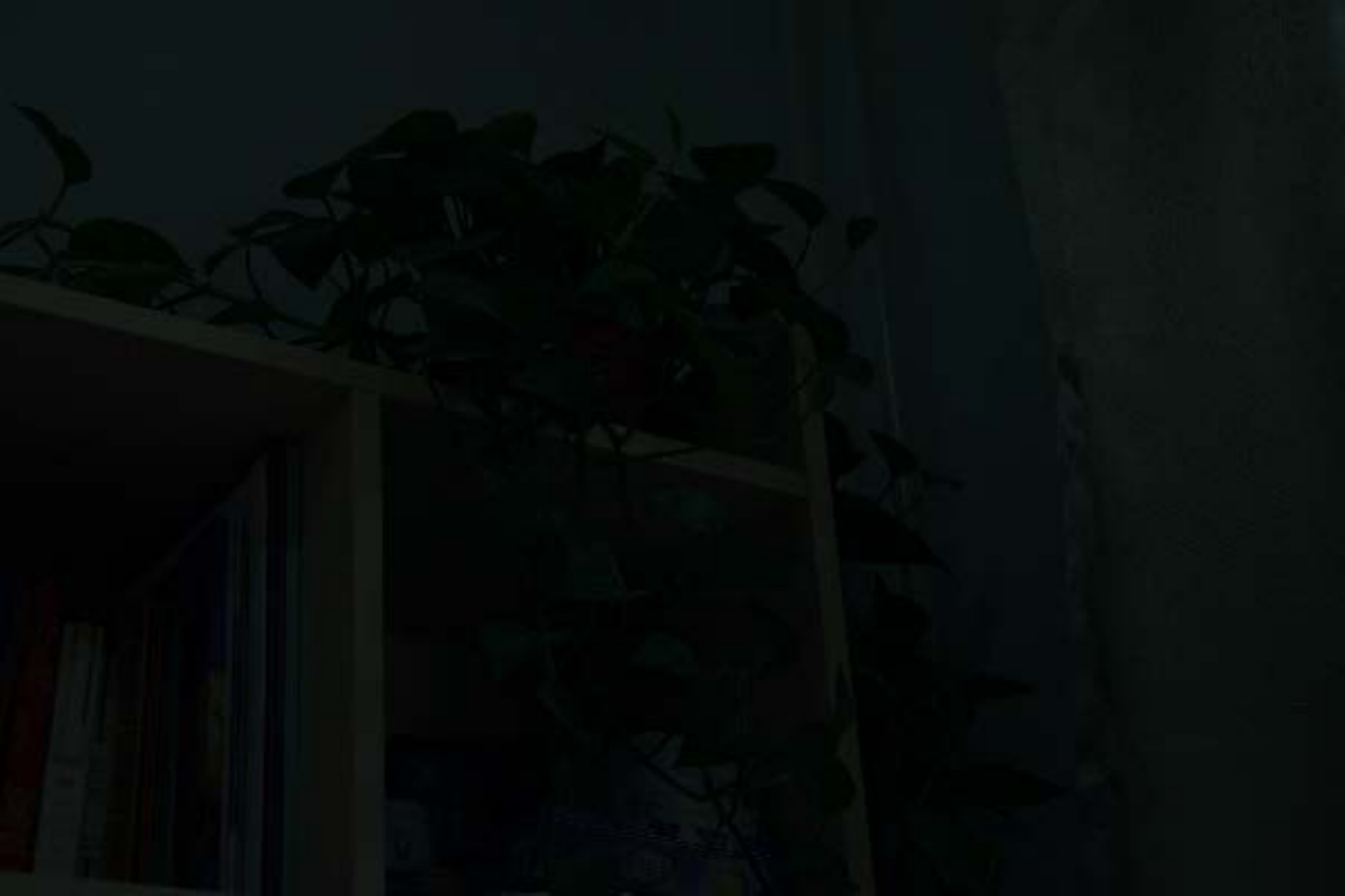}&
			\includegraphics[width=0.18\linewidth]{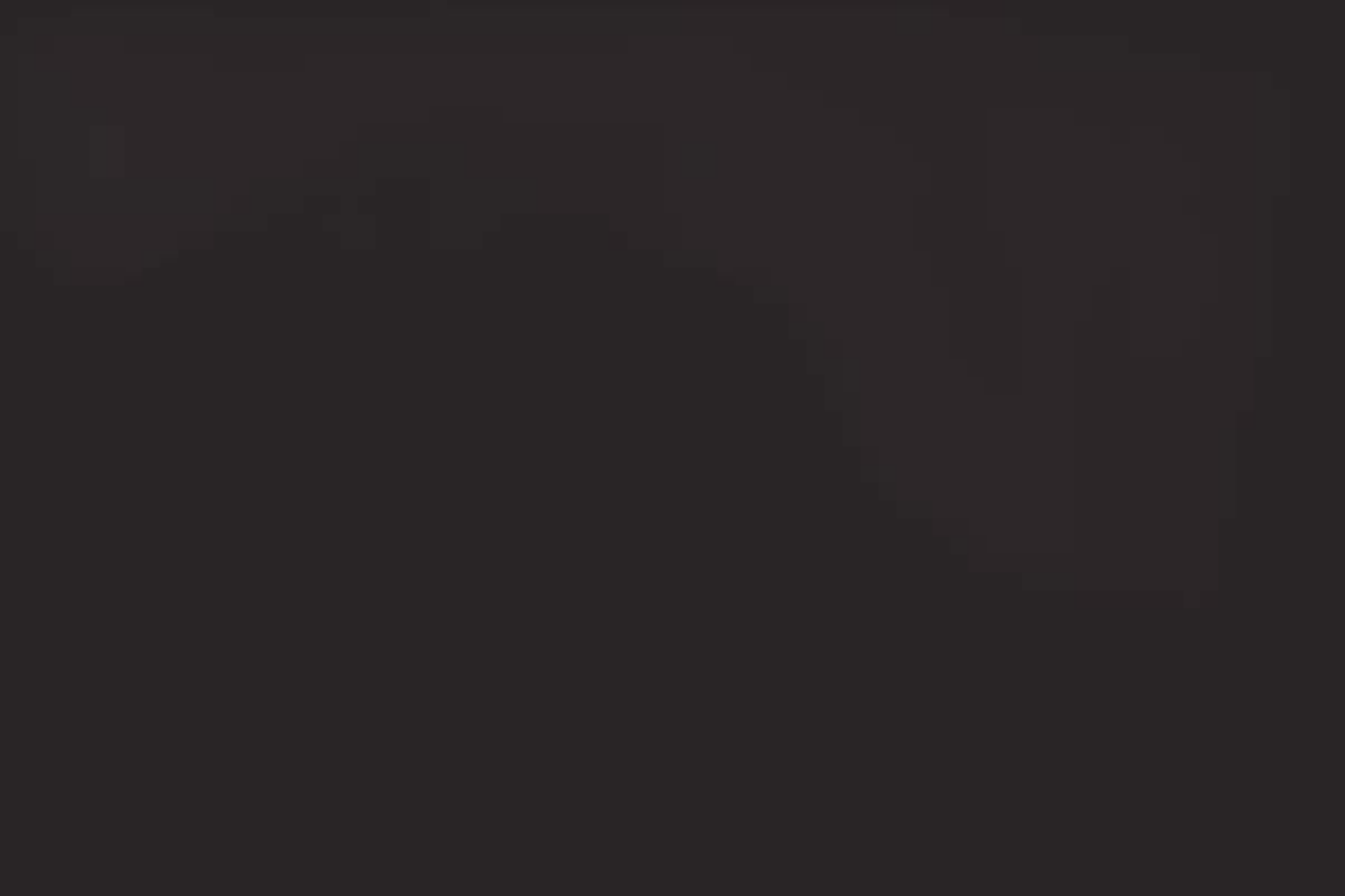}&
			\includegraphics[width=0.18\linewidth]{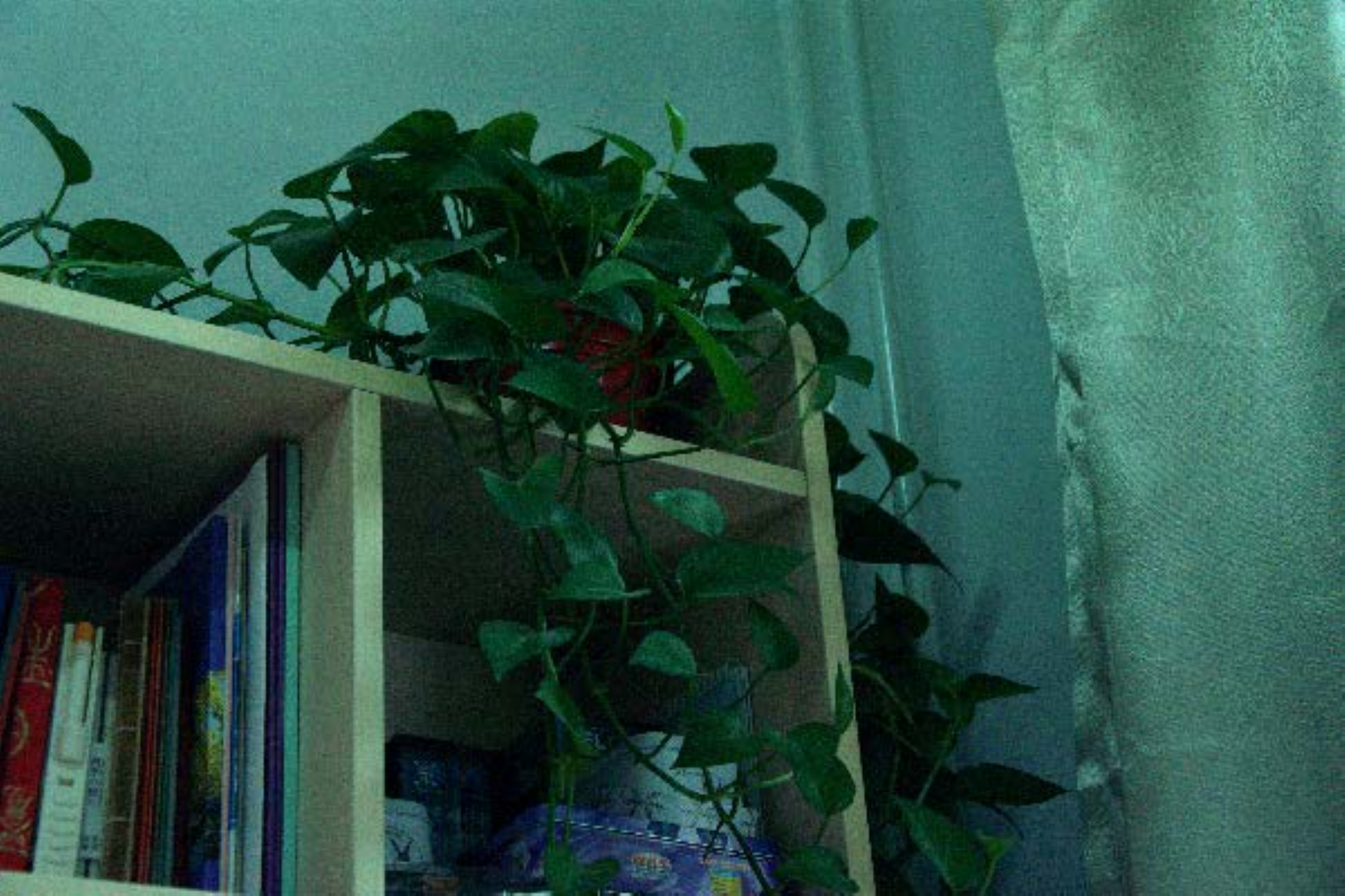}&
			\includegraphics[width=0.18\linewidth]{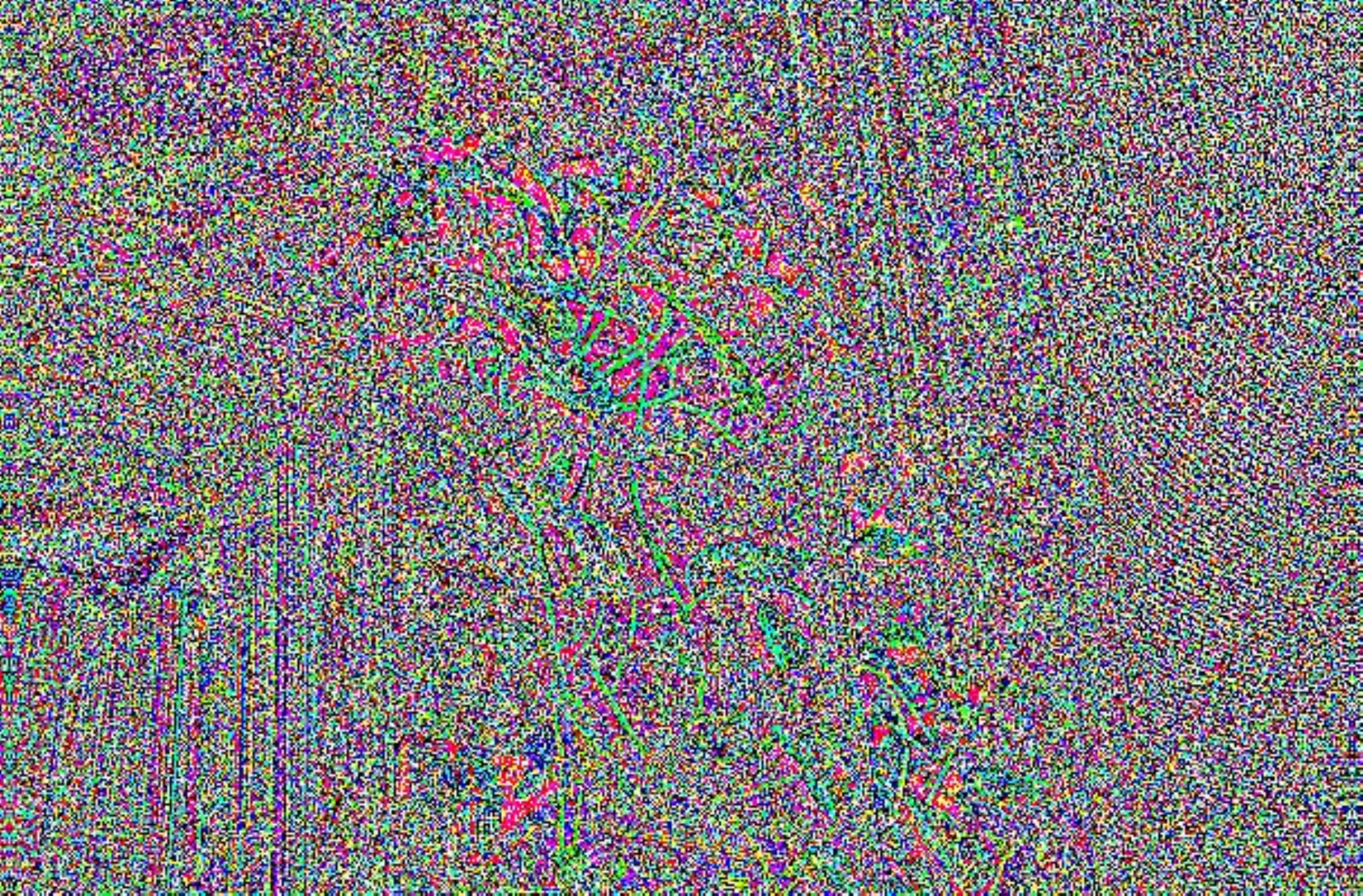}&
			\includegraphics[width=0.18\linewidth]{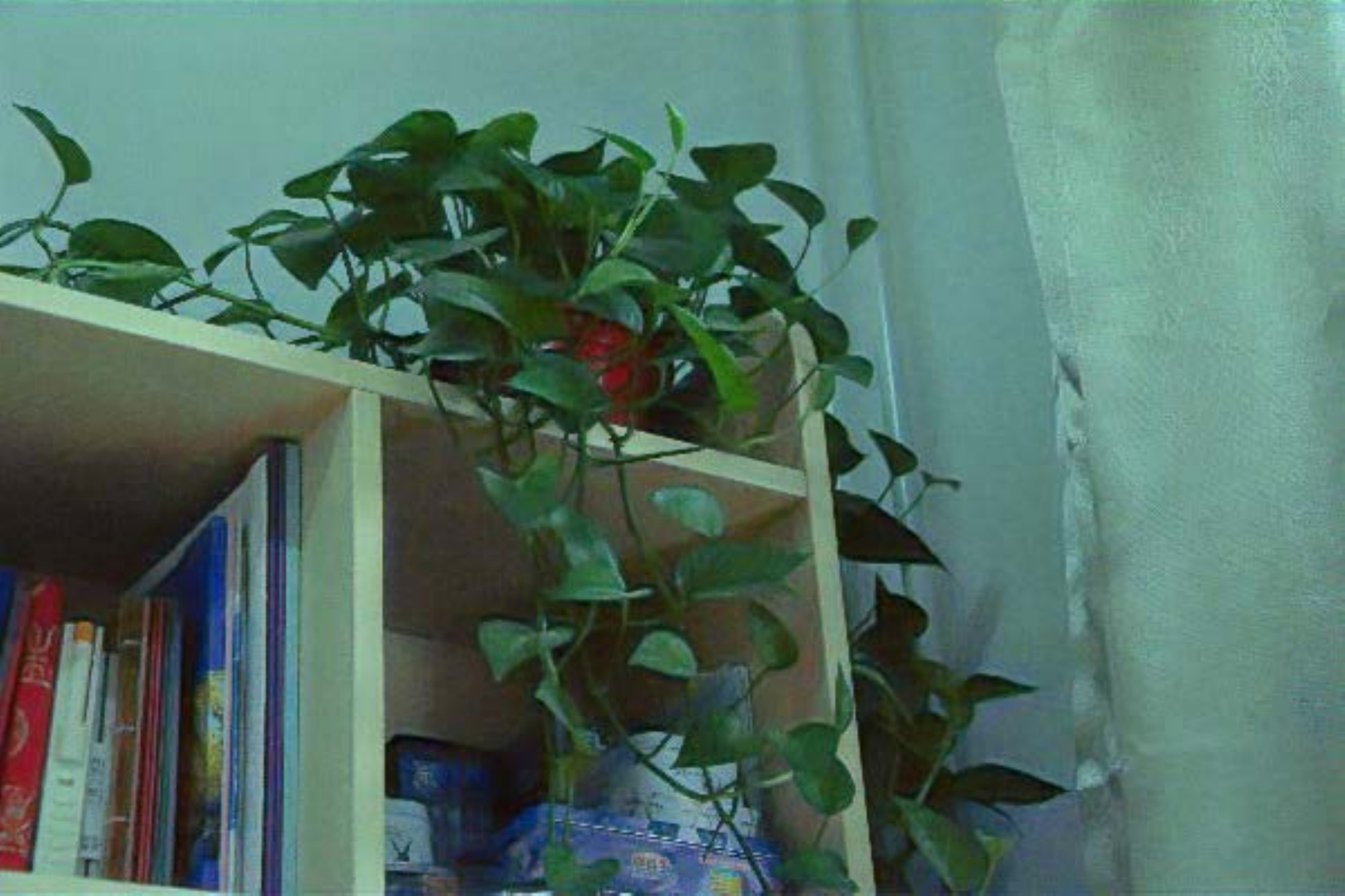}\\			
			\footnotesize $\mathbf{y}$&\footnotesize $\mathbf{x}$&\footnotesize $\mathbf{z}$&\footnotesize $\mathcal{G}(\mathbf{z}; \bm{\Theta}_\mathcal{G})$&\footnotesize $\hat{\mathbf{z}}$\\
		\end{tabular}
		
		%	\vspace{-0.4cm}
		\caption{Intermediate results, including illumination $\mathbf{x}$, enhanced image before denoising $\mathbf{z}$, noise map $\mathcal{G}(\mathbf{z}; \bm{\Theta}_\mathcal{G})$ after normalization,  and final output $\hat{\mathbf{z}}$.}
		\label{fig: Intermediate}
	\end{figure}

	Last but not least, we displayed the intermediate results of our proposed network in Fig.~\ref{fig: Intermediate}.
	%which reflects the input and output results of each module in the process of enhancement and denoising. 
	Although we did not use a certain loss to constrain the illumination, it still seems smooth. Also, the noise map did not contain much marginal information. That is to say, after going through the denoising module, the original image would not be blurred much edge information. Thus, it proved that our training strategy is helpful for the whole network to generate the original properties of every intermediate result, which played a pivotal role in recovering sufficiently accurate image information.
	
	%-------------------------------------------------------------------------
	
	\section{Conclusions and Future Works}
	%In this study, we designed a new bilevel learning scheme to bridge the gap between low-light scenes. It learned the latent correspondence and the respective characteristic among different data distributions so as to fit more complex scenarios. Only a small amount of fine-tuned on the scene-specific decoder was needed to adapt the network to a scene that has never encountered before, which significantly saved the computational cost. Moreover, we established an denoising module which is able to adaptively distinguish noisy images and remove the noise. Extensive experiments further verify our superiority against other state-of-the-art methods. 
	In this study, we proposed a bilevel learning framework for low-light image enhancement to achieve fast adaptation towards different scenes. We explored the underlying relationships among low-light scenes and modeled them from hyperparameter optimization, which enabled us to obtain a scene-independent general encoder. Then we defined a bilevel learning framework to acquire a frozen encoder in the subsequent phases. Furthermore, we designed a reinforced bilevel learning framework to provide the meta-initialization for decoder to further improve the visual quality. Last but not least, a Retinex-induced framework capable of adaptive denoising had been constructed to enhance the practicality of the algorithm, and we applied the proposed learning scheme under different training constraints including supervised and unsupervised forms. A series of evaluated experiments and algorithm analyses were conducted under different sceness thoroughly validated our superiority and effectiveness.
	
	In the future, we will try our best to continue exploring the fast adaptation ability for more computer vision tasks from the perspective of hyperparameter optimization. Moreover, another direction to pay more attention is how to endow fast adaptation ability towards unseen scenes for light-weight or high-efficient models. 
	\begin{acknowledgements}
		This work was supported by the National Natural Science Foundation of China (Nos. U22B2052, 62027826), the LiaoNing Revitalization Talents Program (No. 2022RG04).
	\end{acknowledgements}
	
	\bibliographystyle{spbasic}      % basic style, author-year citations
	\bibliography{egbib} % name your BibTeX data base
	
\end{document}